\documentclass[12pt]{article}
\usepackage{amsmath}
\usepackage{titlesec}
\usepackage{bm}
\usepackage{rotating}
\usepackage{yhmath}
\usepackage{tabularx}
\usepackage{float}
\usepackage{placeins}
\usepackage[linesnumbered,ruled,vlined]{algorithm2e}
\usepackage{pdflscape}

\usepackage{indentfirst}
\DeclareSymbolFont{largesymbol}{OMX}{yhex}{m}{n}
\DeclareMathAccent{\Widehat}{\mathord}{largesymbol}{"62}

\usepackage{graphicx}
\usepackage{url} % not crucial - just used below for the URL 

\usepackage{makecell}
\allowdisplaybreaks[4]
\RequirePackage[colorlinks,citecolor=blue,urlcolor=blue]{hyperref}
\usepackage[round,authoryear]{natbib}
\usepackage{authblk}
\usepackage{amsmath}
\usepackage{amsfonts}
\usepackage{cleveref}
\usepackage{graphicx}
\graphicspath{{fig}}
\usepackage{epstopdf}
\usepackage{amsopn}
\usepackage{mathrsfs}
\usepackage{enumitem}
\usepackage{makecell}
\usepackage{colortbl}
\usepackage{booktabs}
\usepackage{xcolor}
\usepackage{soul}
\usepackage{longtable}
\usepackage{multirow}
\usepackage{tikz}
\usepackage{dsfont}
\usepackage{makecell}
\usepackage{amssymb}
\usepackage{bbm}
\usepackage{enumitem}   
\usepackage{xr}
\usetikzlibrary{decorations.pathreplacing}

\ifpdf
  \DeclareGraphicsExtensions{.eps,.pdf,.png,.jpg}
\else
  \DeclareGraphicsExtensions{.eps}
\fi

\graphicspath{{pics/}}

\newtheorem{assumption}{Assumption}
\newtheorem{lemma}{Lemma}
\newtheorem{theorem}{Theorem}

\newtheorem{definition}{Definition}
\newtheorem{proposition}{Proposition}
\newtheorem{remark}{Remark}
\makeatletter
\newcommand*{\addFileDependency}[1]{% argument=file name and extension
  \typeout{(#1)}% latexmk will find this if $recorder=0 (however, in that case, it will ignore #1 if it is a .aux or .pdf file etc and it exists! if it doesn't exist, it will appear in the list of dependents regardless)
  \@addtofilelist{#1}% if you want it to appear in \listfiles, not really necessary and latexmk doesn't use this
  \IfFileExists{#1}{}{\typeout{No file #1.}}% latexmk will find this message if #1 doesn't exist (yet)
}
\makeatother

\newcommand{\blind}{1}

\hypersetup{linkcolor=blue}
\begin{document}

\def\spacingset#1{\renewcommand{\baselinestretch}%
{#1}\small\normalsize} \spacingset{1}

\date{}

\if1\blind
{
 \title{\bf
 Deep Conditional Distribution Learning via  Conditional F\"{o}llmer Flow} %
  \author[1,2,3]{\small Jinyuan Chang}
\author[4]{\small Zhao Ding}
\author[5,6]{\small Yuling Jiao}
\author[2]{\small Ruoxuan Li}
\author[6,7]{\small Jerry Zhijian Yang}
\affil[1]{\it \small Joint Laboratory of Data Science and Business Intelligence, Institute of Statistical
Interdisciplinary Research, Southwestern University of Finance and Economics, Chengdu, Sichuan, China
}
\affil[2]{\it \small State Key Laboratory of Mathematical Sciences, Academy of
Mathematics and Systems Science, Chinese Academy of Sciences, Beijing, China}
\affil[3]{\it \small 
China Center for Economic Research, Peking University, Beijing, China
}
\affil[4]{\it \small Department of Applied Mathematics, Hong Kong Polytechnic University, Hong Kong SAR}
\affil[5]{\it \small School of Artificial Intelligence, Wuhan University, Wuhan, Hubei, China}
\affil[6]{\it \small Hubei Key Laboratory of Computational Science, Wuhan, Hubei, China}
\affil[7]{\it \small School of Mathematics and Statistics, Wuhan University, Wuhan, Hubei, China}

\setcounter{Maxaffil}{0}
		
		\renewcommand\Affilfont{\itshape\small}

  \maketitle
} \fi

\if0\blind
{
  \bigskip
  \bigskip
  \bigskip
  \begin{center}
    {\LARGE\bf Deep Conditional Distribution Learning via  Conditional F\"{o}llmer Flow}
\end{center}

  \medskip
} \fi
\spacingset{1.5}
\begin{abstract}
We introduce an ordinary differential equation (ODE) based deep generative method for learning  conditional distributions, named {\it Conditional F\"{o}llmer Flow}. 
Starting
 from a standard Gaussian distribution, the proposed flow could approximate the target conditional distribution very well when the time is close to $1$. For effective implementation, we discretize the flow with Euler's method where we estimate  the velocity field nonparametrically using a deep neural network. Furthermore, we also establish the convergence result for the Wasserstein-2 distance between the distribution of the learned samples and the target conditional distribution, providing the first  comprehensive end-to-end error analysis for conditional distribution learning via ODE flow. Our numerical experiments showcase its effectiveness across a range of scenarios, from standard nonparametric conditional density estimation problems to more intricate challenges involving image data, illustrating its superiority over various existing methods.
\end{abstract}

\noindent%
{\it Keywords:}  conditional distribution learning, deep neural networks, end-to-end error bound, ODE flow. %

\spacingset{1.69}
\setlength{\abovedisplayskip}{0.2\baselineskip}
\setlength{\belowdisplayskip}{0.2\baselineskip}
\setlength{\abovedisplayshortskip}{0.2\baselineskip}
\setlength{\belowdisplayshortskip}{0.2\baselineskip}

\section{Introduction}
\label{sec:Introduction}
With the rapid advancements in data storage technology and the accessibility of more powerful computing resources, our human society is taking significant strides into the era of Artificial Intelligence (AI). Recent influential Artificial General Intelligence (AGI) products, like ChatGPT, Stable Diffusion, and Sora have
demonstrated remarkable capabilities in generating high-quality text, image, or video content based on user-provided prompts, which are making a revolutionary shift in the way we live and work. Notably, Statistics plays a crucial role in the development of these AGI products. A fundamental statistical problem involved is how to generate samples efficiently following a learned high-dimensional conditional distribution \citep{liu2024sora, esser2024scaling}.

Intuitively, to solve this fundamental statistical problem, we can first estimate  the conditional distribution and then generate samples from the obtained estimation. A wealth of classical literature has already delved into nonparametric conditional distribution  estimation, including smoothing methods \citep{rosenblatt1969, %
hyndman1996estimating, chen2001,
hall2005}, regression reformulation \citep{fan1996estimation, fan2004crossvalidation}, %
basis function expansion \citep{Sugi2010, izbicki2016nonparametric, izbicki2017converting}. Nevertheless, all these methods suffer from the `curse of dimensionality', where their performance declines drastically as the dimensionality of the related variables increases. Therefore, such a two-step strategy lacks the ability to handle the problem in high-dimensional scenarios. To address this issue, recent years have seen some novel methods. Benefiting from the advancements in deep generative models, these methods mainly focus on  estimating the samplers directly. For instance, \cite{zhou2022deep} introduce a deep generative approach called GCDS utilizing Generative Adversarial Networks (GANs) \citep{goodfellow2014generative}. With a theoretical guarantee, GCDS succeeds in estimating a high-dimensional conditional sampler. However, GANs are known to suffer from training instability \citep{karras2019style} and mode collapse, necessitating considerable engineering efforts and human tuning. Hence, the performance of GAN-based deep generative models is usually less than satisfactory.

The recent breakthrough in deep generative models, known as the diffusion model \citep{ho2020denoising}, gains notable attention for its superior sample quality and significantly more stable training process in comparison with GANs. The basic idea in \cite{ho2020denoising} is training a denoising model to progressively transform noise data to samples following the target distribution, which is equivalent to learning the drift term of a stochastic differential equation (SDE) \citep{Song2021}. 
Following this, research on SDE-based generative models has flourished. A mainstream direction involves utilizing an Ornstein-Uhlenbeck process to transform the target distribution into a Gaussian distribution  \citep{ho2020denoising, Song2021, meng2021sdedit}. Then, solving the time-reversed SDE will yield a sampler. The error analysis of SDE-based generative models can be found in, for example, %
\cite{wang2021deep}, \cite{de2022convergence}, \cite{chen2023improved}, \cite{chen2023sampling}, \cite{oko2023diffusion}, %
\cite{lee2023convergence}, and \cite{benton2023linear}.

Based on the fact that learning the drift term of an SDE corresponds to learning the velocity field of a certain ordinary differential equation (ODE), \cite{Song2021} also consider an ODE-based generative model defined over infinite time interval $(0,\infty)$. However, solving infinite time ODEs usually leads to numerical instability \citep{butcher2016numerical}, which could be improved within the framework of stochastic interpolation. There, finite time ODEs can be constructed to transform standard Gaussian distribution into target distributions. Built on this, several ODE-based generative models have been proposed, see, for example,
\cite{liu2022flow}, \cite{albergo2022building}, \cite{xu2022poisson}, \cite{liu2023flowgrad}, and \cite{gao2024gaussian}.

Despite their achievement, aforementioned SDE/ODE-based methods are not designed to generate samples from conditional distributions. It is important to emphasize that conditional sampling is fundamentally more challenging than unconditional sampling. One might consider learning a corresponding conditional distribution for each fixed value of the conditional variable, thereby reducing the problem to unconditional context. While this approach is feasible for discrete conditional variables, it has significant drawbacks. During the training of each conditional distribution, training data not paired with the current conditional variable value are discarded, leading to inefficient data utilization and reduced training efficiency. For continuous conditional variables, this method is impractical because it essentially requires fitting an infinite number of conditional distributions. Therefore, conditional sampling inherently requires additional methodological innovation. Lately, there have been proposals of several SDE/ODE-based generative models for conditional sampling
\citep{shi2022conditional, Albergo2024, huang2023conditional, zheng2023guided, wildberger2024flow}, while consistency of the learned conditional distribution has not been studied in these works.

In this paper, we introduce a novel ODE-based conditional sampling method named {\it Conditional Föllmer Flow}, which has the following main advantages: 

Firstly, we propose an ODE system over a unit time interval, and both the velocity field and the ODE flow itself exhibit Lipschitz continuity, mathematically ensuring robustness in training and sampling processes. This assures that our method excels in managing high-dimensional problems and can handle both continuous and discrete variables. 
Furthermore, our proposed ODE-based method maps random Gaussian noises to some samples with  distribution arbitrarily close to the target conditional distribution, which allows us to utilize the noise-sample pairs generated by the ODE-based method to train a new end-to-end neural network using least square fitting.
Ideally, the end-to-end neural network takes Gaussian noises as input and produces the corresponding samples as output. 
It is important to note that the sampling time of the ODE-based method on a time grid with size $N$ is of order $\mathcal{O}(N)$. However, once the end-to-end network is trained, the time required to generate a new sample is significantly reduced to $\mathcal{O}(1)$ since only a single network evaluation is needed. See more detailed discussion on this in the last paragraph of Section \ref{sec:scheme}. It is worth emphasizing that the end-to-end generator cannot be derived from the SDE-based methods due to their inherent stochastic nature for each particle.

 Secondly, under some mild conditions, we demonstrate that the distribution of the generated samples will converge in probability to the target conditional distribution with a certain rate, a result rarely seen for ODE-based methods. 
While similar theoretical guarantees exist for SDE-based methods in unconditional scenario, analyses for their ODE-based counterparts in unconditional scenario often rely on some `uncheckable' regularity assumptions. Specifically, prior works for unconditional sampling may directly posit that the true velocity field or score function possesses Lipschitz continuity \citep{chen2023probability}, second-order smoothness \citep{chen2023restoration}, or specific time regularity \citep{gao2024convergence}. Also, some assumptions are made about the estimated score or velocity field, for instance, requiring it to be Lipschitz continuous \citep{chen2023probability, albergo2022building, benton2024error}. Moreover, many existing error analyses do not consider all
sources of error. They may focus on a single source, such as ODE perturbation \citep{albergo2022building, albergo2023stochastic} or numerical discretization \citep{chen2023restoration}, or are conditioned on an already small $L^2$ estimation error of the score or velocity field rather than deriving it from a learning process \citep{chen2023probability, benton2024error, gao2024convergence, li2024towards}. As for the conditional scenario, \cite{huang2023conditional} have extended the SDE/ODE-based methods \citep{albergo2023stochastic} and conducted stability analysis of their proposed SDE. However, their analysis depends on the examination of two separate neural networks and fails to account for errors introduced by sampling algorithms. More importantly, such an analysis framework only works for the SDE-based methods but is not applicable to the ODE-based methods. Our work distinguishes itself by taking a more foundational approach: we first derive the regularity of the true velocity field from elementary properties of the data distribution, and then provide a comprehensive end-to-end analysis that explicitly bounds the velocity field estimation error from data and combines it with perturbation, discretization, and approximation errors to form an overall error bound. To the best of our knowledge, our work presents the first in-depth convergence analysis of ODE-based conditional generative methods.           
    
 Thirdly, we conduct a series of numerical experiments and provide a comprehensive assessment of the versatility and efficacy of the conditional Föllmer flow. In small-scale and low-dimensional scenarios, our method exhibits comparable or superior performance to traditional conditional density estimation methods and some other deep generative methods. Meanwhile, our method can be used to address statistical prediction tasks. As shown in Section \ref{subsec:wine}, we can utilize the conditional sampler derived from the conditional Föllmer flow to construct more precise prediction intervals through repeated sampling, which not only provides more robust decision-making foundations but also offers a generalized framework for solving prediction problems. Furthermore, our approach excels in high-dimensional examples such as image generation and reconstruction where traditional methods encounter difficulties, which highlights its adaptability to modern data challenges. %

The rest of this paper is structured as follows. In Section \ref{sec:Preliminaries}, we provide notation and formally introduce the concept of the conditional F\"{o}llmer flow. In Section \ref{sec:scheme}, we elucidate how to employ the conditional F\"{o}llmer flow to design numerical algorithms for conditional sampling. In Section \ref{sec:Error Analysis}, we analyze the convergence of our numerical scheme, and present a comprehensive error analysis. In Section \ref{sec:numerical}, we conduct numerical experiments to demonstrate its performance. Finally, in Section \ref{sec:discuss}, we discuss our work and outline some future directions. All technical proofs and some additional numerical study results are provided in the supplementary material.

\section{Preliminaries}
\label{sec:Preliminaries}

\subsection{Notation}
For a vector $\mathbf{x}=(x_1,\ldots,x_d)^{\rm{T}} \in \mathbb{R}^{d}$, the $\ell^2$-norm and $\ell^{\infty}$-norm of $\mathbf{x}$ are, respectively, denoted by $|\mathbf{x}|_2 := \sqrt{\sum_{i=1}^d x_i^2}$ and $|\mathbf{x}|_{\infty} := \max_{1\leq i\leq d}|x_i|$. For a probability density $\pi$ and a measurable function $f: \mathbb{R}^{d} \rightarrow \mathbb{R}$, the $L^2(\pi)$-norm of $f$ is defined as $\|f\|_{L^2(\pi)} := \sqrt{\int f^2(\mathbf{x}) \pi(\mathbf{x}) \,\mathrm{d} \mathbf{x}}$. For a vector function $\mathbf{v}: \mathbb{R}^{d} \rightarrow \mathbb{R}^{d}$, its $L^2(\pi)$-norm is defined as $\|\mathbf{v}\|_{L^2(\pi)} := \||\mathbf{v}|_2\|_{L^2(\pi)}$. We denote by ${\rm tr}(\cdot)$ the trace operator on a square matrix. The $d$-dimensional identity matrix is denoted by $\mathbf{I}_d$. We use $\mathcal{U}(a, b)$ to denote the uniform distribution on interval $(a, b)$, and use $\mathcal{N}(\mathbf{0}, \mathbf{I}_{d})$ to denote the $d$-dimensional standard Gaussian distribution. For two positive sequences $\{a_n\}_{n\geq1}$ and $\{b_n\}_{n\geq1}$, 
the asymptotic notation $a_n = \mathcal{O}(b_n)$ means that $a_n \leq C b_n$ for some constant $C > 0$. The notation $\widetilde{\mathcal{O}}(\cdot)$ is used to ignore logarithmic terms. Given two distributions $\mu$ and $\nu$, their Wasserstein-2 distance $W_2(\mu, \nu)$ is defined as $W^2_2(\mu, \nu) := \inf_{\pi \in \Pi(\mu, \nu)} \mathbb{E}_{(\mathbf{x}, \mathbf{y}) \sim \pi}(|\mathbf{x} - \mathbf{y}|_2^2)$, where $\Pi(\mu, \nu)$ is the set of all couplings of $\mu$ and $\nu$. A coupling is a joint distribution on $\mathbb{R}^{d} \times \mathbb{R}^{d}$ whose marginals are $\mu$ and $\nu$ on the first and second factors, respectively.

\subsection{Conditional F\"{o}llmer Flow} \label{sec:CFF}
 
    Suppose that we have a random vector $\mathbf{X} \in \mathbb{R}^{d_x}$, where $d_x$ may be quite large, and a conditional variable $\mathbf{Y} \in \mathbb{R}^{d_y}$ related to $\mathbf{X}$, forming a pair $(\mathbf{X}, \mathbf{Y})$. The marginal densities of $\mathbf{X}$ and $\mathbf{Y}$ are denoted as $p_{x}(\mathbf{x})$ and $p_{y}(\mathbf{y})$, respectively, while the joint density of $(\mathbf{X}, \mathbf{Y})$ is denoted as $p_{x, y}(\mathbf{x}, \mathbf{y})$. Using $p_{x\mkern 2mu|\mkern2muy}(\mathbf{x}\mkern 2mu|\mkern2mu\mathbf{y})$ to represent the conditional density of $\mathbf{X}$ given $\mathbf{Y}=\mathbf{y}$, our interest lies in efficiently sampling from $p_{x\mkern 2mu|\mkern2muy}(\mathbf{x}\mkern 2mu|\mkern2mu\mathbf{y})$. Such interest is underscored by recent advancements in AGI products such as Stable Diffusion and Sora, which specialize in sampling from distributions with $\mathbf{Y}$ representing user-provided multimodal prompts and $\mathbf{X}$ standing for high-dimensional textual, image, or video content corresponding to $\mathbf{Y}$. In our numerical experiments, we will study  class-conditional image generation and image inpainting. These are two representative tasks that fit this framework, with $\mathbf{Y}$ a label prompt or partial observation and $\mathbf{X}$ the generated content. See Section \ref{subsec:mnist} for details.

   Notably, when $\mathbf{X}$ represents the response and $\mathbf{Y}$ the related covariates, we enter the realm of statistical prediction problems. If we are interested in the conditional mean $\mathbb{E}(\mathbf{X} \mkern 2mu | \mkern 2mu \mathbf{Y})$,  we can just repeatedly draw samples from $p_{x\mkern 2mu|\mkern2muy}(\mathbf{x}\mkern 2mu|\mkern2mu\mathbf{y})$ and then use the sample mean to estimate $\mathbb{E}(\mathbf{X} \mkern 2mu | \mkern 2mu \mathbf{Y})$. More importantly, we can also construct the confidence region of $\mathbb{E}(\mathbf{X} \mkern 2mu | \mkern 2mu \mathbf{Y} = \mathbf{y})$ through repeated sampling, thereby obtaining more precise and informative prediction intervals to aid decision-making. See Sections \ref{subsec:simulation2} and \ref{subsec:wine} for details.
    
    It should be noted that, we actually are not concerned about the specific functional form of $p_{x\mkern 2mu|\mkern2muy}(\mathbf{x} \mkern 2mu|\mkern2mu \mathbf{y})$ in these samplings. This is quite reasonable, since even if we are able to obtain such a functional form in high-dimensional scenarios -- given its impracticality -- designing sampling algorithms based on it, e.g. high-dimensional MCMC, still remains intensely challenging. Therefore, instead of focusing on $p_{x\mkern 2mu|\mkern2muy}(\mathbf{x} \mkern 2mu|\mkern2mu\mathbf{y})$ itself, opting to directly learn a sampler proves to be much more practical. In this paper, we introduce the so-called {\it Conditional F\"{o}llmer Flow}, a novel ODE-based method which can map random Gaussian noises to samples with a distribution arbitrarily close to the target conditional distribution.

\begin{definition}[Conditional F\"{o}llmer Flow] \label{def:Conditional Follmer Flow}
If $\mathbf{Z}(t, \mathbf{y})$ solves the following ODE for any $\mathbf{y} \in [0, B]^{d_y}$:
    \begin{equation} \label{equa:CCF ODE}
\mathrm{d} \mathbf{Z}(t, \mathbf{y})=\mathbf{v}_{\rm{F}}(\mathbf{Z}(t, \mathbf{y}), \mathbf{y}, t)\,\mathrm{d} t\, , \quad t \in[0,1)\,,
\end{equation}
with $\mathbf{Z}(0, \mathbf{y}) \sim \mathcal{N}(\mathbf{0}, \mathbf{I}_{d_x})$, then we call 
$\mathbf{Z}(t, \mathbf{y})$ the conditional F\"{o}llmer flow and $\mathbf{v}_{\rm{F}}$ the conditional F\"{o}llmer velocity field associated to $p_{x\mkern 2mu|\mkern2muy}(\mathbf{x}\mkern 2mu|\mkern2mu\mathbf{y})$, respectively, 
where the velocity field $\mathbf{v}_{\rm{F}}$ is defined by
\begin{equation}
\mathbf{v}_{\rm{F}}(\mathbf{x}, \mathbf{y}, t)=\frac{\mathbf{x}+\mathbf{s}(\mathbf{x}, \mathbf{y}, t)}{t}\,, \quad t \in(0,1)\,, 
\end{equation}
for $\mathbf{v}_{\rm{F}}(\mathbf{x}, \mathbf{y}, 0)=\mathbb{E}(\mathbf{X}\mkern 2mu|\mkern2mu\mathbf{Y}=\mathbf{y}),$ and %
\begin{equation} \label{equa:conditional score function}
        \mathbf{s}(\mathbf{x}, \mathbf{y}, t) = \nabla_{\mathbf{x}} \log f_t(\mathbf{x}\mkern 2mu|\mkern2mu\mathbf{y})\,,\quad t\in[0,1)\,,
    \end{equation}  
with $f_t(\mathbf{x}\mkern 2mu|\mkern2mu\mathbf{y})$ denoting the conditional density of $t\mathbf{X}+\sqrt{1-t^2}\mathbf{W}$ given $\mathbf{Y}=\mathbf{y}$, and $\mathbf{W} \sim \mathcal{N}(\mathbf{0}, \mathbf{I}_{d_x})$ independent of $(\mathbf{X}, \mathbf{Y})$.
\end{definition}

It is easy to see that in Definition \ref{def:Conditional Follmer Flow}, $f_0(\mathbf{x}\mkern 2mu|\mkern2mu\mathbf{y})$ is the density of $\mathcal{N}(\mathbf{0}, \mathbf{I}_{d_x})$.
Meanwhile, we will always use $\mathbf{W}$ to denote a standard Gaussian random vector independent of $(\mathbf{X}, \mathbf{Y})$ in the remaining text. For the convenience of later discussion, we introduce the concept of flow map from ODE theory. Simply put, for an ODE system: $\mathrm{d}{\mathbf{x}}_t=\mathbf{v}(\mathbf{x}_t, t) \mkern2mu \mathrm{d}t$, its flow map $\mathbf{\Phi}_t(\cdot)$ is defined as $\mathbf{\Phi}_t(\mathbf{x}_0)=\mathbf{x}_t$, where $\mathbf{x}_0 \in \mathbb{R}^{d}$ is the initial value of the ODE, and $\mathbf{x}_t$ is the ODE solution at time $t$ with initial value $\mathbf{x}_0$. Thus, $\mathbf{\Phi}_t(\cdot)$ determines a mapping from $\mathbb{R}^{d}$ to $\mathbb{R}^{d}$. Based on this, we propose the definition of {\it Conditional F\"{o}llmer Flow Map}.

\begin{definition}[Conditional F\"{o}llmer Flow Map] \label{def:CFFM}
    We refer to the flow map related to the conditional F\"{o}llmer flow $\mathbf{Z}(t, \mathbf{y})$ or the conditional F\"{o}llmer velocity field $\mathbf{v}_{\rm{F}}$ as the conditional F\"{o}llmer flow map, denoted by $\mathbf{F}_t(\cdot, \mathbf{y})$.
\end{definition}

Note that given $\mathbf{y}  \in  [0, B]^{d_{y}}$, $\{\mathbf{Z}(t, \mathbf{y})\}_{t \in [0,1)}$ forms a family of random vectors, and all the randomness originates from the initial point $\mathbf{Z}(0, \mathbf{y})$ which follows the standard Gaussian distribution $\mathcal{N}(\mathbf{0}, \mathbf{I}_{d_x})$, as the subsequent evolution is determined by a deterministic ODE system. 
Theorem \ref{thm:well posedness} ensures that as $t \rightarrow 1$, the density of $\mathbf{Z}(t, \mathbf{y})$, or equivalently expressed as $\mathbf{F}_t(\mathbf{Z}(0, \mathbf{y}), \mathbf{y})$, 
can arbitrarily approach the target conditional density $p_{x\mkern 2mu|\mkern2muy}(\mathbf{x} \mkern 2mu|\mkern2mu \mathbf{y})$. For brevity, we will use $\mathbf{Z}^{\mathbf{y}}_t$ to represent $\mathbf{Z}(t, \mathbf{y})$ in the remaining text. To state Theorem \ref{thm:well posedness}, we need the following two mild assumptions.

\begin{assumption}[Bounded condition]
\label{assump:label}
$p_{y}(\mathbf{y})$ is supported on $[0,B]^{d_y}$, where $B >0$ is a fixed constant.
\end{assumption}

\begin{assumption}[Bounded conditional distribution] \label{assump:bounded support}
  $p_{x\mkern 2mu|\mkern2muy}(\mathbf{x}\mkern 2mu|\mkern2mu\mathbf{y})$ is supported on $[0,1]^{d_x}$ for any $\mathbf{y} \in [0,B]^{d_y}$, resulting in $p_{x}(\mathbf{x})$ also being supported on $[0, 1]^{d_x}$.
\end{assumption}

 These assumptions are common in the literature of generative learning. Many real-world data are bounded by construction \citep{esser2024scaling, liu2024sora}. For example, image pixels take integer values from $0$ to $255$ and are usually normalized to the unit interval before modeling, and audio signals are likewise bounded, so the compact support assumed here is natural. In the nonparametric statistics literature, boundedness is also a mild and commonly used assumption. It is in fact possible to extend our analysis to unbounded response variables, provided their distribution has appropriate tail decay. The standard route is a truncation argument: one fixes a radius $B$, treats the target on $[-B,B]^{d_x}$ exactly as in the present bounded case, and bounds the residual mass beyond this region using the tail condition \citep{gyorfi2002distribution}. To keep the truncation error negligible, $B$ must be allowed to diverge with the sample size $n$. This diverging $B$ may then propagate into the convergence rate of the velocity field, making the theoretical results more involved. We emphasize that our current proof framework can extend to this setting with no essential effort. The only cost is more tedious computation. We therefore retain the boundedness assumptions to highlight our main idea and simplify the presentation.

\begin{theorem} \label{thm:well posedness}
    Let Assumptions {\rm\ref{assump:label}} and {\rm\ref{assump:bounded support}} hold. Then, for any $\mathbf{y} \in [0, B]^{d_y}$, the conditional F\"{o}llmer flow $(\mathbf{Z}^{\mathbf{y}}_t)_{t \in [0,1)}$ associated to $p_{x\mkern 2mu|\mkern2muy}(\mathbf{x} \mkern 2mu|\mkern2mu \mathbf{y})$ is a unique solution to the ODE specified in Definition {\rm\ref{def:Conditional Follmer Flow}}. Also, we have $\mathbf{F}_t(\mathbf{Z}_0, \mathbf{y}) \sim f_t(\mathbf{x}\mkern 2mu|\mkern2mu \mathbf{y})$ for $t \in [0, 1)$,
    where $f_t(\mathbf{x}\mkern 2mu|\mkern2mu \mathbf{y})$ is 
    specified in Definition {\rm\ref{def:Conditional Follmer Flow}} and $\mathbf{Z}_0 \sim \mathcal{N}(\mathbf{0}, \mathbf{I}_{d_x})$. Moreover, for any $\mathbf{y} \in [0, B]^{d_y}$, we have $$W^2_2(f_t(\mathbf{x}\mkern 2mu|\mkern2mu \mathbf{y}), p_{x\mkern 2mu|\mkern2muy}(\mathbf{x} \mkern 2mu|\mkern2mu\mathbf{y})) \leq 4d_x(1-t) \rightarrow 0$$ as $t \rightarrow 1$.
\end{theorem}

Theorem \ref{thm:well posedness}  establishes the theoretical guarantee of conditional F\"{o}llmer flow, whose proof is given in Section B of the supplementary material. 
\begin{remark}
A related concept to conditional F\"{o}llmer flow is the so-called F\"{o}llmer flow, which is also an ODE-based method that can be used to conduct unconditional sampling. As we demonstrate in Section {\rm A} of the supplementary material, using procedures designed for unconditional sampling based on F\"{o}llmer flow to conduct conditional sampling will lead to severe data inefficiency or computational infeasibility. In contrast, our conditional F\"{o}llmer flow provides a unified framework for efficient conditional sampling, which is a non-trivial extension of the ODE-based methods designed for unconditional sampling. See also the discussion below Proposition {\rm\ref{prop:training object}} in Section {\rm\ref{sec:scheme}}.
\end{remark}

\begin{remark}
    In general, F\"{o}llmer flow and some other ODE-based methods {\rm\citep{liu2022flow, Lipman2023Flow}} can be unified within the framework of stochastic interpolant {\rm\citep{albergo2022building}}, which defines a path $\mathbf{W}_t = a_t \mathbf{X} + b_t \mathbf{W}$ from noise to data. The key difference lies in the coefficients. Linear interpolants, such as {\rm\cite{liu2022flow}}, set $(a_t, b_t) = (t, 1-t)$, whereas the F\"ollmer flow uses $(a_t, b_t) = (t, \sqrt{1-t^2})$. Under Assumptions {\rm\ref{assump:label}} and {\rm\ref{assump:bounded support}}, Proposition {\rm P1} in the supplementary material shows that the Lipschitz constants of the  F\"ollmer flow with respect to $\mathbf{x}$ and $t$ are of order $\mathcal{O}\{(1-T)^{-2}\}$ and $\mathcal{O}\{(1-T)^{-3}\}$, respectively. Similarly, one can show that for linear interpolants {\rm\citep{liu2022flow}}, the corresponding orders are, respectively, $\mathcal{O}\{(1-T)^{-3}\}$ and $\mathcal{O}\{(1-T)^{-4}\}$. Therefore, as $T \to 1$, the regularity of the velocity field associated with F\"ollmer flow is better than that of linear interpolants, indicating a more stable training process for estimating the associated velocity field of F\"{o}llmer flow near $T=1$. This is the technical advantage of F\"ollmer flow.
\end{remark}

\section{Sampling Procedure} \label{sec:scheme}

Leveraging insights from \cite{albergo2023stochastic}, we introduce the following proposition, which first offers an alternative expression for $\mathbf{v}_{\rm{F}}$ as a conditional expectation. Based on this, it then constructs a quadratic objective function for which $\mathbf{v}_{\rm{F}}$ stands out as the unique minimizer.

  \begin{proposition} \label{prop:training object}
  For the conditional F\"{o}llmer velocity field $\mathbf{v}_{\rm{F}}$ on $[0, T]$ with $T<1$, the following two assertions are satisfied. 
    \begin{enumerate}
        \item[{\rm(i)}] $\mathbf{v}_{\rm{F}}$ has a conditional expectation form
        \begin{equation} \label{equa: condi expec vF}
          \mathbf{v}_{\rm{F}}(\mathbf{x}, \mathbf{y}, t)=\mathbb{E}\bigg(\mathbf{X}-\frac{t}{\sqrt{1-t^2}}\mathbf{W} \, \bigg| \, t\mathbf{X}+\sqrt{1-t^2}\mathbf{W}=\mathbf{x}, \mathbf{Y}=\mathbf{y}\bigg)\,,
      \end{equation}
        
        \item[{\rm(ii)}] $\mathbf{v}_{\rm{F}}$ is the unique minimizer of the quadratic objective
    \begin{equation} \label{equa:population loss}
    \mathcal{L}(\mathbf{v}):=\frac{1}{T}\int_0^{T}\mathbb{E}\left\{\bigg|\mathbf{X}-\frac{t}{\sqrt{1-t^2}}\mathbf{W}-\mathbf{v}(t\mathbf{X}+\sqrt{1-t^2}\mathbf{W}, \mathbf{Y}, t)\bigg|_2^2\right\}\,\mathrm{d} t\,.
\end{equation}
    \end{enumerate}
\end{proposition}

The proof of Proposition \ref{prop:training object} can be found in Section C of the supplementary material. The objective \eqref{equa:population loss} is of key practical importance, as its reliance on the sample pair $(\mathbf{X}, \mathbf{Y})$ enables learning a joint model for $(\mathbf{x}, \mathbf{y})$. Such a joint model makes our method fundamentally different from the unconditional sampling methods. As stated in Section A of the supplementary material, when we use the unconditional sampling methods to draw samples from $p_{{x} \mkern2mu|\mkern2mu {y}}(\mathbf{x} \mkern2mu|\mkern2mu \mathbf{y})$ for a given $\mathbf{y}$, data points with $\mathbf{Y} \neq \mathbf{y}$ will be discarded, leading to severe data inefficiency. The joint model, in contrast, will exploit the whole dataset during the training process, thus is much more efficient. Also, it can avoid training an infinite number of unconditional ODE-based models if $\mathbf{Y}$ is of continuous type. Based on \eqref{equa:population loss}, we can then design a deep learning algorithm to estimate $\mathbf{v}_{\rm{F}}$ nonparametrically, where we are working with a set of independent and identically distributed (i.i.d.) samples $\{(\mathbf{X}_{i}, \mathbf{Y}_i)\}_{i=1}^n \sim p_{x, y}(\mathbf{x}, \mathbf{y})$ and i.i.d. samples $\{(t_j, \mathbf{W}_{j})\}_{j=1}^m$ with $t_j \sim \mathcal{U}(0, T)$ and $\mathbf{W}_j \sim \mathcal{N}(\mathbf{0}, \mathbf{I}_{d_x})$ independently. To ensure effective learning, the employed deep network class should be expressive enough to approximate the true velocity field.
More specifically, we can choose ReLU-based feed forward neural networks (FNN) defined as Definition \ref{def:FNN}.
\begin{definition} \label{def:FNN}
    Denote by $ \mathrm{FNN}(L, M, J, K, \kappa, \gamma_1, \gamma_2, \gamma_3)$ the set of ReLU neural networks $\mathbf{v}_{{\boldsymbol{\theta}}}: \mathbb{R}^{d_x} \times [0,B]^{d_y} \times [0,T] \rightarrow \mathbb{R}^{d_x}$ with parameter ${{\boldsymbol{\theta}}}$, depth $L$, width $M$ and size $J$ such that

    {\rm (a)}
    $\sup _{\mathbf{x}, \mathbf{y}, t}|\mathbf{v}_{\boldsymbol{\theta}}(\mathbf{x}, \mathbf{y}, t)|_2 \leq K$ and $|\boldsymbol{\theta}|_{\infty} \leq \kappa$,

    {\rm (b)} $|\mathbf{v}_{\boldsymbol{\theta}}(\mathbf{x}_1, \mathbf{y}, t)-\mathbf{v}_{\boldsymbol{\theta}}(\mathbf{x}_2, \mathbf{y}, t)|_{\infty} \leq \gamma_1|\mathbf{x}_1-\mathbf{x}_2|_2$ for any  $t \in[0, T]$ and  $\mathbf{y} \in [0,B]^{d_{y}}$,

    {\rm (c)} $|\mathbf{v}_{\boldsymbol{\theta}}(\mathbf{x}, \mathbf{y}_1, t)-\mathbf{v}_{\boldsymbol{\theta}}(\mathbf{x}, \mathbf{y}_2, t)|_{\infty} \leq \gamma_2|\mathbf{y}_1-\mathbf{y}_2|_2$ for any $t \in[0, T]$  and $\mathbf{x} \in \mathbb{R}^{d_x}$,

    {\rm (d)} $|\mathbf{v}_{\boldsymbol{\theta}}(\mathbf{x}, \mathbf{y}, t_1)-\mathbf{v}_{\boldsymbol{\theta}}(\mathbf{x}, \mathbf{y}, t_2)|_{\infty} \leq \gamma_3|t_1-t_2|$  for any  $\mathbf{x} \in \mathbb{R}^{d_x}$  and  $\mathbf{y} \in [0, B]^{d_{y}}$.
\end{definition}

Here the depth $L$ refers to the number of hidden layers. Let $w_0=d_x+d_y+1$, $w_1,\ldots,w_L$, and $w_{L+1}=d_x$ denote the input width, hidden-layer widths, and output width, respectively. The width $M=\max \{w_1, \ldots, w_L\}$ is the maximum width of the hidden layers. The size $J$ bounds the number of nonzero weights and biases, $|\boldsymbol\theta|_0\leq J$.

\begin{remark}
Two reasons motivate the explicit Lipschitz constraints in our hypothesis class. First, this restriction does not sacrifice approximation power. Approximation theory {\rm\cite[e.g.,][]{chen2023score}} shows that the optimal $L^2$-approximating function for a Lipschitz target, found within a general class of ReLU networks (without imposing the Lipschitz constraints explicitly), is itself Lipschitz. Our constrained subclass is therefore guaranteed to contain this optimal function. Second, these constraints are necessary for our subsequent end-to-end error analysis. The Lipschitz constants of the learned velocity field are critically required to bound the perturbation and discretization errors of our proposed ODE sampler. See Propositions {\rm\ref{prop:estimation error bound}} and {\rm\ref{prop:discretization error bound}} in Section {\rm\ref{sec:Error Analysis}} for details. Note that the requirement for network class regularity is a key feature in current analytical frameworks for diffusion and flow-based
models aiming to establish end-to-end convergence guarantee {\rm\citep{chen2023score, fukumizuflow}}.
\end{remark}

Given the empirical loss function
\begin{equation} \label{equa:empirical loss}
\begin{aligned}
\widehat{\mathcal{L}}(\mathbf{v}) 
& =\frac{1}{mn} \sum_{i=1}^n\sum_{j=1}^m\bigg|\mathbf{X}_{i}-\frac{t_j}{\sqrt{1 - t_j^2}}\mathbf{W}_{j}-\mathbf{v}(t_j \mathbf{X}_{i}+\sqrt{1-t_j^2} \mathbf{W}_{j}, \mathbf{Y}_i, t_j)\bigg|_2^2\,,
\end{aligned}
\end{equation}
 we consider estimating the 
conditional F\"{o}llmer velocity field $\mathbf{v}_{\mathrm{F}}$ as follows:
\begin{equation} \label{equa:ERM}
    \hat{\mathbf{v}} \in \arg\min_{\mathbf{v}_{\boldsymbol{\theta}} \in \mathrm{FNN}(L, M, J, K, \kappa, \gamma_1, \gamma_2, \gamma_3)} \widehat{\mathcal{L}}(\mathbf{v}_{\boldsymbol{\theta}})\,.
\end{equation}
We can employ the stochastic gradient descent algorithm to optimize the empirical loss, which is widely used for optimizing neural networks and has shown significant effectiveness \citep{allen2019convergence, du2019gradient}. When $\mathbf{v}_\mathrm{F}$ is known, Theorem \ref{thm:well posedness} in Section \ref{sec:CFF} indicates that, to sample data from the conditional density $p_{x \mkern 2mu|\mkern2mu y}(\mathbf{x} \mkern 2mu|\mkern2mu \mathbf{y})$, we only need to generate $\mathbf{z}$ from $\mathcal{N}(\mathbf{0}, \mathbf{I}_{d_x})$ and run the ODE dynamics of the conditional F\"{o}llmer flow \eqref{equa:CCF ODE} with the time $t$ near $1$. In practice, based on $\hat{\mathbf{v}}$, the estimate of $\mathbf{v}_\mathrm{F}$ given in \eqref{equa:ERM}, we can obtain the pseudo data via the following Algorithm \ref{algorithm}.

\begin{algorithm}[htbp] \label{algorithm}
\SetAlgoLined
\caption{Sampling pseudo data from $p_{x \mkern2mu|\mkern2mu y}(\mathbf{x} \mkern2mu|\mkern2mu \mathbf{y})$}
\vspace{0.2\baselineskip}
\KwIn{$\tilde{\mathbf{z}}_0 \sim \mathcal{N}(\mathbf{0}, \mathbf{I}_{d_x})$, the estimated velocity field $\hat{\mathbf{v}}$, time steps $N$ and stopping time $T<1$}
\KwOut{$\tilde{\mathbf{Z}}^{\mathbf{y}}_{T}$}
$t_0 = 0$\;
\For{$k=0,1, \ldots, N-1$}{
    Compute $t_{k+1}=t_k+N^{-1}T$\;
    Compute the velocity $\hat{\mathbf{v}}(\tilde{\mathbf{z}}_{t_k}, \mathbf{y}, t_k)$\;
    Update $\tilde{\mathbf{z}}_{t_{k+1}} = \tilde{\mathbf{z}}_{t_{k}} + N^{-1}T\hat{\mathbf{v}}(\tilde{\mathbf{z}}_{t_k}, \mathbf{y}, t_k)$\;
}
$\tilde{\mathbf{Z}}^{\mathbf{y}}_{T} = \tilde{\mathbf{z}}_{t_N}$\;
\end{algorithm}

 Specifically, Algorithm \ref{algorithm} is the Euler's method to discretize the continuous ODE flow with velocity field $\hat{\mathbf{v}}(\mathbf{x}, \mathbf{y}, t)$ on $[0, T]$, where the step size is set as $N^{-1}T$.  With properly selected $T$, Theorem \ref{thm:main thm} in Section \ref{sec:Error Analysis} shows that the Wasserstein-2 distance between $p_{x \mkern 2mu|\mkern2mu y}(\mathbf{x} \mkern 2mu|\mkern2mu \mathbf{y})$ and the density of $\tilde{\mathbf{Z}}^{\mathbf{y}}_{T}$  converges to zero in probability as $n \rightarrow \infty$. Notice that estimating the conditional density $p_{x \mkern 2mu|\mkern2mu y}(\mathbf{x} \mkern 2mu|\mkern2mu \mathbf{y})$ is highly challenging in practice when $d_x$ and $d_y$ are large. Our theoretical analysis shows that, in order to draw samples from $p_{x \mkern 2mu|\mkern2mu y}(\mathbf{x} \mkern 2mu|\mkern2mu \mathbf{y})$, we can just implement Algorithm \ref{algorithm} without estimating $p_{x \mkern 2mu|\mkern2mu y}(\mathbf{x} \mkern 2mu|\mkern2mu \mathbf{y})$. This is the first main advantage of our method.

Furthermore, since the trajectory of ODE flows is deterministic, Algorithm \ref{algorithm} also provides a deterministic sampling procedure. That is, a given starting point $\tilde{\mathbf{z}}_0$ will lead to a unique ending point $\tilde{\mathbf{z}}_{T}$. In fact, when $\hat{\mathbf{v}}$ sufficiently approximates $\mathbf{v}_{\rm F}$ and $N$ is sufficiently large, we have $\tilde{\mathbf{z}}_{T} \approx \mathbf{F}_{T}(\tilde{\mathbf{z}}_0, \mathbf{y})$, where $\mathbf{F}_{t}(\cdot, \mathbf{y})$ is the conditional F\"{o}llmer flow map defined in Definition \ref{def:CFFM}.
Hence, after repeatedly applying Algorithm \ref{algorithm} to obtain the corresponding pseudo data $\tilde{\mathbf{z}}_T^1,\ldots,\tilde{\mathbf{z}}_T^{\scriptscriptstyle \tilde{N}}$ from a set of Gaussian noises $\tilde{\mathbf{z}}_0^1,\ldots,\tilde{\mathbf{z}}_0^{\scriptscriptstyle \tilde{N}} \sim_{\text{i.i.d.}} \mathcal{N}(\mathbf{0},\mathbf{I}_{d_x})$, we can use an additional deep neural network ${\mathbf{G}}_{\boldsymbol{\theta}}(\cdot)$ to directly fit the mapping relationship between noise-sample pairs $\{(\tilde{\mathbf{z}}_0^i, \tilde{\mathbf{z}}_T^i)\}_{i=1}^{\tilde{N}}$. This can be seen as effectively learning the flow map $\mathbf{F}_{T}(\cdot, \mathbf{y})$. It is important to note that, the sampling time of Algorithm \ref{algorithm} on a time grid with size $N$ is of order $\mathcal{O}(N)$. However, once ${\mathbf{G}}_{\boldsymbol{\theta}}(\cdot)$ is trained, the time required to generate a new sample is significantly reduced to $\mathcal{O}(1)$ since only one single network evaluation is needed.
It is worth emphasizing that such an end-to-end generator cannot be derived from the SDE-based methods due to their  inherent stochastic nature for each particle.

\section{Theoretical Analysis} \label{sec:Error Analysis}
Our main interest lies in establishing the validity of Algorithm \ref{algorithm} in generating data from $p_{x \mkern 2mu|\mkern2mu y}(\mathbf{x} \mkern 2mu|\mkern2mu \mathbf{y})$. More specifically, we would like to investigate the convergence rate of the sampling error in Algorithm \ref{algorithm}. For this, besides Assumptions \ref{assump:label} and \ref{assump:bounded support} introduced in Section \ref{sec:CFF}, we need the following regularity assumption.

\begin{assumption}[Lipschitz velocity field with respect to the condition]
\label{assump:Lip in y}
The conditional F\"{o}llmer velocity field $\mathbf{v}_{\rm{F}} (\mathbf{x}, \mathbf{y}, t)$ is locally Lipschitz continuous with respect to the condition $\mathbf{y}$. Specifically, for any $R\geq1$ and $T \in [0,1)$, the Lipschitz constant with respect to $\mathbf{y}$ is bounded by 
$
{\omega}_{R, T} = C_y(d_x, d_y)  R^{\alpha} (1-T)^{-\beta}
$ on $[-R, R]^{d_x} \times [0,B]^{d_y} \times [0,T]$, i.e., $$|\mathbf{v}_{\rm F}(\mathbf{x}, \mathbf{y}_1, t)-\mathbf{v}_{\rm F}(\mathbf{x}, \mathbf{y}_2, t)|_{\infty} \leq \omega_{R, T} |\mathbf{y}_1-\mathbf{y}_2|_2$$ for any $\mathbf{y}_1, \mathbf{y}_2 \in [0,B]^{d_y}$, $t \in[0, T]$  and $\mathbf{x} \in [-R, R]^{d_x}$.
Here, $\alpha, \beta \geq 0$ are constants, and $C_y(d_x, d_y) > 0$ is a constant only depending on $d_x$ and $d_y$.
\end{assumption}

    Under Assumptions \ref{assump:label} and \ref{assump:bounded support}, we will obtain some satisfactory properties of the conditional F\"{o}llmer velocity field $\mathbf{v}_{\rm F}$, such as the Lipschitz properties of $\mathbf{v}_{\rm F}(\mathbf{x}, \mathbf{y}, t)$ with respect to $\mathbf{x}$ and $t$. See Proposition P1 in the supplementary material for details. Furthermore, Assumption \ref{assump:Lip in y} is technically required to demonstrate the effectiveness of $\hat{\mathbf{v}}$ in \eqref{equa:ERM} as a neural network to estimate $\mathbf{v}_{\rm{F}}$. 
     As shown in Section E of the supplementary material, under some regularity conditions on $p_{x, y}(\mathbf{x}, \mathbf{y})$, Assumption \ref{assump:Lip in y} holds automatically with $\alpha=0$, $\beta=1$ and $C_y(d_x, d_y) = U\sqrt{d_xd_y}$ for some universal constant $U>0$. 
    Proposition \ref{prop:generalization} presents the convergence rate of $\hat{\mathbf{v}}$, whose proof is given in Section F of the supplementary material.
\begin{proposition}
   \label{prop:generalization}
    Let Assumptions {\rm \ref{assump:label}--\rm \ref{assump:Lip in y}} hold. Suppose we have i.i.d. samples $\{(\mathbf{X}_{i}, \mathbf{Y}_i)\}_{i=1}^n$ $\sim p_{x, y}(\mathbf{x}, \mathbf{y})$ and  i.i.d. samples $\{(t_j, \mathbf{W}_{j})\}_{j=1}^m$ with $t_j \sim \mathcal{U}(0, T)$ and $\mathbf{W}_j \sim \mathcal{N}(\mathbf{0}, \mathbf{I}_{d_x})$ independently. Choose the network class $\mathrm{FNN}=\mathrm{FNN}(L, M, J, K, \kappa, \gamma_1, \gamma_2, \gamma_3)$ with
\begin{align*}
&M\sim\frac{{C_0^{d_x+(\alpha+1)d_y+1}}\{BC_y(d_x,d_y)\}^{d_y}\log^{(d_x+\alpha d_y+1)/2}\{d_x\varepsilon^{-1}(1-T)^{-1}\}}{(1-T)^{2d_x+\beta d_y+3}\varepsilon^{d_x+d_y+1}d_x^{-3(d_x+1)/2}d_y^{\,-d_y/2}}\,,\\
&J\sim\frac{{C_0^{d_x+(\alpha+1)d_y+1}}\{BC_y(d_x,d_y)\}^{d_y}\log^{(d_x+\alpha d_y+1)/2}\{d_x\varepsilon^{-1}(1-T)^{-1}\}}{(1-T)^{2d_x+\beta d_y+3}\varepsilon^{d_x+d_y+1}{(d_x+d_y+1)^{-1}}d_x^{-3(d_x+1)/2}d_y^{\,-d_y/2}}\,,\\
&\kappa\sim {B^{-1}\vee T^{-1}}\vee\frac{\{{Bd_y^{1/2}C_y(d_x,d_y)}\vee d_x^{3/2}\}\log^{(\alpha\vee1)/2}\{d_x\varepsilon^{-1}(1-T)^{-1}\}}{{\varepsilon}(1-T)^{\beta\vee3}}\,,\\
&{L\sim d_x+d_y+1}\,,\quad
K\sim\frac{\log^{1/2}\{d_x\varepsilon^{-1}(1-T)^{-1}\}}{(1-T)d_x^{-1/2}}\,,\quad
\gamma_1=\frac{{d_x^{3/2}}}{(1-T)^2}\,,\\
&\gamma_2\sim\frac{C_y(d_x,d_y)\log^{\alpha/2}\{d_x\varepsilon^{-1}(1-T)^{-1}\}}{(1-T)^\beta{d_y^{-1/2}}}\,,\quad
\gamma_3\sim\frac{\log^{1/2}\{d_x\varepsilon^{-1}(1-T)^{-1}\}}{(1-T)^3d_x^{-3/2}}\,.
\end{align*}
Here $C_0\geq1$ is a universal constant. Choose $\varepsilon=(1-T)^{-(2d_x+\beta d_y+7)/(d_x+d_y+5)}n^{-1/({d_x}+{d_{y}}+5)}$ with  $1-T \gg n^{-1/(2d_x+\beta d_y + 7)}$. Denote by $g_t(\cdot, \cdot)$ the joint density of $(t \mathbf{X}+\sqrt{1-t^2} \mathbf{W}, \mathbf{Y})$. Let $m=n$. For any fixed $(d_x, d_y)$, we have
$$
\frac{1}{T} \int_0^{{T}}\|\hat{\mathbf{v}}(\mathbf{x}, \mathbf{y}, t)-\mathbf{v}_{\rm{F}} (\mathbf{x}, \mathbf{y}, t)\|_{L^2(g_t)}^2 \, \mathrm{d} t =\widetilde{\mathcal{O}}\left\{\frac{(1-T)^{-(4d_x+2 \beta d_y + 14)/(d_x+d_y+5)}}{n^{2/(d_x+d_y+5)}}\right\}
$$
with probability at least $1-n^{-2}$, where $\widetilde{\mathcal{O}}(\cdot)$ omits the polynomial term of $\log n$.
\end{proposition}
  
Now, we begin to analyze the convergence rate of the sampling error in Algorithm \ref{algorithm}. Notice that the sampling error comes from the following three aspects:
\begin{itemize}
    \item (Approximation Error) As stated in Theorem \ref{thm:well posedness}, $f_{t}(\mathbf{x} \mkern 2mu | \mkern 2mu \mathbf{y})$ can arbitrarily approximate $p_{x \mkern 2mu|\mkern2mu y}(\mathbf{x} \mkern 2mu|\mkern2mu \mathbf{y})$ when $t \rightarrow 1$. In practice, we would select an early stopping time $T<1$, which introduces an error between $f_{_T}(\mathbf{x} \mkern 2mu | \mkern 2mu \mathbf{y})$ and $p_{x \mkern 2mu|\mkern2mu y}(\mathbf{x} \mkern 2mu|\mkern2mu \mathbf{y})$.

    \item (Perturbation Error) Recall that $f_{_T}(\mathbf{x} \mkern 2mu | \mkern 2mu \mathbf{y})$ is the density of the conditional F\"{o}llmer flow $\mathbf{Z}_t^{\mathbf{y}}=\mathbf{Z}(t, \mathbf{y})$ at $t=T$ in \eqref{equa:CCF ODE} with certain unknown velocity field $\mathbf{v}_{\rm F}$. With the nonparametric estimation $\hat{\mathbf{v}}$ in \eqref{equa:ERM} for $\mathbf{v}_{\rm F}$, we can define a new continuous ODE flow
    \begin{align} \label{equa:nnflow}
\mathrm{d} \hat{\mathbf{Z}}^{\mathbf{y}}_t=\hat{\mathbf{v}}  (\hat{\mathbf{Z}}^{\mathbf{y}}_t, \mathbf{y}, t) \mkern2mu \mathrm{d} t \, , \quad \hat{\mathbf{Z}}^{\mathbf{y}}_0 \sim \mathcal{N}(\mathbf{0}, \mathbf{I}_{d_x}) \, , \quad 0 \leq t \leq T \, ,
\end{align}
    which provides an approximation to the original continuous ODE flow \eqref{equa:CCF ODE}. Denote by $\hat{p}_{t}(\mathbf{x}; \mathbf{y})$ the density of  $\hat{\mathbf{Z}}_t^{\mathbf{y}}$. Hence, a perturbation error arises between $f_{_T}(\mathbf{x} \mkern 2mu | \mkern 2mu \mathbf{y})$ and $\hat{p}_{_T}(\mathbf{x}; \mathbf{y})$.

    \item (Discretization Error) The solution of ODE flow  \eqref{equa:nnflow} does not admit a closed form. Let $t_0 = 0$ and $ t_k = t_{k-1} + N^{-1}T$ for $k=1,\ldots,N$. The pseudo data $\tilde{\mathbf{Z}}^{\mathbf{y}}_{T}$ obtained via Algorithm \ref{algorithm} satisfies the following Euler discretization of \eqref{equa:nnflow}:
\begin{align} \label{equa:eulerflow}
\mathrm{d} \tilde{\mathbf{Z}}^{\mathbf{y}}_t=\hat{\mathbf{v}}(\tilde{\mathbf{Z}}^{\mathbf{y}}_{t_k}, \mathbf{y}, t_k) \, \mathrm{d} t \, , \quad \tilde{\mathbf{Z}}_0^{\mathbf{y}} \sim \mathcal{N}(\mathbf{0}, \mathbf{I}_{d_x}) \, , \quad t_k \leq t \leq t_{k+1} \, , \quad 0 \leq k < N \, .
\end{align}
Denote by $\tilde{p}_{t}(\mathbf{x} ; \mathbf{y})$ the density of $\tilde{\mathbf{Z}}^{\mathbf{y}}_t$. The discretization error stems from the difference between $\hat{p}_{_T}(\mathbf{x}; \mathbf{y})$ and $\tilde{p}_{_T}(\mathbf{x}; \mathbf{y})$.
\end{itemize}
We use $W_2  (\mkern0.5mu \tilde{p}_{_{T}}(\mathbf{x} ; \mathbf{y}), \, p_{x \mkern 2mu|\mkern2mu y}(\mathbf{x} \mkern 2mu|\mkern2mu \mathbf{y}))$, the Wasserstein-2 distance between $\tilde{p}_{_{T}}(\mathbf{x} ; \mathbf{y})$ and the target $p_{x \mkern 2mu|\mkern2mu y}(\mathbf{x} \mkern 2mu|\mkern2mu \mathbf{y})$, to characterize the sampling error of Algorithm \ref{algorithm}. By the triangle inequality, it can be decomposed into three terms:
\begin{align*}
    W_2  \big(\mkern0.5mu \tilde{p}_{_{T}}(\mathbf{x} ; \mathbf{y}), \, p_{x \mkern 2mu|\mkern2mu y}(\mathbf{x} \mkern 2mu|\mkern2mu \mathbf{y})\big) \leq & \, \underbrace{W_2  \big(\mkern0.5mu f_{_T}(\mathbf{x} \mkern 2mu|\mkern2mu \mathbf{y}), p_{x \mkern 2mu|\mkern2mu y}(\mathbf{x} \mkern 2mu|\mkern2mu \mathbf{y})\big)}_{\text{Approximation Error}} + \underbrace{W_2 \big(\mkern0.5mu \hat{p}_{_{T}}(\mathbf{x} ; \mathbf{y}), f_{_T}(\mathbf{x} \mkern 2mu|\mkern2mu \mathbf{y})\big)}_{\text{Perturbation Error}} \\
    & + \underbrace{W_2 \big(\mkern0.5mu \tilde{p}_{_{T}}(\mathbf{x} ; \mathbf{y}), \hat{p}_{_{T}}(\mathbf{x}; \mathbf{y})\big)}_{\text{Discretization Error}} \, .
\end{align*}

Theorem \ref{thm:well posedness} presents the convergence rate of Approximation Error, while Propositions \ref{prop:estimation error bound} and \ref{prop:discretization error bound} state the convergence rates of Perturbation Error and Discretization Error, respectively. The proofs of Propositions \ref{prop:estimation error bound} and \ref{prop:discretization error bound} are given in Sections G and H of the supplementary material, respectively.

\begin{proposition} \label{prop:estimation error bound}
    Let Assumptions {\rm \ref{assump:label}--\rm \ref{assump:Lip in y}} hold. Choose $\hat{\mathbf{v}}(\mathbf{x}, \mathbf{y}, t)$ as in Proposition {\rm \ref{prop:generalization}}. For any fixed $(d_x, d_y)$, if $1-T \gg n^{-1/(2d_x+\beta d_y+7)}$, it holds that
\begin{align*}
\int W^2_2 \big( \mkern0.5mu f_{_T}(\mathbf{x} \mkern 2mu|\mkern2mu \mathbf{y}), \mkern2mu \hat{p}_{_T}(\mathbf{x} ; \mathbf{y})\big) p_y(\mathbf{y}) \, \mathrm{d}\mathbf{y} = \widetilde{\mathcal{O}}\left\{\frac{e^{2d_x^2(1-T)^{-2}}}{(1-T)^{(4d_x +2\beta d_y+14)/(d_x+d_y+5)}n^{2/(d_x+d_y+5)}}\right\}
\end{align*}
with probability at least $1-n^{-2}$, where $\widetilde{\mathcal{O}}(\cdot)$ omits the polynomial term of $\log n$.
\end{proposition}

\begin{proposition} \label{prop:discretization error bound}
Let Assumptions {\rm \ref{assump:label}--\rm \ref{assump:Lip in y}} hold. Choose $\hat{\mathbf{v}}(\mathbf{x}, \mathbf{y}, t)$ as in Proposition {\rm \ref{prop:generalization}}. %
For any fixed $(d_x, d_y)$, if $1-T \gg n^{-1/(2d_x+\beta d_y+7)}$, it then holds that
$$
\sup_{\mathbf{y} \in [0, B]^{d_y}} W^2_2\big( \mkern0.5mu \hat{p}_{_T}(\mathbf{x} ; \mathbf{y}), \mkern2mu \tilde{p}_{_T}(\mathbf{x} ; \mathbf{y})\big)=\widetilde{\mathcal{O}}\left\{ \frac{e^{2d_x^2(1-T)^{-2}}}{(1-T)^{6} N^{2}} \right\} \, ,
$$
where $\widetilde{\mathcal{O}}(\cdot)$ omits the polynomial term of $\log n$.
\end{proposition}

For any fixed $(d_x, d_y)$, by Theorem \ref{thm:well posedness}, Propositions \ref{prop:estimation error bound} and \ref{prop:discretization error bound}, it holds with probability $1-n^{-2}$ that
\begin{align*}
&\int W^2_2  \big(\mkern0.5mu \tilde{p}_{_{T}}(\mathbf{x} ; \mathbf{y}), \, p_{x \mkern 2mu|\mkern2mu y}(\mathbf{x} \mkern 2mu|\mkern2mu \mathbf{y})\big) p_y(\mathbf{y}) \, \mathrm{d}\mathbf{y} \\
&~~~~\leq \underbrace{\mathcal{O}(1-T)}_{\text{Error I}}+\underbrace{\widetilde{\mathcal{O}}\left\{\frac{e^{2d_x^2(1-T)^{-2}}}{(1-T)^{(4d_x +2\beta d_y+14)/(d_x+d_y+5)}n^{2/(d_x+d_y+5)}}\right\}}_{\text{Error II}} +\underbrace{\widetilde{\mathcal{O}}\left\{\frac{e^{2d_x^2(1-T)^{-2}}}{(1-T)^{6} N^{2}}\right\}}_{\text{Error III}}
\end{align*}
provided that $1-T \gg n^{-1/(2d_x +\beta d_y+7)}$, where $\widetilde{\mathcal{O}}(\cdot)$ omits the polynomial term of $\log n$. Set
\begin{align} \label{eq:T,N}
T=1-C(\log n)^{-1/2} \, , \quad N \sim n^{\xi} \, ,
\end{align}
with $C>d_x(d_x+d_y+5)^{1/2}$ and $\xi \ge 1 / (d_x+d_y+5)$. Then, for any given $(d_x, d_y)$, we have $\text{Error II} + \text{Error III}  = o_{\rm p}\{(\log n)^{-1/2}\}$, which is negligible in comparison to $\text{Error I}=\mathcal{O}\{(\log n)^{-1/2}\}$. This indicates that the sampling error of Algorithm \ref{algorithm} is mainly dominated by the error from using $f_{_T}(\mathbf{x} \mkern 2mu|\mkern2mu \mathbf{y})$ to approximate $p_{x \mkern 2mu|\mkern2mu y}(\mathbf{x} \mkern 2mu|\mkern2mu \mathbf{y})$.

  Recall that $\tilde{p}_{_T}(\mathbf{x}; \mathbf{y})$ is the density of the pseudo data $\tilde{\mathbf{Z}}^{\mathbf{y}}_T$ produced by Algorithm~\ref{algorithm}. Theorem \ref{thm:main thm} summarizes our above discussion, which establishes the validity of Algorithm \ref{algorithm} in sampling data from $p_{x \mkern 2mu|\mkern2mu y}(\mathbf{x} \mkern 2mu|\mkern2mu \mathbf{y})$, and also highlights the effectiveness of the conditional F\"{o}llmer flow in sampling from the target conditional density.

\begin{theorem} \label{thm:main thm}
    Let Assumptions {\rm \ref{assump:label}--\ref{assump:Lip in y}} hold. Suppose we have i.i.d. samples $\{(\mathbf{X}_{i}, \mathbf{Y}_i)\}_{i=1}^n$ $\sim p_{x, y}(\mathbf{x}, \mathbf{y})$. We choose the hypothesis class $\mathrm{FNN}=\mathrm{FNN}(L, M, J, K, \kappa, \gamma_1, \gamma_2, \gamma_3)$ as specified in Proposition {\rm \ref{prop:generalization}} and then use $\hat{\mathbf{v}}(\mathbf{x}, \mathbf{y}, t)$ in {\rm (\ref{equa:ERM})} to estimate the true velocity field $\mathbf{v}_{\rm F}(\mathbf{x}, \mathbf{y}, t)$. For any fixed $(d_x, d_y)$, when implementing Algorithm {\rm \ref{algorithm}} for sampling pseudo data with $(T, N)$ satisfying \eqref{eq:T,N}, we have
$$
\int W^2_2  \big(\mkern0.5mu \tilde{p}_{_{T}}(\mathbf{x} ; \mathbf{y}), \, p_{x \mkern 2mu|\mkern2mu y}(\mathbf{x} \mkern 2mu|\mkern2mu \mathbf{y})\big) p_y(\mathbf{y}) \, \mathrm{d} \mathbf{y} \rightarrow 0
$$
in probability as $n \rightarrow \infty$.
\end{theorem}

 \section{Numerical Studies}
\label{sec:numerical}
In this section, we carry out numerical experiments to assess the performance of our proposed method.
We conduct our method on both synthetic and real datasets.
In simulation studies, we compare our proposed method (F\"{o}llmer) with the conditional versions of four popular deep generative methods: Wasserstein Generative Adversarial Networks (WGAN) \citep{arjovsky2017wasserstein}, Variational Autoencoders (VAE) \citep{kingma2013auto}, Variance Exploding SDE (VE-SDE) \citep{Song2021}, and Stochastic Interpolants with trigonometric coefficients (Trigonometric) \citep{albergo2022building}.
 Specifically, we follow \cite{mirza2014conditional}, \cite{kingma2014semi} and \cite{ho2021classifierfree}, respectively, to implement the conditional versions of WGAN,  VAE and VE-SDE. Since Trigonometric is an ODE-based method in essence, we can extend it to its conditional version using the same technique as our proposed method.
We also compare with two popular conditional density estimation methods: Nearest Neighbor Kernel Conditional Density Estimation (NNKCDE) \citep{dalmasso2020conditional} and Flexible Conditional Density Estimator (FlexCode) \citep{izbicki2017converting}. The details of NNKCDE and FlexCode are given in Section J of the supplementary material.  The experiments involving deep learning are computed on 3 nodes of an NVIDIA 4xV100 cluster.
Our code is available at the GitHub repository: \url{https://github.com/burning489/ConditionalFollmerFlow}.
 Our simulation studies demonstrate that though our proposed method does not directly provide the conditional density estimation, the samples generated by the method can be effectively utilized to estimate the conditional density and related statistical quantities, such as the conditional mean and the conditional standard deviation.
See Sections \ref{subsec:simulation1} and \ref{subsec:simulation2} for details.
 We use two real data analyses in Sections \ref{subsec:wine} and \ref{subsec:mnist} to demonstrate the advantages of our proposed method beyond other methods.  Throughout the numerical studies, the stopping time $T$ is set to $0.999$ for our method.  In Section L of the supplementary material, we also present extra results for $T=0.9995$ and  $0.9999$, which indicate that our proposed method is robust to the choice of $T$ as long as it is close to $1$. In practice, we suggest selecting $T=0.999$.
For a fair comparison, we adopt the  similar convolution-based neural network architecture for all the deep generative methods. Additionally, we use the same number of discrete time steps when generating samples with both the SDE-based method (VE-SDE) and the ODE-based methods (our proposed method and Trigonometric). For NNKCDE, we choose the bandwidth $h$ and the number of nearest neighbors $k$ by grid search over $h \in \{0.01, 0.02, 0.03, \ldots, 0.10\}$ and $k \in \{3, 6, 9\}$.
For FlexCode, we use the default Fourier basis, and set the max number of bases $l$ as 31, and carry out the regression based on $k$-nearest neighbors with choosing the number of nearest neighbors $k$ by grid search over $k \in \{3, 6, 9\}$. See Section J of the supplementary material for details of the involved tuning parameters for NNKCDE and FlexCode. During the grid search, we measure the discrepancy between the conditional density and its estimate following \cite{dalmasso2020conditional}.
All simulations are implemented in Python.
The Python codes for implementing NNKCDE and FlexCode are provided by the authors of \cite{dalmasso2020conditional} and  \cite{izbicki2017converting} respectively.

 \subsection{Simulation Study I}
\label{subsec:simulation1}
\begin{landscape}
\begin{figure}
  \centering
  \includegraphics[width=\linewidth]{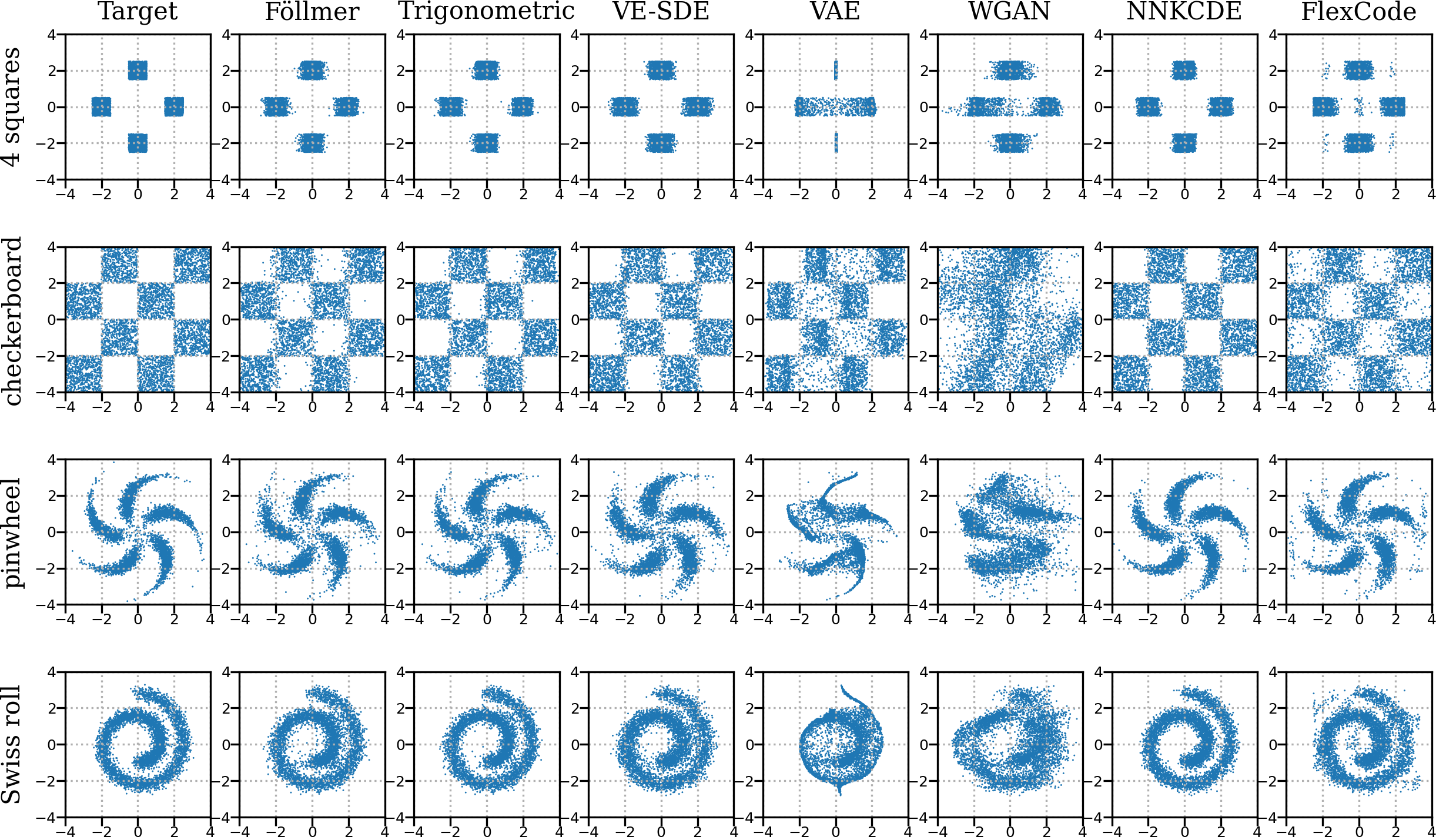}
  \caption{\small Scatter plots of the pairwise samples generated by different methods.}
  \label{fig:simulation1}
\end{figure}
\end{landscape}

We consider several two-dimensional distributions with shapes of 4 squares, checkerboard, pinwheel and Swiss roll, respectively.
 We display scatter plots of $5000$ samples drawn from these target distributions in the first column of Figure \ref{fig:simulation1}.
 For these distributions, we take the $x$-axis variable as $X$ and the $y$-axis variable as $Y$.
We target on generating samples from conditional density $p_{x\mkern 2mu|\mkern2muy}(x \mkern 2mu|\mkern2mu y)$.
 To do this, we use $n=50000$ samples from target distributions for training, and let $\mathcal{Y}$ be a set including 5000 additional samples of $Y$ generated from the associated marginal distribution of the model.

 For all the deep generative methods, we use the Adam algorithm \citep{kingma2015adam} with a learning rate of $0.001$ to train the models. For our proposed method, we first train the velocity estimator on the training set, and then for each given $y_i \in \mathcal{Y}$ we generate an $\hat{X}_i$ by Algorithm \ref{algorithm} based on the velocity estimator.
 We display the scatter plots of the generated 5000 pairs of $(\hat{X}_i, y_i)$ in the second column of Figure \ref{fig:simulation1}.
 Same as what we did in our proposed method, we display 5000 pairs of $(\hat{X}_i, y_i)$ generated by other generative methods in the third to sixth columns of Figure \ref{fig:simulation1}.
 WGAN is prone to mode collapse, a phenomenon illustrated by the checkerboard dataset in which each conditional distribution has two distinct modes. 
 As a result, the generated outputs of WGAN tend to average over these modes rather than capturing them distinctly. 
 VAE, on the other hand, often leads to less diverse generated outputs, as illustrated by the pinwheel and Swiss roll datasets. 
 For NNKCDE and FlexCode, we first fit the conditional density on the training set, and then for each given $y_i \in \mathcal{Y}$ we generate an $\hat{X}_i$ from the estimated conditional density. 
The seventh and eighth columns of Figure \ref{fig:simulation1} display the scatter plots of 5000 pairs of $(\hat{X}_i, y_i)$ generated by NNKCDE and FlexCode.
 Figure \ref{fig:simulation1} demonstrates that our proposed method, Trigonometric, VE-SDE and NNKCDE generate samples close to targets, while FlexCode is less stable for cases such as checkerboard and Swiss roll. 

 Different from NNKCDE and FlexCode, our proposed method, Trigonometric, VE-SDE, VAE and WGAN directly generate samples from  $p_{x\mkern 2mu|\mkern2muy}(x \mkern 2mu|\mkern2mu y)$ without estimating it.
 We are also interested in investigating the performance of our proposed method in estimating $p_{x\mkern 2mu|\mkern2muy}(x \mkern 2mu|\mkern2mu y)$.
 To do this, we draw a subset $\{y_i\}_{i=1}^{100}$ from $\mathcal{Y}$, and generate 500 samples $\{\hat{X}_i^{(j)}\}_{j=1}^{500}$ associated with $y_i$ by each generation-based method. Then, we use kernel density estimation with Gaussian kernel to empirically estimate $p_{x\mkern 2mu|\mkern2muy}(x \mkern 2mu|\mkern2mu y_i)$, denoted by $\hat{p}_{x\mkern 2mu|\mkern2muy}(x \mkern 2mu|\mkern2mu y_i)$.
 For each $y_i$, we compute the total variation distance between $p_{x\mkern 2mu|\mkern2muy}(x \mkern 2mu|\mkern2mu y_i)$ and $\hat{p}_{x\mkern 2mu|\mkern2muy}(x \mkern 2mu|\mkern2mu y_i)$.
 We also consider the total variation distances between $p_{x\mkern 2mu|\mkern2muy}(x \mkern 2mu|\mkern2mu y_i)$ and its estimates based on NNKCDE and FlexCode, respectively.
 Table \ref{tab:simulation1-tv} reports the sample average (AVE) and standard deviation (STD) of 100 obtained total variation distances based on different methods. It indicates that our proposed method achieves the lowest average total variation distance error for the estimated conditional density, except on the Swiss roll dataset, where Trigonometric performs slightly better in terms of average error, but with a higher standard deviation than ours. In Section K of the supplementary material, we also evaluate the performance of our proposed method with the training dataset size $n\in\{1000, 2000, 10000, 40000\}$. Table T1 in the supplementary material shows that our proposed method performs better as $n$ increases,  which is consistent with Theorem \ref{thm:main thm}.
 
\begin{table}[H]
\setlength{\belowcaptionskip}{5pt}
  \centering
  \footnotesize
  \caption{\small The sample average and standard deviation of 100 obtained total variation distances based on different methods in simulation study I.}
  \label{tab:simulation1-tv}
  \renewcommand{\arraystretch}{0.65}
  \begin{tabular}{ccccccccc}
\toprule
 & \multicolumn{2}{c}{4 squares} & \multicolumn{2}{c}{checkerboard} & \multicolumn{2}{c}{pinwheel} & \multicolumn{2}{c}{Swiss roll} \\
method & AVE & STD & AVE & STD & AVE & STD & AVE & STD \\
\midrule
F\"ollmer & 0.054 & 0.015 & 0.110 & 0.075 & 0.116 & 0.041 & 0.093 & 0.027 \\
Trigonometric & 0.067 & 0.010 & 0.120 & 0.079 & 0.117 & 0.043 & 0.089 & 0.035 \\
VE-SDE & 0.089 & 0.010 & 0.125 & 0.062 & 0.123 & 0.041 & 0.101 & 0.037 \\
VAE & 0.153 & 0.010 & 0.179 & 0.046 & 0.244 & 0.067 & 0.230 & 0.090 \\
WGAN & 0.123 & 0.029 & 0.352 & 0.132 & 0.316 & 0.115 & 0.217 & 0.085 \\
NNKCDE & 0.288 & 0.081 & 0.483 & 0.074 & 0.375 & 0.090 & 0.400 & 0.076 \\
FlexCode & 0.107 & 0.056 & 0.177 & 0.071 & 0.275 & 0.101 & 0.257 & 0.081 \\
\bottomrule
\end{tabular}
\end{table}

\subsection{Simulation Study II}
\label{subsec:simulation2}
In this section, we investigate the performance of our proposed method in estimating the conditional mean $\mathbb{E}(X \mkern2mu|\mkern2mu \mathbf{Y})$ and the conditional standard deviation ${\rm std}(X \mkern2mu|\mkern2mu \mathbf{Y})$ via the following three models.
 \begin{itemize}
    \item[(M1)] A nonlinear model with an additive error term:
    \[
    X = Y_1^2 + \exp (Y_2 + 0.25Y_3) + \cos(Y_4 + Y_5) + \varepsilon
    \]
    with  $\varepsilon \sim \mathcal{N}(0, 1)$ and $\mathbf{Y} = (Y_1,\ldots,Y_5)^{\rm T} \sim \mathcal{N}(\mathbf{0}, \mathbf{I}_5)$.    
    \item[(M2)]  A model with an additive error term whose variance depends on the predictors:
    \[
    X = Y_1^2 + \exp (Y_2 + 0.25Y_3) + Y_4 - Y_5 + (0.5 + 0.5Y_2^2 + 0.5Y_5^2)\varepsilon
    \]
    with $\varepsilon \sim \mathcal{N}(0, 1)$ and $\mathbf{Y} =(Y_1,\ldots,Y_5)^{\rm T} \sim \mathcal{N}(\mathbf{0}, \mathbf{I}_5)$.    
    \item[(M3)] Another model with an additive error term whose variance depends on the predictors with higher dimensionality:
    \[
    X = \frac{1}{10} \sum_{i=0}^9 \{ (Y_{5i+1} + Y_{5i+2} -1)^2 + Y_{5i+3} \sin (Y_{5i+4} + 3 Y_{5i+5}) \} + \varepsilon
    \]
    with
\[
\varepsilon\,|\,\mathbf{Y}\sim\mathcal{N}(0, \{1+0.1\sum_{i=0}^9 Y_{5i+1}Y_{5i+2}\cos(2Y_{5i+3}Y_{5i+4}+Y_{5i+5})\}^{2})
\]
and $\mathbf{Y} = (Y_1,\ldots,Y_{50})^{\rm T} \sim \mathcal{N}(\mathbf{0}, \mathbf{I}_{50})$.
\par
    \end{itemize}

For each model, we first prepare a training set of $(X,\mathbf{Y})$ with size $n=10000$ drawn from this model,
and let $\mathcal{Y}$ be a set including 5000 additional samples of $\mathbf{Y}$ generated from the associated marginal distribution of this model. 
For all the deep generative methods, we use the Adam algorithm with a learning rate of $0.001$ to train the models.
For the generation-based methods (our proposed
method, Trigonometric, VE-SDE, VAE and WGAN), we only have access to generated samples.
 For given $\mathbf{y}_i \in \mathcal{Y}$, we generate { 500} samples $\{\hat{X}_i^{(j)}\}_{j=1}^{{ 500}}$ associated with $\mathbf{y}_i$ by each generation-based method, and then estimate $\mathbb{E}(X \mkern2mu|\mkern2mu \mathbf{Y} = \mathbf{y}_i)$ and ${\rm std}(X \mkern2mu|\mkern2mu \mathbf{Y} = \mathbf{y}_i)$ by, respectively, the sample mean and sample standard deviation of $\{\hat{X}_i^{(j)}\}_{j=1}^{{ 500}}$.
 Since NNKCDE and FlexCode output the estimated conditional density $\hat{p}_{x\mkern 2mu|\mkern2muy}(x \mkern 2mu|\mkern2mu \mathbf{y}_i)$ directly, we can then estimate $\mathbb{E}(X \mkern2mu|\mkern2mu \mathbf{Y} = \mathbf{y}_i)$ and ${\rm std}(X \mkern2mu|\mkern2mu \mathbf{Y} = \mathbf{y}_i)$, respectively, by 
\begin{align*}
	&~~~~~~~~~~~~~~ \hat{\mathbb{E}} (X \mkern2mu|\mkern2mu\mathbf{Y}=\mathbf{y}_i) = \int x \hat{p}_{x\mkern 2mu|\mkern2muy}(x \mkern 2mu|\mkern2mu \mathbf{y}_i) \, \mathrm{d}x \, , \\
	& 
 \widehat{\rm std} (X \mkern2mu|\mkern2mu\mathbf{Y}=\mathbf{y}_i) = \sqrt{\int \{x - \hat{\mathbb{E}}(X \mkern 2mu|\mkern2mu \mathbf{Y} = \mathbf{y}_i) \}^2 \hat{p}_{x\mkern 2mu|\mkern2mu y}(x \mkern 2mu|\mkern2mu \mathbf{y}_i) \, \mathrm{d}x} \, .
\end{align*}
For the given estimates $\hat{\mathbb{E}}(X \mkern2mu|\mkern2mu \mathbf{Y} = \mathbf{y}_i)$ and $\widehat{\rm std} \mkern1mu (X \mkern2mu|\mkern2mu \mathbf{Y} = \mathbf{y}_i)$, we compute  
\begin{align*}
{\rm MSE}_1 =&~ \frac{1}{5000} \sum_{i=1}^{5000} \big| \mkern1mu \hat{\mathbb{E}}(X \mkern2mu|\mkern2mu \mathbf{Y} = \mathbf{y}_i) - \mathbb{E}(X \mkern2mu|\mkern2mu \mathbf{Y} = \mathbf{y}_i) \big|^2 \, , \\
{\rm MSE}_2 =&~ \frac{1}{5000} \sum_{i=1}^{5000} \big| \mkern1mu \widehat{\rm std} \mkern1mu (X \mkern2mu|\mkern2mu \mathbf{Y} = \mathbf{y}_i) - {\rm std}(X \mkern2mu|\mkern2mu \mathbf{Y} = \mathbf{y}_i) \big|^2 \, .
\end{align*}
We report ${\rm MSE}_1$ and ${\rm MSE}_2$ for different methods in Table \ref{tab:simulation2}.
Table \ref{tab:simulation2} shows that our proposed method achieves the lowest estimation errors for both the conditional mean and standard deviation, consistently outperforming other methods. In Section K of the supplementary material, we also evaluate the performance of our proposed method with the training dataset size $n\in\{1250, 2500, 5000\}$. As summarized in Table T2 of the supplementary material, as $n$ increases, our proposed method performs better accordingly, which is consistent with Theorem \ref{thm:main thm}.

\begin{table}[H]
\setlength{\belowcaptionskip}{5pt}
  \centering
  \footnotesize
  \caption{\small ${\rm MSE}_1$ and ${\rm MSE}_2$ for different methods.}
  \label{tab:simulation2}
  \renewcommand{\arraystretch}{0.65}
  
  \begin{tabular}{cc ccccccc}
    \toprule
    & & F\"ollmer & Trigonometric & VE-SDE & VAE & WGAN & NNKCDE & FlexCode \\
    \midrule
    \multirow{2}{*}{(M1)} & ${\rm MSE}_1$ & 0.023 & 0.027 & 0.082 & 0.202 & 2.531 & 1.724 & 1.148 \\
    & ${\rm MSE}_2$ & 0.001 & 0.002 & 0.094 & 0.083 & 0.317 & 0.790 & 1.004 \\
    \midrule
    \multirow{2}{*}{(M2)} & ${\rm MSE}_1$ & 0.214 & 0.310 & 0.235 & 0.927 & 4.106 & 2.844 & 1.408 \\
    & ${\rm MSE}_2$ & 0.153 & 0.191 & 0.151 & 0.548 & 1.736 & 1.137 & 0.919 \\
    \midrule
    \multirow{2}{*}{(M3)} & ${\rm MSE}_1$ & 0.277 & 0.325 & 0.332 & 0.565 & 1.413 & 1.406 & 1.382 \\
    & ${\rm MSE}_2$ & 0.051 & 0.054 & 0.562 & 0.063 & 0.345 & 0.267 & 0.618 \\
    \bottomrule
  \end{tabular}
  \vspace{-0.5cm}
\end{table}

\subsection{Real Data Analysis I} %
\label{subsec:wine}
We consider the wine quality dataset \citep{misc_wine_quality_186} in the UCI machine learning repository, which is a combination of two sub-datasets, related to red and white vinho verde wine samples, from the north of Portugal. 
The classes are ordered and not balanced (e.g. there are much more normal wines than excellent or poor ones). This dataset contains 11 continuous features: fixed acidity, volatile acidity, citric acid, residual sugar, chlorides, free sulfur dioxide, total sulfur dioxide, density, pH, sulphates, and alcohol.
The main purpose of this dataset is to rank the wine quality (discrete score between 0 and 10) based on the chemical analysis measurements (the 11 features mentioned above).
The total sample size of this dataset is 6497. 
Denote by $X$ the score of wine quality, and by $\mathbf{Y}$ the vector of the chemical analysis measurements.
We randomly use 90\% of it for training and the rest 10\% for testing. 
Denote by $\mathcal{Y}$ the set including all the feature vectors of the testing set. 

 We compare the prediction intervals of wine quality for given features $\mathbf{Y}$ constructed by the generation-based methods (our proposed method, Trigonometric, VE-SDE, VAE and
WGAN). 
 For all the deep generative methods,
we use the Adam algorithm with a learning rate of $0.0005$ to train the models. 
 For our proposed method, we first train the velocity estimator on the training set, and then for each $\mathbf{y}_i \in \mathcal{Y}$ we generate $N_*$ predictions $\{\hat{X}_i^{(j)}\}_{j=1}^{N_*}$ with $N_*=1000$ associated with $\mathbf{y}_i$ by Algorithm \ref{algorithm}.
Let $\bar{\hat{X}}_i$ and $\hat{s}_i$ be the sample mean and sample standard deviation of $\{\hat{X}_i^{(j)}\}_{j=1}^{1000}$, respectively.
 For each given $i$, we can approximate the distribution of the ancillary statistic $(\hat{s}_i \sqrt{1+N_*^{-1}})^{-1} (X-\bar{\hat{X}}_i)$ by the Student's $t$-distribution with $N_*-1$ degrees of freedom.   Let $z_{1-\alpha/2}$ be the $(1-\alpha/2)$-quantile of the Student's $t$-distribution with $N_*-1$ degrees of freedom. Then
\begin{align*}
\mathcal{C}_{i,1-\alpha}(\mathbf{y}_i)=\big\{x:\bar{\hat{X}}_i - z_{1-\alpha/2} \hat{s}_i \sqrt{1 + N_*^{-1}} \leq x \leq \bar{\hat{X}}_i + z_{1-\alpha/2} \hat{s}_i \sqrt{1 + N_*^{-1}}\big\}
\end{align*}
provides an approximation of the $100(1-\alpha)$\% prediction interval of $X$ when $\mathbf{Y}=\mathbf{y}_i$. 
 We compute
$
{\rm CR}_{1-\alpha}=\frac{1}{650} \sum_{i=1}^{650} \mathbb{I}\{X_i\in\mathcal{C}_{i,1-\alpha}(\mathbf{Y}_i)\}
$
with the samples $(X_i,\mathbf{Y}_i)$ from the testing set.
For each given $\alpha \in (0, 1)$, the closer ${\rm CR}_{1-\alpha}$ is to $1-\alpha$, the more accurate our constructed prediction interval is. 
Similarly, we can apply the above procedure for other generation-based methods. 
With selecting $\alpha =$ 10\%, 5\% and 1\%, 
Table \ref{tab:wine} reports the associated ${\rm CR}_{1-\alpha}$ for all the generation-based methods, indicating that our proposed method yields the most accurate coverage of the prediction intervals for $\alpha=0.05$ and $0.1$. Trigonometric achieves comparable performance to our proposed method, which is little better than WGAN. VAE fails to produce a valid prediction interval in this task.
 While VE-SDE provides the most accurate prediction interval coverage for $\alpha=0.01$, its performance degrades and fails to maintain precision at $\alpha=0.1$.

\begin{table}[H]
\setlength{\belowcaptionskip}{5pt}
\renewcommand{\arraystretch}{0.65}
\footnotesize
  \centering
    \caption{\small Associated ${\rm CR}_{1-\alpha}$ for the generative methods with different selections of $\alpha$.}
    \label{tab:wine}  
\begin{tabular}{cccccc}
\toprule
$\alpha$ & F\"ollmer & Trigonometric & VE-SDE & VAE & WGAN \\
\midrule
0.01 & 98.31\% & 98.31\% & 99.23\% & 42.15\% & 98.77\% \\
0.05 & 94.77\% & 95.38\% & 96.15\% & 36.31\% & 93.85\% \\
0.10 & 90.62\% & 91.08\% & 94.77\% & 32.77\% & 91.38\% \\
\bottomrule
\end{tabular}
\vspace{-0.5cm}
\end{table}

 \subsection{Real Data Analysis II}%
\label{subsec:mnist}
We apply our proposed method to high-dimensional conditional generation problems.
We work on the MNIST handwritten digits dataset, which contains 60000 images for training \citep{lecun1998mnist}.
Each image is represented as a 28 × 28 matrix with gray color intensity from 0 to 1, and paired with a label in $\{0, 1, \ldots , 9\}$ indicating the corresponding digit. 
We flatten the 28 x 28 pixel matrix into a vector to represent $\mathbf{X} \in \mathbb{R}^{784}$.
We perform  on two tasks: generating images by classes and reconstructing missing parts of images from partial observations.  For conditional generation, we utilize one-hot labels as conditions. For image inpainting, we treat the conditional observation as an additional input channel. Both tasks are concrete instances of the conditional AGI problems highlighted in Section~\ref{sec:Introduction}: the class label or the partial image plays the role of the condition $\mathbf{Y}$, while the generated image is the content $\mathbf{X}$. For all the deep generative methods (our proposed method, Trigonometric, VE-SDE, VAE
and WGAN), we use the Adam algorithm with a learning rate of 0.001 to train the models. It is noteworthy that such high-dimensional image-related problems often exceed the scope of traditional density estimation methods.

\subsubsection{Class Conditional Generation}
\label{subsec:class_cond}
We target on generating images of handwritten digits given labels from $\{0, 1, \ldots, 9\}$.
In this problem, the condition is a categorical variable representing one of the ten digits, and we follow the common practice to use one-hot vectors to represent $\mathbf{Y}$ with 
each label being represented by a binary vector in $\mathbb{R}^{10}$ where only one element is hot (set to 1) and all others are cold (set to 0).

\begin{figure}[!b]
\spacingset{1}
\centering
    \makebox[.25\textwidth]{Real}
    \hspace{0.03\textwidth}
    \makebox[.25\textwidth]{F\"ollmer}
    \hspace{0.03\textwidth}
    \makebox[.25\textwidth]{Trigonometric}
    \\
    \includegraphics[width=.25\textwidth]{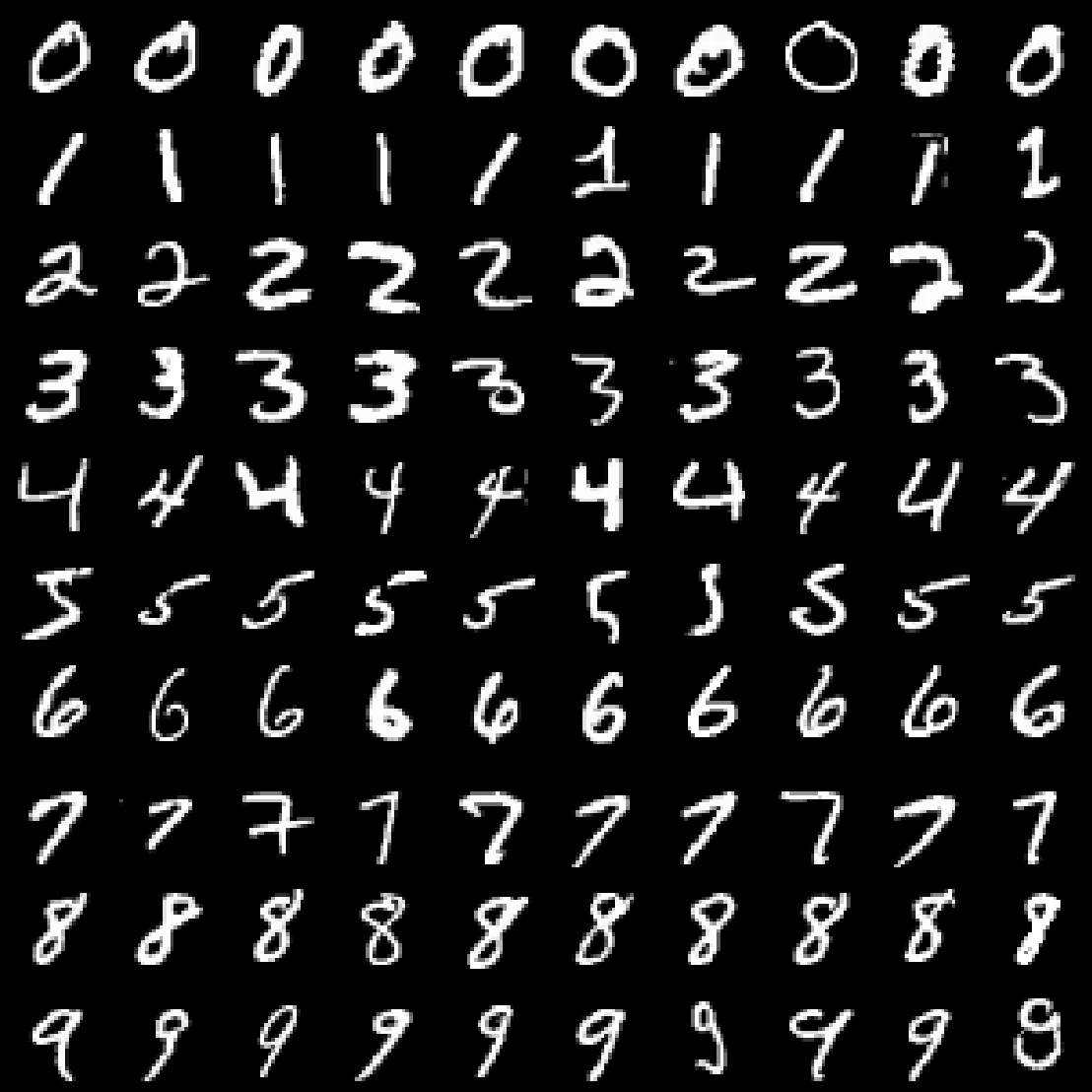}
    \hspace{0.03\textwidth}
    \includegraphics[width=.25\textwidth]{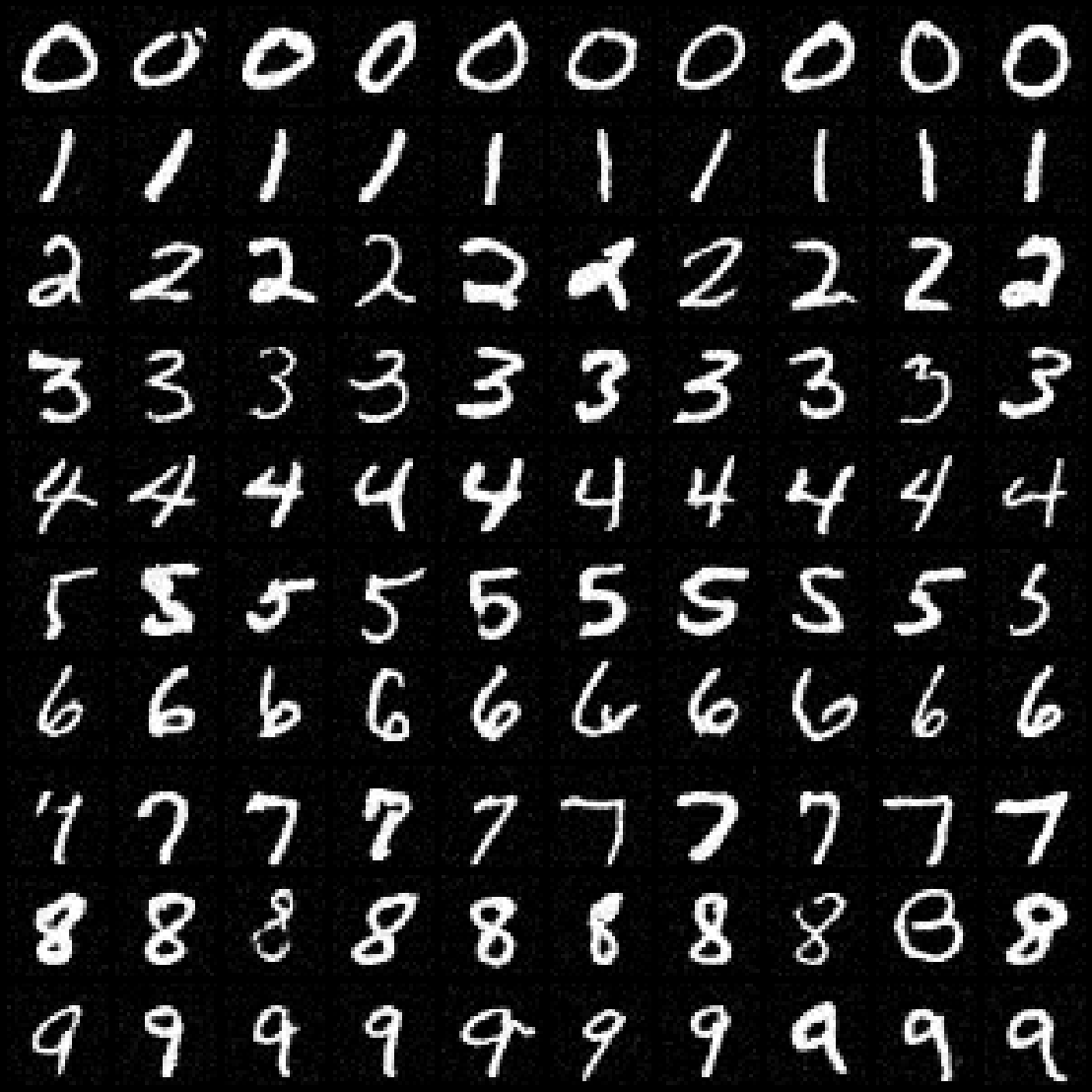}
    \hspace{0.03\textwidth}
    \includegraphics[width=.25\textwidth]{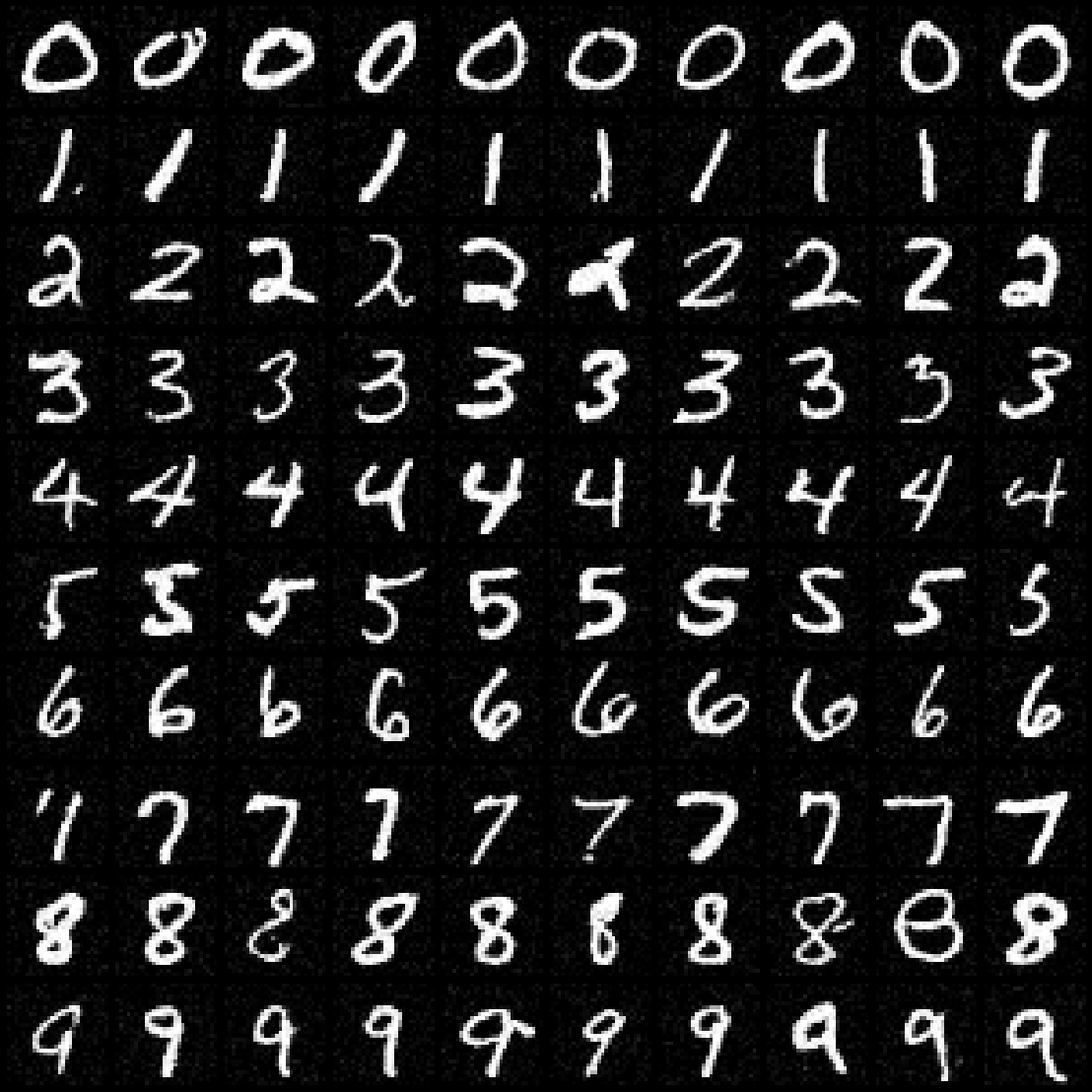}
    \\
    \makebox[.25\textwidth]{VE-SDE}
    \hspace{0.03\textwidth}
    \makebox[.25\textwidth]{VAE}
    \hspace{0.03\textwidth}
    \makebox[.25\textwidth]{WGAN}
    \\
    \includegraphics[width=.25\textwidth]{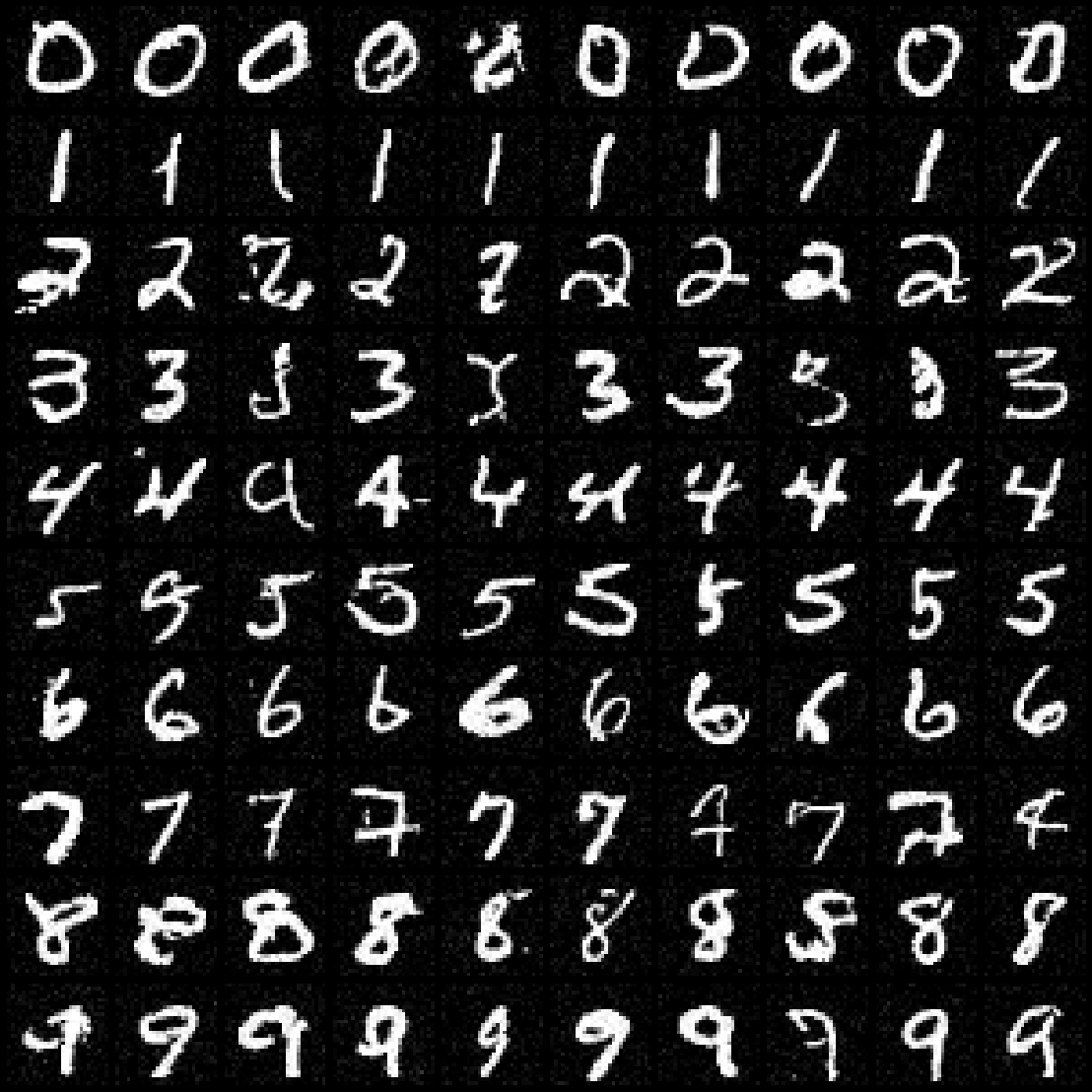}
    \hspace{0.03\textwidth}
    \includegraphics[width=.25\textwidth]{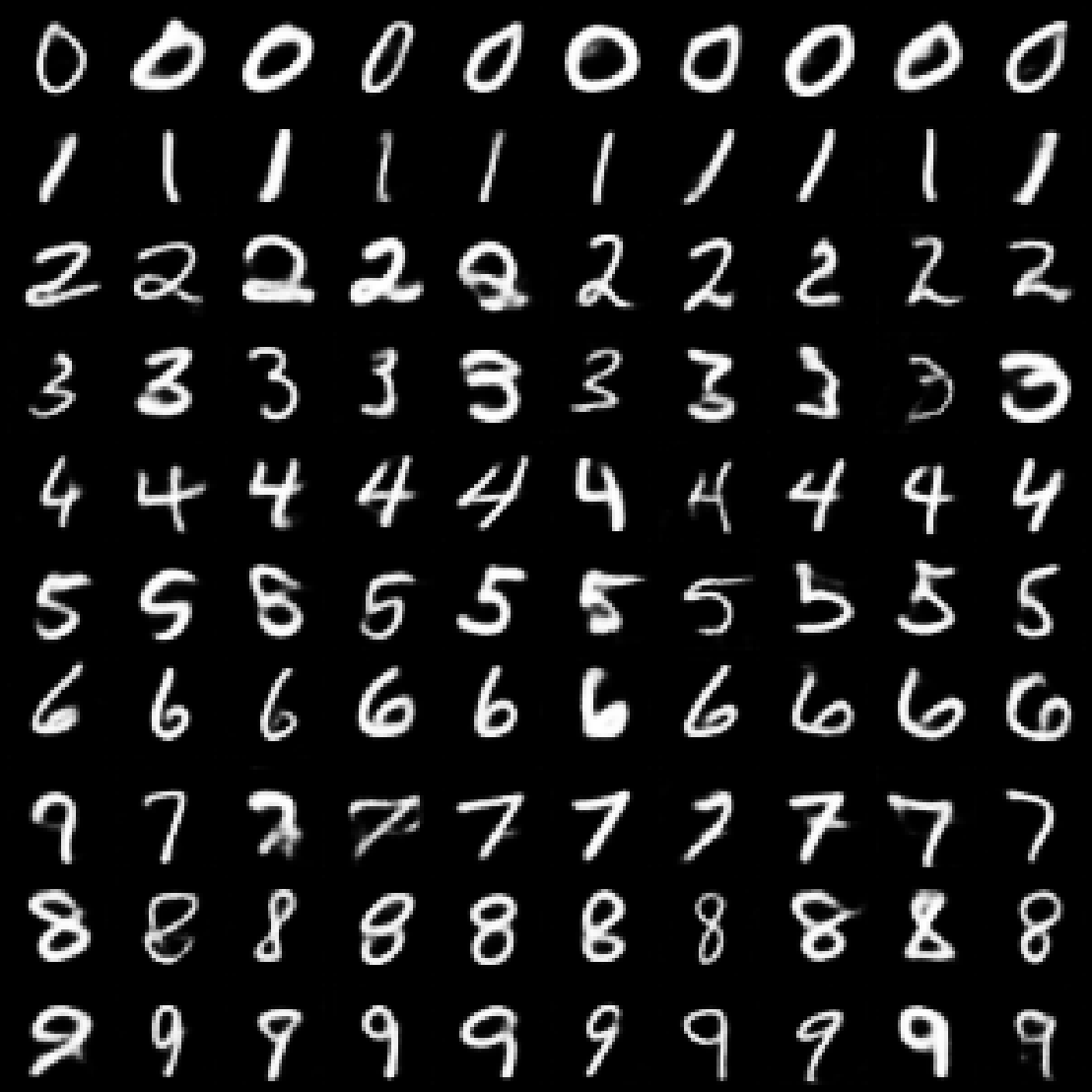}
    \hspace{0.03\textwidth}
    \includegraphics[width=.25\textwidth]{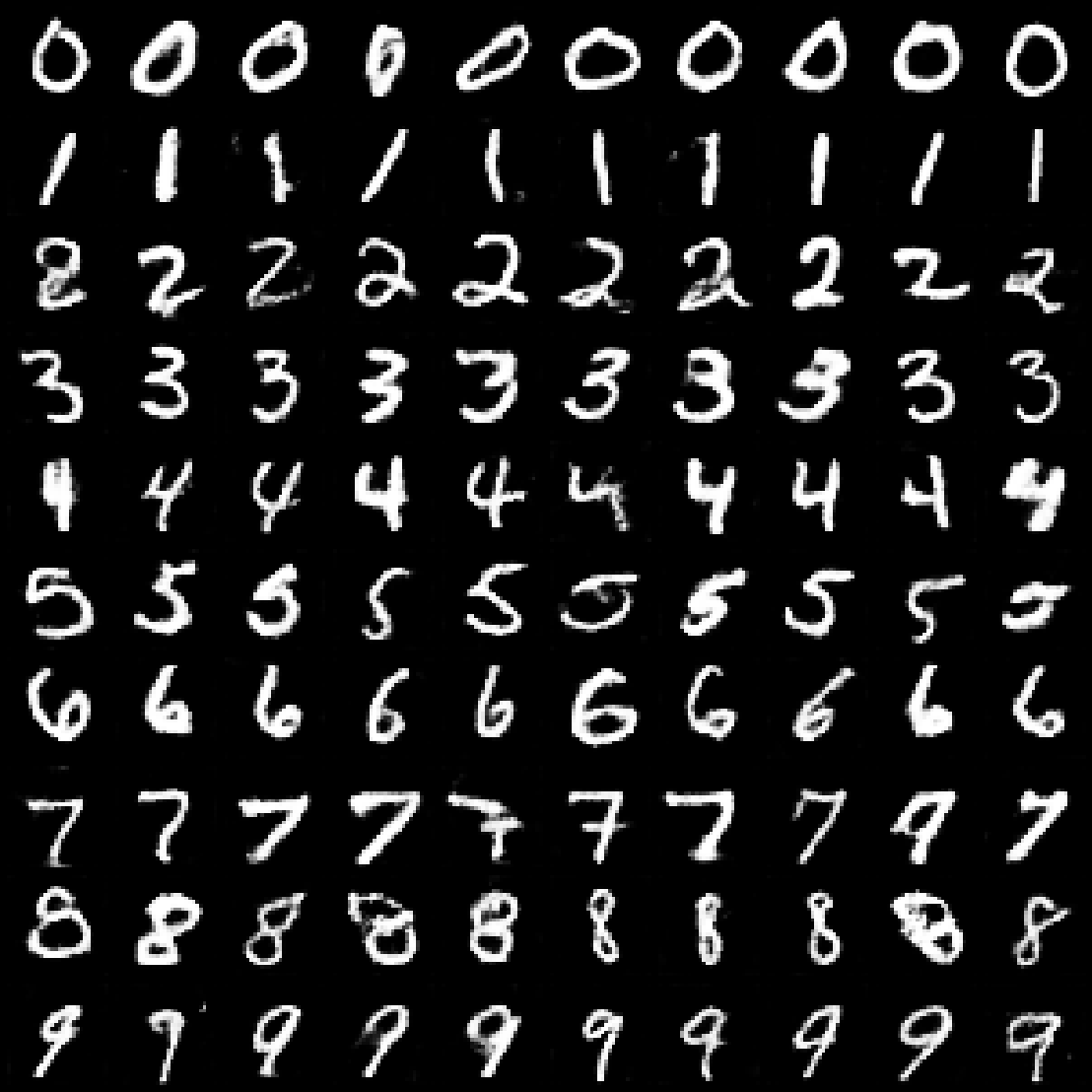}
    
    \caption{\small MNIST: real images (top-left panel) and generated images for given labels by our proposed method (top-middle panel), Trigonometric (top-right panel), VE-SDE (bottom-left panel), VAE (bottom-middle panel), and WGAN (bottom-right panel).}
    \label{fig:mnist-class-cond}
\end{figure}

For our proposed method, we train the velocity estimator on the training set, and then generate synthetic images by Algorithm \ref{algorithm}.
Figure \ref{fig:mnist-class-cond} displays the real images randomly drawn from the training set (top-left panel) and synthetic images by our proposed method (top-middle panel).
 Each row represents 10 images from the same label; and each column, from top to bottom, represents labels ranging from 0 to 9.
We also compare our proposed method with other generation-based methods.
The top-right, bottom-left, bottom-middle, and bottom-right panels of Figure \ref{fig:mnist-class-cond} display, respectively, the synthetic images by Trigonometric, VE-SDE, VAE, and WGAN.
 Generated images are similar to the real images and have differences among columns, indicating the random variations in the generating process, ensuring the richness of the generating capability.

In image generation tasks, the Fréchet inception distance (FID) is a common choice to measure difference between synthetic and real images \citep{heusel2017gans}.
Denote by $\boldsymbol{\mu}_r$ and $\boldsymbol{\Sigma}_r$, respectively, the sample mean and sample covariance matrix of all real images $\mathbf{X}$'s from the training set.
Parallelly, let $\boldsymbol{\mu}_g$ and $\boldsymbol{\Sigma}_g$ be, respectively, the sample mean and sample covariance matrix of the synthetic images $\hat{\mathbf{X}}$'s.
The { pixel space} FID between synthetic and real images is defined as $|\boldsymbol{\mu}_r - \boldsymbol{\mu}_g|_2^2 + {\rm tr}\{\boldsymbol{\Sigma}_r + \boldsymbol{\Sigma}_g - 2(\boldsymbol{\Sigma}_r\boldsymbol{\Sigma}_g)^{1/2}\}$.
For the generation-based methods, we generate 5000 images for each digit and compute the FID between the training set and the overall 50000 synthetic images. Table \ref{tab:class-cond-comp} shows that our proposed method yields the lowest FIDs compared to other generation-based methods, achieving the best generation quality.

\begin{table}[H]
\setlength{\belowcaptionskip}{8pt}
\renewcommand{\arraystretch}{0.65}
\footnotesize
  \centering
  \caption{\small FIDs for different methods in class conditional image generation on MNIST dataset}
  \label{tab:class-cond-comp}
  \begin{tabular}{cccccc}
    \toprule
    & F\"ollmer & Trigonometric & VE-SDE & VAE & WGAN \\
    \midrule
    FID & 0.30 & 0.61 & 1.27 & 1.59 & 1.17 \\
    \bottomrule
  \end{tabular}
\end{table}

 \subsubsection{Image Inpainting}
\begin{figure}[!htbp]
  \centering
\makebox[0.17\textwidth]{ F\"ollmer\phantom{W}}
\hspace{0.01\textwidth}
\makebox[0.17\textwidth]{Trigonometric}
\hspace{0.01\textwidth}
\makebox[0.17\textwidth]{VE-SDE}
\hspace{0.01\textwidth}
\makebox[0.17\textwidth]{\phantom{--}VAE}
\hspace{0.01\textwidth}
\makebox[0.17\textwidth]{\phantom{W-}WGAN}
\hspace{0.01\textwidth}
\\[5pt]
\includegraphics[width=.18\textwidth]{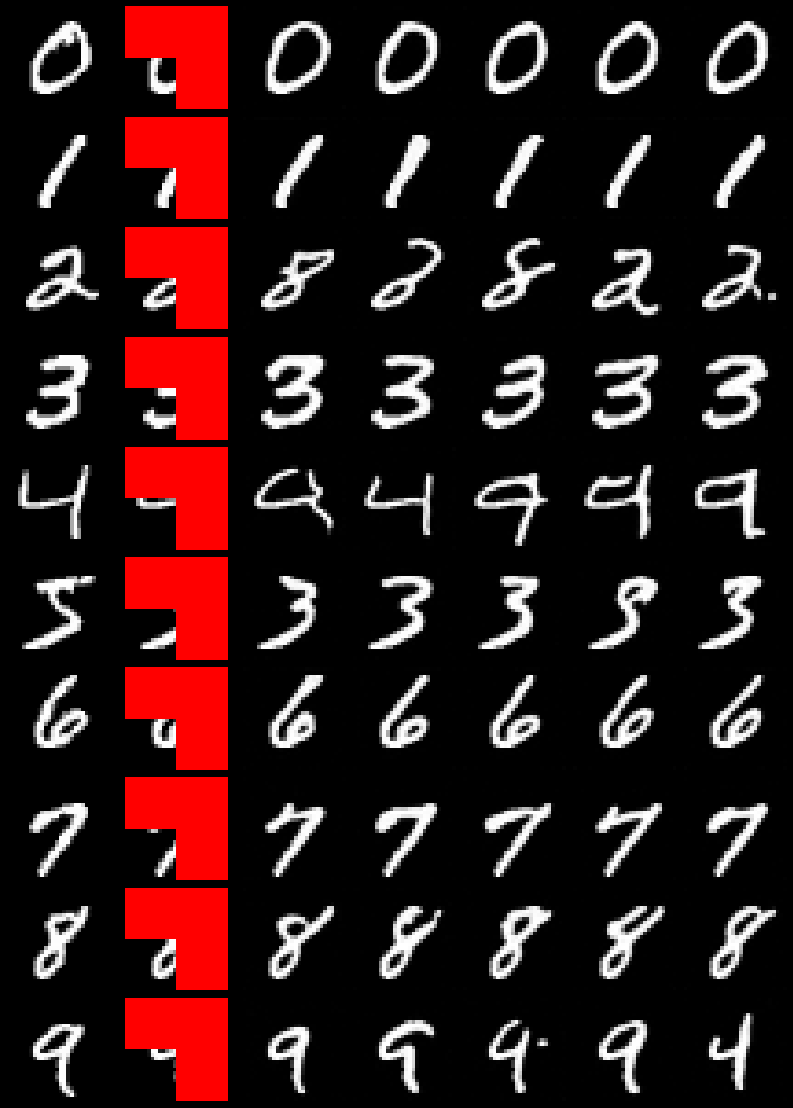}
\hspace{0.01\textwidth}
\includegraphics[width=.18\textwidth]{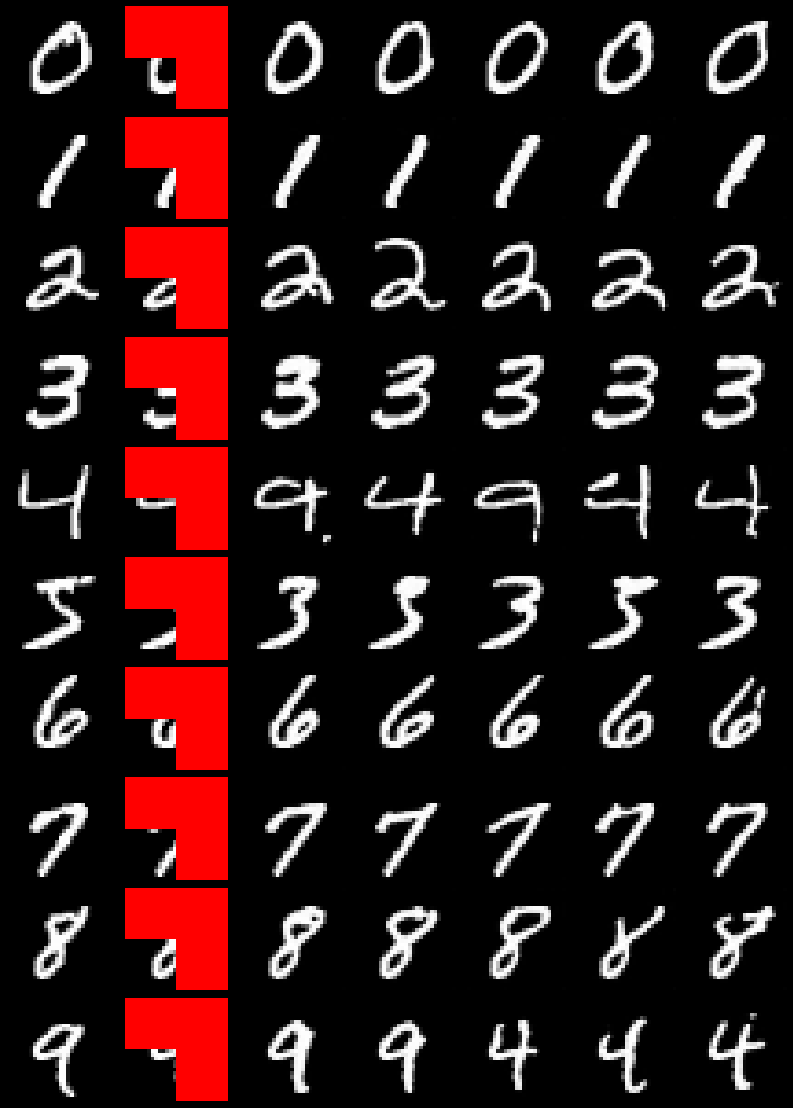}
\hspace{0.01\textwidth}
\includegraphics[width=.18\textwidth]{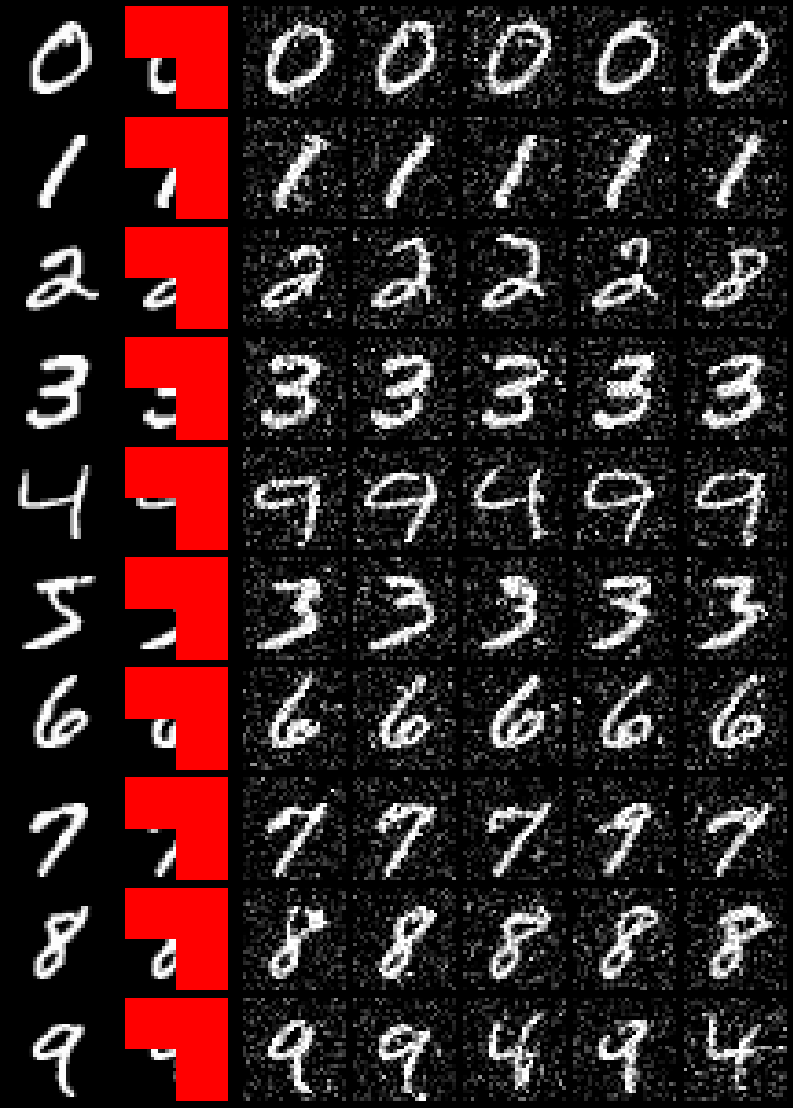}
\hspace{0.01\textwidth}
\includegraphics[width=.18\textwidth]{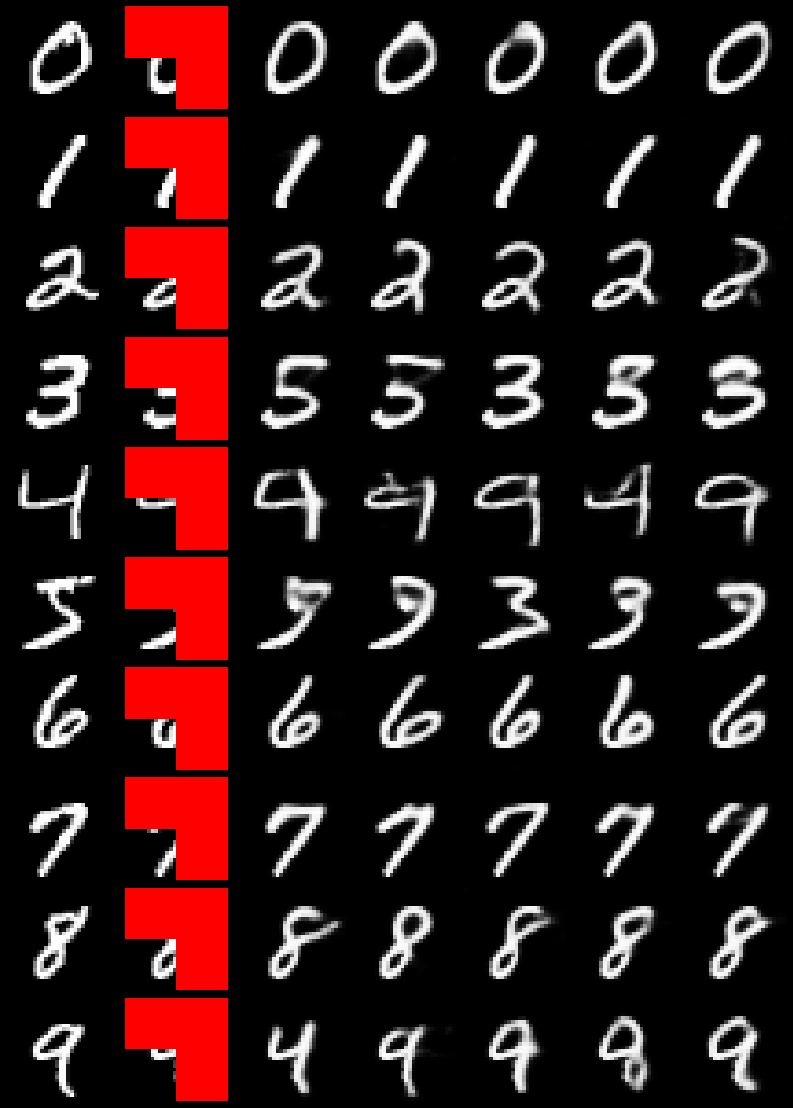}
\hspace{0.01\textwidth}
\includegraphics[width=.18\textwidth]{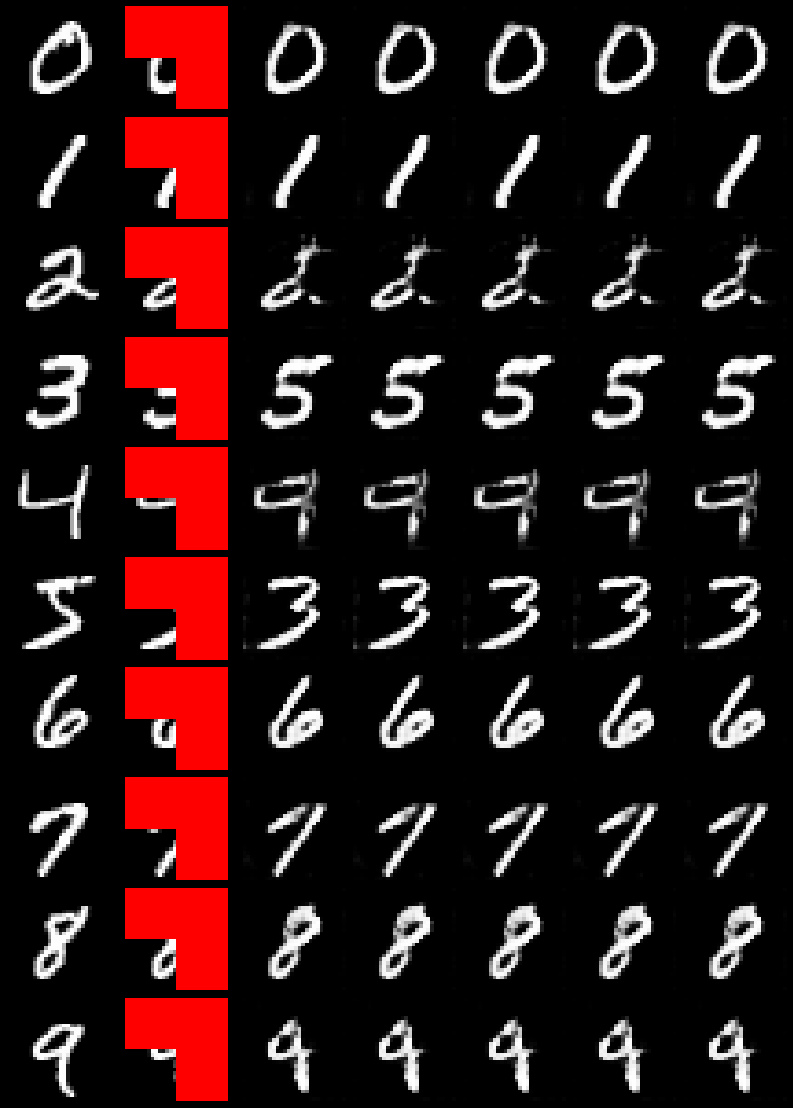}
\\
\includegraphics[width=.18\textwidth]{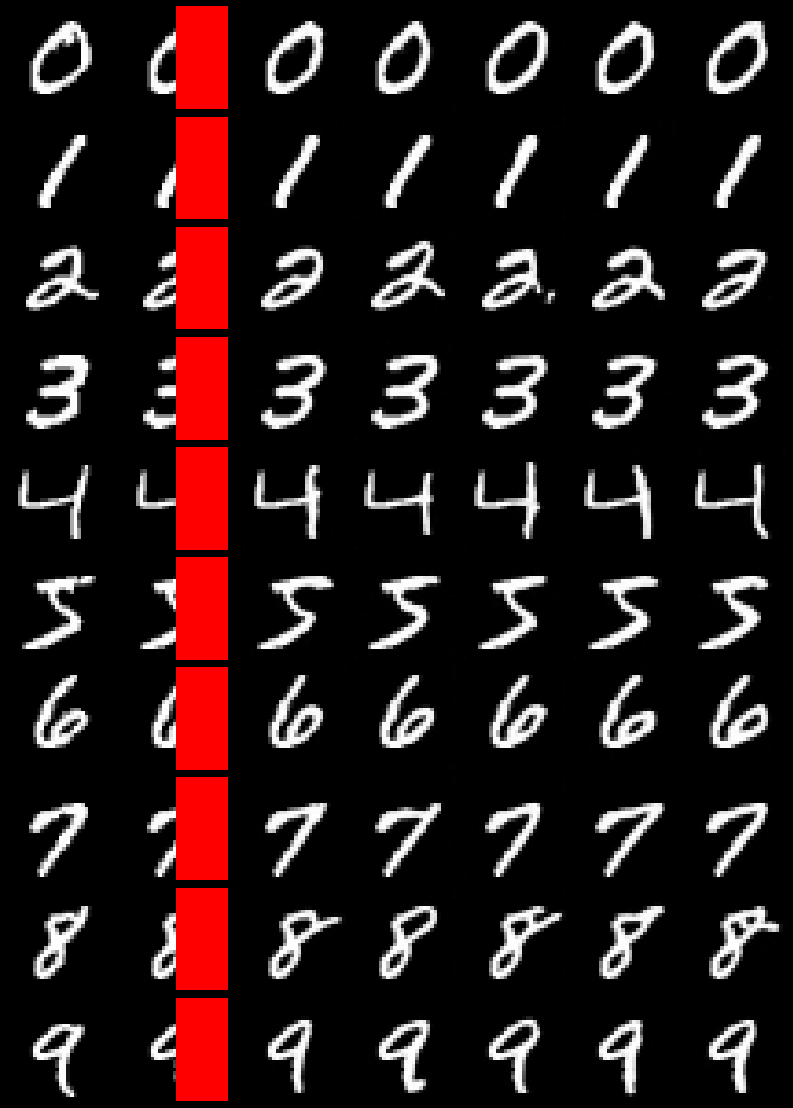}
\hspace{0.01\textwidth}
\includegraphics[width=.18\textwidth]{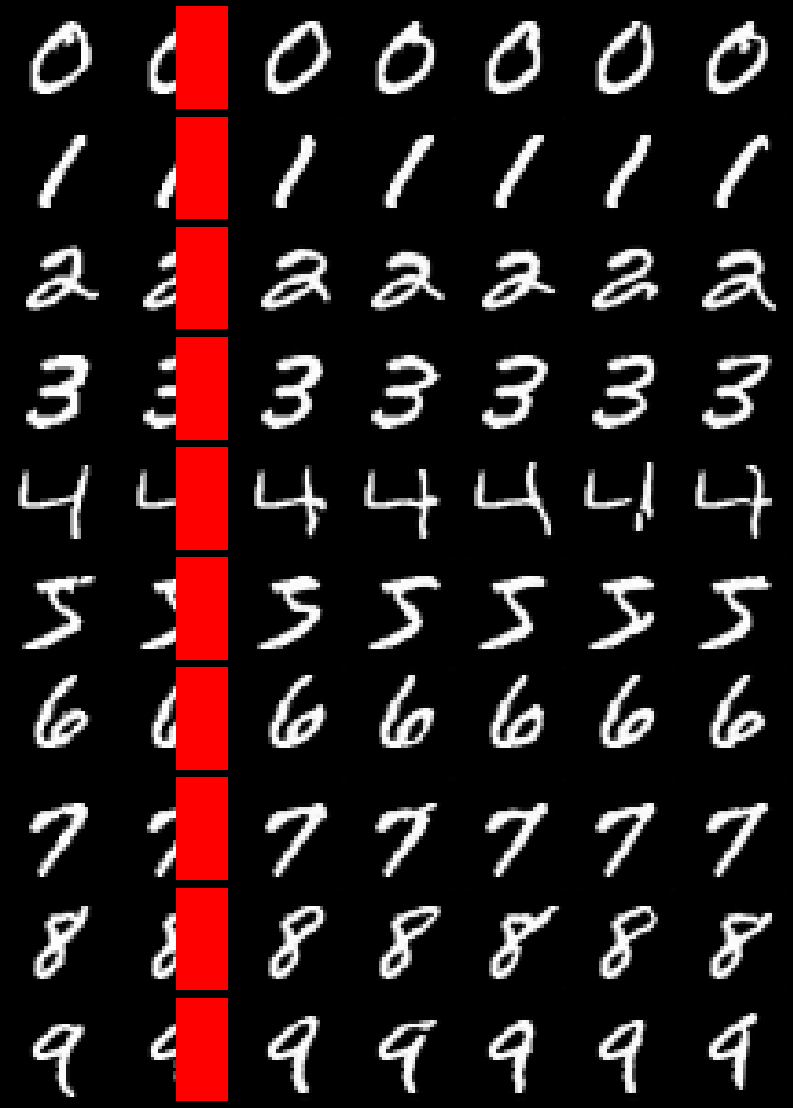}
\hspace{0.01\textwidth}
\includegraphics[width=.18\textwidth]{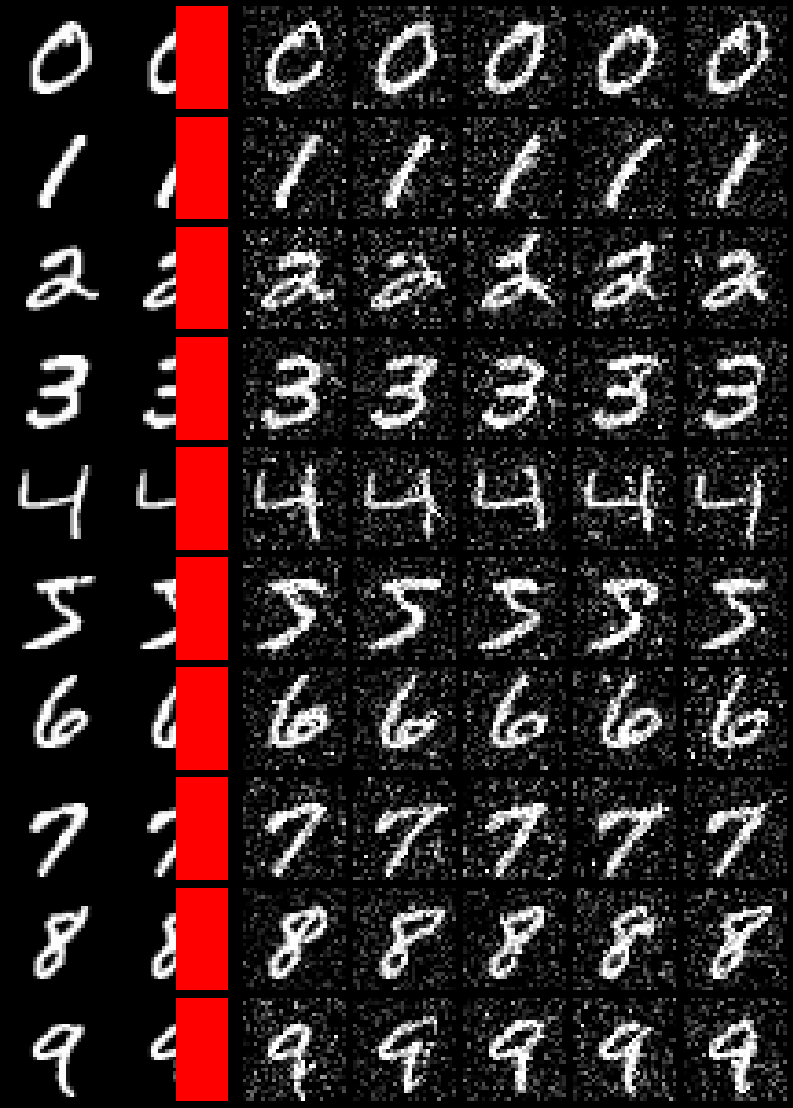}
\hspace{0.01\textwidth}
\includegraphics[width=.18\textwidth]{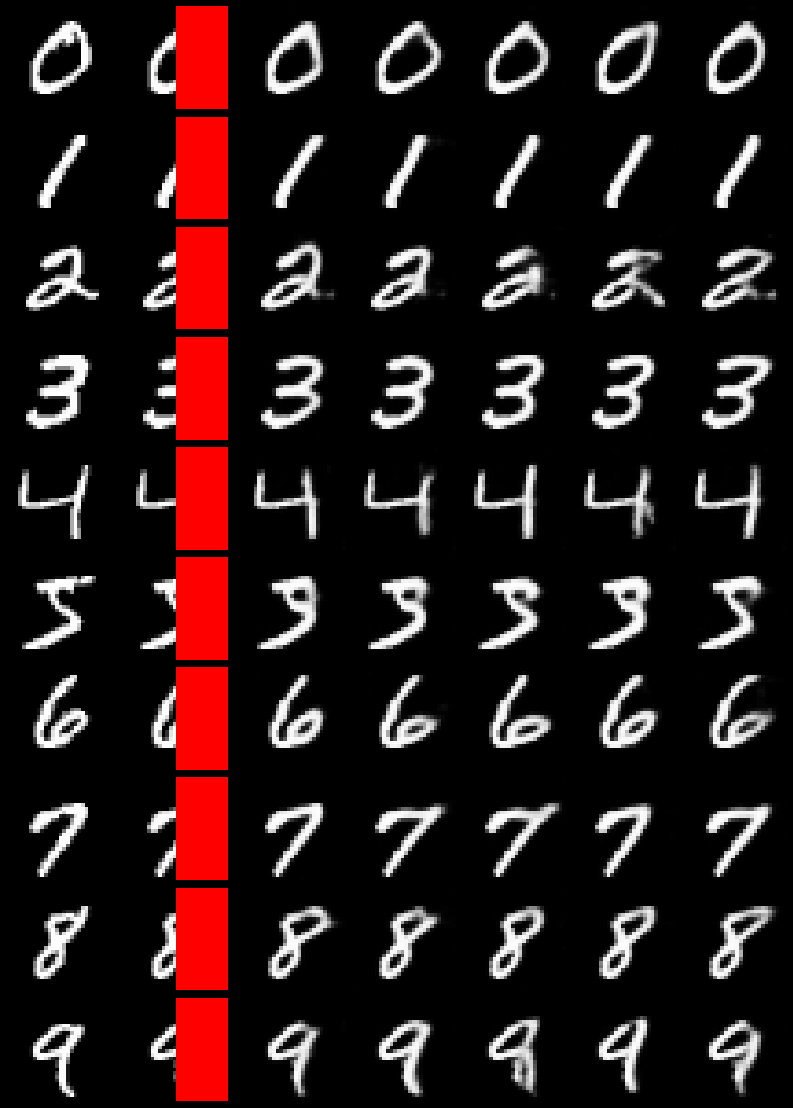}
\hspace{0.01\textwidth}
\includegraphics[width=.18\textwidth]{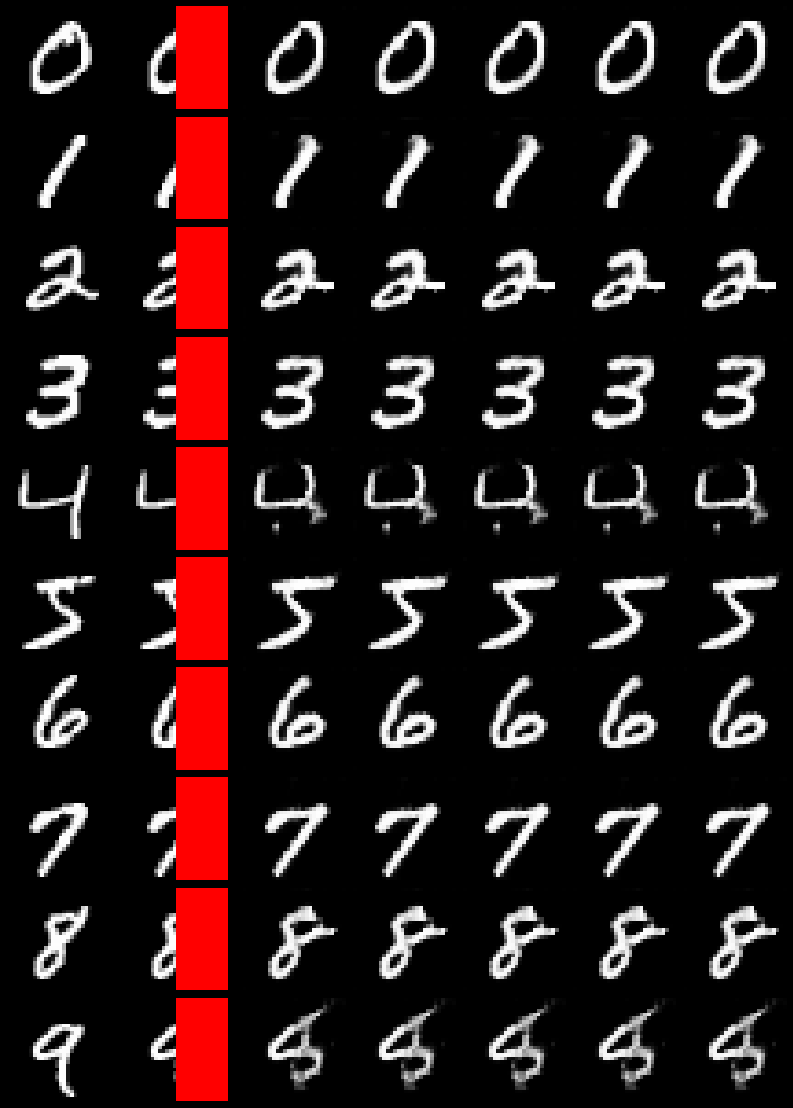}
\\
\includegraphics[width=.18\textwidth]{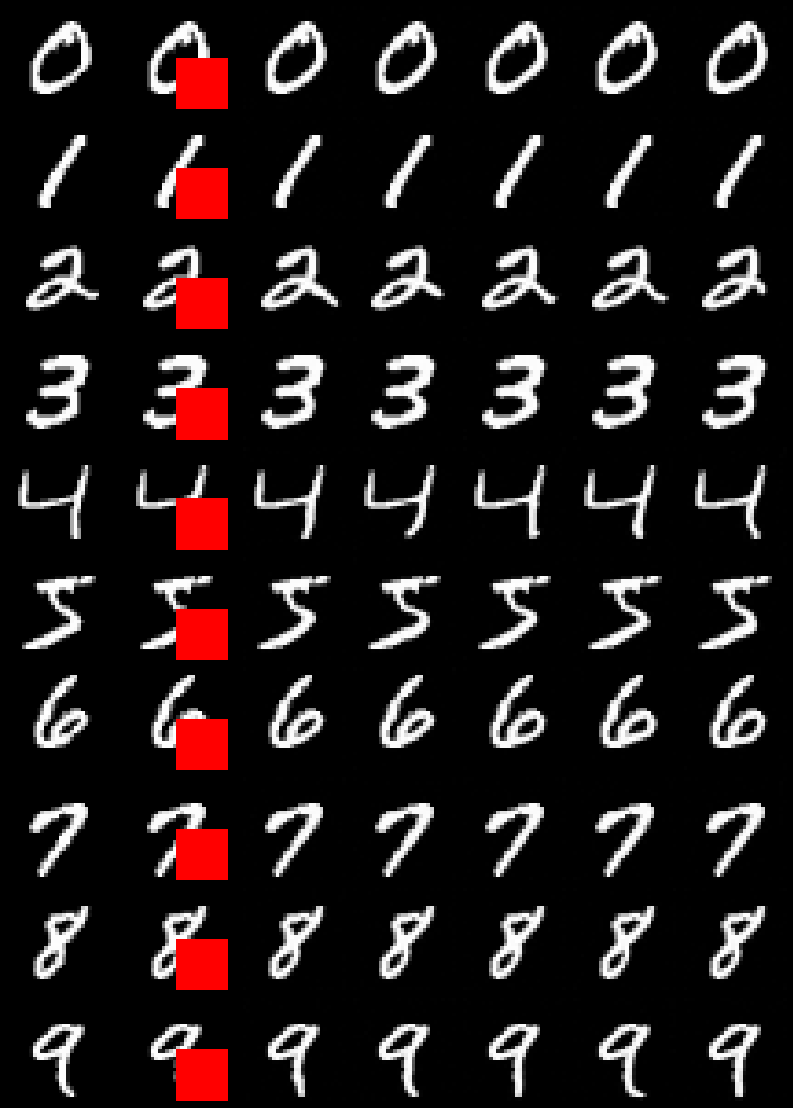}
\hspace{0.01\textwidth}
\includegraphics[width=.18\textwidth]{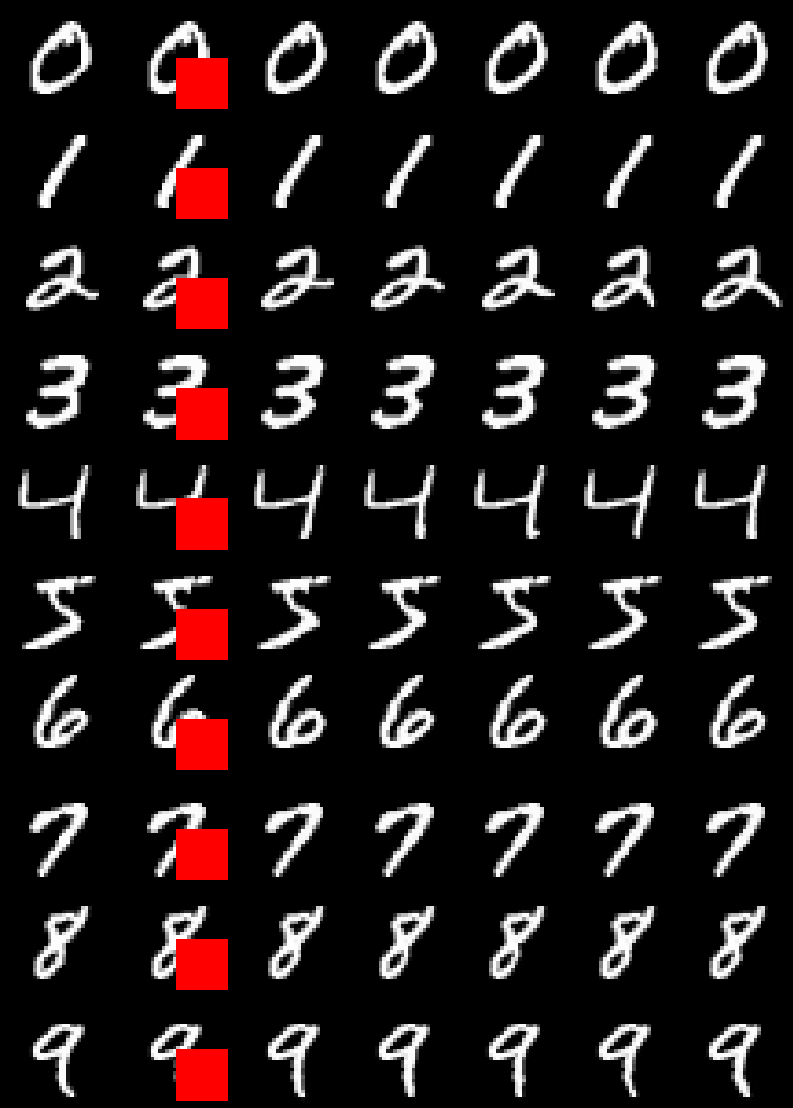}
\hspace{0.01\textwidth}
\includegraphics[width=.18\textwidth]{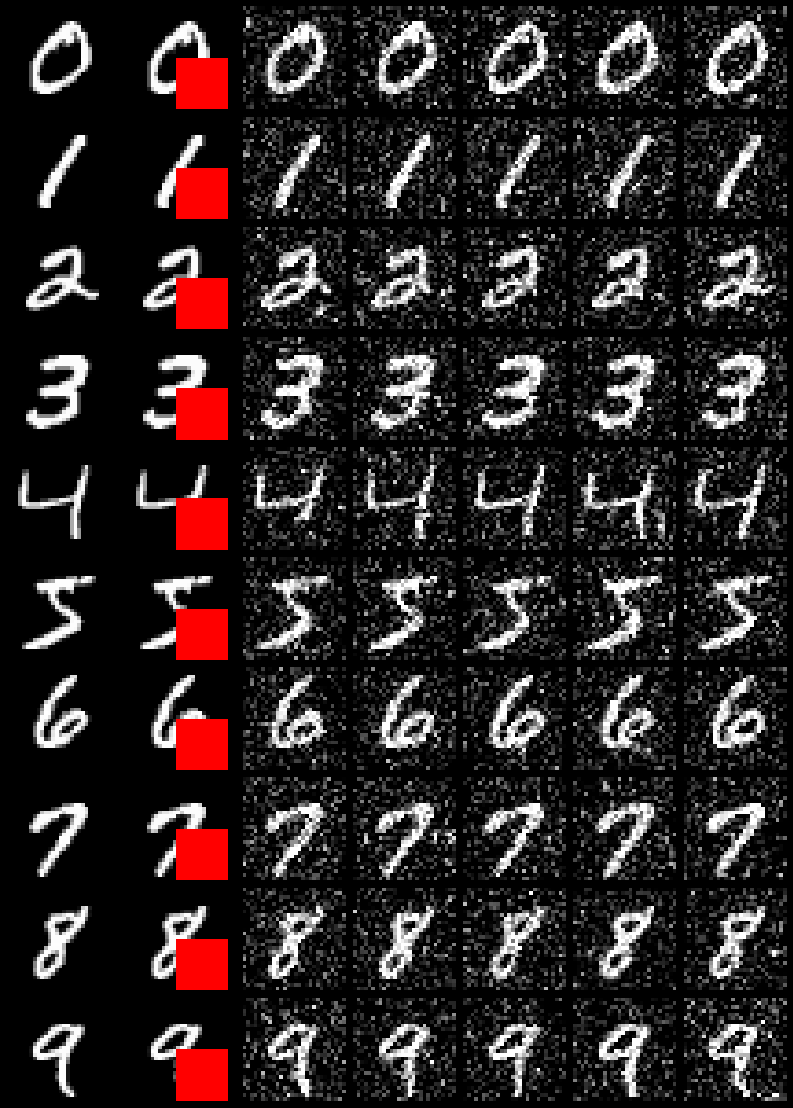}
\hspace{0.01\textwidth}
\includegraphics[width=.18\textwidth]{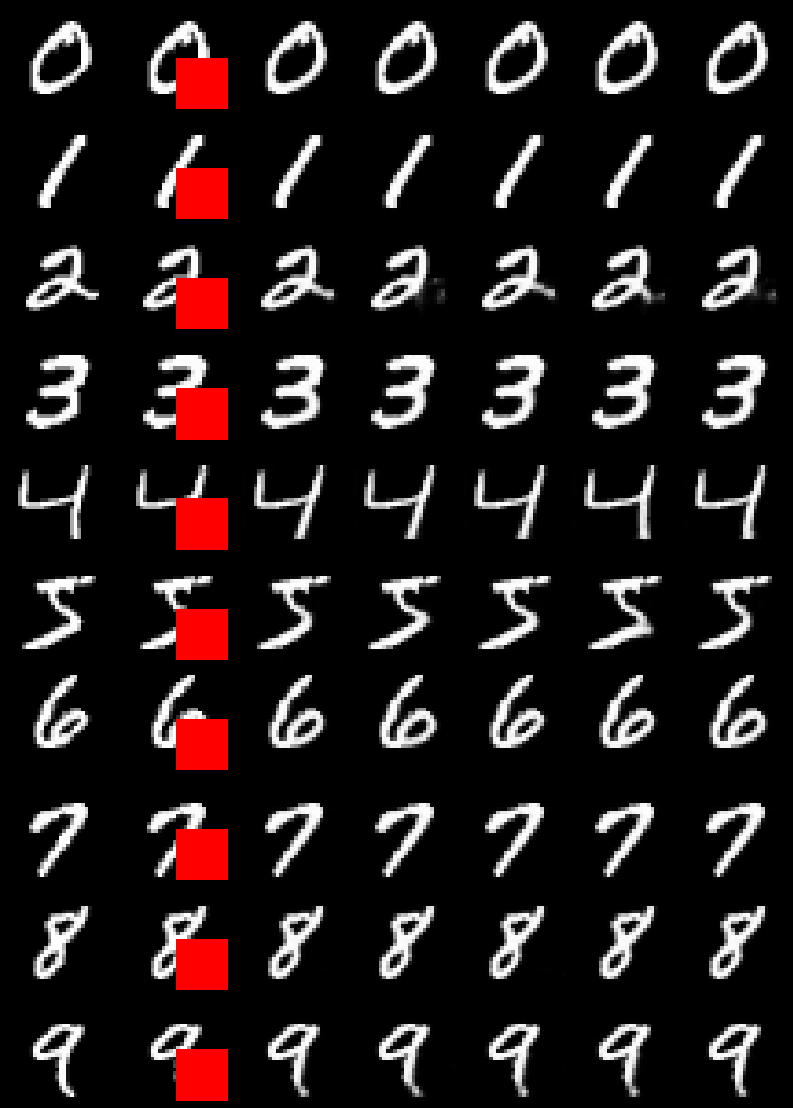}
\hspace{0.01\textwidth}
\includegraphics[width=.18\textwidth]{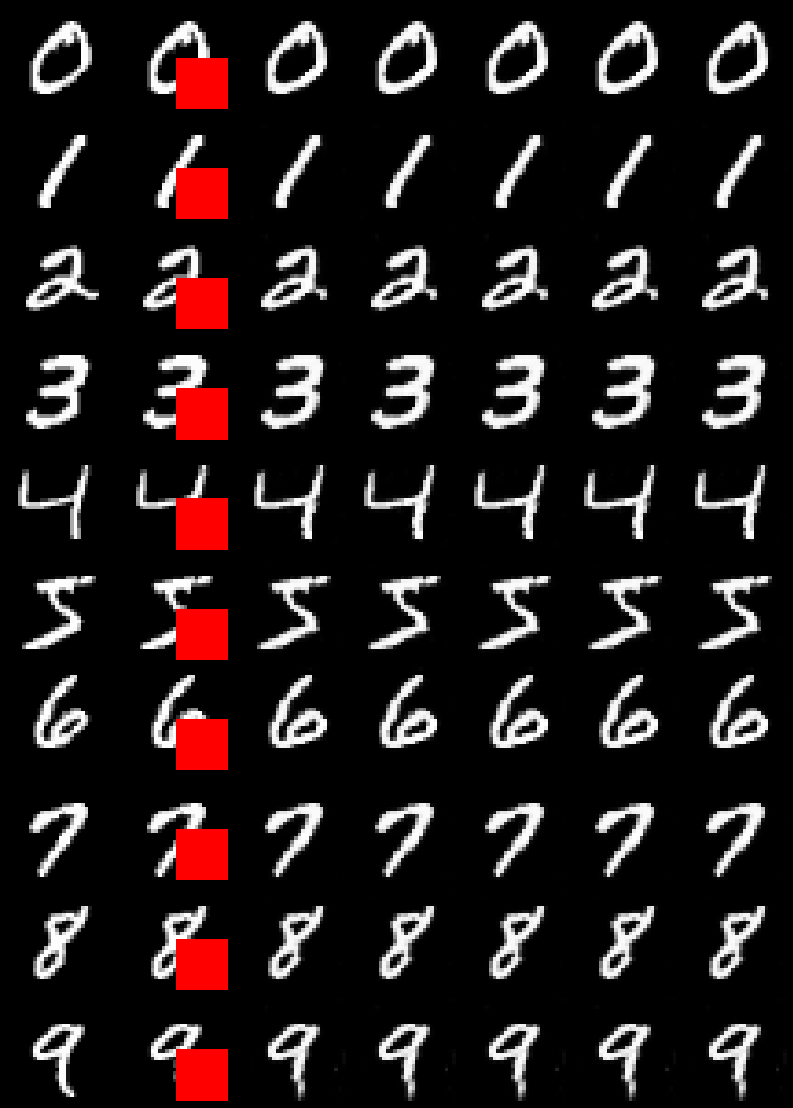}
\\
  \caption{ \small
    Original testing images $\{\mathbf{X}_i\}_{i=1}^{10}$ (first columns), associated conditions $\{\mathbf{Y}_i\}_{i=1}^{10}$ (second columns, with the covered parts shaded in red), and reconstructed images $\{\hat{\mathbf{X}}_i^{(j)}\}_{j=1}^{5}$ with $i=1, \ldots, 10$ by our proposed method, Trigonometric, VE-SDE, VAE and WGAN (from left to right).
  }
  \label{fig:mnist-inpaint}
\end{figure}

We target on reconstructing an image when part of it is covered. 
In this problem, $\mathbf{X} \in \mathbb{R}^{784}$ is the original intact image and $\mathbf{Y}$ is the associated uncovered part of $\mathbf{X}$.
Our goal is to reconstruct the image $\mathbf{X}$ when just its partial observation $\mathbf{Y}$ is given. For each prescribed constant $\delta\in\{3/4, 1/2, 1/4\}$, we first prepare a new training set for inpainting by using the same images $\mathbf{X}$'s as those from the MNIST training set, and then manually cover $\delta$ part of each original image $\mathbf{X}$ to obtain the associated condition $\mathbf{Y}$.
The MNIST dataset also contains a testing set with size 10000.
We randomly draw 10 images $\{\mathbf{X}_i\}_{i=1}^{10}$ in the testing set, with corresponding digits ranging from 0 to 9, and display them in the first column of each panel in Figure \ref{fig:mnist-inpaint}.
The second column of each panel in Figure \ref{fig:mnist-inpaint} displays the associated $\{\mathbf{Y}_i\}_{i=1}^{10}$ of $\{\mathbf{X}_i\}_{i=1}^{10}$, with the covered parts shaded in red.

 For our proposed method, we first train the velocity estimator on the new training set, and then for each $i=1, \ldots, 10$, we generate 5 samples $\{\hat{\mathbf{X}}_i^{(j)}\}_{j=1}^5$ associated with $\mathbf{Y}_i$ by Algorithm \ref{algorithm}.
These 5 samples are displayed from the third to the seventh columns of the related 3 panels in Figure \ref{fig:mnist-inpaint}.
We also compare our proposed method with the other generation-based methods. 
The results reconstructed by the other methods are also displayed in Figure \ref{fig:mnist-inpaint}. The reconstructed results show that
(i) if 3/4 of the images is missing, our proposed method successfully reconstructs images for most digits and fails for some difficult cases, which confuse `2' with `8', `4' with `9', and `5' with `3',
and (ii) if 1/2 or 1/4 of the images is missing, our proposed method is able to reconstruct all images correctly. We also compute the FIDs for all the generation-based methods.
For each prescribed constant $\delta\in\{3/4, 1/2, 1/4\}$, { we generate 5 reconstructed images $\hat{\mathbf{X}}$ for each image $\mathbf{X}$ in the testing set by the generation-based methods}. Same as what we did in Section \ref{subsec:class_cond}, for each generation-based method, we compute the FID between the 60000 original images and { 50000} reconstructed images. The related results are reported in Table \ref{tab:inpaint}, which show that our proposed method yields the lowest FIDs compared to other generation-based methods, achieving the best generation and reconstruction quality.

\begin{table}[!htbp]
\setlength{\belowcaptionskip}{3pt}
\renewcommand{\arraystretch}{0.65}
  \footnotesize
  \centering
   \caption{\small FIDs for different methods when 3/4, 1/2 and 1/4 of the images are covered.}
   \label{tab:inpaint}  
\begin{tabular}{cccccc}
\toprule
$\delta$ & F\"ollmer & Trigonometric & VE-SDE & VAE & WGAN \\
\midrule
$3/4$ & 0.32 & 0.51 & { 0.77} & 1.57 & 2.54 \\
$1/2$ & 0.35 & 0.54 & { 0.76} & 1.02 & 1.58 \\
$1/4$ & 0.40 & 0.57 & { 0.81} & 0.53 & 1.44 \\
\bottomrule
\end{tabular}
\end{table}

\section{Discussion} \label{sec:discuss}
In this paper, we introduce an ODE-based generative method for sampling data from conditional distributions. Such conditional sampling is closely connected to modern AGI, since systems like Stable Diffusion \citep{esser2024scaling} and Sora \citep{liu2024sora} are, at their core, conditional generative models built upon SDE/ODE-based samplers. For such models, our analysis offers, to the best of our knowledge, the first end-to-end convergence guarantee. Two issues deserve further investigation. First, we have demonstrated the regularity of flow map $\mathbf{F}_t(\cdot, \mathbf{y})$, $t \in [0, T]$, and discussed the use of an additional neural network to facilitate one-step generation in Section \ref{sec:scheme}, which can be regarded as directly fitting $\mathbf{F}_T(\cdot, \mathbf{y})$. 
In fact, we can also train a time-dependent neural network to approximate $\mathbf{F}_t(\cdot, \mathbf{y})$ for all $t \in [0, T]$, i.e., predicting the trajectories or characteristic curves of the conditional F\"{o}llmer flow. 
Such research will aid in comprehending the mechanisms of generative models, laying foundation for more efficient and stable sampling methods. 
 Second, an important and challenging direction is to study whether our proposed method has minimax optimality. On one hand, the minimax optimal rate for conditional sampling methods is unclear to the best of our knowledge. On the other hand, the most relevant work to this topic, \cite{fukumizuflow}, proves that in the unconditional case, ODE-based sampling method can nearly achieve minimax optimality, but extending its techniques to the conditional setting is non-trivial. A main difficulty is that their analysis relies on a direct regularity assumption on the velocity field, which is hard to verify based on mild assumptions imposed on the data distribution. We plan to thoroughly investigate the aforementioned problems in the future.

\newpage
\pagenumbering{roman}
\appendix
\setlength{\emergencystretch}{2em}
\renewcommand{\thepage}{S\arabic{page}} %
\setcounter{page}{1} %
\setcounter{equation}{0}
\setcounter{proposition}{0}
\setcounter{definition}{0}
\setcounter{table}{0}
\setcounter{figure}{0}
\renewcommand\theequation{S.\arabic{equation}}
\renewcommand\theproposition{P\arabic{proposition}}
\renewcommand\thedefinition{D\arabic{definition}}
\renewcommand\thetable{T\arabic{table}}
\renewcommand\thefigure{F\arabic{figure}}
\renewcommand{\theHproposition}{supp.\arabic{proposition}}
\renewcommand{\theHdefinition}{supp.\arabic{definition}}
\renewcommand{\theHtable}{supp.\arabic{table}}
\renewcommand{\theHfigure}{supp.\arabic{figure}}
\begin{center}
{\large\bf SUPPLEMENTARY MATERIAL}
\end{center}

We first provide some notation used in the supplementary material. For matrices $\mathbf{A}$ and $\mathbf{B}$, we say $\mathbf{A} \preceq \mathbf{B}$ if $\mathbf{B} - \mathbf{A}$ is positive semi-definite. The $d$-dimensional identity matrix is denoted by $\mathbf{I}_d$. For a vector $\mathbf{x}=(x_1,\ldots,x_d)^{\rm{T}} \in \mathbb{R}^{d}$, we define $\mathbf{x}^{\otimes 2} := \mathbf{x} \mathbf{x}^{\rm{T}}$. The $\ell^2$-norm and the $\ell^{\infty}$-norm of $\mathbf{x}$ are, respectively, denoted by $|\mathbf{x}|_2 := \sqrt{\sum_{i=1}^d x_i^2}$ and $|\mathbf{x}|_{\infty} := \max_{1\leq i\leq d}|x_i|$. We denote by ${\rm tr}(\cdot)$ the trace operator on a square matrix. The operator norm of a matrix $\mathbf{A}$ is defined as $\|\mathbf{A}\|_{\text{op}} := \sup_{|\mathbf{x}|_2 \leq 1}|\mathbf{A}\mathbf{x}|_2$. For a probability density function $\pi$ and a measurable function $f: \mathbb{R}^{d} \rightarrow \mathbb{R}$, the $L^2(\pi)$-norm of $f$ is defined as $\|f\|_{L^2(\pi)} := \sqrt{\int f^2(\mathbf{x}) \pi(\mathbf{x}) \,\mathrm{d} \mathbf{x}}$, and the $L^{\infty}(K)$-norm of $f$ is defined as $\|f\|_{L^{\infty}(K)} := \sup_{\mathbf{x} \in K} |f(\mathbf{x})|$. For a vector function $\mathbf{v}: \mathbb{R}^{d} \rightarrow \mathbb{R}^{d}$, its $L^2(\pi)$-norm is denoted as $\|\mathbf{v}\|_{L^2(\pi)} := \||\mathbf{v}|_2\|_{L^2(\pi)}$, and its $L^{\infty}(K)$-norm is denoted as $\|\mathbf{v}\|_{L^{\infty}(K)} := \||\mathbf{v}|_2\|_{L^{\infty}(K)}$. 
        We use $\mathcal{U}(a, b)$ to denote the uniform distribution on interval $(a, b)$, and use $\mathcal{N}(\mathbf{0}, \mathbf{I}_{d})$ to denote the $d$-dimensional standard Gaussian distribution. For two positive sequences $\{a_n\}_{n\geq1}$ and $\{b_n\}_{n\geq1}$, 
the asymptotic notation $a_n = \mathcal{O}(b_n)$ denotes $a_n \leq C b_n$ for some constant $C > 0$. The notation $\widetilde{\mathcal{O}}(\cdot)$ is used to ignore logarithmic terms. The notation $a \sim b$ means that $a = C  b$ for some constant $C>0$. Given two distributions $\mu$ and $\nu$, the Wasserstein-2 distance $W_2(\mu, \nu)$ is defined as $W^2_2(\mu, \nu) := \inf_{\pi \in \Pi(\mu, \nu)} \mathbb{E}_{(\mathbf{x}, \mathbf{y}) \sim \pi}(|\mathbf{x} - \mathbf{y}|_2^2)$, where $\Pi(\mu, \nu)$ is the set of all couplings of $\mu$ and $\nu$. A coupling is a joint distribution on $\mathbb{R}^{d} \times \mathbb{R}^{d}$ whose marginals are $\mu$ and $\nu$ on the first and second factors, respectively.

\section{Deep Distribution Learning via F\"{o}llmer Flow}
\label{app:challenge_of_conditional}

In the main text, we state that a naive extension of an unconditional ODE sampler to the conditional setting may be practically infeasible. We use the standard, unconditional version of F\"ollmer flow as a representative example to illustrate the challenges that arise from its direct extension to conditional setting.

Let $\mathbf{X} \in \mathbb{R}^{d_x}$ with  $\mathbf{X} \sim p_{x}(\mathbf{x})$ be a random vector. The F\"ollmer flow aims to transform the standard Gaussian $\mathcal{N}(\mathbf{0},\mathbf{I}_{d_x})$, into the target data distribution $p_x(\mathbf{x})$ via an ODE. The definition of F\"{o}llmer flow is given in Definition \ref{def:UnconditionalFollmerFlow_Illustrative}.

\begin{definition}[F\"ollmer Flow] \label{def:UnconditionalFollmerFlow_Illustrative}
If $\mathbf{Z}(t)$ satisfies the following ODE:
\begin{equation} \label{eq:UnconditionalODE_Illustrative}
\mathrm{d} \mathbf{Z}(t) = \mathbf{v}(\mathbf{Z}(t), t)\,\mathrm{d} t\,, \quad t \in[0,1)\,,
\end{equation}
with $\mathbf{Z}(0) \sim \mathcal{N}(\mathbf{0}, \mathbf{I}_{d_x})$, then we call $\mathbf{Z}(t)$ the F\"ollmer flow and $\mathbf{v}(\mathbf{x}, t)$ the F\"ollmer velocity field, respectively, where the velocity field $\mathbf{v}$ is defined by
\begin{equation} \label{eq:UnconditionalVF_Illustrative}
\mathbf{v}(\mathbf{x}, t)=\frac{\mathbf{x}+\mathbf{s}(\mathbf{x}, t)}{t}\,, \quad t \in(0,1)\,, 
\end{equation}
for $\mathbf{v}(\mathbf{x}, 0)=\mathbb{E}(\mathbf{X})$, and $\mathbf{s}(\mathbf{x}, t) = \nabla_{\mathbf{x}} \log f_{t}(\mathbf{x})$ for any $t \in [0, 1)$, with $f_{t}(\mathbf{x})$ denoting the density of $t\mathbf{X}+\sqrt{1-t^2}\mathbf{W}$, and $\mathbf{W} \sim \mathcal{N}(\mathbf{0}, \mathbf{I}_{d_x})$ independent of $\mathbf{X}$.
\end{definition}

Analogous to Proposition \ref{prop:training object}, we can show that the velocity field $\mathbf{v}(\mathbf{x}, t)$ satisfies
$$\mathbf{v}(\mathbf{x}, t)=\mathbb{E}\bigg(\mathbf{X}-\frac{t}{\sqrt{1-t^2}}\mathbf{W} \, \bigg| \, t\mathbf{X}+\sqrt{1-t^2}\mathbf{W}=\mathbf{x}\bigg)\,, $$
and $\mathbf{v}(\mathbf{x}, t)$ minimizes the quadratic objective:
$$\mathcal{L}(\mathbf{v}) := \frac{1}{T}\int_0^{T}\mathbb{E}\left\{\bigg|\mathbf{X}-\frac{t}{\sqrt{1-t^2}}\mathbf{W}-\mathbf{v}(t\mathbf{X}+\sqrt{1-t^2}\mathbf{W}, t)\bigg|_2^2\right\}\,\mathrm{d} t\, ,$$
where $0<T<1$. Practically, given i.i.d. samples $\{\mathbf{X}_i\}_{i=1}^{n} \sim p_x(\mathbf{x})$ and i.i.d. samples $\{(t_j, \mathbf{W}_{j})\}_{j=1}^m$ with $t_j \sim \mathcal{U}(0, T)$ and $\mathbf{W}_j \sim \mathcal{N}(\mathbf{0}, \mathbf{I}_{d_x})$ independently, we can use deep learning methods to estimate the F\"{o}llmer velocity field $\mathbf{v}(\mathbf{x}, t)$ by minimizing an empirical version of $\mathcal{L}(\mathbf{v})$:
\begin{align} \label{eq:uncondi L hat}
    \widehat{\mathcal{L}}(\mathbf{v}) 
& =\frac{1}{mn} \sum_{i=1}^n\sum_{j=1}^m\bigg|\mathbf{X}_{i}-\frac{t_j}{\sqrt{1 - t_j^2}}\mathbf{W}_{j}-\mathbf{v}(t_j \mathbf{X}_{i}+\sqrt{1-t_j^2} \mathbf{W}_{j}, t_j)\bigg|_2^2\, .
\end{align}
 Let $\widehat{\mathbf{v}}$ be the associated estimator to \eqref{eq:uncondi L hat}. Then, given ${\mathbf{z}}_0 \sim \mathcal{N}(\mathbf{0}, \mathbf{I}_{d_x})$, we can use ${\mathbf{z}}_0$ as the start point and numerically solve the ODE $\mathrm{d} \mathbf{z}_t = \widehat{\mathbf{v}}(\mathbf{z}_t, t) \, \mathrm{d}t$ from $t=0$ to $T$ to generate pseudo samples from $p_x(\mathbf{x})$.

To generate samples from $p_{x\mkern 2mu|\mkern2muy}(\mathbf{x}\mkern 2mu|\mkern2mu\mathbf{y})$, a naive idea is to consider the condition $\mathbf{y}$ as a parameter, and apply above unconditional framework for each fixed $\mathbf{y}$. Specifically, the optimization would rely on minimizing an empirical loss defined based on the sub-samples $\{(\mathbf{X}_i, \mathbf{Y}_i)\}_{i=1}^{n_{\mathbf{y}}}$ with $\mathbf{Y}_i\equiv \mathbf{y}$:
$$\widehat{\mathcal{L}}^{\mathbf{y}}(\mathbf{v}) := \frac{1}{n_{\mathbf{y}}m}\sum_{i=1}^{n_{\mathbf{y}}}\sum_{j=1}^m\bigg|\mathbf{X}_{i}-\frac{t_j}{\sqrt{1-t_j^2}}\mathbf{W}_{j}-\mathbf{v}\big(t_j\mathbf{X}_{i}+\sqrt{1-t_j^2}\mathbf{W}_{j}, t_j\big)\bigg|_2^2\,.$$
Unfortunately, this may not be a feasible learning paradigm for conditional sampling in practice because:
\begin{itemize}
    \item If $\mathbf{Y}$ is a discrete random vector, it requires partitioning the dataset based on the value of $\mathbf{y}$. For any given $\mathbf{y}$, all data points $(\mathbf{X}_i, \mathbf{Y}_i)$ where $\mathbf{Y}_i \neq \mathbf{y}$ must be discarded. This leads to severe data inefficiency, especially for conditions with few samples.
    \item If $\mathbf{Y}$ is a continuous random vector, this naive idea is computationally infeasible. Since there are infinitely many possible values for $\mathbf{y}$, it would require training and storing an infinite number of different models, which is impossible.
\end{itemize}
Thus, a simple implementation of the unconditional F\"ollmer flow in the conditional setting is insufficient for conditional sampling. This motivates us to develop a general conditional learning framework, i.e., the Conditional F\"ollmer Flow introduced in Section \ref{sec:CFF}, which can effectively conduct conditional sampling in practice.

\section{Proof of Theorem \ref{thm:well posedness}} \label{append:CCF}

 To prove Theorem \ref{thm:well posedness}, we need the following Proposition \ref{prop:properties of vF}, which characterizes the regularity properties of the conditional F\"{o}llmer velocity field $\mathbf{v}_{\rm F}$. The proof of Proposition \ref{prop:properties of vF} can be found in Section \ref{sec:properties of vF}.
\begin{proposition} \label{prop:properties of vF}
    Let Assumptions {\rm\ref{assump:label}} and {\rm\ref{assump:bounded support}}
    hold. The following two assertions are satisfied: 

    {\rm(i)} There exists some universal constant $C>1$ independent of $(d_x, R, T)$ such that 
    \begin{align*}
        &~~~~\sup_{t \in [0,T]}\sup_{\mathbf{x} \in [-R, R]^{d_x}}\sup_{\mathbf{y} \in [0, B]^{d_{y}}}|\mathbf{v}_{\rm{F}}(\mathbf{x}, \mathbf{y}, t)|_{\infty} \leq \frac{1+TR}{1-T^2} \,, \\
        &\sup _{t \in[0, T]} \sup _{\mathbf{x} \in[-R, R]^{d_x}}\sup_{\mathbf{y} \in [0, B]^{d_{y}}} |\partial_t \mathbf{v}_{\rm{F}} (\mathbf{x}, \mathbf{y}, t)|_2 \leq \frac{C{d^{3/2}_x}(R+1)}{(1-T)^3} \, ,
    \end{align*}
    for any $R>0$ and $T\in(0,1)$.

    {\rm(ii)} 
    For any $\mathbf{y} \in [0, B]^{d_y}$ and $t\in [0, T]$ with $T\in(0,1)$, $\mathbf{v}_{\rm{F}}(\mathbf{x}, \mathbf{y}, t)$ is $d_x(1-T)^{-2}$-Lipschitz continuous with respect to $\mathbf{x}$, i.e., $$|\mathbf{v}_{\rm F}(\mathbf{x}_1, \mathbf{y}, t)-\mathbf{v}_{\rm F}(\mathbf{x}_2, \mathbf{y}, t)|_{\infty} \leq d_x(1-T)^{-2} |\mathbf{x}_1-\mathbf{x}_2|_2$$ for any $\mathbf{x}_1, \mathbf{x}_2 \in \mathbb{R}^{d_x}$, $t \in[0, T]$ and  $\mathbf{y} \in [0,B]^{d_{y}}$, while $\mathbf{F}_t(\mathbf{x}, \mathbf{y})$ is $\exp \{d_x(1-T)^{-2}\}$-Lipschitz continuous with respect to $\mathbf{x}$, i.e.,
    $$|\mathbf{F}_t(\mathbf{x}_1, \mathbf{y})-\mathbf{F}_t(\mathbf{x}_2, \mathbf{y})|_{2} \leq \exp \{d_x(1-T)^{-2} \} |\mathbf{x}_1-\mathbf{x}_2|_2$$ for any $\mathbf{x}_1, \mathbf{x}_2 \in \mathbb{R}^{d_x}$, $t \in[0, T]$ and  $\mathbf{y} \in [0,B]^{d_{y}}$.
\end{proposition}
We will establish the proof of Theorem \ref{thm:well posedness} through the following three steps.

{\bf Step 1.} For any given $\mathbf{y}$ and $\varepsilon\in(0,1)$, we show the existence of a diffusion process $(\bar{\mathbf{Z}}_t^{\mathbf{y}})_{t\in[0,1-\varepsilon]}$ determined by an Itô SDE, which can approximately transform the target conditional density $p_{x \mkern 2mu|\mkern 2mu y}(\mathbf{x} \mkern 2mu|\mkern 2mu \mathbf{y})$ into the density of standard Gaussian distribution $\mathcal{N}(\mathbf{0}, \mathbf{I}_{d_x})$, and then there exists a process $(\check{\mathbf{Z}}_t^{\mathbf{y}})_{t\in[\delta,1-\varepsilon]}$ determined by the associated ODE with any $\delta\in(0,1)$ satisfying $\delta<1-\varepsilon$, such that $\check{\mathbf{Z}}_t^{\mathbf{y}}$ shares the same marginal density with $\bar{\mathbf{Z}}_t^{\mathbf{y}}$ for any $t\in[\delta,1-\varepsilon]$. 

{\bf Step 2.} Under Assumption \ref{assump:bounded support}, we can extend the domain of the ODE involved in Step 1 to the interval $[\delta, 1]$ by supplementing its definition at $t=1$, ensuring an accurate transformation into the density of standard Gaussian distribution $\mathcal{N}(\mathbf{0}, \mathbf{I}_{d_x})$. 

{\bf Step 3.} Under Assumption \ref{assump:bounded support},
we can prove that the ODE involved in Step 2 has a unique solution. Hence, we can time reverse it to obtain the conditional Föllmer flow \eqref{def:Conditional Follmer Flow} over $[0, 1-\delta]$ and further extend it to $[0, 1)$ by letting $\delta \rightarrow 0$. This establishes the well-posedness of the conditional Föllmer flow and its ability to arbitrarily approach the target conditional density $p_{x \mkern 2mu|\mkern 2mu y}(\mathbf{x} \mkern 2mu|\mkern 2mu \mathbf{y})$ from the density of the standard Gaussian distribution $\mathcal{N}(\mathbf{0}, \mathbf{I}_{d_x})$.

\subsection{Step 1}\label{subsec:flow-density-step1}

For any given $\mathbf{y}$ and $\varepsilon \in(0,1)$, we consider a diffusion process $(\bar{\mathbf{Z}}^{\mathbf{y}}_t)_{t \in[0,1-\varepsilon]}$ defined by the following Itô SDE:
\begin{equation} \label{equa:forward SDE}
\mathrm{d} \bar{\mathbf{Z}}^{\mathbf{y}}_t=-\frac{1}{1-t}\bar{\mathbf{Z}}^{\mathbf{y}}_t \mathrm{~d} t+\sqrt{\frac{2}{1-t}} \mathrm{~d} {\mathbf{B}}_t\,, \quad \bar{\mathbf{Z}}^{\mathbf{y}}_0 \sim p_{x \mkern 2mu|\mkern2mu y}(\mathbf{x} \mkern 2mu|\mkern2mu \mathbf{y})\,, \quad t \in[0,1-\varepsilon] \, ,
\end{equation}
where ${\mathbf{B}}_t$ is a standard Brownian motion, and $p_{x \mkern 2mu|\mkern2mu y}(\mathbf{x} \mkern 2mu|\mkern2mu \mathbf{y})$ is the conditional density of $\bf{X}$ given $\mathbf{Y} = \bf{y}$.
The diffusion process defined in (\ref{equa:forward SDE}) has a unique strong solution on $[0,1-\varepsilon]$. The transition probability density of (\ref{equa:forward SDE}) from $\bar{\mathbf{Z}}^{\mathbf{y}}_0$ to $\bar{\mathbf{Z}}^{\mathbf{y}}_t$ is given by
$$
\bar{\mathbf{Z}}^{\mathbf{y}}_t \mkern2mu|\mkern2mu \bar{\mathbf{Z}}^{\mathbf{y}}_0=\mathbf{x} \sim \mathcal{N}\big((1-t) \mathbf{x}, t(2-t) \mkern2mu \mathbf{I}_{d_x}\big)\, , \quad t \in [0,1-\varepsilon] \, .
$$
Denote by $\bar{p}_t(\mathbf{x}  ; \mathbf{y})$ the marginal density of $\bar{\mathbf{Z}}^{\mathbf{y}}_t$ defined in \eqref{equa:forward SDE}. 

Let $f_t(\mathbf{x}\mkern 2mu|\mkern2mu\mathbf{y})$ be the conditional density of $t\mathbf{X}+\sqrt{1-t^2}\mathbf{W}$ given $\mathbf{Y}=\mathbf{y}$, where $\mathbf{W} \sim \mathcal{N}(\mathbf{0}, \mathbf{I}_{d_x})$ is independent of $(\mathbf{X}, \mathbf{Y})$. Write ${\mathbf{W}}_t = t \mathbf{X}+\sqrt{1-t^2} \mathbf{W}$. Then 
\begin{align} \label{equa:function form of ft}
    f_t(\mathbf{x} \mkern2mu|\mkern2mu \mathbf{y}) &=  \int p_{x, w_t \mkern2mu|\mkern2mu y}(\mathbf{u}, \mathbf{x}\mkern2mu|\mkern2mu \mathbf{y}) \, {\rm d} \mathbf{u} = \int p_{x\mkern2mu|\mkern2mu y}(\mathbf{u} \mkern2mu|\mkern2mu \mathbf{y}) p_{w_t \mkern2mu|\mkern2mu x, y}(\mathbf{x} \mkern2mu|\mkern2mu \mathbf{u}, \mathbf{y}) \, {\rm d} \mathbf{u} \\ \notag
    &= \int p_{x\mkern2mu|\mkern2mu y}(\mathbf{u} \mkern2mu|\mkern2mu \mathbf{y}) p_{w_t \mkern2mu|\mkern2mu x}(\mathbf{x} \mkern2mu|\mkern2mu \mathbf{u}) \, {\rm d} \mathbf{u} = \int C \cdot p_{x\mkern2mu|\mkern2mu y}(\mathbf{u} \mkern2mu|\mkern2mu \mathbf{y}) \exp \bigg\{\!-\frac{|\mathbf{x}-t \mathbf{u}|_2^2}{2(1-t^2)}\bigg\} \, {\rm d} \mathbf{u}\,,
\end{align}
where $C = (2 \pi)^{-d_x/2}(1-t^2)^{-d_x/2}$.
Hence, $\bar{p}_t(\mathbf{x}; \mathbf{y}) = f_{1-t}(\mathbf{x} \mkern 2mu|\mkern2mu \mathbf{y})$. Further, $\bar{p}_t(\mathbf{x} ; \mathbf{y})$ satisfies the Fokker-Planck-Kolmogorov equation in an Eulerian framework over $[\delta, 1-\varepsilon]$ with any $\delta > 0$ satisfying $\delta < 1-\varepsilon$ \citep{bogachev2022fokker}, which means on $\mathbb{R}^{d_x} \times [0,B]^{d_{y}} \times [\delta,1-\varepsilon]$, 
$$
\partial_t \bar{p}_t(\mathbf{x} ; \mathbf{y})=\nabla_{\mathbf{x}} \cdot\{\bar{p}_t(\mathbf{x} ; \mathbf{y}) \mkern2mu \mathbf{v}_{\rm{F}}(\mathbf{x}, \mathbf{y}, 1-t)\} \, , \quad \bar{p}_{\delta}(\mathbf{x} ; \mathbf{y})=f_{1-\delta}(\mathbf{x} \mkern 2mu|\mkern2mu \mathbf{y})
$$
in the sense that $\bar{p}_{t}(\mathbf{x} ; \mathbf{y})$ is continuous in $t$ under the weak topology, where the velocity field is defined by
$$
\mathbf{v}_{\rm{F}}(\mathbf{x}, \mathbf{y}, 1-t):=\frac{\mathbf{x}+\mathbf{s}(\mathbf{x}, \mathbf{y}, 1-t)}{1-t}, \quad t \in [\delta,1-\varepsilon]
$$
with
$$
\mathbf{s}(\mathbf{x}, \mathbf{y}, t) := \nabla_{\mathbf{x}} \log \bigg[\int C \cdot p_{x\mkern2mu|\mkern2mu y}(\mathbf{u} \mkern2mu|\mkern2mu \mathbf{y}) \exp \bigg\{\!-\frac{|\mathbf{x}-t \mathbf{u}|_2^2}{2(1-t^2)}\bigg\} \, {\rm d} \mathbf{u}\bigg]=\nabla_{\mathbf{x}} \log f_t(\mathbf{x} \mkern2mu|\mkern2mu \mathbf{y}) \, ,
$$
for any $ t \in [\varepsilon, 1-\delta]$, where $C = (2 \pi)^{-d_x/2}(1-t^2)^{-d_x/2}$. 
Due to the classical Cauchy-Lipschitz theory \citep{ambrosio2014continuity} with a Lipschitz velocity field or the well-established Ambrosio-DiPerna-Lions theory with lower Sobolev regularity assumptions on the velocity fields \citep{ambrosio2004transport, diperna1989ordinary}, we can define a flow $(\check{\mathbf{Z}}^{\mathbf{y}}_t)_{t \in[\delta,1-\varepsilon]}$ in a Lagrangian formulation via the following ODE system
\begin{equation} \label{equa:forward ODE}
\mathrm{d} \check{\mathbf{Z}}^{\mathbf{y}}_t=-\mathbf{v}_{\rm{F}}  (\check{\mathbf{Z}}^{\mathbf{y}}_t, \mathbf{y}, 1-t)\mkern2mu \mathrm{d} t\,, \quad \check{\mathbf{Z}}^{\mathbf{y}}_{\delta} \sim f_{1-\delta}(\mathbf{x} \mkern 2mu|\mkern2mu \mathbf{y})\,, \quad t \in[\delta,1-\varepsilon] \, .
\end{equation}

Denote by $F(\mathbb{R}^{d_x} ; \mathbb{R}^{d_x})$ the set of all functions $f: \mathbb{R}^{d_x} \rightarrow \mathbb{R}^{d_x}$, by $W_{\rm loc}^{1, \infty}(\mathbb{R}^{d_x} ; \mathbb{R}^{d_x})$ the locally bounded Lipschitzian functions $g: \mathbb{R}^{d_x} \rightarrow \mathbb{R}^{d_x}$, and by $L^1([a, b] ; V)$ the set of all Bochner integrable functions $h: [a, b] \rightarrow V$ with $V$ being a Banach space.  Based on Lemma \ref{lemma:check and bar} below, we can conclude that $\check{\mathbf{Z}}^{\mathbf{y}}_t$ in \eqref{equa:forward ODE} and $\bar{\mathbf{Z}}^{\mathbf{y}}_t$ in \eqref{equa:forward SDE} share the same marginal density over $[\delta, 1-\varepsilon]$, which can approximate the density of standard Gaussian distribution $\mathcal{N}(\mathbf{0},\mathbf{I}_{d_x})$ as $t \rightarrow 1$.

\begin{lemma} \label{lemma:check and bar}
For any fixed $\mathbf{y}$, treat the velocity field $\mathbf{v}_{\rm{F}}(\mathbf{x}, \mathbf{y}, t)$ as a map from $[\varepsilon, 1-\delta]$ to $F(\mathbb{R}^{d_x} ; \mathbb{R}^{d_x})$ and assume that $\mathbf{v}_{\rm{F}}(\mathbf{x}, \mathbf{y}, t)$ satisfies
$$\mathbf{v}_{\rm{F}}(\mathbf{x}, \mathbf{y}, t) \in L^1 \big([\varepsilon, 1-\delta] ; W_{\rm loc}^{1, \infty}(\mathbb{R}^{d_x} ; \mathbb{R}^{d_x})\big)\, , \quad \frac{|\mathbf{v}_{\rm{F}}(\mathbf{x}, \mathbf{y}, t)|_2}{1+|\mathbf{x}|_2}  \in L^1\big([\varepsilon, 1-\delta]; L^{\infty}(\mathbb{R}^{d_x})\big) \, .$$ Then,  $\check{\mathbf{Z}}^{\mathbf{y}}_t$ in \eqref{equa:forward ODE} also follows the marginal probability density $\bar{p}_t(\mathbf{x} ; \mathbf{y}) = f_{1-t}(\mathbf{x} \mkern 2mu|\mkern2mu \mathbf{y})$ on $[\delta, 1-\varepsilon]$.  Moreover, the distribution with density $\bar{p}_{1-\varepsilon}(\mathbf{x} ; \mathbf{y})$ converges to $\mathcal{N}(\mathbf{0},\mathbf{I}_{d_x})$ in the Wasserstein-2 distance as $\varepsilon\rightarrow0$. %
\end{lemma}

Lemma \ref{lemma:check and bar} can be regarded as an application of Proposition 3.5 in \cite{dai2023lipschitz}, so we omit the proof here. Under Assumption \ref{assump:bounded support},
the requirements of $\mathbf{v}_{\rm{F}}(\mathbf{x}, \mathbf{y}, t)$ in Lemma \ref{lemma:check and bar} are satisfied automatically. More specifically, as shown in Proposition \ref{prop:properties of vF}(i) and \ref{prop:properties of vF}(ii), $\mathbf{v}_{\rm{F}}(\mathbf{x}, \mathbf{y}, t)$ attains local boundedness and local Lipschitz constant uniformly over $[\varepsilon, 1-\delta]$ on any compact set $\mathcal{K} \subseteq \mathbb{R}^{d_x}$. Further, by Assumption \ref{assump:bounded support} and \eqref{equa:alter condi expec 2}, $|\mathbf{v}_{\rm{F}}(\mathbf{x}, \mathbf{y}, t)|_2/(1+|\mathbf{x}|_2) \leq (d_x^{1/2}+1)/\delta$ for any $\mathbf{x} \in \mathbb{R}^{d_x}$ and $t \in [\varepsilon, 1-\delta]$, thus achieving uniform $L^{\infty}(\mathbb{R}^{d_x})$ norm over $[\varepsilon, 1-\delta]$.

\subsection{Step 2}
In Step 1, we have defined 
$
\mathbf{v}_{\rm{F}}(\mathbf{x}, \mathbf{y}, t):=t^{-1}\{\mathbf{x}+\mathbf{s}(\mathbf{x}, \mathbf{y}, t)\}
$ 
for any $t\in[\varepsilon,1-\delta]$. When $t\in(0,\varepsilon)$, we also define $\mathbf{v}_{\rm{F}}(\mathbf{x}, \mathbf{y}, t)$ in the same manner. Based on Lemma \ref{lemma:vFt at 0} below, we can define $\mathbf{v}_{\rm{F}}(\mathbf{x}, \mathbf{y}, 0)=\mathbb{E}(\mathbf{X} \mkern 2mu|\mkern2mu \mathbf{Y}=\mathbf{y})$.

\begin{lemma} \label{lemma:vFt at 0}
    Let Assumption {\rm \ref{assump:bounded support}} hold. Then 
    $$
\lim _{t \rightarrow 0^{+}} \mathbf{v}_{\rm{F}}(\mathbf{x}, \mathbf{y}, t)=\lim _{t \rightarrow 0^{+}} \partial_t \mathbf{s}(\mathbf{x}, \mathbf{y}, t)=\mathbb{E}(\mathbf{X} \mkern 2mu|\mkern2mu \mathbf{Y}=\mathbf{y}) \, .
$$
\end{lemma}

Lemma \ref{lemma:vFt at 0} can be regarded as a natural extension of Lemma A.1 in \cite{dai2023lipschitz}, so we omit the proof here. Now the process  $(\check{\mathbf{Z}}^{\mathbf{y}}_t)_{t \in[\delta,1)}$ can be extended to time $t = 1$ such that $\check{\mathbf{Z}}^{\mathbf{y}}_1 \sim \mathcal{N}({\bf0}, \mathbf{I}_{d_x})$, which solves the initial value problem  below
\begin{equation} \label{equa:forward extended ODE}
\mathrm{d} \check{\mathbf{Z}}^{\mathbf{y}}_t=-\mathbf{v}_{\rm{F}}(\check{\mathbf{Z}}^{\mathbf{y}}_t, \mathbf{y}, 1-t) \mathrm{d} t\,, \quad \check{\mathbf{Z}}^{\mathbf{y}}_{\delta} \sim f_{1-\delta}(\mathbf{x} \mkern 2mu|\mkern2mu \mathbf{y})\,, \quad t \in[\delta,1] \, ,
\end{equation}
and the marginal density of $\check{\mathbf{Z}}^{\mathbf{y}}_t$ is $\bar{p}_t(\mathbf{x} ; \mathbf{y}) = f_{1-t}(\mathbf{x} \mkern 2mu|\mkern2mu \mathbf{y})$ for any $t \in [\delta, 1]$. 

\subsection{Step 3}
To demonstrate that (\ref{equa:forward extended ODE}) has a unique solution, we also need the Lipschitz property of $\mathbf{v}_{\rm{F}}$, which is provided in Proposition \ref{prop:properties of vF}(ii) under Assumption \ref{assump:bounded support}.
Now, a standard time reversal argument of (\ref{equa:forward extended ODE}) would yield the conditional Föllmer flow \eqref{equa:CCF ODE} over $[0, 1-\delta]$ as given in Definition \ref{def:Conditional Follmer Flow}, i.e.
$$
\mathrm{d}\mathbf{Z}_t^{\mathbf{y}}=\mathbf{v}_{\rm F}(\mathbf{Z}_t^{\mathbf{y}}, \mathbf{y}, t) \, \mathrm{d}t \, , \quad \mathbf{Z}_0^{\mathbf{y}} \sim \mathcal{N}(\mathbf{0}, \mathbf{I}_{d_x})\, , \quad t \in [0, 1-\delta] \, .
$$
Thus, \eqref{equa:CCF ODE} is well-defined on $[0, 1-\delta]$ for any $0<\delta<1$, while letting $\delta \rightarrow 0$ yields its well-posedness over $[0, 1)$. Moreover, $\mathbf{Z}_t^{\mathbf{y}} \sim f_{t}(\mathbf{x} \mkern 2mu|\mkern2mu \mathbf{y})$. 
Recall that $\mathbf{W}_t = t\mathbf{X}+\sqrt{1-t^2}\mathbf{W}$ and $\mathbf{W} \sim \mathcal{N}(\mathbf{0}, \mathbf{I}_{d_x})$ is independent of $(\mathbf{X}, \mathbf{Y})$. Given $\mathbf{Y}=\mathbf{y}$, the conditional densities of $\mathbf{W}_{1-\delta}$ and $\mathbf{X}$ are, respectively, $f_{1-\delta}(\mathbf{x} \mkern 2mu|\mkern2mu \mathbf{y})$ and $p_{x \mkern2mu|\mkern2mu y}(\mathbf{x} \mkern2mu|\mkern2mu \mathbf{y})$. By the definition of Wasserstein-2 distance, we have
\begin{align*}
W^2_2\big( \mkern0.5mu f_{1-\delta}(\mathbf{x} \mkern 2mu|\mkern2mu \mathbf{y}), \mkern2mu p_{x \mkern2mu|\mkern2mu y}(\mathbf{x} \mkern 2mu|\mkern2mu \mathbf{y})\big) & \leq \mathbb{E}(|\mathbf{W}_{1-\delta}-\mathbf{X}|_2^2\mkern2mu|\mkern2mu \mathbf{Y}=\mathbf{y}) \\
&= \mathbb{E}\big\{|\delta\mathbf{X}-\delta^{1/2}(2-\delta)^{1/2}\mathbf{W}|_2^2\mkern2mu|\mkern2mu \mathbf{Y}=\mathbf{y} \big\} \, .
\end{align*}
By Assumption \ref{assump:bounded support}, $|\mathbf{X}|_{\infty} \leq 1$. It holds that
$$
W^2_2\big( \mkern0.5mu f_{1-\delta}(\mathbf{x} \mkern 2mu|\mkern2mu \mathbf{y}), \mkern2mu p_{x \mkern2mu|\mkern2mu y}(\mathbf{x} \mkern 2mu|\mkern2mu \mathbf{y})\big) \leq 4 {d_x}{\delta} \rightarrow 0
$$ 
as $\delta \rightarrow 0$. We then complete the proof of Theorem \ref{thm:well posedness}. $\hfill\Box$

\section{Proof of Proposition \ref{prop:training object}} \label{sec:training object}
The idea of expressing the velocity field $\mathbf{v}_{\rm{F}}$ as a conditional expectation and as the minimizer of a quadratic objective is inspired by \cite{albergo2023stochastic}. While their proof relies on characteristic functions, we provide a more concise proof by directly using the structure of conditional Föllmer flow.
\subsection{Proof of Proposition \ref{prop:training object}(i)}\label{subsec:pf1i}
 Since $\mathbf{W}\sim \mathcal{N}(\mathbf{0}, \mathbf{I}_{d_x})$ is independent of $(\mathbf{X}, \mathbf{Y})$ and $\mathbf{v}_{\rm{F}} (\mathbf{x}, \mathbf{y}, 0) = \mathbb{E}(\mathbf{X} \mkern2mu|\mkern2mu\mathbf{Y}=\mathbf{y})$, then $\mathbf{v}_{\rm{F}} (\mathbf{x}, \mathbf{y}, 0) = \mathbb{E}(\mathbf{X} \mkern2mu|\mkern2mu\mathbf{Y}=\mathbf{y}) = \mathbb{E}(\mathbf{X} \mkern2mu|\mkern2mu \mathbf{W}=\mathbf{x}, \mathbf{Y}=\mathbf{y})$, which implies \eqref{equa: condi expec vF} holds for $t=0$. Write ${\mathbf{W}}_t = t \mathbf{X}+\sqrt{1-t^2} \mathbf{W}$. For any $t \in (0, T]$, we have
      \begin{equation} \label{euqa:alter condi exp}
\begin{aligned}
 &\mathbb{E}\bigg(\mathbf{X}-\frac{t}{\sqrt{1-t^2}}\mathbf{W} \mkern2mu\bigg|\mkern2mu \mathbf{W}_t=\mathbf{x}, \mathbf{Y}=\mathbf{y}\bigg) \\
 &~~~~~=\mathbb{E}\bigg\{\frac{1}{1-t^2}\mathbf{X}-\frac{t}{1-t^2}(t \mathbf{X}+\sqrt{1-t^2}\mathbf{W}) \bigg|\mkern2mu \mathbf{W}_t=\mathbf{x}, \mathbf{Y}=\mathbf{y}\bigg\} \\
 &~~~~~=\frac{1}{1-t^2} \mathbb{E}(\mathbf{X} \mkern2mu|\mkern2mu \mathbf{W}_t=\mathbf{x}, \mathbf{Y}=\mathbf{y})-\frac{t}{1-t^2} \mathbf{x} \, .
\end{aligned}
      \end{equation}
Recall that $f_t(\mathbf{x} \mkern2mu|\mkern2mu \mathbf{y})$ is the conditional density of $\mathbf{W}_t$ given $\mathbf{Y}=\mathbf{y}$. %
Write $C = (2 \pi)^{-d_x/2}(1-t^2)^{-d_x/2}$.  For any $t\in(0,T]$, by \eqref{equa:function form of ft}, it holds that 
\begin{align}\label{eq:condexp1}
&\mathbb{E}(\mathbf{X} \mkern2mu|\mkern2mu \mathbf{W}_t=\mathbf{x}, \mathbf{Y}=\mathbf{y})=\int\mathbf{u}\cdot\frac{p_{x,w_t,y}(\mathbf{u},\mathbf{x},\mathbf{y})}{p_{w_t,y}(\mathbf{x},\mathbf{y})}\,{\rm d}\mathbf{u}=\int \mathbf{u}\cdot\frac{p_{w_t\mkern2mu|\mkern2mu x, y}(\mathbf{x}\mkern2mu|\mkern2mu \mathbf{u},\mathbf{y})p_{x,y}(\mathbf{u},\mathbf{y})}{p_{w_t,y}(\mathbf{x},\mathbf{y})}\,{\rm d}\mathbf{u} \notag\\
 &~~~~~=\int C \mathbf{u} \cdot \frac{  p_{x \mkern2mu | \mkern2mu y}(\mathbf{u}\mkern2mu|\mkern2mu \mathbf{y})}{f_t(\mathbf{x} \mkern 2mu|\mkern2mu \mathbf{y})} \exp \bigg\{\!-\frac{|\mathbf{x}-t \mathbf{u}|_2^2}{2(1-t^2)}\bigg\}\mkern2mu \mathrm{d} \mathbf{u}\notag \\
 &~~~~~= \frac{1-t^2}{t} \int C\cdot \frac{  p_{x \mkern2mu | \mkern2mu y}(\mathbf{u}\mkern2mu| \mkern2mu \mathbf{y})}{f_t(\mathbf{x} \mkern 2mu|\mkern2mu \mathbf{y})} \bigg(\frac{t \mathbf{u}-\mathbf{x}}{1-t^2}+\frac{\mathbf{x}}{1-t^2}\bigg)\exp \bigg\{\!-\frac{|\mathbf{x}-t \mathbf{u}|_2^2}{2(1-t^2)}\bigg\}\mkern2mu \mathrm{d} \mathbf{u} \\
 &~~~~~ = \frac{1-t^2}{t} \int C\cdot \frac{ p_{x \mkern2mu | \mkern2mu y}  (\mathbf{u}\mkern2mu| \mkern2mu \mathbf{y})}{f_t(\mathbf{x} \mkern 2mu|\mkern2mu \mathbf{y})} \nabla_{\mathbf{x}} \exp \bigg\{\!-\frac{|\mathbf{x}-t \mathbf{u}|_2^2}{2(1-t^2)}\bigg\}\mkern2mu \mathrm{d} \mathbf{u}+\frac{1}{t} \mathbf{x} \notag\\
 &~~~~~ =\frac{1-t^2}{t}\frac{\nabla_{\mathbf{x}} f_t(\mathbf{x} \mkern 2mu|\mkern2mu \mathbf{y})}{f_t(\mathbf{x} \mkern 2mu|\mkern2mu \mathbf{y})}+\frac{1}{t} \mathbf{x} = \frac{1-t^2}{t} \nabla_{\mathbf{x}} \log f_t(\mathbf{x} \mkern 2mu|\mkern2mu \mathbf{y})+\frac{1}{t} \mathbf{x}\,,\notag
\end{align}
which implies
\begin{align*}
    \mathbb{E}\bigg(\mathbf{X}-\frac{t}{\sqrt{1-t^2}}\mathbf{W} \mkern2mu\bigg|\mkern2mu \mathbf{W}_t=\mathbf{x}, \mathbf{Y}=\mathbf{y}\bigg) = \frac{\nabla_{\mathbf{x}} \log f_t(\mathbf{x} \mkern 2mu|\mkern2mu \mathbf{y})}{t} +\frac{1}{t} \mathbf{x} = \mathbf{v}_{\rm{F}}(\mathbf{x}, \mathbf{y}, t)
\end{align*}
for any $t\in(0,T]$. We complete the proof of Proposition \ref{prop:training object}(i). $\hfill\Box$

\subsection{Proof of Proposition \ref{prop:training object}(ii)}

Recall that ${\mathbf{W}}_t = t \mathbf{X}+\sqrt{1-t^2} \mathbf{W}$.
For any vector-valued function $\mathbf{v}(\mathbf{x}, \mathbf{y}, t): \mathbb{R}^{d_x} \times \mathbb{R}^{d_{y}} \times[0, T] \rightarrow \mathbb{R}^{d_x}$, we have 
    \begin{align} \label{equa:expanding L(v)}
            &\mathbb{E}\left\{\bigg|\mathbf{X}-\frac{t}{\sqrt{1-t^2}}\mathbf{W}-\mathbf{v}  (  \mathbf{W}_t, \mathbf{Y}, t)\bigg|_2^2\right\} \notag \\
            &~~~~~=\mathbb{E}\left\{\bigg|\mathbf{X}-\frac{t}{\sqrt{1-t^2}}\mathbf{W}-\mathbf{v}_{\rm{F}}  (  \mathbf{W}_t, \mathbf{Y}, t)+\mathbf{v}_{\rm{F}}  (  \mathbf{W}_t, \mathbf{Y}, t)-\mathbf{v}  (  \mathbf{W}_t, \mathbf{Y}, t)\bigg|_2^2\right\} \notag \\
            &~~~~~=\mathbb{E}\left\{\bigg|\mathbf{X}-\frac{t}{\sqrt{1-t^2}}\mathbf{W}-\mathbf{v}_{\rm{F}}  (  \mathbf{W}_t, \mathbf{Y}, t)\bigg|_2^2\right\}+ \mathbb{E}\big\{|\mathbf{v}_{\rm F}(\mathbf{W}_t, \mathbf{Y}, t) - \mathbf{v}(\mathbf{W}_t, \mathbf{Y}, t)|_2^2\big\} \notag \\
            &~~~~~~~~+2\mathbb{E}\bigg\{\bigg\langle\mathbf{v}_{\rm{F}}  (  \mathbf{W}_t, \mathbf{Y}, t)-\mathbf{v}  (  \mathbf{W}_t, \mathbf{Y}, t), \mathbf{X}-\frac{t}{\sqrt{1-t^2}}\mathbf{W}-\mathbf{v}_{\rm{F}}  (  \mathbf{W}_t, \mathbf{Y}, t)\bigg\rangle\bigg\}
    \end{align}
for any $t\in[0,T]$. 
Due to
\begin{align} \label{equa:condiexpeczero}
& \mathbb{E}\bigg\{\bigg\langle\mathbf{v}_{\rm{F}}(\mathbf{W}_t, \mathbf{Y}, t)-\mathbf{v}(\mathbf{W}_t, \mathbf{Y}, t), \mathbf{X}-\frac{t}{\sqrt{1-t^2}}\mathbf{W}-\mathbf{v}_{\rm{F}}(\mathbf{W}_t, \mathbf{Y}, t)\bigg\rangle\bigg\} \notag \\ 
&~~~~~= \mathbb{E}\bigg[\mathbb{E}\bigg\{\bigg\langle\mathbf{v}_{\rm{F}}(  \mathbf{W}_t, \mathbf{Y}, t)-\mathbf{v}(\mathbf{W}_t, \mathbf{Y}, t), \mathbf{X}-\frac{t}{\sqrt{1-t^2}}\mathbf{W}-\mathbf{v}_{\rm{F}}(\mathbf{W}_t, \mathbf{Y}, t)\bigg\rangle \bigg| \mkern2mu \mathbf{W}_t, \mathbf{Y}\bigg\}\bigg] \notag \\ 
&~~~~~=  \mathbb{E}\bigg\{\bigg\langle\mathbf{v}_{\rm{F}}(\mathbf{W}_t, \mathbf{Y}, t)-\mathbf{v}(\mathbf{W}_t, \mathbf{Y}, t), \mathbb{E}\bigg(\mathbf{X}-\frac{t}{\sqrt{1-t^2}}\mathbf{W} \mkern2mu \bigg| \mkern2mu \mathbf{W}_t, \mathbf{Y}\bigg)-\mathbf{v}_{\rm{F}}(\mathbf{W}_t, \mathbf{Y}, t)\bigg\rangle\bigg\} \notag \\ 
&~~~~~= \mathbb{E}\big\{\langle\mathbf{v}_{\rm{F}}(  \mathbf{W}_t, \mathbf{Y}, t)-\mathbf{v}(\mathbf{W}_t, \mathbf{Y}, t),  {\bf 0}\rangle\big\}=0
\end{align}
for any $t\in[0,T]$, 
by \eqref{equa:expanding L(v)}, we have 
        \begin{align*}
            &\mathbb{E}\left\{\bigg|\mathbf{X}-\frac{t}{\sqrt{1-t^2}}\mathbf{W}-\mathbf{v}  (  \mathbf{W}_t, \mathbf{Y}, t)\bigg|_2^2\right\} \geq\mathbb{E}\left\{\bigg|\mathbf{X}-\frac{t}{\sqrt{1-t^2}}\mathbf{W}-\mathbf{v}_{\rm{F}}  (  \mathbf{W}_t, \mathbf{Y}, t)\bigg|_2^2\right\}
        \end{align*}
for any $t\in[0,T]$, where equality holds if and only if $\mathbf{v}=\mathbf{v}_{\rm F}$. Thus, we complete the proof of Proposition \ref{prop:training object}(ii). $\hfill\Box$

\section{Proof of Proposition \ref{prop:properties of vF}} \label{sec:properties of vF}

To prove Proposition \ref{prop:properties of vF}, we need the following Gr\"{o}nwall's inequality,
whose proof can be found in Lemma 1.1 of \cite{bainov1992integral}.

\begin{lemma} [Gr\"{o}nwall's inequality] \label{lemma:gronwall}
 Let $\beta(t)$, $\lambda(t)$ and $v(t)$ be real-valued continuous functions defined on $[a, b]$ with $a < b$. If $v(t)$ is differentiable over the interval $(a, b)$ and satisfies
$
v^{\prime}(t) \leq \beta(t) v(t) + \lambda(t)$ for  any $t \in (a, b)$,
then %
$$
v(t) \leq v(a) \exp \left\{\int_a^t \beta(s) \,\mathrm{d} s\right\} + \int_a^t \lambda(s) \exp\left\{ \int_s^t \beta(\tau) \, \mathrm{d} \tau \right\} \, \mathrm{d} s 
$$
for any $t\in(a,b)$.
\end{lemma}

\subsection{Proof of Proposition \ref{prop:properties of vF}(i)} \label{subsec:proof of vF i}
By Proposition \ref{prop:training object}(i) and \eqref{euqa:alter condi exp}, it holds that 
\begin{equation} \label{equa:alter condi expec 2}
    \mathbf{v}_{\rm{F}}(\mathbf{x}, \mathbf{y}, t) = \frac{1}{1-t^2} \mathbb{E}(\mathbf{X} \mkern2mu|\mkern2mu \mathbf{W}_t=\mathbf{x}, \mathbf{Y}=\mathbf{y})-\frac{t}{1-t^2} \mathbf{x} \, .
\end{equation} 
By Assumption \ref{assump:bounded support}, we have $|\mathbf{X}|_{\infty} \leq 1$, which implies
$$
\sup_{t \in [0,T]}\sup_{\mathbf{x} \in [-R, R]^{d_x}}\sup_{\mathbf{y} \in [0, B]^{d_{y}}}|\mathbf{v}_{\rm F}(\mathbf{x}, \mathbf{y}, t)|_{\infty} \leq \frac{1+TR}{1-T^2}\,.
$$

Recall that $f_t(\mathbf{x} \mkern 2mu|\mkern2mu \mathbf{y})$ is the conditional density of $\mathbf{W}_t$ given $\mathbf{Y}=\mathbf{y}$ and $\mathbf{W}_t = t \mathbf{X} + \sqrt{1-t^2}\mathbf{W}$.
Write
$$
\phi^{\mathbf{y}}_t(\mathbf{x}):=\int p_{x\mkern2mu|\mkern2mu y}(\mathbf{u} \mkern2mu|\mkern2mu \mathbf{y}) \exp \bigg\{\! -\frac{|\mathbf{x}-t \mathbf{u}|_2^2}{2(1-t^2)}\bigg\} \, {\rm d} \mathbf{u} \, .
$$
As we have shown in \eqref{equa:function form of ft} in Section \ref{subsec:flow-density-step1}, $\phi^{\mathbf{y}}_t(\mathbf{x})=C^{-1} f_t(\mathbf{x} \mkern 2mu|\mkern2mu \mathbf{y})$ with $C = (2 \pi)^{-d_x/2}(1-t^2)^{-d_x/2}$. 
Then $\nabla_{\mathbf{x}} \log \phi^{\mathbf{y}}_t(\mathbf{x})=\nabla_{\mathbf{x}} \log f_t(\mathbf{x} \mkern 2mu|\mkern2mu \mathbf{y})$. By the definition of $\mathbf{v}_{\rm{F}}(\mathbf{x}, \mathbf{y}, t)$ given in Definition \ref{def:Conditional Follmer Flow} and \eqref{equa:alter condi expec 2}, we have 
$$
\nabla_{\mathbf{x}} \log f_t(\mathbf{x} \mkern 2mu|\mkern2mu \mathbf{y}) = \frac{t}{1-t^2} \mathbb{E}(\mathbf{X} \mkern 2mu|\mkern2mu \mathbf{W}_t=\mathbf{x}, \mathbf{Y}=\mathbf{y}) - \frac{1}{1-t^2} \mathbf{x} \, .
$$
Furthermore, it holds that
        \begin{align}\label{equa:partial_t v*}
            \partial_t \mathbf{v}_{\rm{F}} (\mathbf{x}, \mathbf{y}, t) &= -\frac{\nabla_{\mathbf{x}} \log f_t(\mathbf{x} \mkern 2mu|\mkern2mu \mathbf{y})}{t^2} + \frac{\partial_t \nabla_{\mathbf{x}} \log f_t(\mathbf{x} \mkern 2mu|\mkern2mu \mathbf{y})}{t} - \frac{\mathbf{x}}{t^2} \\
            &=\frac{\mathbf{x}}{1-t^2} - \frac{\mathbb{E}(\mathbf{X} \mkern 2mu|\mkern2mu \mathbf{W}_t=\mathbf{x}, \mathbf{Y}=\mathbf{y})}{t(1-t^2)} + \frac{1}{t}\bigg[\frac{\partial_t \nabla_{\mathbf{x}} \phi^{\mathbf{y}}_t(\mathbf{x})}{\phi^{\mathbf{y}}_t(\mathbf{x})}-\frac{\partial_t \phi^{\mathbf{y}}_t(\mathbf{x}) \nabla_{\mathbf{x}} \phi^{\mathbf{y}}_t(\mathbf{x})}{\{\phi^{\mathbf{y}}_t(\mathbf{x})\}^2}\bigg]\,. \notag
        \end{align}
    Notice that 
    \begin{equation*} 
        \begin{aligned}
            \frac{\partial_t \nabla_{\mathbf{x}} \phi^{\mathbf{y}}_t(\mathbf{x})}{\phi^{\mathbf{y}}_t(\mathbf{x})} =&\, \int \frac{(1+t^2)\mathbf{u}-2t\mathbf{x}}{(1-t^2)^2} \frac{p_{x\mkern2mu|\mkern2mu y}(\mathbf{u} \mkern2mu|\mkern2mu \mathbf{y})}{\phi^{\mathbf{y}}_t(\mathbf{x})}\exp \bigg\{\!  -\frac{|\mathbf{x}-t \mathbf{u}|_2^2}{2(1-t^2)}\bigg\}  \mkern2mu \mathrm{d} \mathbf{u} \\
            &+\int \frac{(\mathbf{u}^{\rm{T}}\mathbf{x}-t|\mathbf{u}|_2^2)(t\mathbf{u}-\mathbf{x})}{(1-t^2)^2}\frac{p_{x\mkern2mu|\mkern2mu y}(\mathbf{u} \mkern2mu|\mkern2mu \mathbf{y})}{\phi^{\mathbf{y}}_t(\mathbf{x})} \exp \bigg\{\!  -\frac{|\mathbf{x}-t \mathbf{u}|_2^2}{2(1-t^2)}\bigg\}  \mkern2mu \mathrm{d} \mathbf{u} \\
            &+\int \frac{t|\mathbf{x}-t\mathbf{u}|_2^2(\mathbf{x}-t\mathbf{u})}{(1-t^2)^3} \frac{p_{x\mkern2mu|\mkern2mu y}(\mathbf{u} \mkern2mu|\mkern2mu \mathbf{y})}{\phi^{\mathbf{y}}_t(\mathbf{x})}\exp \bigg\{\!  -\frac{|\mathbf{x}-t \mathbf{u}|_2^2}{2(1-t^2)}\bigg\} \mkern2mu \mathrm{d} \mathbf{u} \,.
        \end{aligned}
    \end{equation*}
As we have shown in \eqref{eq:condexp1} in Section \ref{subsec:pf1i}, the conditional density of $\mathbf{X}$ given $(\mathbf{W}_t,\mathbf{Y})=(\mathbf{x},\mathbf{y})$ is given by
\begin{equation}\label{eq:cond11}
p_{x\mkern2mu|\mkern2mu w_t, y}(\mathbf{u}\mkern2mu|\mkern2mu \mathbf{x},\mathbf{y})=C \cdot \frac{  p_{x \mkern2mu | \mkern2mu y}(\mathbf{u}\mkern2mu|\mkern2mu \mathbf{y})}{f_t(\mathbf{x} \mkern 2mu|\mkern2mu \mathbf{y})} \exp \bigg\{\!-\frac{|\mathbf{x}-t \mathbf{u}|_2^2}{2(1-t^2)}\bigg\}
\end{equation}
with $C = (2 \pi)^{-d_x/2}(1-t^2)^{-d_x/2}$. Due to $\phi^{\mathbf{y}}_t(\mathbf{x})=C^{-1} f_t(\mathbf{x} \mkern 2mu|\mkern2mu \mathbf{y})$, then
\begin{align}
\frac{\partial_t \nabla_{\mathbf{x}} \phi^{\mathbf{y}}_t(\mathbf{x})}{\phi^{\mathbf{y}}_t(\mathbf{x})} %
            &=\frac{-2t\mathbf{x}}{(1-t^2)^2} + \frac{1+t^2}{(1-t^2)^2} \mathbb{E}(\mathbf{X} \mkern 2mu|\mkern2mu \mathbf{W}_t = \mathbf{x}, \mathbf{Y} = \mathbf{y})+ \frac{t|\mathbf{x}|_2^2\mathbf{x}}{(1-t^2)^3}\notag\\
            &~~~-\frac{t^2}{(1-t^2)^3}\mathbb{E}(|\mathbf{X}|_2^2{\mathbf{X}} \mkern2mu \big| \mkern2mu \mathbf{W}_t = \mathbf{x}, \mathbf{Y} = \mathbf{y})+ \frac{t(1+t^2)}{(1-t^2)^3}\mathbb{E}({\mathbf{X}}{\mathbf{X}}^{\rm{T}} \mkern2mu \big| \mkern2mu \mathbf{W}_t=\mathbf{x}, \mathbf{Y}=\mathbf{y})\mathbf{x}\notag\\
            &~~~- \frac{1+t^2}{(1-t^2)^3}\mathbb{E}(\mathbf{X}^{\rm{T}}\mathbf{x} \mkern2mu \big| \mkern2mu \mathbf{W}_t=\mathbf{x}, \mathbf{Y}=\mathbf{y})\mathbf{x} + \frac{t}{(1-t^2)^3}\mathbb{E}(|\mathbf{X}|_2^2 \mkern2mu \big| \mkern2mu \mathbf{W}_t=\mathbf{x}, \mathbf{Y}=\mathbf{y})\mathbf{x}\notag\\
            &~~~- \frac{t^2}{(1-t^2)^3}\mathbb{E}(\mathbf{X} \mkern 2mu|\mkern2mu \mathbf{W}_t=\mathbf{x}, \mathbf{Y}=\mathbf{y})|\mathbf{x}|_2^2 \, .\label{equa:(partial_t nabla phi_t)/(phi_t)}
\end{align}
Analogously, we also have
    \begin{align}
    \frac{\partial_t \phi_t^{\mathbf{y}}(\mathbf{x})}{\phi_t^{\mathbf{y}}(\mathbf{x})}&=\frac{1}{\phi^{\mathbf{y}}_t(\mathbf{x})} \int \frac{(\mathbf{u}^{\rm{T}}\mathbf{x}-t|\mathbf{u}|_2^2)}{1-t^2} \exp \bigg\{\!  -\frac{|\mathbf{x}-t \mathbf{u}|_2^2}{2(1-t^2)}\bigg\} p_{x\mkern2mu|\mkern2mu y}(\mathbf{u} \mkern2mu|\mkern2mu \mathbf{y})\,\mathrm{d} \mathbf{u} \notag \\
    &~~~-\frac{1}{\phi^{\mathbf{y}}_t(\mathbf{x})} \int \frac{t|\mathbf{x}-t\mathbf{u}|_2^2}{(1-t^2)^2}\exp \bigg\{\!  -\frac{|\mathbf{x}-t \mathbf{u}|_2^2}{2(1-t^2)}\bigg\} p_{x\mkern2mu|\mkern2mu y}(\mathbf{u} \mkern2mu|\mkern2mu \mathbf{y})\,\mathrm{d} \mathbf{u} \notag\\
    &=\frac{-t|\mathbf{x}|_2^2}{(1-t^2)^2} + \frac{1+t^2}{(1-t^2)^2}\mathbb{E}(\mathbf{X}^{\rm{T}}\mathbf{x} \mkern 2mu|\mkern2mu \mathbf{W}_t=\mathbf{x}, \mathbf{Y}=\mathbf{y}) \notag\\
    &~~~-\frac{t}{(1-t^2)^2}\mathbb{E}(|\mathbf{X}|_2^2\mkern 2mu|\mkern2mu \mathbf{W}_t=\mathbf{x}, \mathbf{Y}=\mathbf{y})\,,\label{equa:(partial_t phi_t)/(phi_t)}\\
        \frac{\nabla_{\mathbf{x}} \phi_t^{\mathbf{y}}(\mathbf{x})}{\phi_t^{\mathbf{y}}(\mathbf{x})} &=\frac{1}{\phi^{\mathbf{y}}_t(\mathbf{x})} \int \frac{t\mathbf{u} - \mathbf{x}}{1-t^2} \exp \bigg\{\!  -\frac{|\mathbf{x}-t \mathbf{u}|_2^2}{2(1-t^2)}\bigg\} p_{x\mkern2mu|\mkern2mu y}(\mathbf{u} \mkern2mu|\mkern2mu \mathbf{y})\,\mathrm{d} \mathbf{u} \notag\\
        &=\frac{-\mathbf{x}}{1-t^2} + \frac{t}{1-t^2}\mathbb{E}(\mathbf{X} \mkern 2mu|\mkern2mu \mathbf{W}_t=\mathbf{x}, \mathbf{Y}=\mathbf{y})\,.\label{equa:(nabla phi_t)/(phi_t)}
    \end{align}
Combining (\ref{equa:partial_t v*}), (\ref{equa:(partial_t nabla phi_t)/(phi_t)}), (\ref{equa:(partial_t phi_t)/(phi_t)}) and (\ref{equa:(nabla phi_t)/(phi_t)}), we obtain
    \begin{align*}
        \partial_t \mathbf{v}_{\rm{F}} (\mathbf{x}, \mathbf{y}, t) &= \frac{1+t^2}{(1-t^2)^3}\operatorname{Cov}(\mathbf{X} \mkern 2mu|\mkern2mu \mathbf{W}_t=\mathbf{x}, \mathbf{Y}=\mathbf{y})\mathbf{x} + \frac{2t}{(1-t^2)^2}\mathbb{E}(\mathbf{X} \mkern 2mu|\mkern2mu \mathbf{W}_t=\mathbf{x}, \mathbf{Y}=\mathbf{y}) \\
        &~~~- \frac{t}{(1-t^2)^3}\mathbb{E}(\mathbf{X}|\mathbf{X}|_2^2 \mkern2mu | \mkern2mu \mathbf{W}_t=\mathbf{x}, \mathbf{Y}=\mathbf{y}) - \frac{1+t^2}{(1-t^2)^2}\mathbf{x}  \\
        &~~~+\frac{t}{(1-t^2)^3}\mathbb{E}(\mathbf{X} \mkern 2mu|\mkern2mu \mathbf{W}_t=\mathbf{x}, \mathbf{Y}=\mathbf{y}) \mathbb{E}(|\mathbf{X}|_2^2 \mkern2mu | \mkern2mu \mathbf{W}_t=\mathbf{x}, \mathbf{Y}=\mathbf{y})\,,
    \end{align*}
which implies
    \begin{align} \label{equa:2-norm of partial_t vF}
        |\partial_t \mathbf{v}_{\rm{F}} (\mathbf{x}, \mathbf{y}, t)|_2 &\leq  \frac{1+t^2}{(1-t^2)^3}\|\operatorname{Cov}(\mathbf{X} \mkern 2mu|\mkern2mu \mathbf{W}_t=\mathbf{x}, \mathbf{Y}=\mathbf{y})\|_{\rm op}|\mathbf{x}|_2 \notag \\
        &~~~+ \frac{2t}{(1-t^2)^2} |\mathbb{E}(\mathbf{X} \mkern 2mu|\mkern2mu \mathbf{W}_t=\mathbf{x}, \mathbf{Y}=\mathbf{y})|_2 + \frac{1+t^2}{(1-t^2)^2}|\mathbf{x}|_2 \notag \\
        &~~~+ \frac{t}{(1-t^2)^3}|\mathbb{E}(\mathbf{X}|\mathbf{X}|_2^2 \mkern2mu | \mkern2mu \mathbf{W}_t=\mathbf{x}, \mathbf{Y}=\mathbf{y})|_2 \notag \\
        &~~~+ \frac{t}{(1-t^2)^3}|\mathbb{E}(\mathbf{X} \mkern 2mu|\mkern2mu \mathbf{W}_t=\mathbf{x}, \mathbf{Y}=\mathbf{y})|_2 \mathbb{E}(|\mathbf{X}|_2^2 \mkern2mu | \mkern2mu \mathbf{W}_t=\mathbf{x}, \mathbf{Y}=\mathbf{y}) \, .
    \end{align}
By Assumption \ref{assump:bounded support}, we have $|\mathbf{X}|_{\infty} \leq 1$, so $|\mathbb{E}(\mathbf{X} \mkern 2mu|\mkern2mu \mathbf{W}_t=\mathbf{x}, \mathbf{Y}=\mathbf{y})|_2 \leq {d^{1 / 2}_x}$, $|\mathbb{E}(|\mathbf{X}|_2^2 \mkern2mu | \mkern2mu \mathbf{W}_t=\mathbf{x}, \mathbf{Y}=\mathbf{y})|_2 \leq {d_x}$ and  $|\mathbb{E}(\mathbf{X}|\mathbf{X}|_2^2 \mkern2mu | \mkern2mu \mathbf{W}_t=\mathbf{x}, \mathbf{Y}=\mathbf{y})|_2 \leq {d^{3 / 2}_x}$. For any $\mathbf{u} \in \mathbb{R}^{d_x}$, due to
\begin{align*}
\mathbf{u}^{\rm{T}} \operatorname{Cov}(\mathbf{X} \mkern2mu|\mkern2mu \mathbf{W}_t=\mathbf{x}, \mathbf{Y}=\mathbf{y}) \mathbf{u} \leq\mathbb{E}\{(\mathbf{u}^{\rm{T}} \mathbf{X})^2 \mkern2mu | \mkern2mu \mathbf{W}_t=\mathbf{x}, \mathbf{Y}=\mathbf{y}\}\leq {d_x}|\mathbf{u}|_2^2 \, ,
\end{align*}
 we have $\|\operatorname{Cov}(\mathbf{X} \mkern2mu|\mkern2mu \mathbf{W}_t=\mathbf{x}, \mathbf{Y}=\mathbf{y})\|_{\mathrm{op}} \leq  {d_x}$. Hence, due to $T<1$, by \eqref{equa:2-norm of partial_t vF}, it holds that
\begin{align*}
&\sup _{t \in[0, T]} \sup _{\mathbf{x} \in[-R, R]^{d_x}}\sup_{\mathbf{y} \in [0, B]^{d_{y}}}|\partial_t \mathbf{v}_{\rm{F}} (\mathbf{x}, \mathbf{y}, t)|_2 \\
&~~~~~~\leq \frac{1+T^2}{(1-T^2)^3}{d^{3/2}_x}R + \frac{2T{d^{1/2}_x}}{(1-T^2)^2} + \frac{2T{d^{3/2}_x}}{(1-T^2)^3} + \frac{1+T^2}{(1-T^2)^2}R{d^{1/2}_x} \\
&~~~~~~\leq  \frac{C_{*} {d^{3 / 2}_x}(R+1)}{(1-T)^3} \, ,
\end{align*}
where $C_{*} > 1$ is some universal constant independent of $(d_x, R, T)$. We complete the proof of Proposition \ref{prop:properties of vF}(i). $\hfill\Box$

\subsection{Proof of Proposition \ref{prop:properties of vF}(ii)}
Recall that $\mathbf{v}_{\rm{F}} (\mathbf{x}, \mathbf{y}, t) = t^{-1} \nabla_{\mathbf{x}} \log f_t(\mathbf{x} \mkern 2mu|\mkern2mu \mathbf{y}) +  t^{-1} \mathbf{x}$. Then we have
    $$
        \nabla_{\mathbf{x}} \mathbf{v}_{\rm{F}} (\mathbf{x}, \mathbf{y}, t) = \frac{1}{t} \nabla^2_{\mathbf{x}} \log f_t(\mathbf{x} \mkern 2mu|\mkern2mu \mathbf{y}) + \frac{1}{t} \mkern2mu \mathbf{I}_{d_x} \, .
    $$
As we have shown in \eqref{equa:function form of ft} in Section {\ref{subsec:flow-density-step1}},  
$$f_t(\mathbf{x} \mkern2mu|\mkern2mu \mathbf{y}) = C \int p_{x\mkern2mu|\mkern2mu y}(\mathbf{u} \mkern2mu|\mkern2mu \mathbf{y}) \exp \bigg\{\!-\frac{|\mathbf{x}-t \mathbf{u}|_2^2}{2(1-t^2)}\bigg\} \, {\rm d} \mathbf{u}\,,$$  
where  $C = (2 \pi)^{-d_x/2}(1-t^2)^{-d_x/2}$. Then it holds that %
\begin{align*} 
        \nabla^2_{\mathbf{x}} \log f_t(\mathbf{x} \mkern 2mu|\mkern2mu \mathbf{y})&=-\int \frac{\mathbf{x}-t\mathbf{u}}{1-t^2}\frac{p_{x\mkern2mu|\mkern2mu y}(\mathbf{u} \mkern2mu|\mkern2mu \mathbf{y})}{C^{-1}f_t(\mathbf{x} \mkern 2mu|\mkern2mu \mathbf{y})}\exp \bigg\{\!-\frac{|\mathbf{x}-t \mathbf{u}|_2^2}{2(1-t^2)}\bigg\} \, {\rm d} \mathbf{u}\\
        &~~~~~~~~~\times \bigg[\int \frac{\mathbf{x}-t\mathbf{u}}{1-t^2}\frac{p_{x\mkern2mu|\mkern2mu y}(\mathbf{u} \mkern2mu|\mkern2mu \mathbf{y})}{C^{-1}f_t(\mathbf{x} \mkern 2mu|\mkern2mu \mathbf{y})}\exp \bigg\{\!-\frac{|\mathbf{x}-t \mathbf{u}|_2^2}{2(1-t^2)}\bigg\} \, {\rm d} \mathbf{u}\bigg]^{\rm T}\\
        &~~~+\int\frac{\mathbf{x}-t\mathbf{u}}{1-t^2}\frac{(\mathbf{x}-t\mathbf{u})^{\rm T}}{1-t^2}\frac{p_{x\mkern2mu|\mkern2mu y}(\mathbf{u} \mkern2mu|\mkern2mu \mathbf{y})}{C^{-1}f_t(\mathbf{x} \mkern 2mu|\mkern2mu \mathbf{y})}\exp \bigg\{\!-\frac{|\mathbf{x}-t \mathbf{u}|_2^2}{2(1-t^2)}\bigg\} \, {\rm d} \mathbf{u}\\
        &~~~-\mathbf{I}_{d_x}\int \frac{1}{1-t^2}\frac{p_{x\mkern2mu|\mkern2mu y}(\mathbf{u} \mkern2mu|\mkern2mu \mathbf{y})}{C^{-1}f_t(\mathbf{x} \mkern 2mu|\mkern2mu \mathbf{y})}\exp \bigg\{\!-\frac{|\mathbf{x}-t \mathbf{u}|_2^2}{2(1-t^2)}\bigg\} \, {\rm d} \mathbf{u}\\
        &=-\int \frac{\mathbf{x}-t\mathbf{u}}{1-t^2}p_{x\mkern2mu|\mkern2mu w_t, y}(\mathbf{u}\mkern2mu|\mkern2mu \mathbf{x},\mathbf{y}) \, {\rm d} \mathbf{u}\times \bigg[\int \frac{\mathbf{x}-t\mathbf{u}}{1-t^2}p_{x\mkern2mu|\mkern2mu w_t, y}(\mathbf{u}\mkern2mu|\mkern2mu \mathbf{x},\mathbf{y}) \, {\rm d} \mathbf{u}\bigg]^{\rm T}\\
        &~~~+\int\frac{\mathbf{x}-t\mathbf{u}}{1-t^2}\frac{(\mathbf{x}-t\mathbf{u})^{\rm T}}{1-t^2}p_{x\mkern2mu|\mkern2mu w_t, y}(\mathbf{u}\mkern2mu|\mkern2mu \mathbf{x},\mathbf{y}) \, {\rm d} \mathbf{u}-\mathbf{I}_{d_x}\int \frac{p_{x\mkern2mu|\mkern2mu w_t, y}(\mathbf{u}\mkern2mu|\mkern2mu \mathbf{x},\mathbf{y})}{1-t^2} \, {\rm d} \mathbf{u}\,,
\end{align*}
where the last step is based on the identity \eqref{eq:cond11} and $p_{x\mkern2mu|\mkern2mu w_t, y}(\mathbf{u}\mkern2mu|\mkern2mu \mathbf{x},\mathbf{y})$ is the conditional density of $\mathbf{X}$ given $(\mathbf{W}_t,\mathbf{Y})=(\mathbf{x},\mathbf{y})$. Hence, 
        \begin{align*}
       \nabla^2_{\mathbf{x}} \log f_t(\mathbf{x} \mkern 2mu|\mkern2mu \mathbf{y}) =\mkern2mu \frac{-1}{1-t^2} \mkern2mu \mathbf{I}_{d_x} + \frac{t^2}{(1-t^2)^2}\operatorname{Cov}(\mathbf{X} \mkern2mu|\mkern2mu \mathbf{W}_t=\mathbf{x}, \mathbf{Y}=\mathbf{y}) \, ,
    \end{align*}
which implies that 
\begin{align}\label{equa:nabla v* = aCov + bI_d}
       \nabla_{\mathbf{x}} \mathbf{v}_{\rm{F}} (\mathbf{x}, \mathbf{y}, t) =\mkern2mu \frac{-t}{1-t^2} \mkern2mu \mathbf{I}_{d_x} + \frac{t}{(1-t^2)^2}\operatorname{Cov}(\mathbf{X} \mkern2mu|\mkern2mu \mathbf{W}_t=\mathbf{x}, \mathbf{Y}=\mathbf{y}) \, .
    \end{align}
    As we have shown in Section \ref{subsec:proof of vF i}, $\|\operatorname{Cov}(\mathbf{X} \mkern2mu|\mkern2mu \mathbf{W}_t=\mathbf{x}, \mathbf{Y}=\mathbf{y})\|_{\mathrm{op}} \leq  {d_x}$, which implies
$
0 \preceq \operatorname{Cov}(\mathbf{X} \mkern2mu|\mkern2mu \mathbf{W}_t=\mathbf{x}, \mathbf{Y}=\mathbf{y}) \preceq {d_x} \mkern2mu \mathbf{I}_{d_x}
$.
Hence, 
\begin{align}\label{eq:covbd11}
-\frac{t}{1-t^2} \mkern2mu \mathbf{I}_{d_x} \preceq \nabla_{\mathbf{x}} \mathbf{v}_{\rm{F}}(\mathbf{x}, \mathbf{y}, t) \preceq \bigg\{\frac{t {d_x}}{(1-t^2)^2}-\frac{t}{1-t^2}\bigg\} \mkern2mu \mathbf{I}_{d_x} \, .
\end{align}
Then it holds that 
\begin{align} \label{equa:g(t)}
\|\nabla_{\mathbf{x}} \mathbf{v}_{\rm F}(\mathbf{x}, \mathbf{y}, t)\|_{\rm op} \leq \bigg|\frac{t}{1-t^2}\bigg| \vee \bigg|\frac{t(d_x-1)+{t^3}}{(1-t^2)^2}\bigg| \leq \frac{d_x}{(1-T)^2} \, .
\end{align}
This means $\mathbf{v}_{\rm F}(\mathbf{x}, \mathbf{y}, t)$ is $d_x(1-T)^{-2}$-Lipschitz w.r.t. $\mathbf{x}$ over $[0, T]$.

In the sequel, we demonstrate that for any $t\in[0,T]$, $\mathbf{F}_t(\mathbf{x}, \mathbf{y})$ exhibits Lipschitz property. %
Recall that the conditional F\"{o}llmer flow map $\mathbf{F}_t(\mathbf{x}, \mathbf{y})$ defined in Definition $\ref{def:CFFM}$ represents the ODE solution of ${\rm d} \mathbf{x}_t = \mathbf{v}_{\rm F}(\mathbf{x}_t, \mathbf{y}, t) \, {\rm d}t$ at time $t$, given initial value ${\bf{x}}_0 = \bf{x}$. Thus, it holds that
\[
{\bf F}_t(\mathbf{x}, \mathbf{y}) - {\bf F}_0(\mathbf{x}, \mathbf{y}) = \int_0^t \mathbf{v}_{\rm F}({\bf F}_s(\mathbf{x}, \mathbf{y}), \mathbf{y}, s) \, {\rm d}s \, ,\quad {\bf F}_0(\mathbf{x}, \mathbf{y}) = \mathbf{x} \, .
\]
Taking the gradient w.r.t. $\mathbf{x}$ on both sides of above equation, we obtain
\[
\nabla_{\mathbf{x}} {\bf F}_t(\mathbf{x}, \mathbf{y}) - \nabla_{\mathbf{x}} {\bf F}_0(\mathbf{x}, \mathbf{y}) = \int_0^t \big\{\nabla_{\mathbf{u}} \mathbf{v}_{\rm F}(\mathbf{u}, \mathbf{y}, s)\big\} \mkern 1mu \big| \mkern 1mu _{\mathbf{u}={\bf F}_s(\mathbf{x}, \mathbf{y})} \nabla_{\mathbf{x}} {\bf F}_s(\mathbf{x}, \mathbf{y}) \, {\rm d}s \, ,
\]
which implies
\begin{align*}
\partial_t \nabla_{\mathbf{x}} {\bf F}_t(\mathbf{x}, \mathbf{y}) = \big\{\nabla_{\mathbf{u}} \mathbf{v}_{\rm F}(\mathbf{u}, \mathbf{y}, t)\big\} \mkern 1mu \big| \mkern 1mu _{\mathbf{u}={\bf F}_t(\mathbf{x}, \mathbf{y})} \nabla_{\mathbf{x}} {\bf F}_t(\mathbf{x}, \mathbf{y}) \, .
\end{align*}
For any given $\mathbf{r} \in \mathbb{R}^{d_x}$ with $|\mathbf{r}|_2=1$, let $u_{\mathbf{r}}(\mathbf{x}, \mathbf{y}, t)=|\nabla_{\mathbf{x}} {\bf F}_t(\mathbf{x}, \mathbf{y}) \mkern 2mu \mathbf{r}|_2^2$. Then we further have
\begin{align*}
\partial_t u_{\mathbf{r}}(\mathbf{x}, \mathbf{y}, t)&=  2 \big \langle \nabla_{\mathbf{x}} {\bf F}_t(\mathbf{x}, \mathbf{y}) \mkern 2mu \mathbf{r}, \partial_t\nabla_{\mathbf{x}} {\bf F}_t(\mathbf{x}, \mathbf{y}) \mkern 2mu \mathbf{r} \big \rangle \\
&= 2 \big \langle \nabla_{\mathbf{x}} {\bf F}_t(\mathbf{x}, \mathbf{y}) \mkern 2mu \mathbf{r}, \{\nabla_{\mathbf{u}} \mathbf{v}_{\rm F}(\mathbf{u}, \mathbf{y}, t)\} \mkern 1mu \big| \mkern 1mu _{\mathbf{u}={\bf F}_t(\mathbf{x}, \mathbf{y})} \nabla_{\mathbf{x}} {\bf F}_t(\mathbf{x}, \mathbf{y}) \mkern 2mu \mathbf{r} \big \rangle \\
&\leq 2\big|\nabla_{\mathbf{x}} {\bf F}_t(\mathbf{x}, \mathbf{y}) \mkern 2mu \mathbf{r}\big|_2 \big| \{\nabla_{\mathbf{u}} \mathbf{v}_{\rm F}(\mathbf{u}, \mathbf{y}, t)\} \mkern 1mu \big| \mkern 1mu _{\mathbf{u}={\bf F}_t(\mathbf{x}, \mathbf{y})} \nabla_{\mathbf{x}} {\bf F}_t(\mathbf{x}, \mathbf{y}) \mkern 2mu \mathbf{r}\big|_2 \\
&\leq 2{\bigl\|\nabla_{\mathbf{u}} \mathbf{v}_{\rm F}(\mathbf{u}, \mathbf{y}, t)\bigr|_{\mathbf{u}={\bf F}_t(\mathbf{x},\mathbf{y})}\bigr\|_{\rm op}} u_{\mathbf{r}}(\mathbf{x}, \mathbf{y}, t) \leq \frac{2d_x}{(1-T)^2} \mkern 2mu u_{\mathbf{r}}(\mathbf{x}, \mathbf{y}, t) \, ,
\end{align*}
where the first inequality follows from the Cauchy-Schwarz inequality. By Lemma \ref{lemma:gronwall}, it holds that  
\begin{align*}
\big\|\nabla_{\mathbf{x}} {\mathbf{F}}_t(\mathbf{x}, \mathbf{y})\big\|_{\rm op} &= \sup_{|\mathbf{r}|_2=1} \sqrt{u_{\mathbf{r}}(\mathbf{x}, \mathbf{y}, t)} \\
&\leq \sup_{|\mathbf{r}|_2=1} \sqrt{u_{\mathbf{r}}(\mathbf{x}, \mathbf{y}, 0)} \exp\{d_x(1-T)^{-2}\} = \exp\{d_x(1-T)^{-2}\} \, .
\end{align*}
which concludes the proof. $\hfill\Box$

 \section{Discussion of Assumption \ref{assump:Lip in y}} \label{sec:discuss Lip y of vF}
{Throughout this section, assume that $p_{x,y}$ is positive on a fixed product support $S\times[0,B]^{d_y}$ and continuously differentiable in $\mathbf y$.} Based on the established results in Section \ref{sec:properties of vF}, we now discuss the reasonableness of our Assumption \ref{assump:Lip in y}. Recall that $\mathbf{v}_{\rm{F}} (\mathbf{x}, \mathbf{y}, t) = t^{-1} \nabla_{\mathbf{x}} \log f_t(\mathbf{x} \mkern 2mu|\mkern2mu \mathbf{y}) +  t^{-1} \mathbf{x}$. Then we have
$\nabla_{\mathbf{y}} \mathbf{v}_{\rm{F}} (\mathbf{x}, \mathbf{y}, t) = t^{-1} \nabla_{\mathbf{y}} \nabla_{\mathbf{x}} \log f_t(\mathbf{x} \mkern 2mu|\mkern2mu \mathbf{y}) \, .$
As we have shown in \eqref{equa:function form of ft} in Section {\ref{subsec:flow-density-step1}},  
$$f_t(\mathbf{x} \mkern2mu|\mkern2mu \mathbf{y}) = C \int p_{x\mkern2mu|\mkern2mu y}(\mathbf{u} \mkern2mu|\mkern2mu \mathbf{y}) \exp \bigg\{\!-\frac{|\mathbf{x}-t \mathbf{u}|_2^2}{2(1-t^2)}\bigg\} \, {\rm d} \mathbf{u}\,,$$  
where  $C = (2 \pi)^{-d_x/2}(1-t^2)^{-d_x/2}$. Thus, it holds that
\begin{align*}
&\nabla_{\mathbf{y}} \nabla_{\mathbf{x}} \log f_t(\mathbf{x} \mkern 2mu|\mkern2mu \mathbf{y}) = \int \frac{t\mathbf{u}-\mathbf{x}}{1-t^2}\frac{\{\nabla_{\mathbf{y}} p_{x\mkern2mu|\mkern2mu y}(\mathbf{u} \mkern2mu|\mkern2mu \mathbf{y})\}^{\rm T}}{C^{-1}f_t(\mathbf{x} \mkern 2mu|\mkern2mu \mathbf{y})}\exp \bigg\{\!-\frac{|\mathbf{x}-t \mathbf{u}|_2^2}{2(1-t^2)}\bigg\} \, {\rm d} \mathbf{u} \\
&~~~~~~~~~~~~~~~~~~~~~~~~~~~ - \int \frac{t\mathbf{u}-\mathbf{x}}{1-t^2}\frac{p_{x\mkern2mu|\mkern2mu y}(\mathbf{u} \mkern2mu|\mkern2mu \mathbf{y})}{C^{-1}f_t(\mathbf{x} \mkern 2mu|\mkern2mu \mathbf{y})}\exp \bigg\{\!-\frac{|\mathbf{x}-t \mathbf{u}|_2^2}{2(1-t^2)}\bigg\} \, {\rm d} \mathbf{u} \\
&~~~~~~~~~~~~~~~~~~~~~~~~~~~~~~~ \times  \int \frac{\{\nabla_{\mathbf{y}} p_{x\mkern2mu|\mkern2mu y}(\mathbf{u} \mkern2mu|\mkern2mu \mathbf{y})\}^{\rm T}}{C^{-1}f_t(\mathbf{x} \mkern 2mu|\mkern2mu \mathbf{y})}\exp \bigg\{\!-\frac{|\mathbf{x}-t \mathbf{u}|_2^2}{2(1-t^2)}\bigg\} \, {\rm d} \mathbf{u} \, .
\end{align*}
 Let $\mathbf{L}(\mathbf{x}, \mathbf{y}) := \nabla_{\bf y} \log p_{x\mkern2mu|\mkern2mu y}(\mathbf{x} \mkern2mu|\mkern2mu \mathbf{y})$. Then, we have $\nabla_{\bf y} p_{x\mkern2mu|\mkern2mu y}(\mathbf{x} \mkern2mu|\mkern2mu \mathbf{y}) = p_{x\mkern2mu|\mkern2mu y}(\mathbf{x} \mkern2mu|\mkern2mu \mathbf{y}) \mathbf{L}(\mathbf{x}, \mathbf{y})$. Recall $\mathbf{W}_t = t \mathbf{X} + \sqrt{1-t^2} \mathbf{W}$ with $\mathbf{W} \sim \mathcal{N}(\mathbf{0}, \mathbf{I}_{d_x})$. It further holds that
\begin{align} \label{eq:nabla_y_x f_t}
& \nabla_{\mathbf{y}} \nabla_{\mathbf{x}} \log f_t(\mathbf{x} \mkern 2mu|\mkern2mu \mathbf{y}) = \int\frac{t\mathbf{u}-\mathbf{x}}{1-t^2} \mathbf{L}^{\rm T}(\mathbf{u}, \mathbf{y}) p_{x\mkern2mu|\mkern2mu w_t, y}(\mathbf{u}\mkern2mu|\mkern2mu \mathbf{x},\mathbf{y}) \, {\rm d} \mathbf{u} \\
&~~~~~~~~~~~~~~~~~~~~~~~~~~~ - \int\frac{t\mathbf{u}-\mathbf{x}}{1-t^2}  p_{x\mkern2mu|\mkern2mu w_t, y}(\mathbf{u}\mkern2mu|\mkern2mu \mathbf{x},\mathbf{y}) \, {\rm d} \mathbf{u} \times \int \mathbf{L}^{\rm T}(\mathbf{u}, \mathbf{y}) p_{x\mkern2mu|\mkern2mu w_t, y}(\mathbf{u}\mkern2mu|\mkern2mu \mathbf{x},\mathbf{y}) \, {\rm d} \mathbf{u} \, , \notag
\end{align}
which is based on the identity \eqref{eq:cond11}, and $p_{x\mkern2mu|\mkern2mu w_t, y}(\mathbf{u}\mkern2mu|\mkern2mu \mathbf{x},\mathbf{y})$ is the conditional density of $\mathbf{X}$ given $(\mathbf{W}_t,\mathbf{Y})=(\mathbf{x},\mathbf{y})$. By \eqref{eq:nabla_y_x f_t}, we get
\begin{align*}
&\nabla_{\mathbf{y}} \nabla_{\mathbf{x}} \log f_t(\mathbf{x} \mkern 2mu|\mkern2mu \mathbf{y}) = \frac{t}{1-t^2}  \mathbb{E}\big\{\mathbf{X}\mathbf{L}^{\rm T}(\mathbf{X}, \mathbf{y}) \mkern 2mu|\mkern2mu \mathbf{W}_t = \mathbf{x}, \mathbf{Y} = \mathbf{y}\big\} \\
&~~~~~~~~~~~~~~~~~~~~~~~~~~ -\frac{t}{1-t^2}  \mathbb{E}\big(\mathbf{X} \mkern 2mu|\mkern2mu \mathbf{W}_t = \mathbf{x}, \mathbf{Y} = \mathbf{y}\big) \mathbb{E}\big\{\mathbf{L}^{\rm T}(\mathbf{X}, \mathbf{y}) \mkern 2mu|\mkern2mu \mathbf{W}_t = \mathbf{x}, \mathbf{Y} = \mathbf{y}\big\} \\
&~~~~~~~~~~~~~~~~~~~~~~~ = \frac{t}{1-t^2} {\rm Cov} \big\{\mathbf{X}, \mathbf{L}(\mathbf{X}, \mathbf{y}) \mkern 2mu|\mkern2mu \mathbf{W}_t = \mathbf{x}, \mathbf{Y} = \mathbf{y} \big\} \, .
\end{align*}
 Thus, it holds that $\nabla_{\mathbf{y}} \mathbf{v}_{\rm{F}} (\mathbf{x}, \mathbf{y}, t ) = (1-t^2)^{-1} {\rm Cov} \{ \mathbf{X}, \mathbf{L}(\mathbf{X}, \mathbf{y}) \mkern 2mu|\mkern2mu \mathbf{W}_t = \mathbf{x}, \mathbf{Y} = \mathbf{y} \}$. To examine the Lipschitz property of $\mathbf{v}_{\rm{F}} (\mathbf{x}, \mathbf{y}, t)$ with respect to $\mathbf{y}$, we need to evaluate the matrix norm of $\nabla_{\mathbf{y}} \mathbf{v}_{\rm{F}} (\mathbf{x}, \mathbf{y}, t)$, such as the Frobenius norm $\|\cdot\|_{\rm Fr}$.
If $\sup_{\mathbf{x}, \mathbf{y}} |\mathbf{L}(\mathbf{x}, \mathbf{y})|_{\infty} \le U$ for some universal constant $U>0$, we could proceed to bound $\|\nabla_{\mathbf{y}} \mathbf{v}_{\rm{F}} (\mathbf{x}, \mathbf{y}, t)\|_{\rm Fr}$ explicitly. 

 Notice that
$$\|\nabla_{\mathbf{y}} \mathbf{v}_{\rm{F}} (\mathbf{x}, \mathbf{y}, t)\|_{\rm Fr} = (1-t^2)^{-1} \|{\rm Cov} \big\{ \mathbf{X}, \mathbf{L}(\mathbf{X}, \mathbf{y}) \mkern 2mu|\mkern2mu \mathbf{W}_t = \mathbf{x}, \mathbf{Y} = \mathbf{y} \big\} \|_{\rm Fr}\,.$$
By the Cauchy-Schwarz inequality, we have $$\|\text{Cov}(\mathbf{A},\mathbf{B})\|_{\rm Fr}^2 \le \text{tr}\{\text{Cov}(\mathbf{A})\}\text{tr}\{\text{Cov}(\mathbf{B})\} \, ,$$
 which implies
\begin{align*}
&\|{\rm Cov} \big\{\mathbf{X}, \mathbf{L}(\mathbf{X}, \mathbf{y}) \mkern 2mu|\mkern2mu \mathbf{W}_t = \mathbf{x}, \mathbf{Y} = \mathbf{y} \big\}\|_{\rm Fr}^2 \\ 
&~~~~~~ \le \text{tr}\big\{{\rm Cov}(\mathbf{X} \mkern 2mu|\mkern2mu \mathbf{W}_t = \mathbf{x}, \mathbf{Y} = \mathbf{y})\big\}\cdot \text{tr}\big[{\rm Cov}\big\{\mathbf{L}(\mathbf{X}, \mathbf{y}) \mkern 2mu|\mkern2mu \mathbf{W}_t = \mathbf{x}, \mathbf{Y} = \mathbf{y}\big\}\big]\,,
\end{align*}
For $\mathbf{X} = (X_1, \ldots, X_{d_x})^{\rm T}$, we have
$$
{\rm tr}\big\{{\rm Cov}(\mathbf{X} \mkern 2mu|\mkern2mu \mathbf{W}_t = \mathbf{x}, \mathbf{Y} = \mathbf{y})\big\} = \sum_{i=1}^{d_x} \text{Var}(X_i \mkern 2mu|\mkern2mu \mathbf{W}_t = \mathbf{x}, \mathbf{Y} = \mathbf{y}) \, .
$$
By Assumption \ref{assump:bounded support}, it holds that $X_i \in [0, 1]$. Thus, 
$$ \text{tr}\big\{{\rm Cov}(\mathbf{X} \mkern 2mu|\mkern2mu \mathbf{W}_t = \mathbf{x}, \mathbf{Y} = \mathbf{y})\big\} \le \sum_{i=1}^{d_x} 1 = d_x \,. $$
  Since $|\mathbf{L}(\mathbf{x}, \mathbf{y})|_{\infty} \le U$, we have
$$\text{tr}\big[{\rm Cov}\big\{\mathbf{L}(\mathbf{X}, \mathbf{y}) \mkern 2mu|\mkern2mu \mathbf{W}_t = \mathbf{x}, \mathbf{Y} = \mathbf{y}\big\}\big] \le \sum_{j=1}^{d_y} U^2 = d_y U^2\,.$$
Combining these, we get  $\|{\rm Cov}\{\mathbf{X}, {\bf L}(\mathbf{X}, \mathbf{y}) \mkern 2mu|\mkern2mu \mathbf{W}_t = \mathbf{x}, \mathbf{Y} = \mathbf{y}\}\|_{\rm Fr} \le {U\sqrt{d_x d_y}}$, which implies
$ \|\nabla_{\mathbf{y}} \mathbf{v}_{\rm{F}} (\mathbf{x}, \mathbf{y}, t)\|_{\rm Fr} \le {U\sqrt{d_x d_y}}(1-t^2)^{-1} \le {U\sqrt{d_x d_y}}(1-t)^{-1}$. Therefore, Assumption \ref{assump:Lip in y} holds with $C_y(d_x, d_y) = U\sqrt{d_xd_y}$, $\alpha=0$ and $\beta=1$. This illustrates that our Assumption \ref{assump:Lip in y} represents a more general case and is thus reasonable.

 The requirement $\sup_{\mathbf{x}, \mathbf{y}} |\mathbf{L}(\mathbf{x}, \mathbf{y})|_{\infty} \le U$ is mild, which can be guaranteed if
\begin{align*}
\sup_{\mathbf{x}, \mathbf{y}} |\nabla_{\mathbf{y}} \log p_{x, y}(\mathbf{x}, \mathbf{y})|_{\infty} \le V 
\end{align*}
for some universal constant $V>0$. More specifically, due to
\begin{align*}
    &\nabla_{\mathbf{y}} \log p_y(\mathbf{y}) = \frac{\int \nabla_{\mathbf{y}} p_{x, y}(\mathbf{x}, \mathbf{y}) \, {\rm d} \mathbf{x}}{p_y(\mathbf{y})} = \int \nabla_{\mathbf{y}} \log p_{x, y}(\mathbf{x}, \mathbf{y}) p_{x\mkern2mu|\mkern2mu y}(\mathbf{x} \mkern2mu|\mkern2mu \mathbf{y}) \, {\rm d} \mathbf{x} \\
    & ~~~~~~~~~~~~~~~~ = \mathbb{E} \big\{\nabla_{\mathbf{y}} \log p_{x, y}(\mathbf{X}, \mathbf{y}) \mkern2mu \big|\mkern2mu \mathbf{Y}=\mathbf{y} \big\} \, ,
\end{align*}
by Jensen's inequality, it holds that
\begin{align*}
    |\nabla_{\mathbf{y}} \log p_y(\mathbf{y})|_{\infty} \le \mathbb{E} \big\{ \big|\nabla_{\mathbf{y}} \log p_{x, y}(\mathbf{X}, \mathbf{y})\big|_{\infty} \mkern2mu |\mkern2mu \mathbf{Y}=\mathbf{y} \big\} \le V \, .
\end{align*}
Thus, by the definition of $\mathbf{L}(\mathbf{x}, \mathbf{y})$, we get
\begin{align*}
    |\mathbf{L}(\mathbf{x}, \mathbf{y})|_{\infty} \le |\nabla_{\bf y} \log p_{x, y}(\mathbf{x}, \mathbf{y})|_{\infty} + |\nabla_{\mathbf{y}} \log p_y(\mathbf{y})|_{\infty} \le 2V
\end{align*}
for any $\mathbf{x}, \mathbf{y}$. In generative learning theory, $\nabla_{(\mathbf{x}, \mathbf{y})} \log p_{x, y}(\mathbf{x}, \mathbf{y})$ is known as the score function \citep{hyvarinen2005estimation} of the data distribution. Requiring its boundedness in technical analysis is a standard practice in this field. For example, see \cite{oko2023diffusion_sm}.  %

\section{Proof of Proposition \ref{prop:generalization}} \label{append:gene}

Denote by $g_t(\mathbf{x}, \mathbf{y})$ the joint density of $t \mathbf{X}+\sqrt{1-t^2} \mathbf{W}$ and $\mathbf{Y}$. For simplicity, we will use ${\mathbf{W}}_t$ to represent $t \mathbf{X}+\sqrt{1-t^2} \mathbf{W}$ and abbreviate $L^2(g_t(\mathbf{x}, \mathbf{y}))$ as $L^2(g_t)$ in the remaining part of this section. For any velocity field $\mathbf{v}: \mathbb{R}^{d_x} \times 
\mathbb{R}^{d_{y}} \times[0, T] \rightarrow \mathbb{R}^{d_x}$, by \eqref{equa:expanding L(v)} and \eqref{equa:condiexpeczero}, we have  
\begin{align*}
    &\mathbb{E}\left\{\bigg|\mathbf{X}-\frac{t}{\sqrt{1-t^2}}\mathbf{W}-\mathbf{v}  (  \mathbf{W}_t, \mathbf{Y}, t)\bigg|_2^2\right\} \\
    &~~~~~=\mathbb{E}\left\{\bigg|\mathbf{X}-\frac{t}{\sqrt{1-t^2}}\mathbf{W}-\mathbf{v}_{\rm{F}}  (  \mathbf{W}_t, \mathbf{Y}, t)\bigg|_2^2\right\}+ \mathbb{E}\big\{|\mathbf{v}_{\rm F}(\mathbf{W}_t, \mathbf{Y}, t) - \mathbf{v}(\mathbf{W}_t, \mathbf{Y}, t)|_2^2\big\}
\end{align*}
for any $t \in [0, T]$. By the definition of $\mathcal{L}(\cdot)$ given in Proposition \ref{prop:training object}(ii), it holds that
\begin{align}
     \label{equa:L(v)-L(v_F)}
\mathcal{L}(\mathbf{v})-\mathcal{L}(\mathbf{v}_{\rm{F}} )=\frac{1}{T} \int_0^{T}\|\mathbf{v}(\mathbf{x}, \mathbf{y}, t)-\mathbf{v}_{\rm{F}} (\mathbf{x}, \mathbf{y}, t)\|_{L^2(g_t)}^2 \mkern2mu \mathrm{d} t \, .
\end{align}
Thus, to construct Proposition \ref{prop:generalization}, we only need to consider $\mathcal{L}(\hat{\mathbf{v}})-\mathcal{L}(\mathbf{v}_{\rm{F}} )$, which can be further decomposed as:
\begin{equation}\label{eq:decomposition}
\mathcal{L}(\hat{\mathbf{v}})-\mathcal{L}(\mathbf{v}_{\rm{F}} )=\underbrace{\mathcal{L}(\hat{\mathbf{v}})-\inf _{\mathbf{v} \in \mathrm{FNN}} \mathcal{L}(\mathbf{v})}_{\text {Generalization Error}}+\underbrace{\inf _{\mathbf{v} \in \mathrm{FNN}}\{\mathcal{L}(\mathbf{v})-\mathcal{L}(\mathbf{v}_{\rm{F}} )\}}_{\text {Approximation Error}} \, .
\end{equation}

To handle the Approximation Error in \eqref{eq:decomposition}, we need the following proposition, whose proof is given in Section \ref{subsec:proof of approx}.
\begin{proposition} 
\label{prop:approx}
    Let Assumptions {\rm \ref{assump:label}}--{\rm \ref{assump:Lip in y}} hold and $\varepsilon_{*}>0$ be a sufficiently small constant. Given an approximation error $\varepsilon \in (0, \varepsilon_{*})$, we choose the hypothesis class $\mathrm{FNN}=\mathrm{FNN}(L, M, J, K, \kappa, \gamma_1, \gamma_2, \gamma_3)$ with
{
\begin{align*}
&M\sim\frac{C_0^{d_x+(\alpha+1)d_y+1}d_x^{\,3(d_x+1)/2}d_y^{\,d_y/2}\{BC_y(d_x,d_y)\}^{d_y}\log^{(d_x+\alpha d_y+1)/2}\{d_x\varepsilon^{-1}(1-T)^{-1}\}}{(1-T)^{2d_x+\beta d_y+3}\varepsilon^{d_x+d_y+1}}\,,\\
&J\sim\frac{C_0^{d_x+(\alpha+1)d_y+1}d_x^{\,3(d_x+1)/2}d_y^{\,d_y/2}\{BC_y(d_x,d_y)\}^{d_y}\log^{(d_x+\alpha d_y+1)/2}\{d_x\varepsilon^{-1}(1-T)^{-1}\}}{(1-T)^{2d_x+\beta d_y+3}\varepsilon^{d_x+d_y+1}(d_x+d_y+1)^{-1}}\,,\\
&\kappa\sim B^{-1}\vee T^{-1}\vee\frac{\{Bd_y^{1/2}C_y(d_x,d_y)\vee d_x^{3/2}\}\log^{(\alpha\vee1)/2}\{d_x\varepsilon^{-1}(1-T)^{-1}\}}{\varepsilon(1-T)^{\beta\vee3}}\,,\\
&L\sim d_x+d_y+1\,,\qquad
K\sim\frac{d_x^{1/2}\log^{1/2}\{d_x\varepsilon^{-1}(1-T)^{-1}\}}{1-T}\,,\qquad
\gamma_1=\frac{d_x^{3/2}}{(1-T)^2}\,,\\
&\gamma_2\sim\frac{d_y^{1/2}C_y(d_x,d_y)\log^{\alpha/2}\{d_x\varepsilon^{-1}(1-T)^{-1}\}}{(1-T)^\beta}\,,\qquad
\gamma_3\sim\frac{d_x^{3/2}\log^{1/2}\{d_x\varepsilon^{-1}(1-T)^{-1}\}}{(1-T)^3}\,.
\end{align*}
Here $C_0\geq1$ is universal.\@
}
{Then, there} exists some $\tilde{\mathbf{v}}(\mathbf{x}, \mathbf{y}, t) \in \mathrm{FNN}(L, M, J, K, \kappa, \gamma_1, \gamma_2, \gamma_3)$ such that
$$
\|\tilde{\mathbf{v}}(\mathbf{x}, \mathbf{y}, t)-\mathbf{v}_{\rm{F}} (\mathbf{x}, \mathbf{y}, t)\|_{L^2(g_t)} \leq \varepsilon {(d_x+1)^{1/2}}
$$
for any $t \in[0, T]$ with ${0<T<1}$.
\end{proposition}

In the sequel, we select $\mathrm{FNN} = \mathrm{FNN}(L, M, J, K, \kappa, \gamma_1, \gamma_2, \gamma_3)$ in Proposition \ref{prop:approx}. By \eqref{equa:L(v)-L(v_F)}, we have
\begin{align} \label{equa:approxbound}
    \inf _{\mathbf{v} \in \mathrm{FNN}}\{\mathcal{L}(\mathbf{v})-\mathcal{L}(\mathbf{v}_{\rm{F}} )\} \leq \frac{1}{T}\int_0^T\|\tilde{\mathbf{v}}(\mathbf{x},\mathbf{y},t)-\mathbf{v}_{\mathbf{F}}(\mathbf{x},\mathbf{y},t)\|^2_{L^2(g_t)}\,{\rm d}t\leq (d_x+1)\varepsilon^2\,.
\end{align}
Now we begin to control the Generalization Error in \eqref{eq:decomposition}. Since $\hat{\mathbf{v}} \in \arg\min_{\mathbf{v} \in \mathrm{FNN}} \widehat{\mathcal{L}}(\mathbf{v})$, then 
\begin{equation} \label{equa:genesup}
\begin{aligned}
\mathcal{L}(\hat{\mathbf{v}})-\inf _{\mathbf{v} \in \mathrm{FNN}} \mathcal{L}(\mathbf{v})&= \mathcal{L}(\hat{\mathbf{v}})-\widehat{\mathcal{L}}(\hat{\mathbf{v}})+\widehat{\mathcal{L}}(\hat{\mathbf{v}})-\widehat{\mathcal{L}}(\check{\mathbf{v}})+\widehat{\mathcal{L}}(\check{\mathbf{v}})-\mathcal{L}(\check{\mathbf{v}}) \\
&\leq \mathcal{L}(\hat{\mathbf{v}})-\widehat{\mathcal{L}}(\hat{\mathbf{v}})+\widehat{\mathcal{L}}(\check{\mathbf{v}})-\mathcal{L}(\check{\mathbf{v}}) \leq 2\sup_{\mathbf{v} \in \mathrm{FNN}}|\mathcal{L}(\mathbf{v})-\widehat{\mathcal{L}}(\mathbf{v})| \, ,
\end{aligned}
\end{equation}
where $\check{\mathbf{v}} \in \arg\min_{\mathbf{v} \in \mathrm{FNN}} \mathcal{L}(\mathbf{v})$. Hence, to control the Generalization Error in \eqref{eq:decomposition}, it suffices to control $\sup_{\mathbf{v} \in \mathrm{FNN}}|\mathcal{L}(\mathbf{v})-\widehat{\mathcal{L}}(\mathbf{v})| $. Define
\begin{align} \label{equa:l,l_hat}
&\ell(\mathbf{x}, \mathbf{y}, \mathbf{v}):=\frac{1}{T}\int_0^{T}\mathbb{E}_{\mathbf{W}}\left\{\bigg|\mathbf{x}-\frac{t}{\sqrt{1-t^2}}\mathbf{W}-\mathbf{v}(t\mathbf{x}+\sqrt{1-t^2}\mathbf{W}, \mathbf{y}, t)\bigg|_2^2\right\} \mkern2mu \mathrm{d} t \, , \notag \\
& \hat{\ell}(\mathbf{x}, \mathbf{y}, \mathbf{v}) :=\frac{1}{m}\sum_{j=1}^m\bigg|\mathbf{x}-\frac{t_j}{\sqrt{1 - t_j^2}}\mathbf{W}_{j}-\mathbf{v}(t_j \mathbf{x}+\sqrt{1-t_j^2} \mathbf{W}_{j}, \mathbf{y}, t_j)\bigg|_2^2\,.
\end{align}
By \eqref{equa:population loss} and \eqref{equa:empirical loss}, we have $\mathcal{L}(\mathbf{v}) = \mathbb{E}_{\mathbf{X}, \mathbf{Y}}\{\ell(\mathbf{X}, \mathbf{Y}, \mathbf{v})\}$ and $\widehat{\mathcal{L}}(\mathbf{v}) =n^{-1}\sum_{i=1}^n \hat{\ell}(\mathbf{X}_i, \mathbf{Y}_i, \mathbf{v})$.
Introducing $\bar{\mathcal{L}}(\mathbf{v}):=n^{-1} \sum_{i=1}^n \ell(\mathbf{X}_i, \mathbf{Y}_i, \mathbf{v})$, we have
\begin{align} 
 \label{equa:gene decomp 1}
\sup_{\mathbf{v} \in \mathrm{FNN}}|\mathcal{L}(\mathbf{v})-\widehat{\mathcal{L}}(\mathbf{v})| \notag & \leq  \sup_{\mathbf{v} \in \mathrm{FNN}}|\mathcal{L}(\mathbf{v})-\bar{\mathcal{L}}(\mathbf{v})+\bar{\mathcal{L}}(\mathbf{v})-\widehat{\mathcal{L}}(\mathbf{v})| \notag  \\
&\leq \underbrace{\frac{1}{n} \sup_{\mathbf{v} \in \mathrm{FNN}}\bigg|\sum_{i=1}^n\big[\mathbb{E}_{\mathbf{X}, \mathbf{Y}}\{\ell(\mathbf{X}, \mathbf{Y}, \mathbf{v})\}-\ell(\mathbf{X}_i, \mathbf{Y}_i, \mathbf{v})\big]\bigg|}_{\text{term 1}} \notag \\
&~~~+\underbrace{\frac{1}{n}\sup_{\mathbf{v} \in \mathrm{FNN}} \bigg| \sum_{i=1}^n\big[\ell(\mathbf{X}_i, \mathbf{Y}_i, \mathbf{v})-\hat{\ell}(\mathbf{X}_i, \mathbf{Y}_i, \mathbf{v})\big]\bigg|}_{\text{term 2}} \, .
\end{align}
Define $\mathcal{H}:=\{\ell(\cdot, \cdot, \mathbf{v}): \mathbf{v} \in \mathrm{FNN}(L, M, J, K, \kappa, \gamma_1, \gamma_2, \gamma_3)\}$. Since $|\mathbf{x}|_{\infty} \leq 1$ and $\mathbf{W} \sim \mathcal{N}(\mathbf{0}, \mathbf{I}_{d_x})$, we have
    \begin{align} \label{equa:sup|l(x,y,v)|}
        \ell(\mathbf{x}, \mathbf{y}, \mathbf{v})&=\frac{1}{T}\int_0^{T}\mathbb{E}_{\mathbf{W}}\left\{\bigg|\mathbf{x}-\frac{t}{\sqrt{1-t^2}}\mathbf{W}-\mathbf{v}(t\mathbf{x}+\sqrt{1-t^2}\mathbf{W}, \mathbf{y}, t)\bigg|_2^2 \right\} \mathrm{d} t \notag \\
        &\leq \frac{2}{T}\int_0^{T}\left[\mathbb{E}_{\mathbf{W}}\left(\bigg|\mathbf{x}-\frac{t}{\sqrt{1-t^2}}\mathbf{W}\bigg|_2^2 \right) + \mathbb{E}_{\mathbf{W}}\left\{\big|\mathbf{v}  (t\mathbf{x}+\sqrt{1-t^2}\mathbf{W}, \mathbf{y}, t)\big|_2^2\right\}\right]\mathrm{d}t \notag \\
        &\leq  \frac{2{d_x}}{T} \int_0^{T}\frac{{\rm d}t}{1-t^2} + 2\sup_{\mathbf{x}, \mathbf{y}, t} |\mathbf{v}(\mathbf{x}, \mathbf{y}, t)|_2^2 \leq \frac{2{d_x}}{1-T} + 2K^2 
    \end{align}
for any function $\ell(\cdot, \cdot, \mathbf{v}) \in \mathcal{H}$ and $(\mathbf{x}, \mathbf{y}) \in[0,1]^{d_x}\times[0,B]^{d_{y}}$. 

We first bound term 1 in \eqref{equa:gene decomp 1}. To do this, we need the following two lemmas.

\begin{lemma} \label{lemma:covering number}
    Write $\mathcal{V}=\mathrm{FNN}(L, M, J, K, \kappa, \gamma_1, \gamma_2, \gamma_3)$. The covering numbers of $\mathcal{V}$ and $\mathcal{H}$ satisfy
\begin{align*} %
&~~~~~~~~\log \mathcal{N}(\delta,   \mathcal{V},\|\cdot\|_{L^{\infty}([-R, R]^{d_x} \times [0, B]^{d_{y}} \times{[0,T]})}) \leq C_1J L \log \{ L M(R \vee B \vee 1) \kappa \delta^{-1}\} \, ,\\
&\log \mathcal{N}(\delta, \mathcal{H},\|\cdot\|_{L^{\infty}([0,1]^{d_x} \times [0,B]^{d_{y}} )})\leq C_2 J L \begingroup\log \Big\{\{K+(1-T)^{-1/2}{d^{1 / 2}_x}\} L M \kappa \delta^{-1}\endgroup\\
&~~~~~~~~~~~~~~~~~~~~~~~~~~~~~~~~~\begingroup\times\bigl(B\vee\log^{1/2} [K {d^{1 / 2}_x}\delta^{-1}\{K+(1-T)^{-1/2}{d^{1 / 2}_x}\} ]\bigr)\Big\} \endgroup
\end{align*}
for some universal constants $C_1, C_2>0$, 
where $B>0$ is specified in Assumption {\rm\ref{assump:label}}. 
\end{lemma}

\begin{lemma} \label{lemma:concentration inequality}
    Let $\mathcal{F}$ be a bounded function class, i.e., there exists a constant $B$ such that for any $f \in \mathcal{F}$ and any $\mathbf{x}$ in its domain, $0 \leq f(\mathbf{x}) \leq B$. Let $\mathbf{X}, \ldots, \mathbf{X}_n \in \mathbb{R}^{d_x}$ be i.i.d. random variables. For any $\varepsilon \in(0,1)$ and $f \in \mathcal{F}$, we have
$$
\mathbb{P}\left(\frac{1}{n}\bigg| \sum_{i=1}^n\big[f(\mathbf{X}_i)-\mathbb{E}\{f(\mathbf{X})\}\big] \bigg|>\varepsilon\right)\leq 2\exp \bigg(\!-\frac{2n \varepsilon^2}{ B^2}\bigg) \, .
$$
\end{lemma}

The proof of Lemma  \ref{lemma:covering number} is given in Section \ref{subsec:covering number}, while Lemma \ref{lemma:concentration inequality} is stated as Theorem 1 of \cite{boucheron2003concentration}. Let $\{\ell(\cdot, \cdot, \mathbf{v}_k)\}_{k=1}^{N_1}$ be a $\delta$-covering of $\mathcal{H}$, i.e. for each $\ell(\cdot, \cdot, \mathbf{v}) \in \mathcal{H}$, there exists a corresponding index $k$, s.t. $\|\ell(\cdot, \cdot, \mathbf{v})-\ell(\cdot, \cdot, \mathbf{v}_k)\|_{L^{\infty}([0,1]^{d_x} \times [0,B]^{d_{y}})} \leq \delta$. Consequently, for any $\mathbf{v} \in \mathrm{FNN}$, we can assert 
$$
\begin{aligned}
&\frac{1}{n} \bigg|\sum_{i=1}^n\big[\mathbb{E}_{\mathbf{X}, \mathbf{Y}}\{\ell(\mathbf{X}, \mathbf{Y}, \mathbf{v})\}-\ell(\mathbf{X}_i, \mathbf{Y}_i, \mathbf{v})\big]\bigg| \\
&~~~~~~\leq   \max _{1\leq k \leq N_1} \frac{1}{n} \bigg|\sum_{i=1}^n\big[\mathbb{E}_{\mathbf{X}, \mathbf{Y}}\{\ell(\mathbf{X}, \mathbf{Y}, \mathbf{v}_k)\}-\ell(\mathbf{X}_i, \mathbf{Y}_i, \mathbf{v}_k)\big] \bigg|+2 \delta \,.
\end{aligned}
$$
Taking supremum over $\mathcal{H}$ on both sides, we have
$$
\begin{aligned}
&\sup _{\ell \in \mathcal{H}} \frac{1}{n} \bigg|\sum_{i=1}^n\big[\mathbb{E}_{\mathbf{X}, \mathbf{Y}}\{\ell(\mathbf{X}, \mathbf{Y}, \mathbf{v})\}-\ell(\mathbf{X}_i, \mathbf{Y}_i, \mathbf{v})\big] \bigg| \\
&~~~~~~\leq \max _{1\leq k \leq N_1} \frac{1}{n} \bigg|\sum_{i=1}^n\big[\mathbb{E}_{\mathbf{X}, \mathbf{Y}}\{\ell(\mathbf{X}, \mathbf{Y}, \mathbf{v}_k)\}-\ell(\mathbf{X}_i, \mathbf{Y}_i, \mathbf{v}_k)\big]\bigg|+2 \delta \, .
\end{aligned}
$$
Thus, it holds that
\begin{align*}
&\mathbb{P}\left(\sup _{\ell \in \mathcal{H}} \frac{1}{n} \bigg|\sum_{i=1}^n\big[\mathbb{E}_{\mathbf{X}, \mathbf{Y}}\{\ell(\mathbf{X}, \mathbf{Y}, \mathbf{v})\}-\ell(\mathbf{X}_i, \mathbf{Y}_i, \mathbf{v})\big]\bigg|>\varepsilon_1+2 \delta\right)\\
&~~~~~~ \leq \mathbb{P}\left(\max _{1\leq k \leq N_1} \frac{1}{n} \bigg|\sum_{i=1}^n\big[\mathbb{E}_{\mathbf{X}, \mathbf{Y}}\{\ell(\mathbf{X}, \mathbf{Y}, \mathbf{v}_k)\}-\ell(\mathbf{X}_i, \mathbf{Y}_i, \mathbf{v}_k)\big] \bigg|>\varepsilon_1 \right) \\
&~~~~~~\leq \sum_{k=1}^{N_1} \mathbb{P}\left(\frac{1}{n} \bigg|\sum_{i=1}^n\big[\mathbb{E}_{\mathbf{X}, \mathbf{Y}}\{\ell(\mathbf{X}, \mathbf{Y}, \mathbf{v}_k)\}-\ell(\mathbf{X}_i, \mathbf{Y}_i, \mathbf{v}_k)\big] \bigg|>\varepsilon_1 \right) \, .
\end{align*}
By \eqref{equa:sup|l(x,y,v)|}, $\mathcal{H}$ is a bounded function class with boundedness constant $B_{\mathcal{H}}=2d_x(1-T)^{-1}+2K^2$. Applying Lemma \ref{lemma:concentration inequality}, we have
$$
\mathbb{P}\left(\sup _{\ell \in \mathcal{H}} \bigg|\frac{1}{n} \sum_{i=1}^n\big[\mathbb{E}_{\mathbf{X}, \mathbf{Y}}\{\ell(\mathbf{X}, \mathbf{Y}, \mathbf{v})\}-\ell(\mathbf{X}_i, \mathbf{Y}_i, \mathbf{v})\big] \bigg|>\varepsilon_1+2 \delta \right) \leq 2N_1 \exp \bigg(\!-\frac{2n \varepsilon_1^2}{ {B^2_{\mathcal{H}}}}\bigg) \, ,
$$ 
Letting $\varepsilon_1 = [2^{-1}n^{-1}B^{2}_{\mathcal{H}} \log (2{\delta}^{-1} N_1)]^{1/2}$ , we get with probability at most $\delta$,
\begin{equation} \label{equa:gene decomp 1, rough probability estimation}
\begin{aligned}
\sup _{\ell \in \mathcal{H}} \bigg|\frac{1}{n} \sum_{i=1}^n\big[\mathbb{E}_{\mathbf{X}, \mathbf{Y}}\{\ell(\mathbf{X}, \mathbf{Y}, \mathbf{v})\}-\ell(\mathbf{X}_i, \mathbf{Y}_i, \mathbf{v})\big] \bigg|> \sqrt{\frac{ {B_{\mathcal{H}}^2} \log (2N_1 / \delta)}{2n}}+2 \delta \, .
\end{aligned}
\end{equation}
Then we complete the analysis of term 1 in \eqref{equa:gene decomp 1}.

To bound term 2 in \eqref{equa:gene decomp 1}, we need some additional truncation arguments. More specifically, we define
\begin{align} \label{equa:r_D,l_D}
    &r_{D}  (\mathbf{w}, t, \mathbf{x}, \mathbf{y}, \mathbf{v}):=\bigg|\mathbf{x}-\frac{t}{\sqrt{1-t^2}}\mathbf{w}-\mathbf{v}(t \mathbf{x}+\sqrt{1-t^2} \mathbf{w}, \mathbf{y},  t)\bigg|_2^2 \mathbb{I}(|\mathbf{w}|_{\infty} \leq D) \notag \, ,\\
    &\ell_{D}(\mathbf{x}, \mathbf{y}, \mathbf{v}):=\mathbb{E}_{t, \mathbf{W}}\{r_{D}(\mathbf{W}, t, \mathbf{x}, \mathbf{y}, \mathbf{v})\} \, , \quad  \hat{\ell}_{D}(\mathbf{x}, \mathbf{y}, \mathbf{v}):=\frac{1}{m} \sum_{j=1}^m r_{D}  (\mathbf{W}_j, t_j, \mathbf{x}, \mathbf{y}, \mathbf{v})\,,
\end{align}
with some $D>0$ determined later. Setting $\Delta = \Delta_1 + \Delta_2 + \Delta_3$, term 2 has the following decomposition:
\begin{align} \label{equa:gene decomp 2, probability version}
        &\mathbb{P}\left[\frac{1}{n}\sup_{\mathbf{v} \in \mathrm{FNN}}\bigg| \sum_{i=1}^n \big\{\ell(\mathbf{X}_i, \mathbf{Y}_i, \mathbf{v})-\hat{\ell}(\mathbf{X}_i, \mathbf{Y}_i, \mathbf{v})\big\} \bigg|>\Delta\right] \notag \\
        &~~~~~~\leq  \underbrace{\mathbb{P}\bigg\{\frac{1}{n}\sum_{i=1}^n\sup_{\mathbf{v} \in \mathrm{FNN}}\big|\ell(\mathbf{X}_i, \mathbf{Y}_i, \mathbf{v})-\ell_{D}(\mathbf{X}_i, \mathbf{Y}_i, \mathbf{v})\big|>\Delta_1 \bigg\}}_{\text {Truncation Error (I) }} \notag \\        &~~~~~~~~~+\underbrace{\sum_{i=1}^n\mathbb{P}\bigg\{\sup_{\mathbf{v} \in \mathrm{FNN}}\big|\ell_{D}(\mathbf{X}_i, \mathbf{Y}_i, \mathbf{v})-\hat{\ell}_{D}(\mathbf{X}_i, \mathbf{Y}_i, \mathbf{v})\big|>\Delta_2\bigg\}}_{\text {Statistical Error }} \notag \\        &~~~~~~~~~+\underbrace{\sum_{i=1}^n\mathbb{P}\bigg\{ \sup_{\mathbf{v} \in \mathrm{FNN}}\big|\hat{\ell}(\mathbf{X}_i, \mathbf{Y}_i, \mathbf{v})-\hat{\ell}_{D}(\mathbf{X}_i, \mathbf{Y}_i, \mathbf{v})\big|>\Delta_3\bigg\}}_{\text {Truncation Error (II) }} \, .
    \end{align}
We first control Truncation Error (I) in \eqref{equa:gene decomp 2, probability version}. By \eqref{equa:l,l_hat} and \eqref{equa:r_D,l_D}, we have
\begin{align*}
&\ell(\mathbf{X}_i, \mathbf{Y}_i, \mathbf{v})-\ell_{D}(\mathbf{X}_i, \mathbf{Y}_i, \mathbf{v})  \\
&~~~=\mathbb{E}_{t, \mathbf{W}} \! \left\{ \bigg|\mathbf{X}_i-\frac{t}{\sqrt{1-t^2}}\mathbf{W}-\mathbf{v}  \big(t \mathbf{X}_i+\sqrt{1-t^2} \mathbf{W}, \mathbf{Y}_i, t\big)\bigg|_2^2 \mathbb{I}(|\mathbf{W}|_{\infty}>D)\right\}  \\
 &~~~\leq \mathbb{E}_t\left[\mathbb{E}^{1/2}_{\mathbf{W}}\left\{\bigg|\mathbf{X}_i-\frac{t}{\sqrt{1-t^2}}\mathbf{W}-\mathbf{v}  \big(t \mathbf{X}_i+\sqrt{1-t^2} \mathbf{W},\mathbf{Y}_i, t\big)\bigg|_2^4 \right\} \mathbb{P}^{1/2}(|\mathbf{W}|_{\infty}>D)\right] \, . 
\end{align*}
Since $|\mathbf{v}(\mathbf{x}, \mathbf{y}, t)|_2 \leq K$ for any $\mathbf{v} \in \mathrm{FNN}(L, M, J, K, \kappa, \gamma_1, \gamma_2, \gamma_3)$, $\mathbf{W} \sim \mathcal{N}(\mathbf{0}, \mathbf{I}_{d_x})$ and $|\mathbf{X}_i|_{\infty}\leq1$ by Assumption \ref{assump:bounded support}, {using Minkowski inequality,} it holds that
    \begin{align*}
        &\mathbb{E}_{\mathbf{W}}^{1/2}\left\{\bigg|\mathbf{X}_i-\frac{t}{\sqrt{1-t^2}}\mathbf{W}-\mathbf{v}  \big(t \mathbf{X}_i+\sqrt{1-t^2} \mathbf{W},\mathbf{Y}_i, t\big)\bigg|_2^4 \right\} \\
        &~~~~\leq  {2 \Bigg[\mathbb{E}_{\mathbf{W}}^{1/2}\Bigg( \bigg|\mathbf{X}_i-\frac{t}{\sqrt{1-t^2}}\mathbf{W}\bigg|_2^4\Bigg)+\mathbb{E}_{\mathbf W}^{1/2}\Bigg\{\bigg|\mathbf{v} \big(t \mathbf{X}_i+\sqrt{1-t^2} \mathbf{W}, \mathbf{Y}_i, t\big)\bigg|_2^4\Bigg\}\Bigg]} \\
        &~~~~\leq  2\bigg\{4|\mathbf{X}_i|_2^4+\frac{4t^4}{(1-t^2)^2}\mathbb{E}(|\mathbf{W}|_2^4)\bigg\}^{1 / 2}\begingroup+2K^2\endgroup \leq  2\bigg\{4 {d^2_x} + \frac{4t^4}{(1-t^2)^2}{d_x}({d_x}+2)\bigg\}^{1/2}\begingroup+2K^2\endgroup \\
        &~~~~\leq  2\bigg\{\frac{8({d_x}+2)^2}{(1-t^2)^2}\bigg\}^{1/2}\begingroup+2K^2\endgroup \leq \frac{4\sqrt{2}({d_x}+2)}{1-t^2}+2K^2 \, .
    \end{align*}
Write $\mathbf{W} = (W_1, \ldots, W_{d_x})^{\rm T}$. It holds that
$$
\mathbb{P}(|\mathbf{W}|_{\infty}>D)
\leq \sum_{k=1}^{d_x} \mathbb{P}\big(|W_{k}|>D\big)
 \leq 2 {d_x} \exp \bigg(\!-\frac{D^2}{2}\bigg) \, .
$$
Combining above estimations and {noticing} $t\leq T<1$, we have
\begin{align*}
\ell(\mathbf{X}_i, \mathbf{Y}_i, \mathbf{v})-\ell_{D}(\mathbf{X}_i, \mathbf{Y}_i, \mathbf{v}) \leq \bigg\{\frac{8({d_x}+2)^{3/2}}{1-T}+4{d^{1/2}_x}K^2\bigg\}  \exp \bigg(\!-\frac{D^2}{4}\bigg) \, .
\end{align*}
Letting $\Delta_1=\{8({d_x}+2)^{3/2}(1-T)^{-1}+4{d^{1/2}_x}K^2\} \exp (-D^2/4)$, it then holds that
\begin{align} \label{equa:geneterm2trunc1}
\mathbb{P}\bigg\{\frac{1}{n}\sum_{i=1}^n\sup_{\mathbf{v} \in \mathrm{FNN}}\big|\ell(\mathbf{X}_i, \mathbf{Y}_i, \mathbf{v})-\ell_{D}(\mathbf{X}_i, \mathbf{Y}_i, \mathbf{v})\big|>\Delta_1 \bigg\} = 0 \, .
\end{align}
Next, we deal with Truncation Error (II) in \eqref{equa:gene decomp 2, probability version}. We have
\begin{align} \label{equa:condiexp1}
&\sum_{i=1}^n\mathbb{P}\bigg\{ \sup_{\mathbf{v} \in \mathrm{FNN}}\big|\hat{\ell}(\mathbf{X}_i, \mathbf{Y}_i, \mathbf{v})-\hat{\ell}_{D}(\mathbf{X}_i, \mathbf{Y}_i, \mathbf{v})\big|>\Delta_3\bigg\} \notag \\
&~~~~=\sum_{i=1}^n\mathbb{E}_{\mathbf{X}_i, \mathbf{Y}_i}\bigg[\mathbb{P}\bigg\{ \sup_{\mathbf{v} \in \mathrm{FNN}}\big|\hat{\ell}(\mathbf{X}_i, \mathbf{Y}_i, \mathbf{v})-\hat{\ell}_{D}(\mathbf{X}_i, \mathbf{Y}_i, \mathbf{v})\big|>\Delta_3 \mkern2mu \Big| \mkern2mu \mathbf{X}_i, \mathbf{Y}_i \bigg\}\bigg] \notag \\
&~~~~=\sum_{i=1}^n\mathbb{E}_{\mathbf{X}_i, \mathbf{Y}_i}\bigg[\mathbb{P}\bigg\{ \sup_{\mathbf{v} \in \mathrm{FNN}}\big|\hat{\ell}(\mathbf{x}, \mathbf{y}, \mathbf{v})-\hat{\ell}_{D}(\mathbf{x}, \mathbf{y}, \mathbf{v})\big|>\Delta_3 \bigg\}\Big|_{(\mathbf{x}, \mathbf{y})=(\mathbf{X}_i, \mathbf{Y}_i)}\bigg] \, .
\end{align}
By \eqref{equa:l,l_hat} and \eqref{equa:r_D,l_D}, it holds that
\begin{align*}
    &\hat{\ell}(\mathbf{x}, \mathbf{y}, \mathbf{v})-\hat{\ell}_{D}(\mathbf{x}, \mathbf{y}, \mathbf{v}) \\
    &~~~~~~=\frac{1}{m} \sum_{j=1}^m \bigg|\mathbf{x}-\frac{{t_j}}{\sqrt{1-{t_j^2}}}\mathbf{W}_j-\mathbf{v}({t_j} \mathbf{x}+\sqrt{1-{t_j^2}} \mathbf{W}_j, \mathbf{y},  {t_j})\bigg|_2^2 \mathbb{I}(|\mathbf{W}_j|_{\infty} > D) \, .
\end{align*}
Thus, if $|\mathbf{W}_j|_{\infty} \leq D$ for all $j=1, \ldots, m$, we have
$$
\sup_{\mathbf{v} \in \mathrm{FNN}}\big|\hat{\ell}(\mathbf{x}, \mathbf{y}, \mathbf{v})-\hat{\ell}_{D}(\mathbf{x}, \mathbf{y}, \mathbf{v})\big|=0 \, .
$$
Since $\mathbf{W}_j \sim \mathcal{N}(\mathbf{0}, \mathbf{I}_{d_x})$, this implies
\begin{align} \label{equa:truncerr2}
&\mathbb{P}\bigg\{\sup_{\mathbf{v} \in \mathrm{FNN}}\big|\hat{\ell}(\mathbf{x}, \mathbf{y}, \mathbf{v})-\hat{\ell}_{D}(\mathbf{x}, \mathbf{y}, \mathbf{v})\big|>0\bigg\} \notag \\
&~~~~~~~~~~~~~~~~~~~~~~~~ \leq \sum_{j=1}^m \mathbb{P}(|\mathbf{W}_j|_{\infty} > D) \leq2 m {d_x} \exp \bigg(\!-\frac{D^2}{2}\bigg) \, .
\end{align}
Combining \eqref{equa:condiexp1} and \eqref{equa:truncerr2}, and letting $\Delta_3=0$, it holds that
\begin{equation} \label{equa:geneterm2trunc2}
\sum_{i=1}^n\mathbb{P}\bigg\{ \sup_{\mathbf{v} \in \mathrm{FNN}}\big|\hat{\ell}(\mathbf{X}_i, \mathbf{Y}_i, \mathbf{v})-\hat{\ell}_{D}(\mathbf{X}_i, \mathbf{Y}_i, \mathbf{v})\big|>\Delta_3\bigg\} \leq 2nm{d_x}\exp \bigg(\!-\frac{D^2}{2}\bigg) \, .
\end{equation}
Finally, we control the Statistical Error in \eqref{equa:gene decomp 2, probability version}. We have
\begin{align} \label{equa:condiexp2}
&\sum_{i=1}^n\mathbb{P}\bigg\{ \sup_{\mathbf{v} \in \mathrm{FNN}}\big|\ell_{D}(\mathbf{X}_i, \mathbf{Y}_i, \mathbf{v})-\hat{\ell}_{D}(\mathbf{X}_i, \mathbf{Y}_i, \mathbf{v})\big|>\Delta_2\bigg\} \notag \\
&~~~~=\sum_{i=1}^n\mathbb{E}_{\mathbf{X}_i, \mathbf{Y}_i}\bigg[\mathbb{P}\bigg\{ \sup_{\mathbf{v} \in \mathrm{FNN}}\big|{\ell}_{D}(\mathbf{X}_i, \mathbf{Y}_i, \mathbf{v})-\hat{\ell}_{D}(\mathbf{X}_i, \mathbf{Y}_i, \mathbf{v})\big|>\Delta_2 \mkern2mu \Big| \mkern2mu \mathbf{X}_i, \mathbf{Y}_i \bigg\}\bigg] \notag \\
&~~~~=\sum_{i=1}^n\mathbb{E}_{\mathbf{X}_i, \mathbf{Y}_i}\bigg[\mathbb{P}\bigg\{ \sup_{\mathbf{v} \in \mathrm{FNN}}\big|{\ell}_{D}(\mathbf{x}, \mathbf{y}, \mathbf{v})-\hat{\ell}_{D}(\mathbf{x}, \mathbf{y}, \mathbf{v})\big|>\Delta_2 \bigg\}\Big|_{(\mathbf{x}, \mathbf{y})=(\mathbf{X}_i, \mathbf{Y}_i)}\bigg] \, .
\end{align}
By \eqref{equa:r_D,l_D}, it holds that
\begin{align} \label{equa:ld-ld_hat}
    &{\ell}_{D}(\mathbf{x}, \mathbf{y}, \mathbf{v})-\hat{\ell}_{D}(\mathbf{x}, \mathbf{y}, \mathbf{v})\\
    &~~~~~~~~~~~~= \mathbb{E}_{t, \mathbf{W}}\big\{r_{D} (\mathbf{W}, t, \mathbf{x}, \mathbf{y}, \mathbf{v})\big\}-\frac{1}{m} \sum_{j=1}^m r_{D}  (\mathbf{W}_j, t_j, \mathbf{x}, \mathbf{y}, \mathbf{v}) \, . \notag
\end{align}
For any fixed $(\mathbf{x}, \mathbf{y}) \in [0, 1]^{d_x} \times [0, B]^{d_y}$, we define
$$
\mathcal{R}^{D}_{\mathbf{x}, \mathbf{y}}:=\{r_{D} (\cdot, \cdot, \mathbf{x}, \mathbf{y}, \mathbf{v}): \mathbf{v} \in \mathrm{FNN}(L, M, J, K, \kappa, \gamma_1, \gamma_2, \gamma_3)\} \, .
$$
Denote by $\mathbb{I}_D(\mathbf{w}):=\mathbb{I}(|\mathbf{w}|_{\infty}\leq D)$. Since $0\leq t\leq T <1$, we have
\begin{align} \label{equa:upper bound of r_D}
    &r_{D} (\mathbf{w}, t, \mathbf{x}, \mathbf{y}, \mathbf{v})=\bigg|\mathbf{x}-\frac{t}{\sqrt{1-t^2}}\mathbf{w}-\mathbf{v}(t \mathbf{x}+\sqrt{1-t^2} \mathbf{w}, \mathbf{y},  t)\bigg|_2^2 \mathbb{I}_D(\mathbf{w})  \\
    &~~~~\leq  4|\mathbf{x}|_2^2 + \frac{4}{1-T}|\mathbf{w}|_2^2 \, \mathbb{I}_D(\mathbf{w})+2 \sup_{\mathbf{x}, \mathbf{y}, t}|\mathbf{v}(\mathbf{x}, \mathbf{y}, t)|_2^2 \leq \frac{4d_x(D+1)^2}{1-T}+2K^2 \notag
\end{align}
for any $r_D(\cdot, \cdot, \mathbf{x}, \mathbf{y}, \mathbf{v}) \in \mathcal{R}^{D}_{\mathbf{x}, \mathbf{y}}$ and $(\mathbf{w}, t) \in \mathbb{R}^{d_x}\times[0, T]$.

Note when $|\mathbf{w}|_{\infty} \leq D$ and $|\mathbf{x}|_{\infty}\leq1$, we have $t \mathbf{x}+\sqrt{1-t^2} \mathbf{w} \in[-D-1, D+1]^{d_x}$. Denote by $\mathbf{w}_t=t \mathbf{x}+\sqrt{1-t^2}\mathbf{w}$ and $\mathcal{K}_D=[-D-1, D+1]^{d_x} \times [0,B]^{d_{y}} \times[0, T]$. Let $\{\mathbf{v}_i\}_{i=1}^{N_2}$ be a $G(\delta)$-covering of $\mathrm{FNN}$ w.r.t. $\|\cdot\|_{L^{\infty}(\mathcal{K}_D)}$, i.e. for any $\mathbf{v} \in \mathrm{FNN}$, there exists a corresponding index $k$, s.t. $\|\mathbf{v}-\mathbf{v}_k\|_{L^{\infty}(\mathcal{K}_D)} \leq G(\delta)$. For any fixed $(\mathbf{x}, \mathbf{y}) \in [0, 1]^{d_x} \times [0, B]^{d_y}$, it holds that 
\begin{align} \label{equa:delta r_D}
&|r_{D}  (\mathbf{w}, t, \mathbf{x}, \mathbf{y}, \mathbf{v}_k)-r_{D}  (\mathbf{w}, t, \mathbf{x}, \mathbf{y} ,\mathbf{v})| \notag \\
&~~~=  \bigg|\bigg\langle \bigg\{2\mathbf{x}-\frac{2t}{\sqrt{1-t^2}} \mathbf{w}-\mathbf{v}_k(\mathbf{w}_t, \mathbf{y}, t)-\mathbf{v}(\mathbf{w}_t, \mathbf{y}, t)\bigg\}\mathbb{I}_D(\mathbf{w}), \notag \\
& ~~~~~~~~~~~~~~~~~~~~~~~~~~~~~~~~~~~~~~~~~~~\{\mathbf{v}_k(\mathbf{w}_t, \mathbf{y}, t)-\mathbf{v}(\mathbf{w}_t, \mathbf{y}, t)\}\mathbb{I}_D(\mathbf{w})\bigg\rangle\bigg|  \\
&~~~\leq  \bigg| \bigg\{2\mathbf{x}-\frac{2t}{\sqrt{1-t^2}} \mathbf{w}-\mathbf{v}_k(\mathbf{w}_t, \mathbf{y}, t)-\mathbf{v}(\mathbf{w}_t, \mathbf{y}, t)\bigg\}\mathbb{I}_D(\mathbf{w})\bigg|_2 \notag \\
& ~~~~~~~~~~~~~~~~~~~~~~~~~~~~~~~~~~~~~~~~~~~ \cdot \big|\{\mathbf{v}_k(\mathbf{w}_t, \mathbf{y}, t)-\mathbf{v}(\mathbf{w}_t, \mathbf{y}, t)\}\mathbb{I}_D(\mathbf{w})\big|_2 \notag \\
&~~~\leq  \bigg\{2|\mathbf{x}|_2+\frac{2t}{\sqrt{1-t^2}}|\mathbf{w}|_2 \,\mathbb{I}_D(\mathbf{w})+|\mathbf{v}(\mathbf{w}_t, \mathbf{y}, t)|_2+|\mathbf{v}_k(\mathbf{w}_t, \mathbf{y}, t)|_2\bigg\} \|\mathbf{v}-\mathbf{v}_k\|_{L^{\infty}(\mathcal{K}_D)} \notag \\
&~~~\leq  \{2 (1-T)^{-1/2}{d^{1/2}_x}(D+1)+2 K\} \cdot G(\delta) \notag \, .
\end{align}
for any $(\mathbf{w}, t) \in \mathbb{R}^{d_x}\times[0, T]$. Set $G(\delta)=\{2 (1-T)^{-1/2}{d^{1/2}_x}(D+1)+2 K\}^{-1}\delta$. By \eqref{equa:delta r_D}, we can assert that a $G(\delta)$-covering of $\mathrm{FNN}$ w.r.t. $\|\cdot\|_{L^{\infty}(\mathcal{K}_D)}$ induces a $\delta$-covering of $\mathcal{R}^{D}_{\mathbf{x}, \mathbf{y}}$ w.r.t. $\|\cdot\|_{L^{\infty}(\mathbb{R}^{d_x}\times[0, T])}$.

By \eqref{equa:ld-ld_hat}, for any $\mathbf{v} \in \mathrm{FNN}$, we have
$$
\begin{aligned}
& \big|\ell_{D}  (\mathbf{x}, \mathbf{y}, \mathbf{v})-\hat{\ell}_{D}  (\mathbf{x}, \mathbf{y}, \mathbf{v})\big| \\
&~~~~~~=  
\bigg|\mathbb{E}_{t, \mathbf{W}}\big\{r_{D}  (\mathbf{W}, t, \mathbf{x}, \mathbf{y}, \mathbf{v})\big\}-\frac{1}{m} \sum_{j=1}^m r_{D}  (\mathbf{W}_j, t_j, \mathbf{x}, \mathbf{y}, \mathbf{v})\bigg| \\
&~~~~~~\leq  \max _{1 \leq k \leq N_2}\bigg|\mathbb{E}_{t, \mathbf{W}}\big\{r_{D}  (\mathbf{W}, t, \mathbf{x}, \mathbf{y}, \mathbf{v}_k)\big\}-\frac{1}{m} \sum_{j=1}^m r_{D}  (\mathbf{W}_j, t_j, \mathbf{x}, \mathbf{y}, \mathbf{v}_k)\bigg|+2 \delta \, .
\end{aligned}
$$
Taking supremum on both sides, it holds that
$$
\begin{aligned}
& \sup _{\mathbf{v} \in \mathrm{FNN}}\big|\ell_{D}  (\mathbf{x}, \mathbf{y}, \mathbf{v})-\hat{\ell}_{D}  (\mathbf{x}, \mathbf{y}, \mathbf{v})| \\
&~~~~~~\leq  2 \delta+\max _{1 \leq k \leq N_2}\bigg|\mathbb{E}_{t, \mathbf{W}}\big\{r_{D}  (\mathbf{W}, t, \mathbf{x}, \mathbf{y}, \mathbf{v}_k)\big\}-\frac{1}{m} \sum_{j=1}^m r_{D}  (\mathbf{W}_j, t_j, \mathbf{x}, \mathbf{y}, \mathbf{v}_k)\bigg| \, .
\end{aligned}
$$
Thus, we have
\begin{align} \label{equa:boundsupl_D}
& \mathbb{P}\bigg\{\sup _{\mathbf{v} \in \mathrm{FNN}}\big|\ell_{D}  (\mathbf{x}, \mathbf{y}, \mathbf{v})-\hat{\ell}_{D}  (\mathbf{x}, \mathbf{y}, \mathbf{v})\big|>\varepsilon_2+2 \delta \bigg\} \\
 &~~~~~\mathrel{\leq} \mathbb{P}\left[\max _{1 \leq k \leq N_2}\bigg|\mathbb{E}_{t, \mathbf{W}}\big\{r_{D}  (\mathbf{W}, t, \mathbf{x}, \mathbf{y}, \mathbf{v}_k)\big\}-\frac{1}{m} \sum_{j=1}^m r_{D}  (\mathbf{W}_j, t_j, \mathbf{x}, \mathbf{y}, \mathbf{v}_k)\bigg|>\varepsilon_2\right] \notag \\
&~~~~~\leq \sum_{k=1}^{N_2}\mathbb{P}\left[\bigg|\mathbb{E}_{t, \mathbf{W}}\big\{r_{D}  (\mathbf{W}, t, \mathbf{x}, \mathbf{y}, \mathbf{v}_k)\big\}-\frac{1}{m} \sum_{j=1}^m r_{D}  (\mathbf{W}_j, t_j, \mathbf{x}, \mathbf{y}, \mathbf{v}_k)\bigg|>\varepsilon_2\right] \,. \notag
\end{align}
By \eqref{equa:upper bound of r_D}, for any fixed $(\mathbf{x}, \mathbf{y}) \in [0, 1]^{d_x} \times [0, B]^{d_y}$, $\mathcal{R}^D_{\mathbf{x}, \mathbf{y}}$ is a bounded function class with boundedness constant $B_D=4 {d_x}(D+1)^2(1-T)^{-1}+2K^2$. Applying Lemma \ref{lemma:concentration inequality}, we obtain
\begin{align} \label{equa:concentration r_D}
&\mathbb{P}\left[\bigg|\mathbb{E}_{t, \mathbf{W}}\big\{r_{D}  (\mathbf{W}, t, \mathbf{x}, \mathbf{y}, \mathbf{v}_k)\big\}-\frac{1}{m} \sum_{j=1}^m r_{D}  (\mathbf{W}_j, t_j, \mathbf{x}, \mathbf{y}, \mathbf{v}_k)\bigg|  >\varepsilon_2 \right] \notag \\
&~~~~~~ \leq 2 \exp \bigg(\!-\frac{2m \varepsilon_2^2}{B_D^2}\bigg)\,.
\end{align}
Let $\Delta_2=\varepsilon_2+2\delta$. Combining \eqref{equa:condiexp2}, \eqref{equa:boundsupl_D} and \eqref{equa:concentration r_D}, we have
\begin{align} \label{equa:geneterm2sta}
\sum_{i=1}^n\mathbb{P}\bigg\{\sup _{\mathbf{v} \in \mathrm{FNN}}\big|\ell_{D} 
 (\mathbf{X}_i, \mathbf{Y}_i, \mathbf{v})-\hat{\ell}_{D} 
 (\mathbf{X}_i, \mathbf{Y}_i, \mathbf{v})\big|>\Delta_2 \bigg\} \leq  2nN_2 \exp \bigg(\!-\frac{2m \varepsilon_2^2}{B_D^2}\bigg) \,.
\end{align}

Now return to \eqref{equa:gene decomp 2, probability version}. Combining \eqref{equa:geneterm2trunc1}, \eqref{equa:geneterm2trunc2} and \eqref{equa:geneterm2sta}, by \eqref{equa:gene decomp 2, probability version}, it holds that, with probability at most $2nm{d_x}\exp (-2^{-1}D^2)+2nN_2 \exp (-2m \varepsilon_2^2B_D^{-2})$,
\begin{align}
    &\frac{1}{n}\sup_{\mathbf{v} \in \mathrm{FNN}}\bigg| \sum_{i=1}^n \big\{\ell(\mathbf{X}_i, \mathbf{Y}_i, \mathbf{v})-\hat{\ell}(\mathbf{X}_i, \mathbf{Y}_i, \mathbf{v})\big\} \bigg| \notag \\    &~~~~~~~~>\varepsilon_2+2\delta+\bigg\{\frac{8({d_x}+2)^{3/2}}{1-T}+4{d^{1/2}_x}K^2\bigg\}  \exp \bigg(\!-\frac{D^2}{4}\bigg) \, .
\end{align}
Recall $B_D=4 {d_x}(D+1)^2(1-T)^{-1}+2K^2$. Letting
$$
2nm{d_x}\exp \bigg(\!-\frac{D^2}{2}\bigg) = \delta \, ,\quad 2nN_2 \exp \bigg(\!-\frac{2m \varepsilon_2^2}{B_D^2}\bigg)=\delta
\, ,$$
we have that, with probability at most $2\delta$,
\begin{equation} \label{equa:gene decomp 1, term 2 probability estimation}
\begin{aligned}
    &\frac{1}{n}\sup_{\mathbf{v} \in \mathrm{FNN}}\bigg| \sum_{i=1}^n \big\{\ell(\mathbf{X}_i, \mathbf{Y}_i, \mathbf{v})-\hat{\ell}(\mathbf{X}_i, \mathbf{Y}_i, \mathbf{v})\big\} \bigg| \\
    &~~~~~~>\bigg[\frac{4 {d_x}\{\sqrt{2}\log^{1/2}(2nmd_x\delta^{-1})+1\}^2}{1-T}+2K^2\bigg]\sqrt{\frac{\log(2n\delta^{-1}N_2)}{2m}}+2\delta \\
    &~~~~~~~~~+\bigg\{\frac{8({d_x}+2)^{3/2}}{1-T}+4{d^{1/2}_x}K^2\bigg\} \cdot \sqrt{\frac{\delta}{2nm{d_x}}} \, .
\end{aligned}
\end{equation}
Combining \eqref{equa:gene decomp 1, rough probability estimation} and \eqref{equa:gene decomp 1, term 2 probability estimation}, by \eqref{equa:gene decomp 1} and \eqref{equa:genesup}, it holds that, with probability at least $1-3\delta$,
\begin{align} \label{equa:final gene error}
        \mathcal{L}(\hat{\mathbf{v}})- \inf_{\mathbf{v} \in \mathrm{FNN}}\mathcal{L}({\mathbf{v}}) 
        \leq & \, \bigg[\frac{2 {d_x}\{\sqrt{2}\log^{1/2}(2nmd_x\delta^{-1})+1\}^2}{1-T}+K^2\bigg]\sqrt{\frac{8\log(2n\delta^{-1}N_2)}{m}} \notag \\
    &+\bigg\{\frac{2({d_x}+2)^{3/2}}{1-T}+{d^{1/2}_x}K^2\bigg\} \cdot \sqrt{\frac{32\delta}{mn{d_x}}} \notag \\
    & + \sqrt{\frac{2 {B^2_{\mathcal{H}}} \log (2N_1 \delta^{-1})}{n}} +8\delta  \, .
\end{align}
Recall that $\{\ell(\cdot, \cdot, \mathbf{v}_i)\}_{i=1}^{N_1}$ is a $\delta$-covering of $\mathcal{H}$ w.r.t. $\|\cdot\|_{L^{\infty}([0, 1]^{d_x} \times [0, B]^{d_y})}$, $\{\mathbf{v}_j\}_{j=1}^{N_2}$ is a $\{2 (1-T)^{-1/2}{d^{1/2}_x}(D+1)+2 K\}^{-1}\delta$-covering of $\mathrm{FNN}$ w.r.t. $\|\cdot\|_{L^{\infty}(\mathcal{K}_D)}$ where $\mathcal{K}_D=[-D-1, D+1]^{d_x} \times [0,B]^{d_{y}} \times[0, T]$ and $2nm{d_x}\exp (-2^{-1}D^2)=\delta$.
By Proposition \ref{prop:approx} and Lemma \ref{lemma:covering number}, it holds that
\begin{align*}
&\log N_1\leq
\frac{C_3{C_0^{d_x+(\alpha+1)d_y+1}d_x^{3(d_x+1)/2}d_y^{d_y/2}}\{BC_y(d_x,d_y)\}^{d_y}\log^{(d_x+\alpha d_y+1)/2}\{d_x\varepsilon^{-1}(1-T)^{-1}\}}
{(1-T)^{2d_x+\beta d_y+3}\varepsilon^{d_x+d_y+1}}\\
&~~~~~~~~~~~~~\qquad\times {(d_x+d_y+1)^2}\bigg[\log\frac{1}{\delta}
+{(1+\alpha+\beta)(d_x+d_y+1)}\\
&~~~~~~~~~~~~~~~~~~~~~~~~~~\qquad\qquad\qquad {\times\log\bigg\{\frac{d_x+d_y+B+B^{-1}+C_y(d_x,d_y)}{\varepsilon(1-T)}\bigg\}}\bigg] \, ,\\
&\log N_2\leq
\frac{C_4{C_0^{d_x+(\alpha+1)d_y+1}d_x^{3(d_x+1)/2}d_y^{d_y/2}}\{BC_y(d_x,d_y)\}^{d_y}\log^{(d_x+\alpha d_y+1)/2}\{d_x\varepsilon^{-1}(1-T)^{-1}\}}
{(1-T)^{2d_x+\beta d_y+3}\varepsilon^{d_x+d_y+1}}\\
&~~~~~~~~~~~~~\qquad\times {(d_x+d_y+1)^2}\bigg[\log\frac{1}{\delta}
+{(1+\alpha+\beta)(d_x+d_y+1)}\\
&~~~~~~~~~~~~~~~~~\qquad\qquad\qquad {\times\log\bigg\{\frac{d_x+d_y+B+B^{-1}+C_y(d_x,d_y)}{\varepsilon(1-T)}\bigg\}}
+\log\log(nm)\bigg] \, .
\end{align*}
when $\delta, \varepsilon \rightarrow 0$ and $T \rightarrow 1$. Here, $C_{3}, C_{4} > 0$ are two universal constants.

Letting $m=n$ and $\delta=3^{-1}n^{-2}$, \eqref{equa:final gene error} implies that, with probability at least $1 - n^{-2}$,
\begin{align}\label{equa:finalfinalgene}
&\mathcal{L}(\hat{\mathbf v})-\inf_{\mathbf v\in\mathrm{FNN}}\mathcal{L}(\mathbf v)\notag\\
&~~~~\leq \frac{C_5{C_0^{\{d_x+(\alpha+1)d_y+1\}/2}d_x^{(3d_x+7)/4}d_y^{d_y/4}}\{BC_y(d_x,d_y)\}^{d_y/2}\log^{(d_x+\alpha d_y+5)/4}\{d_x\varepsilon^{-1}(1-T)^{-1}\}}
{(1-T)^{(2d_x+\beta d_y+7)/2}\varepsilon^{(d_x+d_y+1)/2}n^{1/2}}\notag\\
&~~~~~\qquad\times {(d_x+d_y+1)}(\log n)\bigg[{\{(1+\alpha+\beta)(d_x+d_y+1)\}^{1/2}}\notag\\
&~~~~~~~~~~~~~\qquad\qquad\qquad {\times\log^{1/2}\bigg\{\frac{d_x+d_y+B+B^{-1}+C_y(d_x,d_y)}{\varepsilon(1-T)}\bigg\}}+\log^{1/2}n\bigg] \, .
\end{align}
where $C_5 > 0$ is a universal {constant}. Combining \eqref{equa:approxbound} and \eqref{equa:finalfinalgene}, by \eqref{equa:L(v)-L(v_F)} and \eqref{eq:decomposition}, we have that, with probability at least $1 - n^{-2}$,
\begin{align*}
&\frac{1}{T}\int_0^T\|\hat{\mathbf v}(\mathbf x,\mathbf y,t)-\mathbf v_{\rm F}(\mathbf x,\mathbf y,t)\|_{L^2(g_t)}^2\,\mathrm{d}t\\
&~~~\leq \frac{C_5{C_0^{\{d_x+(\alpha+1)d_y+1\}/2}d_x^{(3d_x+7)/4}d_y^{d_y/4}}\{BC_y(d_x,d_y)\}^{d_y/2}\log^{(d_x+\alpha d_y+5)/4}\{d_x\varepsilon^{-1}(1-T)^{-1}\}}
{(1-T)^{(2d_x+\beta d_y+7)/2}\varepsilon^{(d_x+d_y+1)/2}n^{1/2}}\\
&~~~~~~~~\qquad\times {(d_x+d_y+1)}(\log n)\bigg[{\{(1+\alpha+\beta)(d_x+d_y+1)\}^{1/2}}\\
&~~~~~~~\qquad\qquad\qquad {\times\log^{1/2}\bigg\{\frac{d_x+d_y+B+B^{-1}+C_y(d_x,d_y)}{\varepsilon(1-T)}\bigg\}}+\log^{1/2}n\bigg] +2d_x\varepsilon^2 \, .
\end{align*}
Let $\varepsilon=(1-T)^{-(2d_x+\beta d_y + 
 7)/(d_x+d_y+5)}n^{-1/({d_x}+{d_{y}}+5)}$ with $1-T \gg n^{-1/(2d_x+\beta d_y + 7)}$. Therefore, for given $(d_x, d_y)$, it holds that
 \begin{align*}
\frac{1}{T} \int_0^{{T}}\|\hat{\mathbf{v}}(\mathbf{x}, \mathbf{y}, t)-\mathbf{v}_{\rm{F}} (\mathbf{x}, \mathbf{y}, t)\|_{L^2(g_t)}^2 \, \mathrm{d} t =\widetilde{\mathcal{O}}\left\{\frac{(1-T)^{-(4d_x+2\beta d_y+14)/(d_x+d_y+5)}}{n^{2/(d_x+d_y+5)}}\right\}
\end{align*}
with probability at least $1 - n^{-2}$. Here, $\widetilde{\mathcal{O}}(\cdot)$ omits the polynomial term of $\log n$. We complete the proof of Proposition \ref{prop:generalization}. $\hfill\Box$

\subsection{Proof of Proposition \ref{prop:approx}} \label{subsec:proof of approx}
    The goal is to find a network $\tilde{\mathbf{v}}(\mathbf{x}, \mathbf{y}, t)$ in $\mathrm{FNN}=\mathrm{FNN}(L, M, J, K, \kappa, \gamma_1, \gamma_2, \gamma_3)$ to approximate the true vector field $\mathbf{v}_{\rm{F}}(\mathbf{x}, \mathbf{y}, t)$, where a major difficulty is that the input space $\mathbb{R}^{d_x} \times [0, B]^{d_{y}} \times[0, T]$ is unbounded. To address this difficulty, we partition $\mathbb{R}^{d_x}$ into a compact subset $\mathcal{K}$ and its complement $\mathcal{K}^{\rm c}$.

\color{black}
    We first consider the approximation on $\mathcal{K} \times [0, B]^{d_{y}} \times[0, T]$. Let $\mathcal{K}=\{\mathbf{x} \in \mathbb{R}^{d_x} :|\mathbf{x}|_{\infty} \leq R\}$ be a $d_x$-dimensional hypercube with edge length $2 R>0$, where $R$ will be determined later. Write $\mathbf{v}_{\rm{F}}(\mathbf{x}, \mathbf{y}, t) =({v}_{\scriptscriptstyle \rm{F}}^{1}(\mathbf{x}, \mathbf{y}, t), \ldots, {v}_{\scriptscriptstyle \rm{F}}^{d_x}(\mathbf{x}, \mathbf{y}, t))^{\rm{T}}$. On $\mathcal{K} \times [0, B]^{d_{y}} \times[0, T]$, we approximate each coordinate map ${v}_{\scriptscriptstyle \rm{F}}^{k}(\mathbf{x}, \mathbf{y}, t)$ separately.

    We first rescale the input by $\mathbf{x}^{\prime}=(\mathbf{x}+R \mathbf{1}) / (2 R)$, $\mathbf{y}^{\prime} = \mathbf{y} / B$ and $t^{\prime}=t / T$, where $\mathbf{1}:=(1, \ldots, 1)^{\rm{T}}$, so that the transformed space is $[0,1]^{{d_x}+{d_{y}}+1}$. Such a transformation can be exactly implemented by a single ReLU layer. By Proposition \ref{prop:properties of vF}(ii), $\mathbf{v}_{\rm{F}} (\mathbf{x}, \mathbf{y}, t)$ is $d_x(1-T)^{-2}$-Lipschitz w.r.t. $\mathbf{x}$. For simplicity, we write $d_x(1-T)^{-2}$ as $\zeta_T$ in the rest of this section. We now define the rescaled function on the transformed input space as $\mathbf{v}(\mathbf{x}^{\prime}, \mathbf{y}^{\prime},  t^{\prime}):=\mathbf{v}_{\rm{F}} (2 R \mathbf{x}^{\prime}-R \mathbf{1}, B\mathbf{y}^{\prime}, T t^{\prime})$. Such defined $\mathbf{v}$ is $(2 R \zeta_T)$-Lipschitz w.r.t. $\mathbf{x}^{\prime}$. Further by Assumption \ref{assump:Lip in y}, $\mathbf{v}$ is $(B {\omega}_{R,T})$-Lipschitz w.r.t. $\mathbf{y}^{\prime}$, where ${\omega}_{R,T} = C_y(d_x, d_y) R^{\alpha} (1-T)^{-\beta}$. By Proposition \ref{prop:properties of vF}(i), we know that $\mathbf{v}(\mathbf{x}^{\prime}, \mathbf{y}^{\prime}, t^{\prime})$ is $(T  \tau_{R, T})$-Lipschitz w.r.t $t^{\prime}$, where $ \tau_{R, T}:=\sup _{t \in[0, T]} \sup _{\mathbf{y} \in[0, B]^{d_{y}}} \sup _{\mathbf{x} \in[-R, R]^{d_x}}|\partial_t \mathbf{v}_{\rm{F}}(\mathbf{x}, \mathbf{y}, t)|_2 \leq {C_1d^{3 / 2}_x}(R+1)(1-T)^{-3}$ for some universal constant $C_1>1$. Now the goal becomes approximating $\mathbf{v}(\mathbf{x}^{\prime}, \mathbf{y}^{\prime}, t^{\prime})$ on $[0,1]^{{d_x}+{d_{y}}+1}$. We partition $[0,1]^{d_x}$ into non-overlapping hypercubes with equal edge length $e_1$, and $[0,1]^{d_{y}}$ into equal edge length $e_2$. We also partition the time interval $[0,1]$ into non-overlapping sub-intervals of length $e_3$. We will choose $e_1$, $e_2$ and $e_3$ later depending on the desired approximation level. Write $N_1=\lceil e_1^{-1}\rceil$, $N_2=\lceil e_2^{-1}\rceil$ and $N_3=\lceil e_3^{-1}\rceil$.

    We denote $[N] := \{0, \ldots, N-1\}$ for any positive integer $N$. Let $\mathbf{p}=(p_1, \ldots, p_{d_x})^{\rm{T}} \in{[N_1+1]^{d_x}}$, $\mathbf{q}=(q_1, \ldots, q_{d_y})^{\rm{T}} \in{[N_2+1]^{d_{y}}}$ be multi-indexes. {The grid nodes are
\[
\left(\frac{\mathbf{p}}{N_1},\frac{\mathbf{q}}{N_2},\frac{j}{N_3}\right)\,,\qquad j\in[N_3+1]\,.
\]
} {Define
\[
\bar{\mathbf v}(\mathbf x',\mathbf y',t')
:=\sum_{\mathbf p\in[N_1+1]^{d_x},\ \mathbf q\in[N_2+1]^{d_y},\ j\in[N_3+1]}
\mathbf v\!\left(\frac{\mathbf p}{N_1},\frac{\mathbf q}{N_2},\frac{j}{N_3}\right)
\Psi_{\mathbf p,\mathbf q,j}(\mathbf x',\mathbf y',t')\,,
\]
}
where { for $s=d_x+d_y+1$ and
\[
\mathbf a=\bigl((N_1\mathbf x'-\mathbf p)^{\rm T},\,(N_2\mathbf y'-\mathbf q)^{\rm T},\,N_3t'-j\bigr)^{\rm T}\,,
\]
we define
\[
\Psi_{\mathbf p,\mathbf q,j}(\mathbf x',\mathbf y',t')
=\max\!\left\{0,\,1-\max\{0,a_1,\ldots,a_s\}-\max\{0,-a_1,\ldots,-a_s\}\right\}\,.
\]
Fix $(\mathbf x',\mathbf y',t')$ in the unit cube, and let $\Theta$ be uniformly distributed on $[0,1)$. Define the random grid indices
\[
\widehat p_i=\lfloor N_1x_i'+\Theta\rfloor\,,\qquad
\widehat q_i=\lfloor N_2y_i'+\Theta\rfloor\,,\qquad
\widehat j=\lfloor N_3t'+\Theta\rfloor\,.
\]
They belong to the stated index ranges, since $0\leq N_1x_i'+\Theta<N_1+1$, and similarly for the other coordinates. For each spatial coordinate,
\[
\widehat p_i=p_i
\iff p_i\leq N_1x_i'+\Theta<p_i+1
\iff 0\leq N_1x_i'-p_i+\Theta<1\,.
\]
Applying the same equivalence to the remaining coordinates gives
\[
\begin{aligned}
&\mathbb P\bigl(\widehat{\mathbf p}=\mathbf p,\,
\widehat{\mathbf q}=\mathbf q,\,\widehat j=j\bigr)
=\mathbb P\bigl(0\leq a_\ell+\Theta<1\text{ for all }\ell\bigr)\\
&\quad=\mathbb P\!\left(\max\{0,-a_1,\ldots,-a_s\}\leq\Theta<1-\max\{0,a_1,\ldots,a_s\}\right)
=\Psi_{\mathbf p,\mathbf q,j}(\mathbf x',\mathbf y',t')\,.
\end{aligned}
\]
The last equality follows because a uniform random variable assigns each subinterval of $[0,1)$ its length, with an empty interval having probability zero. Summing over all possible grid indices yields}
\[
\sum_{\mathbf{p} \in{[N_1+1]^{d_x}},\ \mathbf{q} \in{[N_2+1]^{d_y}},\ j \in{[N_3+1]}} \Psi_{\mathbf{p}, \mathbf{q}, j}(\mathbf{x}^{\prime}, \mathbf{y}^{\prime}, t^{\prime})
\begingroup=\sum_{\mathbf p,\mathbf q,j}\mathbb P\bigl(\widehat{\mathbf p}=\mathbf p,\,\widehat{\mathbf q}=\mathbf q,\,\widehat j=j\bigr)=1\endgroup
\]
for any $(\mathbf{x}^{\prime}, \mathbf{y}^{\prime}, t^{\prime}) \in [0,1]^{{d_x}} \times [0,1]^{d_{y}} \times[0,1]$.

We claim that { $\bar{\mathbf v}(\mathbf x',\mathbf y',t')$ approximates $\mathbf v(\mathbf x',\mathbf y',t')$. Whenever $\Psi_{\mathbf p,\mathbf q,j}(\mathbf x',\mathbf y',t')>0$, its definition implies $|N_1x_i'-p_i|<1$, $|N_2y_i'-q_i|<1$, and $|N_3t'-j|<1$. Hence,
\[
\Bigl|\mathbf x'-\frac{\mathbf p}{N_1}\Bigr|_2\leq\frac{d_x^{1/2}}{N_1}\,,\qquad
\Bigl|\mathbf y'-\frac{\mathbf q}{N_2}\Bigr|_2\leq\frac{d_y^{1/2}}{N_2}\,,\qquad
\Bigl|t'-\frac{j}{N_3}\Bigr|\leq\frac{1}{N_3}\,.
\]
Using the nonnegativity and partition of unity of the weights, together with the Lipschitz bounds for $\mathbf v$, we obtain
\[
\begin{aligned}
\bigl|\bar{\mathbf v}(\mathbf x',\mathbf y',t')-\mathbf v(\mathbf x',\mathbf y',t')\bigr|_\infty
&\leq\sum_{\mathbf p,\mathbf q,j}\Psi_{\mathbf p,\mathbf q,j}(\mathbf x',\mathbf y',t')
\left|\mathbf v\!\left(\frac{\mathbf p}{N_1},\frac{\mathbf q}{N_2},\frac{j}{N_3}\right)-\mathbf v(\mathbf x',\mathbf y',t')\right|_\infty\\
&\leq\frac{2R\zeta_Td_x^{1/2}}{N_1}+\frac{B\omega_{R,T}d_y^{1/2}}{N_2}+\frac{T\tau_{R,T}}{N_3}\,.
\end{aligned}
\]
\par}
Recall $\varepsilon_{*}>0$ is a sufficiently small constant. Given an approximation error $\varepsilon \in (0, \varepsilon_{*})$, we select
{
\[
N_1=1\vee\bigl\lceil6R\zeta_Td_x^{1/2}/\varepsilon\bigr\rceil\,,\qquad
N_2=1\vee\bigl\lceil3B\omega_{R,T}d_y^{1/2}/\varepsilon\bigr\rceil\,,\qquad
N_3=1\vee\bigl\lceil3T\tau_{R,T}/\varepsilon\bigr\rceil\,.
\]
Each term in the preceding error bound is then at most $\varepsilon/3$, yielding
\[
\sup_{(\mathbf x',\mathbf y',t')\in[0,1]^{d_x}\times[0,1]^{d_y}\times[0,1]}
\bigl|\bar{\mathbf v}(\mathbf x',\mathbf y',t')-\mathbf v(\mathbf x',\mathbf y',t')\bigr|_\infty
\leq\varepsilon\,.
\]
\par}
{
Since the weights are nonnegative and sum to one, Proposition \ref{prop:properties of vF}(i) gives, for any $(\mathbf x',\mathbf y',t')\in[0,1]^{d_x}\times[0,1]^{d_y}\times[0,1]$,
\[
\bigl|\bar{\mathbf v}(\mathbf x',\mathbf y',t')\bigr|_2
\leq\sum_{\mathbf p,\mathbf q,j}\Psi_{\mathbf p,\mathbf q,j}(\mathbf x',\mathbf y',t')
\Bigl|\mathbf v\!\left(\frac{\mathbf p}{N_1},\frac{\mathbf q}{N_2},\frac{j}{N_3}\right)\Bigr|_2
\leq {\color{black}\frac{{d_x^{1/2}}(1+TR)}{1-T^2}}\,.
\]
\par}
{
The probability representation of $\Psi_{\mathbf p,\mathbf q,j}$ yields
\[
\bar{\mathbf v}(\mathbf x',\mathbf y',t')
=\mathbb E_\Theta\Bigl[\mathbf v\Bigl(\frac{\widehat{\mathbf p}}{N_1},\frac{\widehat{\mathbf q}}{N_2},\frac{\widehat j}{N_3}\Bigr)\Bigr]\,.
\]
For $u=m+r$ with $m\in\mathbb Z$ and $r\in[0,1)$, $\lfloor u+\Theta\rfloor$ equals $m$ or $m+1$ with probabilities $1-r$ and $r$. Thus $\mathbb E_\Theta\lfloor u+\Theta\rfloor=(1-r)m+r(m+1)=u$. Monotonicity of the floor function then gives, using the same $\Theta$ for any $u,w\in\mathbb R$,
\[
\mathbb E_\Theta\bigl|\lfloor u+\Theta\rfloor-\lfloor w+\Theta\rfloor\bigr|
=\bigl|\mathbb E_\Theta\lfloor u+\Theta\rfloor-\mathbb E_\Theta\lfloor w+\Theta\rfloor\bigr|
=|u-w|\,.
\]

Fix $\mathbf y',t'$ and $\mathbf x_1',\mathbf x_2'\in[0,1]^{d_x}$. Set $\widehat p_\ell^{(a)}=\lfloor N_1x_{a,\ell}'+\Theta\rfloor$, $a=1,2$, with the same $\Theta$, $\widehat{\mathbf q}$ and $\widehat j$ for both inputs. The triangle inequality, the spatial Lipschitz bound for $\mathbf v$ and the preceding identity yield
\[
\begin{aligned}
&\bigl|\bar{\mathbf v}(\mathbf x_1',\mathbf y',t')-\bar{\mathbf v}(\mathbf x_2',\mathbf y',t')\bigr|_\infty
=\Bigl|\mathbb E_\Theta\Bigl[
\mathbf v\Bigl(\frac{\widehat{\mathbf p}^{(1)}}{N_1},\frac{\widehat{\mathbf q}}{N_2},\frac{\widehat j}{N_3}\Bigr)
-\mathbf v\Bigl(\frac{\widehat{\mathbf p}^{(2)}}{N_1},\frac{\widehat{\mathbf q}}{N_2},\frac{\widehat j}{N_3}\Bigr)
\Bigr]\Bigr|_\infty\\
&~~\leq\mathbb E_\Theta\Bigl[\Bigl|
\mathbf v\Bigl(\frac{\widehat{\mathbf p}^{(1)}}{N_1},\frac{\widehat{\mathbf q}}{N_2},\frac{\widehat j}{N_3}\Bigr)
-\mathbf v\Bigl(\frac{\widehat{\mathbf p}^{(2)}}{N_1},\frac{\widehat{\mathbf q}}{N_2},\frac{\widehat j}{N_3}\Bigr)
\Bigr|_\infty\Bigr]
\leq\frac{2R\zeta_T}{N_1}\,\mathbb E_\Theta\bigl|\widehat{\mathbf p}^{(1)}-\widehat{\mathbf p}^{(2)}\bigr|_2\\
&~~\leq\frac{2R\zeta_T}{N_1}\sum_{\ell=1}^{d_x}\mathbb E_\Theta\bigl|\widehat p_\ell^{(1)}-\widehat p_\ell^{(2)}\bigr|
=2R\zeta_T\sum_{\ell=1}^{d_x}|x_{1,\ell}'-x_{2,\ell}'|
\leq2R\zeta_Td_x^{1/2}\,|\mathbf x_1'-\mathbf x_2'|_2\,.
\end{aligned}
\]
The same argument gives Lipschitz constants $B\omega_{R,T}d_y^{1/2}$ and $T\tau_{R,T}$ with respect to $\mathbf y'$ and $t'$, respectively.
\par}
{
\begin{lemma}\label{lemma:exact-local-realization}
Let $R\geq1$ and $0<\varepsilon<\varepsilon_*$. On $[0,1]^{d_x}\times[0,1]^{d_y}\times[0,1]$, $\bar{\mathbf v}$ admits an exact ReLU representation with $L$ hidden layers, width $M$, and at most $J$ nonzero weights and biases bounded in absolute value by $\kappa$, satisfying $|\bar{\mathbf v}|_2\leq K$, where
\begin{align*}
&L\lesssim d_x+d_y+1\,,\qquad
M\lesssim (1\vee T\tau_{R,T})(8R\zeta_Td_x^{1/2})^{d_x}(5B\omega_{R,T}d_y^{1/2})^{d_y}\varepsilon^{-d_x-d_y-1}\,,\\
&J\lesssim\frac{(1\vee T\tau_{R,T})(8R\zeta_Td_x^{1/2})^{d_x}(5B\omega_{R,T}d_y^{1/2})^{d_y}}{\varepsilon^{d_x+d_y+1}(d_x+d_y+1)^{-1}}\,,\\
&\kappa\lesssim\frac{R\zeta_Td_x^{1/2}\vee B\omega_{R,T}d_y^{1/2}\vee T\tau_{R,T}}{\varepsilon}\,,\qquad
K=\frac{d_x^{1/2}(1+TR)}{1-T^2}\,.
\end{align*}
The implicit constants are universal.
\end{lemma}
}
Let $\mathring{\mathbf{v}}(\mathbf{x}, \mathbf{y}, t) = {\bar{\mathbf v}}(\mathbf{x}^{\prime}, \mathbf{y}^{\prime}, t^{\prime})$ with $\mathbf{x}^{\prime}=(\mathbf{x}+R \mathbf{1}) / (2 R)$, $\mathbf{y}^{\prime} = \mathbf{y} / B$ and $t^{\prime}=t / T$.
{This change of variables adds one ReLU layer with at most $2d_x+d_y+1$ nonzero parameters. The orders of $L,M,J$ in Lemma \ref{lemma:exact-local-realization} are preserved, with
\[
\kappa\lesssim B^{-1}\vee T^{-1}\vee\frac{R\zeta_Td_x^{1/2}\vee B\omega_{R,T}d_y^{1/2}\vee T\tau_{R,T}}{\varepsilon}\,.
\]
}
{By the change of variables, we have}
\begin{align*}
&\sup _{(\mathbf{x}, \mathbf{y}, t) \in \mathcal{K} \times[0,B]^{d_{y}} \times[0, T]}|\mathring{\mathbf{v}}(\mathbf{x}, \mathbf{y}, t)-\mathbf{v}_{\rm{F}} (\mathbf{x}, \mathbf{y}, t)|_{\infty}
\leq \varepsilon \, ,\\
&\begingroup\sup _{(\mathbf{x}, \mathbf{y}, t) \in \mathcal{K} \times[0,B]^{d_{y}} \times[0, T]}|\mathring{\mathbf{v}}(\mathbf{x}, \mathbf{y}, t)|_2
\leq\endgroup\frac{d_x^{1/2}(1+TR)}{1-T^2} \, .
\end{align*}
Furthermore,
{
\[
\begin{aligned}
\bigl|\mathring{\mathbf v}(\mathbf x_1,\mathbf y,t)-\mathring{\mathbf v}(\mathbf x_2,\mathbf y,t)\bigr|_\infty
&\leq d_x^{1/2}\,\zeta_T|\mathbf x_1-\mathbf x_2|_2\,,\\
\bigl|\mathring{\mathbf v}(\mathbf x,\mathbf y_1,t)-\mathring{\mathbf v}(\mathbf x,\mathbf y_2,t)\bigr|_\infty
&\leq d_y^{1/2}\,\omega_{R,T}|\mathbf y_1-\mathbf y_2|_2\,,\\
\bigl|\mathring{\mathbf v}(\mathbf x,\mathbf y,t_1)-\mathring{\mathbf v}(\mathbf x,\mathbf y,t_2)\bigr|_\infty
&\leq\tau_{R,T}|t_1-t_2|\,,
\end{aligned}
\]
for all arguments in $\mathcal K\times[0,B]^{d_y}\times[0,T]$.
\par}
Consider the following univariate real-valued function
\begin{align*}
	T_R(x):=\left\{
	\begin{aligned}	R\,,~~~\,~&\textrm{if}~x>R\,,\\
		x\,,~~~\,~&\textrm{if}~ x \in [-R, R]\,,\\
  -R\,,~~\,~&\textrm{if}~ x<-R\, .
	\end{aligned}
	\right.
\end{align*}
Define $\mathcal{T}_{\mathcal{K}}(\mathbf{x})=(T_R(x_1), \ldots, T_R(x_{d_x}))^{\rm T}$ with $\mathbf{x}=(x_1,\ldots,x_{d_x})^{\rm T}$. Immediately, it yields that $\mathcal{T}_{\mathcal{K}}(\mathbf{x}) = \mathbf{x}$ for any $\mathbf{x} \in \mathcal{K}$, and $\mathcal{T}_{\mathcal{K}}(\mathbf{x}) \in \partial\mathcal{K}$ for any $\mathbf{x} \in \mathcal{K}^{\rm c}$.
Simple calculation tells us
\begin{align*}
|\mathcal{T}_{\mathcal{K}}(\mathbf{x})-\mathcal{T}_{\mathcal{K}}(\mathbf{y})|_{2} &= \bigg\{\sum^{d_x}_{i=1} |T_R(x_i)-T_R(y_i)|^2\bigg\}^{1/2} \\
& \leq \bigg(\sum^{d_x}_{i=1} |x_i-y_i|^2\bigg)^{1/2} = |\mathbf{x}-\mathbf{y}|_2 \, .
\end{align*}
Also, it's easy to check
$
T_R(x) = \text{ReLU}(x)-\text{ReLU}(-x)+\text{ReLU}(-x-R)-\text{ReLU}(x-R)$.
Due to 
\begin{align*}
&\sup _{(\mathbf{x}, \mathbf{y}, t) \in \mathcal{K} \times[0,B]^{d_{y}} \times[0, T]}|\mathring{\mathbf{v}}(\mathcal{T}_{\mathcal{K}}(\mathbf{x}), \mathbf{y}, t)-\mathbf{v}_{\rm{F}}  (\mathbf{x}, \mathbf{y}, t)|_{\infty} \\
&~~~~~~= \sup _{(\mathbf{x}, \mathbf{y}, t) \in \mathcal{K} \times[0,B]^{d_{y}} \times[0, T]}|\mathring{\mathbf{v}}(\mathbf{x}, \mathbf{y}, t)-\mathbf{v}_{\rm{F}}  (\mathbf{x}, \mathbf{y}, t)|_{\infty}\leq \varepsilon \, ,
\end{align*}

we know that $\mathring{\mathbf{v}}(\mathcal{T}_{\mathcal{K}}(\mathbf{x}), \mathbf{y}, t)$ preserves the approximation capability of $\mathring{\mathbf{v}}(\mathbf{x}, \mathbf{y}, t)$. Meanwhile, 
\begin{equation} \label{equa:globalupperbound}
\begin{aligned}
\sup _{(\mathbf{x}, \mathbf{y}, t) \in \mathbb{R}^{d_x} \times[0,B]^{d_{y}} \times[0, T]}|\mathring{\mathbf{v}}(\mathcal{T}_{\mathcal{K}}(\mathbf{x}), \mathbf{y}, t)|_2 &= \sup _{(\mathbf{x}, \mathbf{y}, t) \in \mathcal{K} \times[0,B]^{d_{y}} \times[0, T]}|\mathring{\mathbf{v}}(\mathbf{x}, \mathbf{y}, t)|_2 \\
&\leq  {K =\frac{d_x^{1/2}(1+TR)}{1-T^2}\leq\frac{d_x^{1/2}R}{1-T}}
\end{aligned}
\end{equation}
{for $R\geq1$.} Furthermore, for any $\mathbf{x}_1, \mathbf{x}_2 \in \mathbb{R}^{d_x}$, $\mathbf{y} \in [0, B]^{d_{y}}$ and $t \in[0, T]$, it holds that 
$$
\begin{aligned}
|\mathring{\mathbf{v}}(\mathcal{T}_{\mathcal{K}}(\mathbf{x}_1), \mathbf{y}, t)-\mathring{\mathbf{v}}(\mathcal{T}_{\mathcal{K}}(\mathbf{x}_2), \mathbf{y}, t)|_{\infty} \leq {d_x^{1/2}} \zeta_T|\mathcal{T}_{\mathcal{K}}(\mathbf{x}_1)-\mathcal{T}_{\mathcal{K}}(\mathbf{x}_2)|_2 \leq {d_x^{1/2}} \zeta_T |\mathbf{x}_1-\mathbf{x}_2|_2 \, .
\end{aligned}
$$
So $\mathring{\mathbf{v}} 
 (\mathcal{T}_{\mathcal{K}}(\mathbf{x}), \mathbf{y}, t)$ acquires global Lipschitz continuity w.r.t. $\mathbf{x}$.

 Let 
 \begin{align} \label{equa:def of tilde_v}
 \tilde{\mathbf{v}}(\mathbf{x}, \mathbf{y}, t)=\mathring{\mathbf{v}}(\mathcal{T}_{\mathcal{K}}(\mathbf{x}), \mathbf{y}, t) 
 \, .
 \end{align}
The $L^2$ approximation error of $\tilde{\mathbf{v}}$ now can be decomposed into two terms,
$$
\begin{aligned}
\|\mathbf{v}_{\rm{F}} (\mathbf{x}, \mathbf{y}, t)-\tilde{\mathbf{v}}(\mathbf{x}, \mathbf{y},  t)\|^{2}_{L^2(g_t)}= \, &  \|\{\mathbf{v}_{\rm{F}} (\mathbf{x}, \mathbf{y}, t)-\tilde{\mathbf{v}}(\mathbf{x}, \mathbf{y}, t)\} \mathbb{I}(|\mathbf{x}|_{\infty} \leq R)\|^{2}_{L^2(g_t)} \\
& +\|\{\mathbf{v}_{\rm{F}} (\mathbf{x}, \mathbf{y}, t)-\tilde{\mathbf{v}}(\mathbf{x}, \mathbf{y}, t)\} \mathbb{I}(|\mathbf{x}|_{\infty}>R)\|^{2}_{L^2(g_t)} \, .
\end{aligned}
$$
The first term on the right-hand side of the last display is bounded by
$$
\begin{aligned}
& \|\{\mathbf{v}_{\rm{F}}(\mathbf{x}, \mathbf{y}, t) -\tilde{\mathbf{v}}(\mathbf{x}, \mathbf{y}, t)\} \mathbb{I}(|\mathbf{x}|_{\infty} \leq R)\|^{2}_{L^2(g_t)} \\
&~~~~~~\leq {d_x} \sup _{(\mathbf{x}, \mathbf{y}, t) \in \mathcal{K} \times[0,B]^{d_{y}} \times[0, T]}|\mathbf{v}_{\rm{F}} (\mathbf{x}, \mathbf{y}, t)-\tilde{\mathbf{v}}(\mathbf{x}, \mathbf{y}, t)|^{2}_{\infty} \leq {d_x} \varepsilon^2 \, .
\end{aligned}
$$
The following lemma gives an upper bound for the second term whose proof is presented in Section \ref{subsec:approx trunc error}.

 \begin{lemma} \label{lemma:approx trunc error}
    Let Assumptions {\rm \ref{assump:label}} and {\rm \ref{assump:bounded support}} hold. There exists some universal constant $C>0$ such that 
$$
\|\{\mathbf{v}_{\rm{F}} (\mathbf{x}, \mathbf{y}, t)-\tilde{\mathbf{v}}(\mathbf{x}, \mathbf{y}, t)\} \mathbb{I}(|\mathbf{x}|_{\infty}>R)\|^{2}_{L^2(g_t)}  \leq \varepsilon^2 
$$
for any $\varepsilon\in(0,\varepsilon_*)$ and $t \in [0, T]$ with selecting $R\geq C\log^{1/2} \{d_x \varepsilon^{-1}(1-T)^{-1}\}$, where $\varepsilon_*$ is specified in Proposition {\rm \ref{prop:approx}}.
\end{lemma}

With selecting $R\geq C\log^{1/2} \{d_x \varepsilon^{-1}(1-T)^{-1}\}$, it holds that
$$
\|\mathbf{v}_{\rm{F}} (\mathbf{x}, \mathbf{y}, t)-\tilde{\mathbf{v}}(\mathbf{x}, \mathbf{y}, t)\|_{L^2(g_t)} \leq \varepsilon {(d_x+1)^{1/2}}  \, . 
$$
{
{\color{black}Substituting $R$ into the network configuration, we obtain}
\begin{align*}
&M\lesssim\frac{C_0^{d_x+(\alpha+1)d_y+1}d_x^{\,3(d_x+1)/2}d_y^{\,d_y/2}\{BC_y(d_x,d_y)\}^{d_y}\log^{(d_x+\alpha d_y+1)/2}\{d_x\varepsilon^{-1}(1-T)^{-1}\}}{(1-T)^{2d_x+\beta d_y+3}\varepsilon^{d_x+d_y+1}}\,,\\
&J\lesssim\frac{C_0^{d_x+(\alpha+1)d_y+1}d_x^{\,3(d_x+1)/2}d_y^{\,d_y/2}\{BC_y(d_x,d_y)\}^{d_y}\log^{(d_x+\alpha d_y+1)/2}\{d_x\varepsilon^{-1}(1-T)^{-1}\}}{(1-T)^{2d_x+\beta d_y+3}\varepsilon^{d_x+d_y+1}(d_x+d_y+1)^{-1}}\,,\\
&\kappa\lesssim B^{-1}\vee T^{-1}\vee\frac{\{Bd_y^{1/2}C_y(d_x,d_y)\vee d_x^{3/2}\}\log^{(\alpha\vee1)/2}\{d_x\varepsilon^{-1}(1-T)^{-1}\}}{\varepsilon(1-T)^{\beta\vee3}}\,,\\
&L\lesssim d_x+d_y+1\,,\qquad
K\lesssim\frac{d_x^{1/2}\log^{1/2}\{d_x\varepsilon^{-1}(1-T)^{-1}\}}{1-T}\,,\qquad
\gamma_1=\frac{d_x^{3/2}}{(1-T)^2}\,,\\
&\gamma_2\lesssim\frac{d_y^{1/2}C_y(d_x,d_y)\log^{\alpha/2}\{d_x\varepsilon^{-1}(1-T)^{-1}\}}{(1-T)^\beta}\,,\qquad
\gamma_3\lesssim\frac{d_x^{3/2}\log^{1/2}\{d_x\varepsilon^{-1}(1-T)^{-1}\}}{(1-T)^3}\,.
\end{align*}
}
We complete the proof of Proposition \ref{prop:approx}. $\hfill\Box$

\section{Proof of  Proposition \ref{prop:estimation error bound}} \label{append:W2 estimate}
Recall that the flow map related to $\mathbf{Z}_t^{\mathbf{y}}$ is $\mathbf{F}_t(\cdot, \mathbf{y})$. Denote by $\hat{\mathbf{F}}_t(\cdot, 
\mathbf{y})$ the flow map related to $\hat{\mathbf{Z}}_t^{\mathbf{y}}$ in \eqref{equa:nnflow}. Then, given $\mathbf{Z}_0 \sim \mathcal{N}(\mathbf{0}, \mathbf{I}_{d_x})$, we have $\mathbf{F}_t  (\mathbf{Z}_0, \mathbf{y}) \sim f_t(\mathbf{x} \mkern 2mu|\mkern2mu \mathbf{y})$ and $\hat{\mathbf{F}}_t(\mathbf{Z}_0, \mathbf{y}) \sim \hat{p}_t(\mathbf{x};\mathbf{y})$.
For simplicity, we abbreviate $\mathbf{F}_t  (\cdot, \mathbf{y})$ and $\hat{\mathbf{F}}_t  (\cdot, \mathbf{y})$ as $\mathbf{Z}^{\mathbf{y}}_t(\cdot)$ and $\hat{\mathbf{Z}}_t^{\mathbf{y}}(\cdot)$, respectively. Now, $\mathbf{Z}^{\mathbf{y}}_t({\mathbf{Z}_0})$ and $\hat{\mathbf{Z}}^{\mathbf{y}}_t({\mathbf{Z}_0})$ form a coupling of $f_t(\mathbf{x} \mkern 2mu|\mkern2mu \mathbf{y})$ and $\hat{p}_t(\mathbf{x};\mathbf{y})$. Denote by ${\nu}_0(\mathbf{x})$ the density of $\mathcal{N}(\mathbf{0}, \mathbf{I}_{d_x})$. By the definition of Wasserstein-2 distance, we have
$$
W_2^2\big(f_t(\mathbf{x} \mkern2mu|\mkern2mu \mathbf{y}), \mkern2mu \hat{p}_t(\mathbf{x};\mathbf{y})\big) \leq \int\big|\mathbf{Z}^{\mathbf{y}}_t(\mathbf{x})-\hat{\mathbf{Z}}^{\mathbf{y}}_t(\mathbf{x})\big|_2^2 \,{\nu}_0(\mathbf{x}) \mkern2mu \mathrm{d} \mathbf{x} =: R_t^{\mathbf{y}} \, .
$$
Due to $
    \mathrm{d} \mathbf{Z}^{\mathbf{y}}_t(\mathbf{x}) = \mathbf{v}_{\rm{F}}  (\mathbf{Z}^{\mathbf{y}}_t(\mathbf{x}), \mathbf{y}, t) \, \mathrm{d}t $ and $ \mathrm{d} \hat{\mathbf{Z}}^{\mathbf{y}}_t(\mathbf{x}) = \hat{\mathbf{v}}  (\hat{\mathbf{Z}}^{\mathbf{y}}_t(\mathbf{x}), \mathbf{y}, t) \, \mathrm{d}t$, 
we have
\begin{align*}
\frac{\mathrm{d} R_t^{\mathbf{y}}}{\mathrm{d} t} & =2 \int\big\langle\mathbf{v}_{\rm{F}}  (\mathbf{Z}^{\mathbf{y}}_t(\mathbf{x}), \mathbf{y}, t)-\hat{\mathbf{v}}  (\hat{\mathbf{Z}}^{\mathbf{y}}_t(\mathbf{x}), \mathbf{y}, t), \mathbf{Z}^{\mathbf{y}}_t(\mathbf{x})-\hat{\mathbf{Z}}^{\mathbf{y}}_t(\mathbf{x})\big\rangle {\nu}_0(\mathbf{x}) \mkern2mu \mathrm{d} \mathbf{x} \\
& =2 \int\big\langle\mathbf{v}_{\rm{F}}  (\mathbf{Z}^{\mathbf{y}}_t(\mathbf{x}), \mathbf{y}, t)-\hat{\mathbf{v}}  (\mathbf{Z}^{\mathbf{y}}_t(\mathbf{x}), \mathbf{y}, t), \mathbf{Z}^{\mathbf{y}}_t(\mathbf{x})-\hat{\mathbf{Z}}^{\mathbf{y}}_t(\mathbf{x})\big\rangle {\nu}_0(\mathbf{x}) \mkern2mu \mathrm{d} \mathbf{x} \\
&~~~+2 \int\big\langle\hat{\mathbf{v}}  (\mathbf{Z}^{\mathbf{y}}_t(\mathbf{x}), \mathbf{y}, t)-\hat{\mathbf{v}}  (\hat{\mathbf{Z}}_t^{\mathbf{y}}(\mathbf{x}), \mathbf{y}, t), \mathbf{Z}^{\mathbf{y}}_t(\mathbf{x})-\hat{\mathbf{Z}}^{\mathbf{y}}_t(\mathbf{x})\big\rangle {\nu}_0(\mathbf{x}) \mkern2mu \mathrm{d} \mathbf{x} \, .
\end{align*}
Notice that
\begin{align*}
    &2\big\langle\mathbf{v}_{\rm{F}}  (\mathbf{Z}^{\mathbf{y}}_t(\mathbf{x}), \mathbf{y}, t)-\hat{\mathbf{v}}  (\mathbf{Z}^{\mathbf{y}}_t(\mathbf{x}), \mathbf{y}, t), \mathbf{Z}^{\mathbf{y}}_t(\mathbf{x})-\hat{\mathbf{Z}}^{\mathbf{y}}_t(\mathbf{x})\big\rangle\\
    &~~~~~~~~~~~~~~~\leq \big|\mathbf{v}_{\rm{F}}  (\mathbf{Z}^{\mathbf{y}}_t(\mathbf{x}), \mathbf{y}, t)-\hat{\mathbf{v}}  (\mathbf{Z}^{\mathbf{y}}_t(\mathbf{x}), \mathbf{y}, t)\big|_2^2+\big|\mathbf{Z}^{\mathbf{y}}_t(\mathbf{x})-\hat{\mathbf{Z}}^{\mathbf{y}}_t(\mathbf{x})\big|_2^2 \, .
\end{align*}
Since $\hat{\mathbf{v}}(\mathbf{x}, \mathbf{y}, t) \in \mathrm{FNN}$ defined in Proposition \ref{prop:generalization} is $\gamma_1$-Lipschitz continuous w.r.t. $\mathbf{x}$, we have
$$
\big\langle\hat{\mathbf{v}}  (\mathbf{Z}^{\mathbf{y}}_t(\mathbf{x}), \mathbf{y}, t)-\hat{\mathbf{v}}  (\hat{\mathbf{Z}}_t^{\mathbf{y}}(\mathbf{x}), \mathbf{y}, t), \mathbf{Z}^{\mathbf{y}}_t(\mathbf{x})-\hat{\mathbf{Z}}^{\mathbf{y}}_t(\mathbf{x})\big\rangle \leq  {d^{1/2}_x} \gamma_1 \big|\mathbf{Z}^{\mathbf{y}}_t(\mathbf{x})-\hat{\mathbf{Z}}^{\mathbf{y}}_t(\mathbf{x})\big|_2^2 \, .
$$
Therefore,
$$
\frac{\mathrm{{d}} R_t^{\mathbf{y}}}{\mathrm{d} t} \leq(1+2 {d^{1/2}_x}\gamma_1) R_t^{\mathbf{y}}+\int \big|\mathbf{v}_{\rm{F}}  (\mathbf{Z}^{\mathbf{y}}_t(\mathbf{x}), \mathbf{y}, t)-\hat{\mathbf{v}}  (\mathbf{Z}^{\mathbf{y}}_t(\mathbf{x}), \mathbf{y}, t)\big|_2^2 \, {\nu}_0(\mathbf{x}) \mkern2mu \mathrm{d} \mathbf{x} \, .
$$
Denote by $g_t(\cdot, \cdot)$ the joint density of $(t \mathbf{X}+\sqrt{1-t^2} \mathbf{W}, \mathbf{Y})$. By Lemma \ref{lemma:gronwall} and $R^{\mathbf{y}}_0=0$, it holds that
$$
\begin{aligned}
&\int W^2_2 \big( \mkern0.5mu f_{_T}(\mathbf{x} \mkern 2mu|\mkern2mu \mathbf{y}), \mkern2mu \hat{p}_{_T}(\mathbf{x} ; \mathbf{y})\big) p_y(\mathbf{y}) \, \mathrm{d}\mathbf{y} \leq \int R_T^{\mathbf{y}} \mkern3mu p_y(\mathbf{y}) \, \mathrm{d}\mathbf{y} \\
&~~~\leq e^{1+2 {d^{1/2}_x} \gamma_1} \int_0^{{T}} \int \int \big|\mathbf{v}_{\rm{F}}  (\mathbf{Z}^{\mathbf{y}}_t(\mathbf{x}), \mathbf{y}, t)-\hat{\mathbf{v}}  (\mathbf{Z}^{\mathbf{y}}_t(\mathbf{x}), \mathbf{y}, t)\big|_2^2 \, {\nu}_0(\mathbf{x}) p_y( \mathbf{y}) \mkern2mu \mathrm{d}\mathbf{x} \mkern1mu \mathrm{d} \mathbf{y} \mkern1mu \mathrm{d} t \\
&~~~=e^{1+2 {d^{1/2}_x} \gamma_1} \int_0^{{T}} \int \int |\mathbf{v}_{\rm{F}}  (\mathbf{x}, \mathbf{y}, t)-\hat{\mathbf{v}}  (\mathbf{x}, \mathbf{y}, t)|_2^2 \, f_t(\mathbf{x} \mkern 2mu|\mkern2mu \mathbf{y}) p_y(\mathbf{y}) \mkern2mu \mathrm{d}\mathbf{x} \mkern1mu \mathrm{d} \mathbf{y} \mkern1mu \mathrm{d} t \\
&~~~=e^{1+2 {d^{1/2}_x} \gamma_1} \int_0^{{T}} \|\mathbf{v}_{\rm{F}} (\mathbf{x}, \mathbf{y}, t)-\hat{\mathbf{v}}(\mathbf{x}, \mathbf{y}, t)\|^2_{L^2(g_t)} \, \mathrm{d}t \, .
\end{aligned}
$$
 As given in Proposition \ref{prop:generalization}, $\gamma_1={d_x^{3/2}}(1-T)^{-2}$. By Proposition \ref{prop:generalization}, with probability at least $1-n^{-2}$, we have
 \begin{align*}
   \int W^2_2 \big( \mkern0.5mu f_{_T}(\mathbf{x} \mkern 2mu|\mkern2mu \mathbf{y}), \mkern2mu \hat{p}_{_T}(\mathbf{x} ; \mathbf{y})\big) p_y(\mathbf{y}) \, \mathrm{d}\mathbf{y} = \widetilde{\mathcal{O}}\left\{{e^{2d_x^2(1-T)^{-2}}}\frac{(1-T)^{-(4d_x+2\beta d_y + 14)/(d_x+d_y+5)}}{n^{2/(d_x+d_y+5)}}\right\} \, .
\end{align*}
We complete the proof of Proposition \ref{prop:estimation error bound}. $\hfill\Box$

\section{Proof of Proposition \ref{prop:discretization error bound}} \label{append:W2 discre}
Recall that the flow map related to $\hat{\mathbf{Z}}_t^{\mathbf{y}}$ in \eqref{equa:nnflow} is defined as $\hat{\mathbf{F}}_t(\cdot, \mathbf{y})$ in Section \ref{append:W2 estimate}. Denote by $\tilde{\mathbf{F}}_t  (\cdot, \mathbf{y})$ the flow map related to $\tilde{\mathbf{Z}}^{\mathbf{y}}_t$ in \eqref{equa:eulerflow}. Then, given $\mathbf{Z}_0 \sim \mathcal{N}(\mathbf{0}, \mathbf{I}_{d_x})$, we have $\hat{\mathbf{F}}_t(\mathbf{Z}_0, \mathbf{y}) \sim \hat{p}_t(\mathbf{x};\mathbf{y})$ and $\tilde{\mathbf{F}}_t(\mathbf{Z}_0, \mathbf{y}) \sim \tilde{p}_t(\mathbf{x};\mathbf{y})$.
For simplicity, we abbreviate $\hat{\mathbf{F}}_t  (\cdot, \mathbf{y})$ and $\tilde{\mathbf{F}}_t  (\cdot, \mathbf{y})$ as $\hat{\mathbf{Z}}_t^{\mathbf{y}}(\cdot)$ and $\tilde{\mathbf{Z}}_t^{\mathbf{y}}(\cdot)$, respectively. Now, $\hat{\mathbf{Z}}^{\mathbf{y}}_t({\mathbf{Z}_0})$ and $\tilde{\mathbf{Z}}^{\mathbf{y}}_t({\mathbf{Z}_0})$ form a coupling of $\hat{p}_t(\mathbf{x};\mathbf{y})$ and $\tilde{p}_t(\mathbf{x};\mathbf{y})$. Denote by ${\nu}_0(\mathbf{x})$ the density of $\mathcal{N}(\mathbf{0}, \mathbf{I}_{d_x})$. By the definition of Wasserstein-2 distance, we have
    $$
    W^2_2\big( \mkern0.5mu \hat{p}_t(\mathbf{x};\mathbf{y}), \mkern2mu \tilde{p}_t(\mathbf{x} ; \mathbf{y})\big) \leq \int \big|\hat{\mathbf{Z}}^{\mathbf{y}}_t(\mathbf{x})-\tilde{\mathbf{Z}}^{\mathbf{y}}_t(\mathbf{x})\big|_2^2 \, {\nu}_0(\mathbf{x}) \mkern2mu \mathrm{d} \mathbf{x} =: L_t^{\mathbf{y}} \, .
    $$
Since $\tilde{\mathbf{Z}}_t(\mathbf{x})$ is piece-wise linear, we consider the evolution of $L_t^{\mathbf{y}}$ on each split interval $(t_k, t_{k+1})$. Due to $
    \mathrm{d} \hat{\mathbf{Z}}^{\mathbf{y}}_t(\mathbf{x}) = \hat{\mathbf{v}}  (\hat{\mathbf{Z}}^{\mathbf{y}}_t(\mathbf{x}), \mathbf{y}, t) \, \mathrm{d}t $ and $ \mathrm{d} \tilde{\mathbf{Z}}^{\mathbf{y}}_t(\mathbf{x}) = \hat{\mathbf{v}} (\tilde{\mathbf{Z}}^{\mathbf{y}}_{t_k}(\mathbf{x}), \mathbf{y}, t_k) \, \mathrm{d}t$ for any $t \in (t_k, t_{k+1})$, it holds that 
\begin{align*}
\frac{\mathrm{d} L_t^{\mathbf{y}}}{\mathrm{d} t}= & \int 2\big\langle \hat{\mathbf{v}} (\hat{\mathbf{Z}}^{\mathbf{y}}_t(\mathbf{x}), \mathbf{y}, t)-\hat{\mathbf{v}}(\tilde{\mathbf{Z}}^{\mathbf{y}}_{t_k}(\mathbf{x}), \mathbf{y}, t_k), \hat{\mathbf{Z}}^{\mathbf{y}}_t(\mathbf{x})-\tilde{\mathbf{Z}}^{\mathbf{y}}_t(\mathbf{x})\big\rangle {\nu}_0(\mathbf{x}) \, \mathrm{d} \mathbf{x} \\
= & \int 2\big\langle\hat{\mathbf{v}} (\hat{\mathbf{Z}}^{\mathbf{y}}_t(\mathbf{x}), \mathbf{y}, t)-\hat{\mathbf{v}} (\tilde{\mathbf{Z}}^{\mathbf{y}}_t(\mathbf{x}), \mathbf{y}, t), \hat{\mathbf{Z}}^{\mathbf{y}}_t(\mathbf{x})-\tilde{\mathbf{Z}}^{\mathbf{y}}_t(\mathbf{x})\big\rangle {\nu}_0(\mathbf{x}) \, \mathrm{d} \mathbf{x} \\
& +\int 2\big\langle\hat{\mathbf{v}}  (\tilde{\mathbf{Z}}^{\mathbf{y}}_t(\mathbf{x}), \mathbf{y}, t)-\hat{\mathbf{v}} (\tilde{\mathbf{Z}}^{\mathbf{y}}_{t_k}(\mathbf{x}), \mathbf{y}, t), \hat{\mathbf{Z}}^{\mathbf{y}}_t(\mathbf{x})-\tilde{\mathbf{Z}}^{\mathbf{y}}_t(\mathbf{x})\big\rangle {\nu}_0(\mathbf{x}) \, \mathrm{d} \mathbf{x} \\
& +\int 2\big\langle\hat{\mathbf{v}}  (\tilde{\mathbf{Z}}^{\mathbf{y}}_{t_k}(\mathbf{x}), \mathbf{y}, t)-\hat{\mathbf{v}}(\tilde{\mathbf{Z}}^{\mathbf{y}}_{t_k}(\mathbf{x}), \mathbf{y}, t_k), \hat{\mathbf{Z}}^{\mathbf{y}}_t(\mathbf{x})-\tilde{\mathbf{Z}}^{\mathbf{y}}_t(\mathbf{x})\big\rangle {\nu}_0(\mathbf{x}) \, \mathrm{d} \mathbf{x}
\end{align*}
for any $t \in (t_k, t_{k+1})$. Since $\hat{\mathbf{v}}(\mathbf{x}, \mathbf{y}, t)$ is $\gamma_1$-Lipschitz continuous w.r.t. $\mathbf{x}$, we have
$$
\begin{aligned}
&\int \big\langle\hat{\mathbf{v}} (\hat{\mathbf{Z}}^{\mathbf{y}}_t(\mathbf{x}), \mathbf{y}, t)-\hat{\mathbf{v}}  (\tilde{\mathbf{Z}}^{\mathbf{y}}_t(\mathbf{x}), \mathbf{y}, t), \hat{\mathbf{Z}}^{\mathbf{y}}_t(\mathbf{x})-\tilde{\mathbf{Z}}^{\mathbf{y}}_t(\mathbf{x})\big\rangle {\nu}_0(\mathbf{x}) \, \mathrm{d} \mathbf{x} \\
&~~~~~~~~~~~\leq  {d^{1/2}_x}\gamma_1 \int \big|\hat{\mathbf{Z}}^{\mathbf{y}}_t(\mathbf{x})-\tilde{\mathbf{Z}}^{\mathbf{y}}_t(\mathbf{x})\big|_2^2 \,{\nu}_0(\mathbf{x}) \mkern2mu \mathrm{d} \mathbf{x} = d^{1/2}_x\gamma_1L_t^{\mathbf{y}} \, .
\end{aligned}
$$
Note that $\tilde{\mathbf{Z}}^{\mathbf{y}}_t(\mathbf{x})=\tilde{\mathbf{Z}}^{\mathbf{y}}_{t_k}(\mathbf{x})+(t-t_k) \hat{\mathbf{v}}  (\tilde{\mathbf{Z}}^{\mathbf{y}}_{t_k}(\mathbf{x}), \mathbf{y}, t_k)$. So it holds that
\begin{align*}
& 2 \int \big\langle\hat{\mathbf{v}}  (\tilde{\mathbf{Z}}^{\mathbf{y}}_t(\mathbf{x}), \mathbf{y}, t)-\hat{\mathbf{v}}  (\tilde{\mathbf{Z}}^{\mathbf{y}}_{t_k}(\mathbf{x}), \mathbf{y}, t), \hat{\mathbf{Z}}^{\mathbf{y}}_t(\mathbf{x})-\tilde{\mathbf{Z}}^{\mathbf{y}}_t(\mathbf{x})\big\rangle {\nu}_0(\mathbf{x}) \, \mathrm{d} \mathbf{x} \\
&~~~~\leq  \int \big|\hat{\mathbf{v}}  (\tilde{\mathbf{Z}}_t^{\mathbf{y}}(\mathbf{x}), \mathbf{y}, t)-\hat{\mathbf{v}}(\tilde{\mathbf{Z}}_{t_k}^{\mathbf{y}}(\mathbf{x}), \mathbf{y}, t)\big|_2^2 \,{\nu}_0(\mathbf{x}) \mkern2mu \mathrm{d} \mathbf{x}+\int \big|\hat{\mathbf{Z}}^{\mathbf{y}}_t(\mathbf{x})-\tilde{\mathbf{Z}}^{\mathbf{y}}_t(\mathbf{x})\big|_2^2 \,{\nu}_0(\mathbf{x}) \mkern2mu \mathrm{d} \mathbf{x} \\
&~~~~\leq  {d_x}\gamma_1^2(t-t_k)^2 \sup_{\mathbf{x}, \mathbf{y}, t} |\hat{\mathbf{v}}(\mathbf{x}, \mathbf{y}, t)|_2^2+L_t^{\mathbf{y}} \leq {d_x}\gamma_1^2(t-t_k)^2 K^2+L_t^{\mathbf{y}} \, .
\end{align*}
Finally, since $\hat{\mathbf{v}}(\mathbf{x}, \mathbf{y}, t)$ is $\gamma_3$-Lipschitz continuous w.r.t. $t$, we have
\begin{align*}
& \int 2\big\langle\hat{\mathbf{v}}  (\tilde{\mathbf{Z}}^{\mathbf{y}}_{t_k}(\mathbf{x}), \mathbf{y}, t)-\hat{\mathbf{v}}(\tilde{\mathbf{Z}}^{\mathbf{y}}_{t_k}(\mathbf{x}), \mathbf{y}, t_k), \hat{\mathbf{Z}}^{\mathbf{y}}_t(\mathbf{x})-\tilde{\mathbf{Z}}^{\mathbf{y}}_t(\mathbf{x})\big\rangle {\nu}_0(\mathbf{x}) \mkern2mu \mathrm{d} \mathbf{x} \\
&~~~~~~\leq  d_x\gamma_3^2(t-t_k)^2 + \int\big|\hat{\mathbf{Z}}^{\mathbf{y}}_t(\mathbf{x})-\tilde{\mathbf{Z}}^{\mathbf{y}}_t(\mathbf{x})\big|_2^2 \,{\nu}_0(\mathbf{x}) \mkern2mu \mathrm{d} \mathbf{x} = d_x\gamma_3^2(t-t_k)^2 + L_t^{\mathbf{y}} \, .
\end{align*}
All above tells us
$$
\frac{\mathrm{d} L_t^{\mathbf{y}}}{\mathrm{d} t} \leq 2( {d^{1/2}_x}\gamma_1+1) L_t^{\mathbf{y}}+{d_x}(\gamma_1^2 K^2+\gamma_3^2)(t-t_k)^2, \quad t \in (t_k, t_{k+1}) \, .
$$
By Lemma \ref{lemma:gronwall}, we obtain
$$
L^{\mathbf{y}}_{t_{k+1}} \leq L^{\mathbf{y}}_{t_{k}} e^{2( {d^{1/2}_x}\gamma_1+1) (t_{k+1}-t_k)} + e^{2( {d^{1/2}_x}\gamma_1+1) (t_{k+1}-t_k)}{d_x}(\gamma_1^2 K^2+\gamma_3^2)(t_{k+1}-t_k)^3 \, ,
$$
which implies
$$
e^{-2({d^{1/2}_x}\gamma_1+1) t_{k+1}} L^{\mathbf{y}}_{t_{k+1}}-e^{-2({d^{1/2}_x}\gamma_1+1) t_k} L^{\mathbf{y}}_{t_k} \leq {d_x}(\gamma_1^2 K^2+\gamma_3^2)(t_{k+1}-t_k)^3 \, .
$$
Recall that $t_{k+1}-t_k=\Delta t=N^{-1}T$. Since $t_0 = 0$, $t_N=T<1$ and $L^{\mathbf{y}}_0=0$, we have
$$
L^{\mathbf{y}}_T \leq {d_x}e^{2({d^{1/2}_x}\gamma_1+1)}(\gamma_1^2 K^2+\gamma_3^2) \sum_{k=0}^{N-1}(t_{k+1}-t_k)^3 \leq {d_x}e^{2({d^{1/2}_x}\gamma_1+1)}(\gamma_1^2 K^2+\gamma_3^2) N^{-2} \, .
$$
As shown in Proposition \ref{prop:generalization}, for fixed $(d_x, d_y)$, we have $K=\widetilde{\mathcal{O}}\{(1-T)^{-1}\}$, $\gamma_1={d_x^{3/2}}(1-T)^{-2}$ and $\gamma_3=\widetilde{\mathcal{O}}\{(1-T)^{-3}\}$, where $\widetilde{\mathcal{O}}(\cdot)$ omits the polynomial term of $\log n$. Hence,
\begin{align*}
\sup_{\mathbf{y} \in [0, B]^{d_y}} W^2_2\big( \mkern0.5mu \hat{p}_{_T}(\mathbf{x} ; \mathbf{y}), \mkern2mu \tilde{p}_{_T}(\mathbf{x} ; \mathbf{y})\big) \leq \sup_{\mathbf{y} \in [0, B]^{d_y}} L^{\mathbf{y}}_T=\widetilde{\mathcal{O}}\big\{{e^{2d_x^2(1-T)^{-2}}}(1-T)^{-6} N^{-2} \big\}  \, .
\end{align*}
We complete the proof of Proposition \ref{prop:discretization error bound}. $\hfill\Box$

\section{Proofs of Lemmas}
\subsection{Proof of Lemma \ref{lemma:covering number}} \label{subsec:covering number}
The first result is directly obtained from  Lemma 7 of \cite{chen2022nonparametric}, with a slight modification of the input region. To prove the second result, it suffices to show there exists a function $C(\cdot)$ such that 
\begin{align}\label{eq:elll}
\|\ell(\mathbf{x}, \mathbf{y}, \mathbf{v}_1)-\ell(\mathbf{x}, \mathbf{y}, \mathbf{v}_2)\|_{L^{\infty}([0,1]^{d_x} \times [0,B]^{d_{y}} )} \leq \delta
\end{align}
for any 
     $\mathbf{v}_1, \mathbf{v}_2\in{\rm FNN}$ satisfying 
     \begin{align}\label{eq:vbddd}
     \|\mathbf{v}_1(\mathbf{x}, \mathbf{y}, t)-\mathbf{v}_2(\mathbf{x}, \mathbf{y}, t)\|_{{L^{\infty}([-R,R]^{d_x} \times[0,B]^{d_{y}}\times{[0,T]})}} \leq C(\delta)\,,
     \end{align}
     where $R>0$ will be specified later. We rewrite $\ell(\mathbf{x}, \mathbf{y}, \mathbf{v})$ as follows:
    $$
    \begin{aligned}
    \ell(\mathbf{x}, \mathbf{y}, \mathbf{v})&=\frac{1}{T}\int_0^{{T}}\left[\mathbb{E}_{\mathbf{W}}\left(\bigg|\mathbf{x}-\frac{t}{\sqrt{1-t^2}}\mathbf{W}\bigg|_2^2\right) + \mathbb{E}_{\mathbf{W}}\big\{|\mathbf{v}  (t\mathbf{x}+\sqrt{1-t^2}\mathbf{W}, \mathbf{y}, t)|_2^2\big\}\right]\mathrm{d}t\\
    &~~~- \frac{2}{T}\int_0^{{T}}\mathbb{E}_{\mathbf{W}}\left\{\bigg(\mathbf{x}-\frac{t}{\sqrt{1-t^2}}\mathbf{W}\bigg)^{\rm{T}}\mathbf{v}  (t\mathbf{x}+\sqrt{1-t^2}\mathbf{W}, \mathbf{y}, t)\right\}\mathrm{d} t \, .
    \end{aligned}
    $$
    Then
\begin{equation} \label{equa:lv1-lv2}
\begin{aligned}
|\ell(\mathbf{x}, \mathbf{y}, \mathbf{v}_1)-\ell(\mathbf{x}, \mathbf{y}, \mathbf{v}_2)| &\leq  \underbrace{\frac{2}{T} \int_0^{{T}} \mathbb{E}_{\mathbf{W}}\Bigg(\bigg|\mathbf{x}-\frac{t}{\sqrt{1-t^2}}\mathbf{W}\bigg|_2 \cdot|\mathbf{v}_1-\mathbf{v}_2|_2\Bigg)\mkern2mu \mathrm{d} t}_{\text {(A) }} \\
&~~~ +\underbrace{\frac{1}{T} \int_0^{{T}} \mathbb{E}_{\mathbf{W}}(|\mathbf{v}_1-\mathbf{v}_2|_2 \cdot |\mathbf{v}_1+\mathbf{v}_2|_2)\,\mathrm{d} t}_{\text {(B) }} \, ,
\end{aligned}
\end{equation}
where we omit the input of $\mathbf{v}_1$ and $\mathbf{v}_2$ for brevity. In the sequel, we always assume \eqref{eq:vbddd} holds.

We first focus on the upper bound for term (A). By {Cauchy-Schwarz} inequality, 
    \begin{align}\label{equa:|B_Y(x,v1)-B_Y(x,v2)| term A}
        &\frac{1}{T} \int_0^{{T}} \mathbb{E}_{\mathbf{W}}\Bigg(\bigg|\mathbf{x}-\frac{t}{\sqrt{1-t^2}}\mathbf{W}\bigg|_2 \cdot|\mathbf{v}_1-\mathbf{v}_2|_2\Bigg)\, {\rm d} t\notag \\
        &~~~~\leq \frac{1}{T}\left\{\int_0^{{T}} \mathbb{E}_{\mathbf{W}}\left(\bigg|\mathbf{x}-\frac{t}{\sqrt{1-t^2}}\mathbf{W}\bigg|_2^2\right)\,\mathrm{d}t\right\}^{1/2}\bigg\{\int_0^{{T}} \mathbb{E}_{\mathbf{W}}(|\mathbf{v}_1-\mathbf{v}_2|_2^2)\,\mathrm{d}t\bigg\}^{1/2} \\
        &~~~~\leq \frac{{d^{1/2}_x}}{(1-T)^{1/2}}\bigg\{\frac{1}{T}\int_0^{{T}} \mathbb{E}_{\mathbf{W}}(|\mathbf{v}_1-\mathbf{v}_2|_2^2)\,\mathrm{d}t\bigg\}^{1/2} \, .\notag
    \end{align}
Recall $\mathbf{W}_t=t\mathbf{X}+\sqrt{1-t^2}\mathbf{W}$. Denote by $p_{w_t \mkern2mu|\mkern2mu x}(\mathbf{u}\mkern2mu|\mkern2mu\mathbf{x})$ the conditional density of $\mathbf{W}_t$ given $\mathbf{X}=\mathbf{x}$. Since $\mathbf{v}_1, \mathbf{v}_2 \in \mathrm{FNN}=\mathrm{FNN}(L, M, J, K, \kappa, \gamma_1, \gamma_2, \gamma_3)$, then $|\mathbf{v}_1(\mathbf{x}, \mathbf{y}, t)|_2 \leq K$ and $|\mathbf{v}_2(\mathbf{x}, \mathbf{y}, t)|_2 \leq K$. Furthermore, we have 
\begin{align*}
    &\mathbb{E}_{\mathbf{W}}\big\{|\mathbf{v}_1 
 (t\mathbf{x}+\sqrt{1-t^2}\mathbf{W},\mathbf{y},t)-\mathbf{v}_2 
 (t\mathbf{x}+\sqrt{1-t^2}\mathbf{W},\mathbf{y},t)|_2^2\big\} \\
    &~~~~~~=\int_{|\mathbf{u}|_{\infty}\leq R}|\mathbf{v}_1(\mathbf{u}, \mathbf{y}, t) - \mathbf{v}_2(\mathbf{u}, \mathbf{y}, t)|_2^2 \, p_{w_t \mkern2mu|\mkern2mu x}(\mathbf{u}\mkern2mu|\mkern2mu\mathbf{x}) \mkern2mu \mathrm{d} \mathbf{u} \\    &~~~~~~~~~+\int_{|\mathbf{u}|_{\infty}>R}|\mathbf{v}_1(\mathbf{u}, \mathbf{y}, t) - \mathbf{v}_2(\mathbf{u}, \mathbf{y}, t)|_2^2 \, p_{w_t \mkern2mu|\mkern2mu x}(\mathbf{u}\mkern2mu|\mkern2mu\mathbf{x}) \mkern2mu \mathrm{d} \mathbf{u} \\
    &~~~~~~ \leq \mkern2mu C^2(\delta)+{4K^2} \mathbb{P}\big(|\mathbf{W}_t|_{\infty}>R \mkern2mu|\mkern2mu \mathbf{X}=\mathbf{x}\big) \\
    &~~~~~~= \mkern2mu C^2(\delta)+{4K^2} \mathbb{P}\big(|t \mathbf{x}+\sqrt{1-t^2} \mathbf{W}|_{\infty}>R\big)\,.
\end{align*}
Notice that $\mathbf{W} \sim \mathcal{N}(\mathbf{0}, \mathbf{I}_{d_x})$. Write $\mathbf{W}=(W_1, \ldots, W_{d_x})^{\rm T}$ and $\mathbf{x}=(x_1, \ldots, x_{d_x})^{\rm T}$. Then
\begin{align*}
    \mathbb{P}\big(|t \mathbf{x}+\sqrt{1-t^2} \mathbf{W}|_{\infty}>R\big) &\leq \sum_{i=1}^{d_x} \mathbb{P}\big(|t x_i+\sqrt{1-t^2} W_i|>R\big) \\
    &\leq d_x \mathbb{P}\bigg(|W_1|>\frac{R-1}{\sqrt{1-t^2}} \bigg) \leq 2 d_x \exp \bigg\{\!-\frac{(R-1)^2}{2(1-t^2)}\bigg\} 
\end{align*}
for any $R>1$, 
which implies
\begin{align} \label{equa:int ||v1-v2|| estimation}
\mathbb{E}_{\mathbf{W}}\big\{|\mathbf{v}_1 
 &(t\mathbf{x}+\sqrt{1-t^2}\mathbf{W},\mathbf{y},t)-\mathbf{v}_2 
 (t\mathbf{x}+\sqrt{1-t^2}\mathbf{W},\mathbf{y},t)|_2^2\big\}  \\
 &\leq C^2(\delta)+ {8K^2 {d_x}} \exp \bigg\{\!-\frac{(R-1)^2}{2(1-t^2)}\bigg\} \leq C^2(\delta)+ {8K^2 {d_x}} \exp \bigg\{\!-\frac{(R-1)^2}{2}\bigg\} \, . \notag
\end{align}
Combining (\ref{equa:|B_Y(x,v1)-B_Y(x,v2)| term A}) and (\ref{equa:int ||v1-v2|| estimation}), we get
    \begin{align}\label{equa:term(A) estimation}
        &\frac{1}{T} \int_0^{{T}} \mathbb{E}_{\mathbf{W}}\Bigg(\bigg|\mathbf{x}-\frac{t}{\sqrt{1-t^2}}\mathbf{W}\bigg|_2 \cdot|\mathbf{v}_1-\mathbf{v}_2|_2\Bigg) \mathrm{d} t \notag\\
        &~~~~~~ \leq \mkern2mu \frac{{d^{1/2}_x}}{(1-T)^{1/2}}\left[C^2(\delta)+{8K^2 {d_x}} \exp \bigg\{\!-\frac{(R-1)^2}{2}\bigg\}\right]^{1/2} \\
        &~~~~~~ \leq \mkern2mu \frac{{d^{1/2}_x}}{(1-T)^{1/2}}\left[C(\delta)+{2^{3/2}K {d^{1/2}_x}} \exp \bigg\{\!-\frac{(R-1)^2}{4}\bigg\}\right]\,. \notag
    \end{align}

Now we consider term (B). Again, using {Cauchy-Schwarz} inequality, we have
\begin{equation} \label{equa:term(B) estimation}
    \begin{aligned}
        &\frac{1}{T} \int_0^{{T}} \mathbb{E}_{\mathbf{W}}(|\mathbf{v}_1-\mathbf{v}_2|_2 \cdot|\mathbf{v}_1+\mathbf{v}_2|_2)\mathrm{d} t \\
        &~~~~~~\leq \mkern2mu  \frac{1}{T} \bigg\{\int_0^{{T}} \mathbb{E}_{\mathbf{W}}(|\mathbf{v}_1-\mathbf{v}_2|_2^2) \mkern2mu \mathrm{d}t\bigg\}^{1/2}\bigg\{\int_0^{{T}}\mathbb{E}_{\mathbf{W}}(|\mathbf{v}_1+\mathbf{v}_2|_2^2) \mkern2mu \mathrm{d} t\bigg\}^{1/2} \\
        &~~~~~~\leq \mkern2mu 2K\left[C(\delta)+ {2^{3/2}K {d^{1 / 2}_x}} \exp \bigg\{\!-\frac{(R-1)^2}{4}\bigg\}\right] \, .
    \end{aligned}
\end{equation}
Combining \eqref{equa:lv1-lv2}, (\ref{equa:term(A) estimation}) and (\ref{equa:term(B) estimation}), we obtain
\begin{equation} \label{equa:R, C}
\begin{aligned}
    &\sup_{\mathbf{x}, \mathbf{y} \in [0,1]^{d_x} \times [0,B]^{d_{y}}} \mkern 2mu|\mkern2mu \ell(\mathbf{x}, \mathbf{y}, \mathbf{v}_1)-\ell(\mathbf{x}, \mathbf{y}, \mathbf{v}_2)| \\
    & ~~~~~~~~~\leq  \left\{2K+\frac{2d^{1/2}_x}{(1-T)^{1/2}}\right\}\left[C(\delta)+ {2^{3/2}K {d^{1 / 2}_x}} \exp \bigg\{\!-\frac{(R-1)^2}{4}\bigg\}\right]\,.
\end{aligned}
\end{equation}
Letting
\begin{align*}
&R=2\log^{1/2}\big[{2^{7/2}} K\delta^{-1} {d^{1 / 2}_x}\{K+(1-T)^{-1/2}{d^{1 / 2}_x}\}\big]+1 \, ,\\
&~~~~~~~~~~~~~ C(\delta)=\frac{\delta}{4\{K+(1-T)^{-1/2}{d^{1 / 2}_x}\}} \, ,
\end{align*}
by \eqref{equa:R, C}, we have \eqref{eq:elll} holds
for any $\|\mathbf{v}_1(\mathbf{x}, \mathbf{y}, t)-\mathbf{v}_2(\mathbf{x}, \mathbf{y}, t)\|_{L^{\infty}([-R, R]^{d_x} \times [0,B]^{d_{y}} \times[0, T])} \leq C(\delta)$. 
Hence, a $C(\delta)$-covering of FNN w.r.t. $\|\cdot\|_{L^{\infty}([-R, R]^{d_x} \times[0,B]^{d_{y}}\times[0, T])}$ induces a $\delta$-covering of $\mathcal{H}$, which implies
$$
\begin{aligned}
&\log \mathcal{N}(\delta, \mathcal{H},\|\cdot\|_{L^{\infty}([0,1]^{d_x} \times [0,B]^{d_{y}})}) \\
&~~~~\leq \log \mathcal{N}\big\{C(\delta), \mathrm{FNN},\|\cdot\|_{L^{\infty}([-R, R]^{d_x} \times [0,B]^{d_{y}} \times[0, T])}\big\} \\
&~~~~\leq C J L \begingroup\log \Big\{\{K+(1-T)^{-1/2}{d^{1 / 2}_x}\} L M \kappa \delta^{-1}\endgroup\\
&~~~~~~~~~~\begingroup\times \bigl(B\vee\log^{1/2} [K {d^{1 / 2}_x}\delta^{-1}\{K+(1-T)^{-1/2}{d^{1 / 2}_x}\} ]\bigr)\Big\} \, ,\endgroup
\end{aligned}
$$
where $C>1$ is a universal constant. We complete the proof of Lemma \ref{lemma:covering number}. $\hfill\Box$

\subsection{Proof of Lemma \ref{lemma:approx trunc error}} \label{subsec:approx trunc error}
By Proposition \ref{prop:training object}(i), we have
$$
\mathbf{v}_{\rm{F}}(\mathbf{x}, \mathbf{y}, t)  = \mathbb{E}\bigg(\mathbf{X}-\frac{t}{\sqrt{1-t^2}}\mathbf{W} \mkern2mu \Big| \mkern2mu  \mathbf{W}_t=\mathbf{x}, \mathbf{Y}=\mathbf{y}\bigg) \, ,
$$
where $\mathbf{W}_t=t\mathbf{X}+\sqrt{1-t^2}\mathbf{W}$. By \eqref{equa:globalupperbound} and \eqref{equa:def of tilde_v}, we define
$$
\tilde{K} := \sup_{(\mathbf{x}, \mathbf{y}, t) \in \mathbb{R}^{d_x} \times[0,B]^{d_{y}} \times[0, T]}|\tilde{\mathbf{v}}(\mathbf{x}, \mathbf{y}, t)|_2^2 \leq \frac{{R^2 d_x}}{(1-T)^2} \, .
$$
For any $R>1$, it holds that
    \begin{align} \label{equa:expand of int||v*-v_theta||}
        &\|\{\mathbf{v}_{\rm{F}} (\mathbf{x}, \mathbf{y}, t)-\tilde{\mathbf{v}}(\mathbf{x}, \mathbf{y}, t)\} \mathbb{I}(|\mathbf{x}|_{\infty}>R)\|^{2}_{L^2(g_t)} \notag \\
        &~~~~~~=\int \int_{|\mathbf{x}|_{\infty}>R}  |\mathbf{v}_{\rm{F}} (\mathbf{x}, \mathbf{y}, t)-\tilde{\mathbf{v}}(\mathbf{x}, \mathbf{y}, t)|_2^2 \, g_t(\mathbf{x}, \mathbf{y}) \mkern2mu \mathrm{d}\mathbf{x} \mkern2mu \mathrm{d}\mathbf{y} \notag \\
        &~~~~~~ \leq 2 \int \int_{|\mathbf{x}|_{\infty}>R} \big\{|\mathbf{v}_{\rm{F}} (\mathbf{x}, \mathbf{y}, t)|_2^2 + |\tilde{\mathbf{v}}(\mathbf{x}, \mathbf{y}, t)|_2^{2}\big \} \, g_t(\mathbf{x}, \mathbf{y}) \mkern2mu \mathrm{d}\mathbf{x} \mkern2mu \mathrm{d}\mathbf{y} \notag \\
        &~~~~~~ \leq 2 \int \int_{|\mathbf{x}|_{\infty}>R}  \bigg|\mathbb{E}\bigg(\mathbf{X}-\frac{t}{\sqrt{1-t^2}}\mathbf{W} \mkern2mu {\Big|} \mkern2mu \mathbf{W}_t=\mathbf{x}, \mathbf{Y}=\mathbf{y}\bigg)\bigg|_2^2g_t(\mathbf{x}, \mathbf{y}) \mkern2mu \mathrm{d}\mathbf{x} 
\mkern2mu \mathrm{d}\mathbf{y} \notag \\
 &~~~~~~~~~ + {2\tilde{K}}\mathbb{P}(|\mathbf{W}_t|_{\infty}>R) \notag \\
        &~~~~~~\leq 2 \int \int_{|\mathbf{x}|_{\infty}>R} \mathbb{E}\bigg(\Big|\mathbf{X}-\frac{t}{\sqrt{1-t^2}}\mathbf{W}\Big|_2^2 \mkern2mu \Big | \mkern2mu \mathbf{W}_t=\mathbf{x}, \mathbf{Y}=\mathbf{y} \bigg)g_t(\mathbf{x}, \mathbf{y}) 
\mkern2mu \mathrm{d}\mathbf{x} \mkern2mu \mathrm{d}\mathbf{y} \notag \\
&~~~~~~~~~+{2\tilde{K}}\mathbb{P}(|\mathbf{W}_t|_{\infty}>R) \, .
    \end{align}

     By Cauchy-Schwarz inequality and Jensen's inequality, it holds that
    \begin{align} \label{equa:condi term in approx trunc}
        &\int \int_{|\mathbf{x}|_{\infty}>R} \mathbb{E}\bigg(\Big|\mathbf{X}-\frac{t}{\sqrt{1-t^2}}\mathbf{W}\Big|_2^2 \mkern2mu \Big| \mkern2mu \mathbf{W}_t=\mathbf{x}, \mathbf{Y}=\mathbf{y}\bigg)g_t(\mathbf{x}, \mathbf{y}) \mkern2mu \mathrm{d}\mathbf{x} \mkern2mu \mathrm{d}\mathbf{y} \notag \\
        &~~~~~=\int \int \mathbb{E}\bigg(\Big|\mathbf{X}-\frac{t}{\sqrt{1-t^2}}\mathbf{W}\Big|_2^2 \mkern2mu \Big| \mkern2mu \mathbf{W}_t=\mathbf{x}, \mathbf{Y}=\mathbf{y}\bigg)  \mathbb{I} (|\mathbf{x}|_{\infty}>R) g_t(\mathbf{x}, \mathbf{y}) \mkern2mu \mathrm{d}\mathbf{x} \mkern2mu \mathrm{d}\mathbf{y} \notag \\ 
        &~~~~~\leq \bigg[\mathbb{E}\bigg\{ \Big| \mathbb{E}\Big(\Big|\mathbf{X}-\frac{t}{\sqrt{1-t^2}}\mathbf{W}\Big|_2^2 \mkern2mu \Big| \mkern2mu \mathbf{W}_t, \mathbf{Y}\Big) \Big|^2 \bigg\} \bigg]^{1/2} \{\mathbb{P}(|\mathbf{W}_t|_{\infty}>R)\}^{1 / 2} \notag \\ 
        &~~~~~\leq \Big\{\mathbb{E}\Big(\Big|\mathbf{X}-\frac{t}{\sqrt{1-t^2}}\mathbf{W}\Big|_2^4\Big)\Big\}^{1 / 2} \{\mathbb{P}(|\mathbf{W}_t|_{\infty}>R)\}^{1 / 2} \,.
    \end{align}

  By Assumption \ref{assump:bounded support}, it holds that $|\mathbf{X}|_{\infty} \leq 1$.
Write $\mathbf{W} = (W_1, \ldots, W_{d_x})^{\rm T}$. Using the inequality $(a+b)^2 \leq 2 a^2+2 b^2$, we have
\begin{equation} \label{equa:fourth moment of X_t}
\begin{aligned}
\mathbb{E}\Big(\Big|\mathbf{X}-\frac{t}{\sqrt{1-t^2}}\mathbf{W}\Big|_2^4\Big) &  \leq \mathbb{E}\bigg\{4|\mathbf{X}|_2^4+\frac{4t^4}{(1-t^2)^2}|\mathbf{W}|_2^4\bigg\} \\
& \leq4 {d^2_x}+\frac{4t^4}{(1-t^2)^2} \mathbb{E}\bigg(\sum_{k=1}^{d_x} W_{k}^4+\sum_{i \neq j} W_{i}^2 W_j^2\bigg) \\
& \leq4 {d^2_x} + \frac{4t^4}{(1-t^2)^2}{d_x}({d_x}+2) \leq \frac{8(d_x+2)^2}{(1-t^2)^2}
\end{aligned}
\end{equation}
for any $t \in [0, 1)$. Write $\mathbf{X} = (X_1, \ldots, X_{d_x})^{\rm T}$ and $\mathbf{W}_t = (W_{t, 1}, \ldots, W_{t, d_x})^{\rm T}$. Since $W_{i}$ is a standard Gaussian, using the union inequality, we have
\begin{align} \label{equa:tail probability bound}
\mathbb{P}(|\mathbf{W}_t|_{\infty}>R) & =\mathbb{P}\bigg(\bigcup_{i=1}^{d_x}\{|W_{t, i}|>R\}\bigg) \leq \sum_{i=1}^{d_x} \mathbb{P}(|W_{t, i}|>R) \notag \\
& \leq \sum_{i=1}^{d_x} \mathbb{P}\big(t|X_i|+\sqrt{1-t^2}|W_{i}|>R\big) \notag \\
& \leq \sum_{i=1}^{d_x} \mathbb{P}\bigg(|W_{i}|>\frac{R-1}{\sqrt{1-t^2}}\bigg)  \leq 2 d_x \exp \bigg\{\!-\frac{(R-1)^2}{2(1-t^2)}\bigg\}
\end{align}
for any $R>1$.

Combining (\ref{equa:expand of int||v*-v_theta||}), (\ref{equa:condi term in approx trunc}), (\ref{equa:fourth moment of X_t}) and \eqref{equa:tail probability bound} for $t \in [0, T]$ with $T<1$, we have
\begin{align*}
&\|(\mathbf{v}_{\rm{F}} (\mathbf{x}, \mathbf{y}, t)-\tilde{\mathbf{v}}(\mathbf{x}, \mathbf{y}, t)) \mathbb{I}\{|\mathbf{x}|_{\infty}>R\}\|^{2}_{L^2(g_{t})} \\
&~~~~\leq\frac{8({d_x}+2)^{3/2}}{1-t^2}\exp \bigg\{\!-\frac{(R-1)^2}{4(1-t^2)}\bigg\} + \frac{{4R^2d_x^2}}{(1-T)^2} \exp \bigg\{\!-\frac{(R-1)^2}{2(1-t^2)}\bigg\} \\
&~~~~\leq\frac{8(R+1)^2({d_x}+2)^{2}}{(1-T)^2}\exp \bigg\{\!-\frac{(R-1)^2}{4(1-t^2)}\bigg\} \leq {\bar{C}} (R+1)^2 \exp \bigg\{\!-\frac{(R-1)^2}{4}\bigg\} \, ,
\end{align*}
where ${\bar{C}=8(d_x+2)^2(1-T)^{-2}}$.

Letting the right-hand side in the above inequality be smaller than $\varepsilon^2$, we need to choose $R$ satisfying
\begin{align*}
    \log {\bar{C}} + 2\log(R+1) - \frac{(R-1)^2}{4} \leq 2 \log \varepsilon\,.
\end{align*}
Since $\log (R+1) \leq R$, it suffices to require
\begin{align*}
    \log {\bar{C}} + 2R - \frac{(R-1)^2}{4} \leq 2 \log \varepsilon\,,
\end{align*}
which leads to
$$
R \geq 5+2\sqrt{\log {\bar{C}} + 2 \log(\varepsilon^{-1}) + 6}\,.
$$
Hence,  $\|\{\mathbf{v}_{\rm{F}}(\mathbf{x}, \mathbf{y}, t) -\tilde{\mathbf{v}}(\mathbf{x}, \mathbf{y}, t)\} \mathbb{I}(|\mathbf{x}|_{\infty}>R)\|^{2}_{L^2(g_{t})} \leq \varepsilon^2$ if  $R\geq C \sqrt{\log \{d_x \varepsilon^{-1}(1-T)^{-1}\}}$ for some sufficiently large universal constant $C>0$.
$\hfill\Box$

{
\subsection{Proof of Lemma \ref{lemma:exact-local-realization}}
The proof proceeds in two steps.

\noindent{\bf Step 1. Exact representation of $\Psi_{\mathbf p,\mathbf q,j}$.}
Fix $(\mathbf p,\mathbf q,j)$. Let $s=d_x+d_y+1$, with
\[
\mathbf a
=\bigl((N_1\mathbf x'-\mathbf p)^{\rm T},
       (N_2\mathbf y'-\mathbf q)^{\rm T},
       N_3t'-j\bigr)^{\rm T}
\]
as defined in the proof of Proposition \ref{prop:approx}. Write
\[
\mathbf z=\bigl((\mathbf x')^{\rm T},(\mathbf y')^{\rm T},t'\bigr)^{\rm T}
\in[0,1]^s\,.
\]

For $\ell=1,\ldots,s$, let $b_\ell=\max\{0,a_1,\ldots,a_\ell\}\geq0$, with $b_0=0$. Then
\[
\begin{aligned}
b_\ell&=\max\{a_\ell,b_{\ell-1}\}
=\operatorname{ReLU}(a_\ell-b_{\ell-1})+b_{\ell-1}\\
&=\operatorname{ReLU}\!\left\{\operatorname{ReLU}(a_\ell-b_{\ell-1})+\operatorname{ReLU}(b_{\ell-1})\right\}\,.
\end{aligned}
\]
Writing $r_\ell=\operatorname{ReLU}(a_\ell-b_{\ell-1})$, we have $b_\ell=\operatorname{ReLU}(r_\ell+b_{\ell-1})$. Each update is implemented by two successive ReLU layers. The first computes $r_\ell$ while copying $\mathbf z$ and $b_{\ell-1}$. The second adds $r_\ell$ to $b_{\ell-1}$ while copying $\mathbf z$:
\[
\begin{pmatrix}
\mathbf z\\
b_{\ell-1}
\end{pmatrix}
\xrightarrow{\text{layer }2\ell-1}
\operatorname{ReLU}\!\begin{pmatrix}
\mathbf z\\
a_\ell-b_{\ell-1}\\
b_{\ell-1}
\end{pmatrix}
=\begin{pmatrix}
\mathbf z\\
r_\ell\\
b_{\ell-1}
\end{pmatrix}\,,
\]
\[
\begin{pmatrix}
\mathbf z\\
r_\ell\\
b_{\ell-1}
\end{pmatrix}
\xrightarrow{\text{layer }2\ell}
\operatorname{ReLU}\!\begin{pmatrix}
\mathbf z\\
r_\ell+b_{\ell-1}
\end{pmatrix}
=\begin{pmatrix}
\mathbf z\\
b_\ell
\end{pmatrix}\,.
\]
The copied values remain unchanged because they are nonnegative. Applying the same recursion to $-a_\ell$ in parallel computes $\max\{0,-a_1,\ldots,-a_s\}$. One final ReLU yields
\[
\Psi_{\mathbf p,\mathbf q,j}(\mathbf x',\mathbf y',t')
=\operatorname{ReLU}\!\left(1-\max\{0,a_1,\ldots,a_s\}-\max\{0,-a_1,\ldots,-a_s\}\right)\,.
\]
The two recursions share the $s$ input coordinates. Their per-update bounds are
\[
\begin{array}{c|cc}
&\text{width}&\text{nonzero parameters}\\ \hline
\text{layer }2\ell-1&s+4&s+8\\
\text{layer }2\ell&s+2&s+4
\end{array}
\]
Including the final ReLU and a scalar linear output layer, we obtain
\[
\begin{aligned}
\text{hidden layers}&\leq2s+1\,,\qquad \text{width}\leq s+4\,,\\
\text{nonzero parameters}
&\leq s\bigl[(s+8)+(s+4)\bigr]+3+1
=s(2s+12)+4\,.
\end{aligned}
\]
All weights and biases have absolute values at most $N_1\vee N_2\vee N_3$.

\medskip
\noindent{\bf Step 2. Exact representation of $\bar{\mathbf v}$.}
For the $Q=(N_1+1)^{d_x}(N_2+1)^{d_y}(N_3+1)$ grid nodes, parallelize the subnetworks from Step 1 with shared input $\mathbf z$ and use the grid values as coefficients in a linear output layer to obtain
\[
\bar{\mathbf v}(\mathbf x',\mathbf y',t')
=\sum_{\mathbf p,\mathbf q,j}
\mathbf v\!\left(\frac{\mathbf p}{N_1},\frac{\mathbf q}{N_2},\frac{j}{N_3}\right)
\Psi_{\mathbf p,\mathbf q,j}(\mathbf x',\mathbf y',t')\,.
\]
The layerwise bounds in Step 1 and $Q\geq2^s\geq s$ give
\begin{align*}
&L\leq2s+2\,,\qquad M\leq s+4Q\leq5Q\,,\\
&J\leq s(2s+12Q)+3Q+d_xQ\leq18sQ\,.
\end{align*}
For sufficiently small $\varepsilon_*$, the choices of $N_1,N_2,N_3$ give
\[
Q\leq5(1\vee T\tau_{R,T})(8R\zeta_Td_x^{1/2})^{d_x}(5B\omega_{R,T}d_y^{1/2})^{d_y}\varepsilon^{-d_x-d_y-1}\,.
\]

As established above, the grid values and $\bar{\mathbf v}$ on $[0,1]^s$ have Euclidean norms bounded by $K\leq R\zeta_Td_x^{1/2}$, so
\[
\kappa\leq K\vee N_1\vee N_2\vee N_3
\lesssim\frac{R\zeta_Td_x^{1/2}\vee B\omega_{R,T}d_y^{1/2}\vee T\tau_{R,T}}{\varepsilon}\,.
\]
This completes the proof.
$\hfill\Box$
}

\section{Brief Review of NNKCDE and FlexCode}\label{sec:review-baseline}
In this section, we review two non-parametric methods for conditional density estimation that serve as baseline in our numerical experiments: Nearest-Neighbors Kernel Conditional Density Estimation (NNKCDE) \citep{dalmasso2020conditional} and Flexible Conditional Density Estimation (FlexCode) \citep{izbicki2017converting}.

 NNKCDE is a straightforward and interpretable approach for conditional density estimation. It constructs a kernel density estimate by leveraging the $k$ nearest neighbors of a given condition $\mathbf{y}$, which is governed by two primary hyper-parameters: the number of nearest neighbors $k$, and the kernel bandwidth $h$. 
Specifically, the conditional density of $\mathbf{X}$ given $\mathbf{Y} = \mathbf{y}$ is estimated as follows:
\begin{align*}
 \hat{p}_{x \mkern2mu|\mkern2mu y}(\mathbf{x} \mkern2mu|\mkern2mu \mathbf{y}) = \frac{1}{k} \sum_{i=1}^k K_h \{ \rho(\mathbf{x}, \mathbf{x}_{N_i(\mathbf{y})})\} \, ,
\end{align*}
where $K_h(u) = h^{-1}K(u/h)$ with some kernel function (typically Gaussian) $K(\cdot)$, $\rho$ denotes a distance metric, and $N_i(\mathbf{y})$ refers to the index of the $i$-th nearest neighbor of $\mathbf{y}$. 
Due to its simplicity, NNKCDE is often more interpretable and easier to implement than more sophisticated conditional density estimation methods, particularly in scenarios involving limited training data.

 FlexCode formulates the conditional density estimation through a basis expansion of the univariate response variable $X$, transforming the density estimation task into a series of univariate regression problems. 
One of the main strengths of FlexCode lies in its adaptability, as it allows for any suitable regression technique, thus enabling alignment with the specific structure or characteristics of the dataset.
Specifically, let $\{\phi_j(x)\}_{j=1}^\infty$ be an orthonormal basis for square-integrable functions, such as a Fourier or wavelet basis. For each fixed $\mathbf{y}$, the true conditional density $p_{x\mkern2mu|\mkern2mu y}(x \mkern2mu|\mkern2mu \mathbf{y})$ can be expanded as:
$$p_{x\mkern2mu|\mkern2mu y}(x\mkern2mu|\mkern2mu\mathbf{y}) = \sum_{j=1}^\infty \beta_j(\mathbf{y}) \phi_j(x) \, .$$
Due to the orthonormality of the basis, each coefficient $\beta_j(\mathbf{y})$ is the conditional expectation of the transformed response variable $\phi_j(X)$ given $\mathbf{Y}=\mathbf{y}$:
$$\beta_j(\mathbf{y}) = \int p_{x\mkern2mu|\mkern2mu y}(x\mkern2mu|\mkern2mu\mathbf{y}) \phi_j(x) \, {\rm d}x = \mathbb{E}\{\phi_j(X) \mkern2mu|\mkern2mu \mathbf{Y}=\mathbf{y} \} \, .$$
Thus, given i.i.d. samples $\{(X_i, \mathbf{Y}_i)\}_{i=1}^n \sim p_{x, y}(x, \mathbf{y})$, we can use regression methods  to obtain an estimator $\hat{\beta}_j(\mathbf{y})$ for each coefficient ${\beta}_j(\mathbf{y})$. In practice, we truncate the series to a finite number of $l$ terms. The FlexCode estimator is then constructed using the estimated coefficients ${\{\hat{\beta}_j(\mathbf{y})\}_{j=1}^l}$:
\begin{align*}
\hat{p}_{x\mkern2mu|\mkern2mu y}(x\mkern2mu|\mkern2mu \mathbf{y}) = {\sum_{j=1}^l \hat{\beta}_j(\mathbf{y}) \phi_j(x)} \, .
\end{align*}
 FlexCode effectively reframes the problem of density estimation as a regression task, allowing for considerable modeling flexibility. 
The method includes two main hyperparameters: the number of basis terms retained in the expansion, and the choice of regression algorithm.

 \section{Practical Influence of the Sample Size \texorpdfstring{$n$}{n}} \label{sec:Influence n}
 In this section, we examine the performance of our proposed method with respect to the sample size $n$ of training dataset through the settings of simulation studies I and II mentioned in Sections \ref{subsec:simulation1} and \ref{subsec:simulation2} of the main paper. The relevant results are shown in Tables \ref{tab:sim1-convergence-n} and \ref{tab:sim2-convergence-n} below. These results indicate that as $n$ increases, our proposed method performs better accordingly, which is consistent with Theorem \ref{thm:main thm}.

\begin{table}[H]
\centering
\setlength{\belowcaptionskip}{5pt}
\renewcommand{\arraystretch}{0.75}
\caption{Influence of $n$ on the total variation distance error of estimated conditional density obtained by our proposed method in simulation study I.}
\label{tab:sim1-convergence-n}
\begin{tabular}{ccccccccc}
\toprule
 & \multicolumn{2}{c}{4 squares} & \multicolumn{2}{c}{checkerboard} & \multicolumn{2}{c}{pinwheel} & \multicolumn{2}{c}{Swiss roll} \\
 $n$ & AVE & STD & AVE & STD & AVE & STD & AVE & STD \\
\midrule
1000 & 0.093 & 0.040 & 0.214 & 0.130 & 0.157 & 0.084 & 0.156 & 0.073 \\
2000 & 0.085 & 0.041 & 0.152 & 0.070 & 0.147 & 0.051 & 0.122 & 0.030 \\
10000 & 0.068 & 0.022 & 0.130 & 0.059 & 0.132 & 0.041 & 0.112 & 0.036 \\
40000 & 0.053 & 0.007 & 0.122 & 0.054 & 0.126 & 0.035 & 0.096 & 0.030 \\
\bottomrule
\end{tabular}
\end{table}

\begin{table}[H]
\centering
\setlength{\belowcaptionskip}{5pt}
\renewcommand{\arraystretch}{0.75}
\caption{Influence of $n$ on the MSE of estimated conditional mean (${\rm MSE}_1$) and standard deviation (${\rm MSE}_2$) of our proposed method in simulation study II.}
\label{tab:sim2-convergence-n}
\begin{tabular}{ccccccccccc}
\toprule
 & \multicolumn{2}{c}{M1} & \multicolumn{2}{c}{M2} & \multicolumn{2}{c}{M3} \\
 $n$ & ${\rm MSE}_1$ & ${\rm MSE}_2$ & ${\rm MSE}_1$ & ${\rm MSE}_2$ & ${\rm MSE}_1$ & ${\rm MSE}_2$ \\
\midrule
1250 & 0.338 & 0.011 & 1.022 & 0.333 & 3.956 & 0.144 \\
2500 & 0.085 & 0.003 & 0.673 & 0.151 & 2.997 & 0.097 \\
5000 & 0.042 & 0.001 & 0.177 & 0.096 & 1.557 & 0.068 \\
\bottomrule
\end{tabular}
\end{table}

 \section{Practical Influence of the Stopping Time \texorpdfstring{$T$}{T}} \label{sec:Influence T}
\begin{figure}[H]
  \centering
  \includegraphics[width=0.7\linewidth]{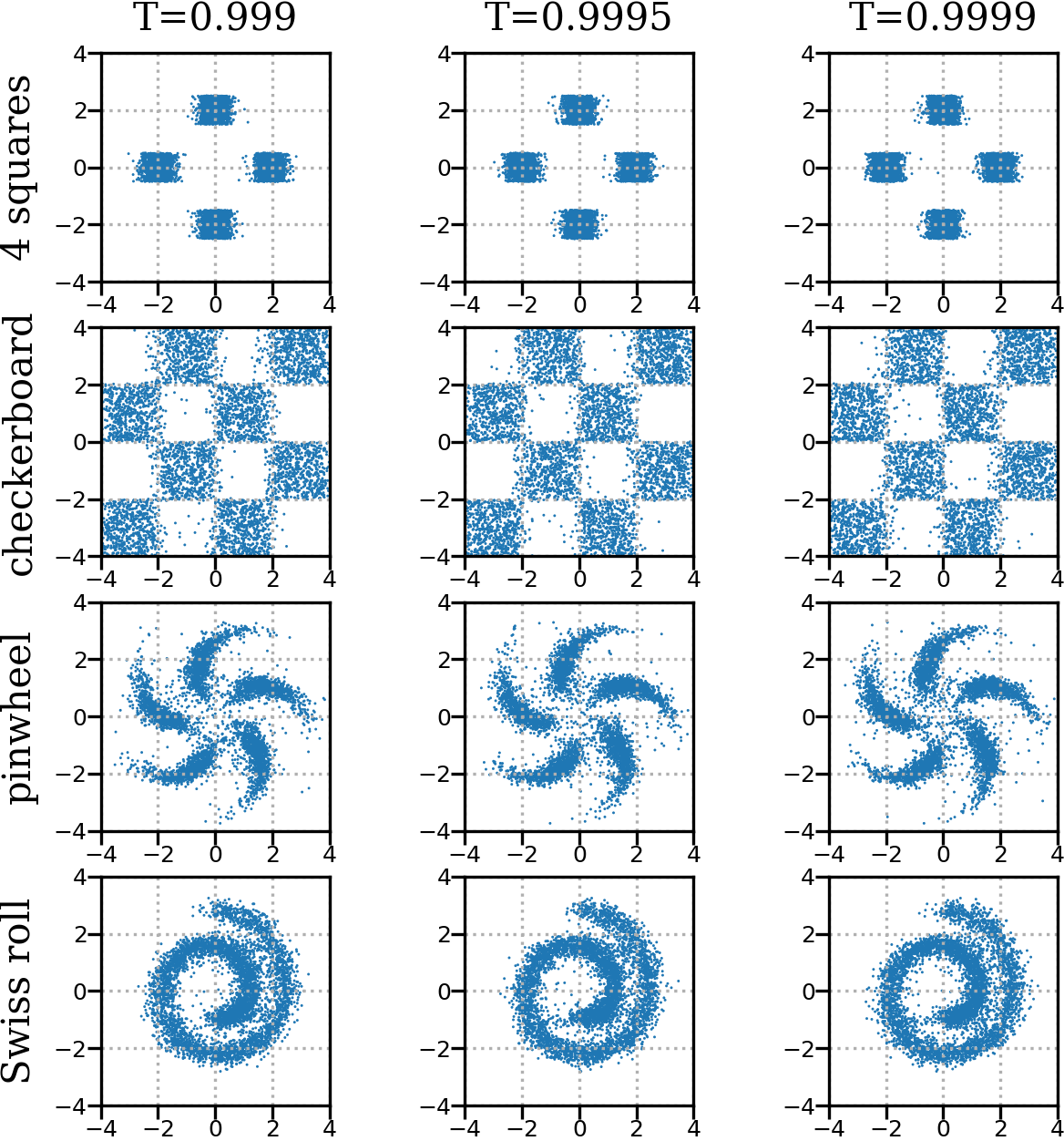}
  \caption{Scatter plots of the pairwise samples generated under different stopping {times} $T=0.999$, $0.9995$ and $0.9999$.}
  \label{fig:sim1-T}
\end{figure}

 When $T$ approaches 1, the training dynamics tend to become less stable due to weaker regularization of the velocity field near $t$=1. 
More precisely, as $T$ approaches 1, optimization of the velocity neural network will be difficult due to the term $t/\sqrt{1 - t^2}$ in the objective function becoming unbounded. However, in practice, the choice of $T$ is generally not an issue. We empirically examine the sensitivity of $T$ across all the numerical studies when $T$ increases from $0.999$ to $0.9999$. Tables \ref{tab:sim1-T}--\ref{tab:mnist-inpaint-T} summarize the results. In simulation study I,
Figure \ref{fig:sim1-T} shows that the generated samples are visually indistinguishable with different stopping times $T$. These results suggest that the quality of the generated conditional samples is not sensitive to the precise choice of $T$, provided that $T$ is sufficiently close to 1. 
{ We use the same training checkpoint, and test with different stopping times $T$ to show its influence on the generated results.}

\begin{table}[H]
\setlength{\belowcaptionskip}{5pt}
\renewcommand{\arraystretch}{0.75}
\centering
\caption{The sample average and standard deviation of 100 obtained total variation distances based on our proposed method with different $T$ in simulation study I.}
\label{tab:sim1-T}
\begin{tabular}{ccccccccccc}
\toprule
 & \multicolumn{2}{c}{4 squares} & \multicolumn{2}{c}{checkerboard} & \multicolumn{2}{c}{pinwheel} & \multicolumn{2}{c}{Swiss roll} \\
$T$ & AVE & STD & AVE & STD & AVE & STD & AVE & STD \\
\midrule
0.999 & 0.054 & 0.015 & 0.110 & 0.075 & 0.116 & 0.041 & 0.093 & 0.027 \\
0.9995 & 0.053 & 0.015 & 0.110 & 0.075 & 0.116 & 0.041 & 0.093 & 0.028 \\
0.9999 & 0.053 & 0.015 & 0.110 & 0.075 & 0.116 & 0.041 & 0.093 & 0.028 \\
\bottomrule
\end{tabular}
\end{table}

\begin{table}[H]
\setlength{\belowcaptionskip}{5pt}
\renewcommand{\arraystretch}{0.75}
\centering
\caption{Influence of $T$ on the MSE of estimated conditional mean (${\rm MSE}_1$) and standard deviation (${\rm MSE}_2$) of our proposed method in simulation study II. }
\label{tab:sim2-T}
\begin{tabular}{ccccccccccc}
\toprule
 & \multicolumn{2}{c}{M1} & \multicolumn{2}{c}{M2} & \multicolumn{2}{c}{M3} \\
$T$ & ${\rm MSE}_1$ & ${\rm MSE}_2$ & ${\rm MSE}_1$ & ${\rm MSE}_2$ & ${\rm MSE}_1$ & ${\rm MSE}_2$ \\
\midrule
0.999 & { 0.023} & 0.001 & { 0.214} & { 0.153} & { 0.277} & { 0.051} \\
0.9995 & { 0.023} & 0.001 & { 0.214} & { 0.153} & { 0.277} & { 0.051} \\
0.9999 & { 0.023} & 0.001 & { 0.213} & { 0.152} & { 0.277} & { 0.051} \\
\bottomrule
\end{tabular}
\end{table}

\begin{table}[H]
\setlength{\belowcaptionskip}{5pt}
\renewcommand{\arraystretch}{0.75}
\centering
\caption{Comparison of prediction interval coverage by our proposed method with different $T$ in real data analysis I. }
\label{tab:wine-T}
\begin{tabular}{cccc}
\toprule
$T$ & $\alpha=0.01$ & $\alpha=0.05$ & $\alpha=0.10$ \\
\midrule
0.999 & 98.31\% & 94.77\% & 90.62\% \\
0.9995 & 98.31\% & 94.46\% & 90.00\% \\
0.9999 & 98.31\% & 94.46\% & 89.85\% \\
\bottomrule
\end{tabular}
\end{table}

\begin{table}[H]
\setlength{\belowcaptionskip}{5pt}
\renewcommand{\arraystretch}{0.75}
\centering
\caption{FIDs for our proposed method with different $T$ in class conditional image generation on MNIST dataset.}
\label{tab:mnist-class-T}
\begin{tabular}{cccccc}
\toprule
$T$  & 0.999 & 0.9995 & 0.9999 \\
\midrule
FID  & 0.30 & 0.26 & 0.24 \\
\bottomrule
\end{tabular}
\end{table}

\begin{table}[H]
\setlength{\belowcaptionskip}{5pt}
\renewcommand{\arraystretch}{0.75}
\centering
\caption{Comparison of FIDs for our proposed method with different $T$ in image inpainting on MNIST dataset.}
\label{tab:mnist-inpaint-T}
\begin{tabular}{cccc}
\toprule
$T$ & $\delta=3/4$ & $\delta=1/2$ & $\delta=1/4$ \\
\midrule
0.999 & 0.32 & 0.35 & 0.40 \\
0.9995 & 0.29 & 0.31 & 0.36 \\
0.9999 & 0.28 & 0.29 & 0.34 \\
\bottomrule
\end{tabular}
\end{table}


\begin{thebibliography}{xx}
\spacingset{1.2}
\harvarditem[Albergo et~al.(2023)]{Albergo, Boffi \harvardand\ Vanden-Eijnden}{2023}{albergo2023stochastic}
Albergo, M.~S., Boffi, N.~M. \harvardand\ Vanden-Eijnden, E.  \harvardyearleft 2023\harvardyearright .
\newblock Stochastic interpolants: A unifying framework for flows and diffusions, {\it arXiv:2303.08797}.

\harvarditem[Albergo et~al.(2024)]{Albergo, Goldstein, Boffi, Ranganath \harvardand\ Vanden-Eijnden}{2024}{Albergo2024}
Albergo, M.~S., Goldstein, M., Boffi, N.~M., Ranganath, R. \harvardand\ Vanden-Eijnden, E.  \harvardyearleft 2024\harvardyearright .
\newblock Stochastic Interpolants with Data-Dependent Couplings, {\it International Conference on Machine Learning}.

\harvarditem{Albergo \harvardand\ Vanden-Eijnden}{2023}{albergo2022building}
Albergo, M.~S. \harvardand\ Vanden-Eijnden, E.  \harvardyearleft 2023\harvardyearright .
\newblock Building normalizing flows with stochastic interpolants, {\it International Conference on Learning Representations}.

\bibitem[Allen-Zhu et~al.(2019)]{allen2019convergence}
Allen-Zhu, Z., Li, Y. and Song, Z. (2019). A convergence theory for deep learning via over-
parameterization, \textit{International Conference on Machine Learning}.

\bibitem[Arjovsky et~al.(2017)]{arjovsky2017wasserstein}
Arjovsky, M., Chintala, S. and Bottou, L. (2017). Wasserstein generative adversarial networks, \textit{International Conference on Machine Learning}.

\harvarditem[Benton et~al.]{Benton, De~Bortoli, Doucet \harvardand\ Deligiannidis}{2023}{benton2023linear}
Benton, J., De~Bortoli, V., Doucet, A. \harvardand\ Deligiannidis, G.  \harvardyearleft 2023\harvardyearright .
\newblock Linear convergence bounds for diffusion models via stochastic localization, {\it arXiv:2308.03686}.

\bibitem[Benton et~al.(2024)]{benton2024error}
Benton, J., Deligiannidis, G. and Doucet, A. (2024). Error bounds for flow matching methods, {\it Transactions on Machine Learning Research}.

\harvarditem{Butcher}{2016}{butcher2016numerical}
Butcher, J.~C.  \harvardyearleft 2016\harvardyearright .
\newblock {\it Numerical methods for ordinary differential equations}, John Wiley \& Sons.

\harvarditem[Chen et~al.(2023a)]{Chen, Lee \harvardand\ Lu}{2023}{chen2023improved}
Chen, H., Lee, H. \harvardand\ Lu, J.  \harvardyearleft 2023a\harvardyearright .
\newblock Improved analysis of score-based generative modeling: User-friendly bounds under minimal smoothness assumptions, {\it International Conference on Machine Learning}.

\harvarditem[Chen et~al.(2023b)]{Chen, Huang, Zhao \harvardand\ Wang}{2023}{chen2023score}
Chen, M., Huang, K., Zhao, T. \harvardand\ Wang, M.  \harvardyearleft 2023b\harvardyearright .
\newblock Score approximation, estimation and distribution recovery of diffusion models on low-dimensional data, {\it International Conference on Machine Learning}.

\harvarditem[Chen et~al.(2023c)]{Chen, Chewi, Lee, Li, Lu \harvardand\ Salim}{2023}{chen2023probability}
Chen, S., Chewi, S., Lee, H., Li, Y., Lu, J. \harvardand\ Salim, A.  \harvardyearleft 2023c\harvardyearright .
\newblock The probability flow ode is provably fast, {\it Advances in Neural Information Processing Systems} {\bf 36}.

\harvarditem[Chen et~al.(2023d)]{Chen, Chewi, Li, Li, Salim \harvardand\ Zhang}{2023}{chen2023sampling}
Chen, S., Chewi, S., Li, J., Li, Y., Salim, A. \harvardand\ Zhang, A.~R.  \harvardyearleft 2023d\harvardyearright .
\newblock Sampling is as easy as learning the score: theory for diffusion models with minimal data assumptions, {\it International Conference on Learning Representations}.

\harvarditem[Chen et~al.(2023e)]{Chen, Daras \harvardand\ Dimakis}{2023}{chen2023restoration}
Chen, S., Daras, G. \harvardand\ Dimakis, A.  \harvardyearleft 2023e\harvardyearright .
\newblock Restoration-degradation beyond linear diffusions: A non-asymptotic analysis for ddim-type samplers, {\it International Conference on Machine Learning}.

\harvarditem{Chen \harvardand\ Linton}{2001}{chen2001}
Chen, X. \harvardand\ Linton, O.  \harvardyearleft 2001\harvardyearright .
\newblock The estimation of conditional densities, {\it In Asymptotics in Statistics and Probability, Festschrift for George Roussas, ed. M.L. Puri.}

\bibitem[Cortez et~al.(2009)]{misc_wine_quality_186}
Cortez, P., Cerdeira, A., Almeida, F., Matos, T. and Reis, J. (2009). Wine Quality, UCI
Machine Learning Repository. DOI: https://doi.org/10.24432/C56S3T.

\harvarditem[Dalmasso et~al.]{Dalmasso, Pospisil, Lee, Izbicki, Freeman \harvardand\ Malz}{2020}{dalmasso2020conditional}
Dalmasso, N., Pospisil, T., Lee, A.~B., Izbicki, R., Freeman, P.~E. \harvardand\ Malz, A.~I.  \harvardyearleft 2020\harvardyearright .
\newblock Conditional density estimation tools in python and r with applications to photometric redshifts and likelihood-free cosmological inference, {\it Astronomy and Computing} {\bf 30}:~100362.

\harvarditem{De~Bortoli}{2022}{de2022convergence}
De~Bortoli, V.  \harvardyearleft 2022\harvardyearright .
\newblock Convergence of denoising diffusion models under the manifold hypothesis, {\it Transactions on Machine Learning Research} .

\bibitem[Du et~al.(2019)]{du2019gradient}
Du, S., Lee, J., Li, H., Wang, L. and Zhai, X. (2019). Gradient descent finds global
minima of deep neural networks, \textit{International Conference on Machine Learning}.

\harvarditem[Esser et~al.]{Esser, Kulal, Blattmann, Entezari, M{\"u}ller, Saini, Levi, Lorenz, Sauer, Boesel et~al.}{2024}{esser2024scaling}
Esser, P., Kulal, S., Blattmann, A., Entezari, R., M{\"u}ller, J., Saini, H., Levi, Y., Lorenz, D., Sauer, A., Boesel, F. et~al.  \harvardyearleft 2024\harvardyearright .
\newblock Scaling rectified flow transformers for high-resolution image synthesis, {\it arXiv preprint arXiv:2403.03206}.

\harvarditem[Fan et~al.]{Fan, Yao \harvardand\ Tong}{1996}{fan1996estimation}
Fan, J., Yao, Q. \harvardand\ Tong, H.  \harvardyearleft 1996\harvardyearright .
\newblock Estimation of conditional densities and sensitivity measures in nonlinear dynamical systems, {\it Biometrika} {\bf 83}(1):~189--206.

\harvarditem{Fan \harvardand\ Yim}{2004}{fan2004crossvalidation}
Fan, J. \harvardand\ Yim, T.~H.  \harvardyearleft 2004\harvardyearright .
\newblock A crossvalidation method for estimating conditional densities, {\it Biometrika} {\bf 91}(4):~819--834.

\bibitem[Fukumizu et~al.(2025)]{fukumizuflow}
Fukumizu, K., Suzuki, T., Isobe, N., Oko, K. and Koyama, M. (2025). Flow matching achieves almost minimax optimal convergence, {\it The Thirteenth International Conference on Learning Representations}.

\bibitem[Gao and Zhu, 2024]{gao2024convergence}
Gao, X. and Zhu, L. (2024). Convergence analysis for general probability flow {ODE}s of diffusion models in {W}asserstein distances, {\it arXiv:2401.17958}.

\harvarditem[Gao et~al.]{Gao, Huang \harvardand\ Jiao}{2024}{gao2024gaussian}
Gao, Y., Huang, J. \harvardand\ Jiao, Y.  \harvardyearleft 2024\harvardyearright .
\newblock Gaussian interpolation flows, {\it Journal of Machine Learning Research}.

\harvarditem[Goodfellow et~al.]{Goodfellow, Pouget-Abadie, Mirza, Xu, Warde-Farley, Ozair, Courville \harvardand\ Bengio}{2014}{goodfellow2014generative}
Goodfellow, I.~J., Pouget-Abadie, J., Mirza, M., Xu, B., Warde-Farley, D., Ozair, S., Courville, A. \harvardand\ Bengio, Y.  \harvardyearleft 2014\harvardyearright .
\newblock Generative adversarial nets, {\it Advances in Neural Information Processing Systems} \textbf{27}.

\bibitem[Gy\"orfi et~al.(2002)]{gyorfi2002distribution}
Gy\"orfi, L., Kohler, M., Krzyzak, A., Walk, H. et al. (2002). \textit{A distribution-free theory of
nonparametric regression}, Vol. 1, Springer.

\harvarditem{Hall \harvardand\ Yao}{2005}{hall2005}
Hall, P. \harvardand\ Yao, Q.  \harvardyearleft 2005\harvardyearright .
\newblock Approximating conditional distribution functions using dimension reduction, {\it Annals of {Statistics}} {\bf 33}(3):~1404--1421.

\harvarditem[Heusel et~al.]{Heusel, Ramsauer, Unterthiner, Nessler \harvardand\ Hochreiter}{2017}{heusel2017gans}
Heusel, M., Ramsauer, H., Unterthiner, T., Nessler, B. \harvardand\ Hochreiter, S.  \harvardyearleft 2017\harvardyearright .
\newblock Gans trained by a two time-scale update rule converge to a local nash equilibrium, {\it Advances in Neural Information Processing Systems} \textbf{30}.

\harvarditem[Ho et~al.]{Ho, Jain \harvardand\ Abbeel}{2020}{ho2020denoising}
Ho, J., Jain, A. \harvardand\ Abbeel, P.  \harvardyearleft 2020\harvardyearright .
\newblock Denoising diffusion probabilistic models, {\it Advances in Neural Information Processing Systems} \textbf{33}.

\harvarditem[Ho \harvardand\ Salimans]{Ho \harvardand\ Salimans}{2021}{ho2021classifierfree}
Ho, J. \harvardand\ Salimans, T.  \harvardyearleft 2021\harvardyearright .
\newblock Classifier-Free Diffusion Guidance, {\it NeurIPS 2021 Workshop on Deep Generative Models and Downstream Applications}.
\newblock \url{https://openreview.net/forum?id=qw8AKxfYbI}

\harvarditem[Huang et~al.]{Huang, Huang, Li \harvardand\ Shen}{2023}{huang2023conditional}
Huang, D., Huang, J., Li, T. \harvardand\ Shen, G.  \harvardyearleft 2023\harvardyearright .
\newblock Conditional stochastic interpolation for generative learning, {\it arXiv:2312.05579}.

\harvarditem[Hyndman et~al.]{Hyndman, Bashtannyk \harvardand\ Grunwald}{1996}{hyndman1996estimating}
Hyndman, R.~J., Bashtannyk, D.~M. \harvardand\ Grunwald, G.~K.  \harvardyearleft 1996\harvardyearright .
\newblock Estimating and visualizing conditional densities, {\it Journal of Computational and Graphical Statistics} {\bf 5}(4):~315--336.

\harvarditem{Izbicki \harvardand\ Lee}{2016}{izbicki2016nonparametric}
Izbicki, R. \harvardand\ Lee, A.~B.  \harvardyearleft 2016\harvardyearright .
\newblock Nonparametric conditional density estimation in a high-dimensional regression setting, {\it Journal of Computational and Graphical Statistics} {\bf 25}(4):~1297--1316.

\harvarditem{Izbicki \harvardand\ Lee}{2017}{izbicki2017converting}
Izbicki, R. \harvardand\ Lee, A.~B.  \harvardyearleft 2017\harvardyearright .
\newblock Converting high-dimensional regression to high-dimensional conditional density estimation, {\it Electronic Journal of Statistics} {\bf 11}(2):~2800--2831.

\harvarditem[Karras et~al.]{Karras, Laine \harvardand\ Aila}{2019}{karras2019style}
Karras, T., Laine, S. \harvardand\ Aila, T.  \harvardyearleft 2019\harvardyearright .
\newblock A style-based generator architecture for generative adversarial networks, {\it Proceedings of the IEEE/CVF Conference on Computer Vision and Pattern Recognition}, pp.~4401--4410.

\bibitem[Kingma and Welling(2013)]{kingma2013auto}
Kingma, D. P. and Welling, M. (2013). Auto-encoding variational Bayes, \textit{arXiv preprint arXiv:1312.6114}.

\bibitem[Kingma and Ba(2015)]{kingma2015adam}
Kingma, D. P. and Ba, J. (2015). Adam: A method for stochastic optimization, \textit{International Conference on Learning Representations}.

\harvarditem[Kingma et~al.]{Kingma, Rezende, Mohamed \harvardand\ Welling}{2014}{kingma2014semi}
Kingma, D.~P., Rezende, D.~J., Mohamed, S. \harvardand\ Welling, M.  \harvardyearleft 2014\harvardyearright .
\newblock Semi-supervised learning with deep generative models, {\it Advances in Neural Information Processing Systems} {\bf 27}.

\harvarditem{LeCun}{1998}{lecun1998mnist}
LeCun, Y.  \harvardyearleft 1998\harvardyearright .
\newblock The mnist database of handwritten digits. {\hypersetup{urlcolor=black}\urlstyle{same}\harvardurl{\url{http://yann.lecun.com/exdb/mnist/}}}

\harvarditem[Lee et~al.]{Lee, Lu \harvardand\ Tan}{2023}{lee2023convergence}
Lee, H., Lu, J. \harvardand\ Tan, Y.  \harvardyearleft 2023\harvardyearright .
\newblock Convergence of score-based generative modeling for general data distributions, {\it International Conference on Algorithmic Learning Theory}.

\bibitem[Li et~al.(2024)]{li2024towards}
Li, G., Wei, Y., Chen, Y. and Chi, Y. (2024). Towards non-asymptotic convergence for diffusion-based generative models, {\it The Twelfth International Conference on Learning Representations}.

\bibitem[Lipman et~al.(2023)]{Lipman2023Flow}
Lipman, Y., Chen, R. T., Ben-Hamu, H., Nickel, M. and Le, M. (2023). Flow matching for generative modeling, \textit{The Eleventh International Conference on Learning Representations}.

\bibitem[Liu et~al.(2020)]{Liu2020On}
Liu, L., Jiang, H., He, P., Chen, W., Liu, X., Gao, J. and Han, J. (2020). On the variance
of the adaptive learning rate and beyond, \textit{International Conference on Learning Representations}.

\harvarditem[Liu et~al.(2023a)]{Liu, Gong \harvardand\ Liu}{2023}{liu2022flow}
Liu, X., Gong, C. \harvardand\ Liu, Q.  \harvardyearleft 2023a\harvardyearright .
\newblock Flow straight and fast: Learning to generate and transfer data with rectified flow, {\it International Conference on Learning Representations}.

\harvarditem[Liu et~al.(2023b)]{Liu, Wu, Zhang, Gong, Ping \harvardand\ Liu}{2023}{liu2023flowgrad}
Liu, X., Wu, L., Zhang, S., Gong, C., Ping, W. \harvardand\ Liu, Q.  \harvardyearleft 2023b\harvardyearright .
\newblock Flowgrad: Controlling the output of generative odes with gradients, {\it Proceedings of the IEEE/CVF Conference on Computer Vision and Pattern Recognition}, pp.~24335--24344.

\harvarditem[Liu et~al.]{Liu, Zhang, Li, Yan, Gao, Chen, Yuan, Huang, Sun, Gao et~al.}{2024}{liu2024sora}
Liu, Y., Zhang, K., Li, Y., Yan, Z., Gao, C., Chen, R., Yuan, Z., Huang, Y., Sun, H., Gao, J. et~al.  \harvardyearleft 2024\harvardyearright .
\newblock Sora: A review on background, technology, limitations, and opportunities of large vision models, {\it arXiv preprint arXiv:2402.17177}.

\harvarditem[Meng et~al.]{Meng, He, Song, Song, Wu, Zhu \harvardand\ Ermon}{2021}{meng2021sdedit}
Meng, C., He, Y., Song, Y., Song, J., Wu, J., Zhu, J.-Y. \harvardand\ Ermon, S.  \harvardyearleft 2021\harvardyearright .
\newblock Sdedit: Guided image synthesis and editing with stochastic differential equations, {\it International Conference on Learning Representations}.

\harvarditem[Mirza \harvardand\ Osindero]{Mirza \harvardand\ Osindero}{2014}{mirza2014conditional}
Mirza, M. \harvardand\ Osindero, S.  \harvardyearleft 2014\harvardyearright .
\newblock Conditional generative adversarial nets, {\it arXiv preprint arXiv:1411.1784}.

\harvarditem[Oko et~al.]{Oko, Akiyama \harvardand\ Suzuki}{2023}{oko2023diffusion}
Oko, K., Akiyama, S. \harvardand\ Suzuki, T.  \harvardyearleft 2023\harvardyearright .
\newblock Diffusion models are minimax optimal distribution estimators, {\it International Conference on Machine Learning}.

\harvarditem{Rosenblatt}{1969}{rosenblatt1969}
Rosenblatt, M.  \harvardyearleft 1969\harvardyearright .
\newblock Conditional probability density and regression estimators, {\it in} P.~R. Krishnaiah (ed.), {\it Multivariate analysis, {II}}, Academic Press, New York, pp.~25--31.

\harvarditem[Shi et~al.]{Shi, De~Bortoli, Deligiannidis \harvardand\ Doucet}{2022}{shi2022conditional}
Shi, Y., De~Bortoli, V., Deligiannidis, G. \harvardand\ Doucet, A.  \harvardyearleft 2022\harvardyearright .
\newblock Conditional simulation using diffusion schr{\"o}dinger bridges, {\it The 38th Conference on Uncertainty in Artificial Intelligence}.

\harvarditem[Song et~al.]{Song, Sohl-Dickstein, Kingma, Kumar, Ermon \harvardand\ Poole}{2021}{Song2021}
Song, Y., Sohl-Dickstein, J., Kingma, D.~P., Kumar, A., Ermon, S. \harvardand\ Poole, B.  \harvardyearleft 2021\harvardyearright .
\newblock Score-based generative modeling through stochastic differential equations, {\it International Conference on Learning Representations}.

\harvarditem[Sugiyama et~al.]{Sugiyama, Takeuchi, Suzuki, Kanamori, Hachiya \harvardand\ Okanohara}{2010}{Sugi2010}
Sugiyama, M., Takeuchi, I., Suzuki, T., Kanamori, T., Hachiya, H. \harvardand\ Okanohara, D.  \harvardyearleft 2010\harvardyearright .
\newblock Least-squares conditional density estimation, {\it IEICE Transactions on Information and Systems} {\bf 93}(3):~583--594.

\harvarditem[Wang et~al.]{Wang, Jiao, Xu, Wang \harvardand\ Yang}{2021}{wang2021deep}
Wang, G., Jiao, Y., Xu, Q., Wang, Y. \harvardand\ Yang, C.  \harvardyearleft 2021\harvardyearright .
\newblock Deep generative learning via schr{\"o}dinger bridge, {\it International Conference on Machine Learning}.

\harvarditem[Wildberger et~al.]{Wildberger, Dax, Buchholz, Green, Macke \harvardand\ Sch\"{o}lkopf}{2023}{wildberger2024flow}
Wildberger, J., Dax, M., Buchholz, S., Green, S., Macke, J.~H. \harvardand\ Sch\"{o}lkopf, B.  \harvardyearleft 2023\harvardyearright .
\newblock Flow matching for scalable simulation-based inference, {\it Advances in Neural Information Processing Systems} \textbf{36}.

\harvarditem[Xu et~al.]{Xu, Liu, Tegmark \harvardand\ Jaakkola}{2022}{xu2022poisson}
Xu, Y., Liu, Z., Tegmark, M. \harvardand\ Jaakkola, T.  \harvardyearleft 2022\harvardyearright .
\newblock Poisson flow generative models, {\it Advances in Neural Information Processing Systems} {\bf 35}.

\harvarditem[Zheng et~al.]{Zheng, Le, Shaul, Lipman, Grover \harvardand\ Chen}{2023}{zheng2023guided}
Zheng, Q., Le, M., Shaul, N., Lipman, Y., Grover, A. \harvardand\ Chen, R.~T.  \harvardyearleft 2023\harvardyearright .
\newblock Guided flows for generative modeling and decision making, {\it arXiv preprint arXiv:2311.13443}.

\harvarditem[Zhou et~al.]{Zhou, Jiao, Liu \harvardand\ Huang}{2023}{zhou2022deep}
Zhou, X., Jiao, Y., Liu, J. \harvardand\ Huang, J.  \harvardyearleft 2023\harvardyearright .
\newblock A deep generative approach to conditional sampling, {\it Journal of the American Statistical Association} {\bf 118}:~1837--1848.

\end{thebibliography}

\begin{thebibliography}{99}
\spacingset{1.2}

\bibitem[Ambrosio, 2004]{ambrosio2004transport}
Ambrosio, L. (2004).   Transport equation and cauchy problem for BV vector fields, {\it Inventiones Mathematicae} {\bf 158}(2): 227–260.

\bibitem[Ambrosio and Crippa, 2014]{ambrosio2014continuity}
Ambrosio,~ L. ~and ~Crippa,~ G. ~(2014).  \quad Continuity equations and ODE flows with non-smooth velocity, {\it Proceedings of the Royal Society of Edinburgh Section A: Mathematics}
{\bf 144}(6): 1191–1244.

\bibitem[Bainov and Simeonov(1992)]{bainov1992integral}
Bainov, D. and Simeonov, P. (1992). {\it Integral Inequalities and Applications}, Vol. 57,
Springer Science \& Business Media.

\bibitem[Bogachev et~al., 2022]{bogachev2022fokker}
Bogachev, V. I., Krylov, N. V., R\"{o}ckner, M. and Shaposhnikov, S. V. (2022).  {\it Fokker–
Planck–Kolmogorov Equations}, Vol. 207, American Mathematical Society.

\bibitem[Boucheron et~al.(2003)]{boucheron2003concentration}
Boucheron, S., Lugosi, G. and Bousquet, O. (2003). Concentration inequalities, {\it Summer
school on machine learning}, Springer, pp. 208–240.

\bibitem[Chen, Jiang, Liao and Zhao (2022)]{chen2022nonparametric}
Chen, M., Jiang, H., Liao, W. and Zhao, T. (2022). Nonparametric regression on {low-dimensional}
manifolds using deep ReLU networks: Function approximation and statistical
recovery, {\it Information and Inference: A Journal of the IMA} {\bf 11}(4): 1203–1253.

\harvarditem[Chen et~al.(2023)]{Chen, Huang, Zhao \harvardand\ Wang}{2023}{chen2023score_sm}
Chen, M., Huang, K., Zhao, T. \harvardand\ Wang, M.  \harvardyearleft 2023\harvardyearright .
\newblock Score approximation, estimation and distribution recovery of diffusion models on low-dimensional data, {\it International Conference on Machine Learning}.

\bibitem[Dai et~al.(2023)]{dai2023lipschitz}
Dai, Y., Gao, Y., Huang, J., Jiao, Y., Kang, L. and Liu, J. (2023). Lipschitz transport
maps via the F\"ollmer flow, {\it arXiv:2309.03490}.

\bibitem[DiPerna and Lions, 1989]{diperna1989ordinary}
DiPerna, R. J. and Lions, P.-L. (1989).   {Ordinary} differential equations, transport theory
and sobolev spaces, {\it Inventiones Mathematicae} {\bf 98}(3): 511–547.

\bibitem[Hyv{\"a}rinen and Dayan(2005)]{hyvarinen2005estimation}
Hyv{\"a}rinen, A. and Dayan, P. (2005). Estimation of non-normalized statistical models by score matching, \textit{Journal of Machine Learning Research}.

\harvarditem[Oko et~al.]{Oko, Akiyama \harvardand\ Suzuki}{2023}{oko2023diffusion_sm}
Oko, K., Akiyama, S. \harvardand\ Suzuki, T.  \harvardyearleft 2023\harvardyearright .
\newblock Diffusion models are minimax optimal distribution estimators, {\it International Conference on Machine Learning}.
\end{thebibliography}
\end{document}